\def\year{2022}\relax
\documentclass[letterpaper]{article} 
\usepackage{aaai22}  
\usepackage{times}  
\usepackage{helvet}  
\usepackage{courier}  
\usepackage[hyphens]{url}  
\usepackage{graphicx} 
\usepackage{natbib}  
\usepackage{caption} 
\DeclareCaptionStyle{ruled}{labelfont=normalfont,labelsep=colon,strut=off} 
\usepackage[utf8]{inputenc} 
\usepackage[T1]{fontenc}    
\usepackage{booktabs}       
\usepackage{amsfonts}       
\usepackage{nicefrac}       
\usepackage{microtype}      
\usepackage{xcolor}         

\usepackage{amsmath}
\usepackage{amsthm}
\usepackage{amssymb}
\usepackage{dsfont}
\usepackage{bbm}
\usepackage{mathtools}
\usepackage{mathrsfs}
\usepackage{theoremref}
\usepackage{multicol}

\usepackage[ruled,vlined, noend]{algorithm2e}
\usepackage{adjustbox}
\usepackage{graphicx,wrapfig,lipsum}

\usepackage{floatrow}
\newfloatcommand{capbtabbox}{table}[][\FBwidth]

\usepackage{blindtext}
\usepackage{caption}
\usepackage{subcaption}

\DeclareMathOperator*{\argmax}{arg\,max}

\title{On Optimizing Interventions in Shared Autonomy}
\author{
    Weihao Tan\textsuperscript{\rm 1}\thanks{Corresponding author}\equalcontrib,
    David Koleczek\textsuperscript{\rm 1}\textsuperscript{\rm 3}\equalcontrib,
    Siddhant Pradhan\textsuperscript{\rm 1}\equalcontrib,
    Nicholas Perello\textsuperscript{\rm 1},
    Vivek Chettiar\textsuperscript{\rm 2},
    Vishal Rohra\textsuperscript{\rm 2},
    Aaslesha Rajaram\textsuperscript{\rm 2},
    Soundararajan Srinivasan\textsuperscript{\rm 2},
    H M Sajjad Hossain\textsuperscript{\rm 2}\footnote{Equal Advising},
    Yash Chandak \textsuperscript{\rm 1}\textsuperscript{$\ddagger$}
}
\affiliations{
    \textsuperscript{\rm 1} University of Massachusetts Amherst, 
    \textsuperscript{\rm 2} Microsoft, 
    \textsuperscript{\rm 3} MassMutual Data Science  \\
    \{weihaotan, dkoleczek, sspradhan, nperello, ychandak\}@umass.edu,  hivishalrohra@gmail.com \\
    \{Vivek.Chettiar, Aaslesha.Rajaram, Soundararajan.Srinivasan, Hossain.H\}@microsoft.com \\
    
}

\usepackage{bibentry}

\begin{document}

\maketitle

\begin{abstract}
Shared autonomy refers to approaches for enabling an autonomous agent to collaborate with a human with the aim of improving human performance.
However, besides improving performance, it may often also be beneficial that the agent concurrently accounts for preserving the user’s experience or satisfaction of collaboration. 
In order to address this additional goal, we examine approaches for improving the user experience by constraining the number of interventions by the autonomous agent.
We propose two model-free reinforcement learning methods that can account for both hard and soft constraints on the number of interventions. 
We show that not only does our method outperform the existing baseline, but also eliminates the need to manually tune a black-box hyperparameter for controlling the level of assistance.
We also provide an in-depth analysis of intervention scenarios in order to further illuminate system understanding.
\end{abstract}

\section{Introduction}

\noindent 
Human-AI collaboration forms an integral part of advancing science in domains where neither advances in independent AI nor human experts are able to fully realize the solution to a problem.
It has shown promising and practical advances in deploying synergistic systems to assist humans. With the assist of these powerful artificial systems, human can perform more safely and smoothly or even above their own ability in many complex and challenging tasks, such as aircraft operations \cite{airplane1}, semi-autonomous driving \cite{cars1}, flying drone control \cite{AssistDrone}, teleoperation \cite{kofman2005teleoperation, aarno2005adaptive} and enabling people with disabilities to enjoy playing video games \cite{XBOX}. A key question that we further explore is how to balance the control between a human and the assistive system in order to not only help the human in achieving a task successfully, but just as critically maintain their satisfaction of collaboration, or level of autonomy.

Shared autonomy provides a framework to allow human and autonomous agents to interact congruently to improve performance at the task, while still allowing the human to maintain control \cite{ATopology}.
Several prior works on shared autonomy rely on specific knowledge of the environment’s dynamics, or the ability to model them accurately \cite{HITLRobotics, HindsightOpt, AssistDrone}, or having data a priori in order to determine the goals of the collaborating human \cite{HumanObjectives, WhereDo, OntheUtility, MinimizingUser}. However, for many tasks the computational dynamics model may be unavailable, or determining user goals accurately, in advance of system design, may not be feasible. Therefore, similar to the works by \cite{SharedAutonomy}, \cite{ResidualPolicy}, and \cite{empowerment} we focus on a model-free approach and do not aim to model user’s goals explicitly. 

An important aspect not addressed by the aforementioned prior works is how to allow humans to use the full depth of their expertise towards achieving their goals. Because the agents serves as assistants, a natural assumption is that the agent should only assist the human minimally or intervene only when necessary, while maintaining good performance. This consideration is especially important in domains such as assistive robotics \cite{HindsightOpt, HITLRobotics, cars1} in reality and assistive agent in video games where users may want to feel as independent as possible while performing the task at hand \cite{verdonck2011irish, palmer2005evaluation}, AI assisted employee training \cite{seo2021} where the AI agent is expected to gradually reduce assistance to ensure the user is learning the objective. Our work focuses on addressing this gap by incentivizing the agent to minimize interventions, while making sure that the human-agent collaboration achieves near-optimal performance. For many tasks, this objective is quite natural. Interventions are intuitively a reasonable proxy to capture user experience because by definition we are optimizing for human control (since more human actions need to be realized), while also achieving greater reward than they could have achieved without assistance.

We introduce two new, model-free reinforcement learning methods: one based on soft-constraints and the other based on hard-constraints on the number of interventions permitted by the user. In the cases where the user is flexible in the number of interventions acceptable, the soft constraint method provides an approach to consider an intervention penalty along with the reward function, and solves it using the dual formulation \cite{puterman2014markov}. On the other hand, if a user desires strict upper-bounds on the acceptable number of interventions, the hard constraint method provides a procedure to include the intervention constraint directly in the state representation ensuring that the constraint is \textit{never} violated.

To benchmark our algorithms, we conduct experiments using simulated human agents in the Lunar Lander \cite{OpenAIGym} and \textit{Super Mario Bros.} \cite{gym-smb} environments. Additionally, for the Lunar Lander environment, we provide an interactive demo with our assistive agents.\footnote{Demo and code used  for  experiments can be found at https://github.com/DavidKoleczek/human\_marl.} We show that our method outperforms the previous baseline \cite{SharedAutonomy} in terms of intervention rate and environment returns. Finally, we also provide analysis of the states where the agent takes control as an additional means of validation of the proposed methods.

\section{Related Work}

\begin{figure}
  \centering
  \includegraphics[scale=0.56]{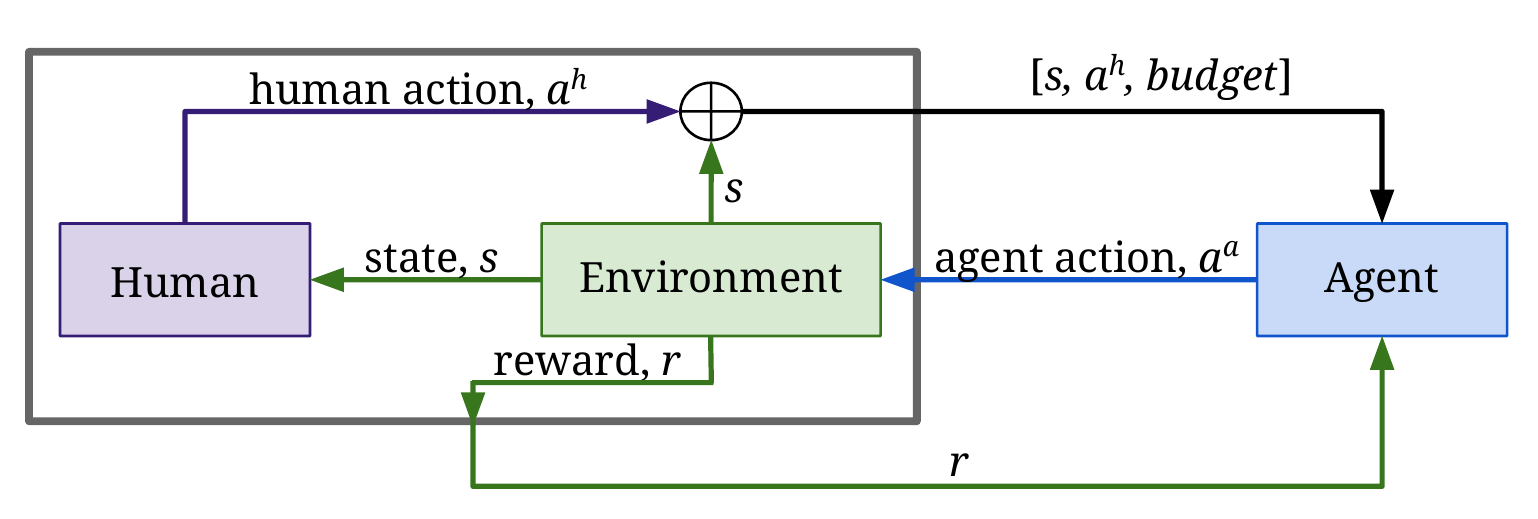}%
  \caption{Overview of the proposed human in the loop reinforcement learning framework.}
  \label{fig:agentenvdiag}
\end{figure}

\textbf{Shared Autonomy.} 
\noindent In shared autonomy, the control of a system is shared by human and agent to accomplish a common goal. Earlier works \cite{crandall2002characterizing, kofman2005teleoperation, you2012assisted} assumed that the agent is aware of the common goal. However, this is a strong assumption in new and previously unseen applications, and as a result, more recent works have proposed agents that can infer the user's intent and predict the user's goal from the user's action and environmental dynamics. They do so by formulating this problem as a partially observable Markov decision process (POMDP). Bayesian inference \cite{HindsightOpt, muelling2017autonomy, sadigh2016information, sadigh2018planning}, inverse reinforcement learning \cite{ng2000algorithms, ratliff2006maximum, levine2012continuous} and hindsight optimization \cite{javdani2015shared}
are the most common approaches to predicting the user's goal in this formulation. These approaches are based on an assumption that the environment dynamics, the goal space, and the user’s policy are known to the agent, which is often violated in practice.
\citet{SharedAutonomy} proposed a model-free shared autonomy framework to relax these assumptions and formulate the problem as a Markov decision process (MDP). This method does not require prior knowledge of environment dynamics, the human's policy, the goal space, and goal representation. Instead, it uses human-in-the-loop reinforcement learning with functional approximation to formulate a policy for the agent learned only from environmental observation and user input. Based on the aforementioned work, \cite{bragg2020fake} proposed Stochastic Q Bumpers. Rather than simply focusing on improving the current performance of a human, they consider the additional objective of helping a human learn, or optimizing \textit{future} performance.

\textbf{Intervention optimization.} In the setting of user assistance, there are only a few works that emphasise the influence of the assistive agent on the user. An agent that intervenes too often will reduce a user's advocacy and ultimately affect their experience in the system.To tackle this challenge,~\citet{SharedAutonomy} employed deep Q-learning to train the assistive agent. Instead of taking the action with the highest q-value, they first sort the actions in terms of similarity to the user provided action, in descending order, using an action-similarity function. A final action is then selected as the one that is closest to the user's action while ensuring its q-value is not significantly worse than the optimal action. This action selection  addresses the intervention problem implicitly.

By tuning the tolerance of the system to suboptimal human suggestions, which is a black-box hyperparameter, this method can provide different levels of assistance. However, designing such an action-similarity function makes this approach an environment-dependent problem. 
\citet{broad2019highly} proposed to minimize the interventions in a similar way. They accept the user's action only when it is close enough to the optimal action as determined by an optimal controller. However, developing such an optimal controller may require complete knowledge of the environment dynamics and the user's goal. Similar to \citet{broad2019highly}, 
\citet{ResidualPolicy} formulate the problem as a constrained MDP where they consider minimizing the amount of interventions an agent makes, subject to the returns of the agent's learned policy being greater than a given constant. In contrast, our work describes a method to maximize returns of the human-AI collaboration, subject to the number of interventions taken being less than a given constant. Additionally, their method only works with on-policy RL algorithms and it needs to set a target return as the constraint, which requires a deep understanding of the environment and user.  \cite{hoque2021lazydagger} and \cite{hoque2021thriftydagger} used imitation learning and supervised learning to predict action discrepancy or the novelty and the risk of the state to decide when to intervene, which have shown promising results on robotic tasks.

\section{Method} \label{sec:method}
\noindent We first introduce the problem setup based on the shared autonomy framework \cite{SharedAutonomy}. We then present our proposed methods that enable the agent to optimize the frequency of interventions, while maximizing the collaborative performance. Our method is model-free and does \textit{not} assume the knowledge of the environment dynamics, the goal space, or, the user’s policy. 

Shared autonomy can be formulated as a MDP, given by a tuple $\{\mathcal{S}, \mathcal{A}_h, \mathcal{A}_a, \mathcal{T}, \mathcal{R}, \gamma \}$, where $\mathcal{S}$ is the set of environment states and $\mathcal A_h$ is the set of actions that can be taken by the human. The set of actions available to the agent is denoted using $\mathcal A_a$. Dynamics, the reward function, and discount factor are represented using $\mathcal{T}$, $\mathcal{R}$ and $\gamma$, respectively. Before choosing an action according to its policy, the agent observes both the environment state $s \in \mathcal S$ and the user's action $a^h \in \mathcal A_h$. Therefore, the effective set of states for the agent is $\overline{S} = \mathcal S \times \mathcal A_h$, and, we define an instance of this set as $\overline{s} \coloneqq [s, a^h]$, a simple concatenation. The MDP that our agent acts in then can be formalized as $\mathcal{M} = \{\overline{\mathcal{S}}, \mathcal{A}_a, \overline{\mathcal{T}}, \mathcal{R}, \gamma \}$, where $\overline{\mathcal{T}}(\overline{s}, a^a, \overline{s}')=\mathcal{T}(s, a^a, s')\pi_h(s', a^{h'})$, where $a^a \in \mathcal{A}_a$, and, $\pi_h$ is the policy of the human user. The goal of the agent is to find the optimal policy $\pi_a^*$ that maximizes the expected discounted sum of rewards:
{\small$\pi_a^* = \argmax \limits_{\pi_a} \mathbb{E}_{\pi}\bigg[\sum_{t=0}^T \gamma^t\mathcal{R}(\overline{s}_t,a_t) \bigg]$.} We provide an overview of the proposed agent-human-environment interaction in Figure \ref{fig:agentenvdiag}.

\subsection{Hard Constrained Shared Autonomy}
A simple way to control the level of assistance is to limit the discrete amount of instances where the assistive agent can intervene, or override user's input in an episode. Thus, we propose to set a \textit{budget}, which is the maximum number of times an agent can intervene in an episode, as a hard constraint to limit the behaviour of agent. When the budget is greater than 0, the agent can take any corrective action. However when the budget is exhausted, the agent is only able to accept the user's action. Furthermore, if the agent still attempts to intervene, we add a penalty to the reward in order to reinforce the need for the agent to assist optimally while there is a residual budget.

We introduce budget as part of the environment observed by the agent. The hard constrained shared autonomy MDP can be formalized as $\mathcal{M}_b = \{\overline{\mathcal{S}}_b,  \mathcal{A}_a, \mathcal{T}_b, \mathcal{R}_b, \gamma \}$, where $s^b \in \overline{\mathcal{S}}_b \coloneqq [s, a^h, b]$, $\mathcal{T}_b(s^b, a^a, {s^b}')=\mathcal{T}(s, a^a, s')\pi_h(s', {a^h}')\mathbb{P}(b'|b, a^a, a^h)$. The reward function $\mathcal{R}_b$ is defined in the following equation:
{\small 
$$\mathcal{R}_b(s^b, a^a, {s^b}') =  R(s, a^a, s') - \begin{cases}
    \lambda& \text{$b = 0 $ and $a^a \neq a^h$} \\
    0 & \text{else}\\
\end{cases}$$
}
Here, the hyperparameter $\lambda \geq 0$ refers to the penalty for intervening after the budget is exhausted. The goal of the agent is still to find the optimal policy $\pi_a^*$ that maximizes the expected discounted sum of rewards:
{\small $\pi_a^* = \argmax \limits_{\pi_a} \mathbb{E}_{\pi}\bigg[\sum_{t=0}^T \gamma^t R(s_t^b,a_t) \bigg]$}. We also define $I(a^a, a^h) = \begin{cases}
        1& \text{$a^a \neq a^h$}\\
        0& \text{$a^a = a^h$}
\end{cases}$ as an indicator variable for the occurrence of an intervention.
The method is summarized in Algorithm \ref{alg:hard constrained}. We refer to this method as the budget method in the rest of the paper.
\begin{algorithm}
\SetAlgoLined
\DontPrintSemicolon

Initialize agent's policy: $\pi_a$, penalty: $\lambda$

\textup{Observe} $s_0$ \\
\While{\textit{not converged}}{
    Initialize budget: $b \in \mathbb{Z}^{0+}$ \\
    \For{$t\gets0 \dots T$}{
        \textup{Sample} $a_t^h \sim \pi_h (a_t^h|s_t)$ \\
        $s_t^b = [s_t,a_t^h,b]$ \\
        \textup{Sample} $a_t^a \sim \pi_a (a_t^a|s_t^b)$ \\
        \uIf {$I(a_t^a, a_t^h) = 1$ and $b > 0$}{
                $b \leftarrow b - 1$ \\
        }
        \uElse{
                $a_t^a \leftarrow a_t^h$
        }
        Execute action $a_t^a$ and observe $s_{t+1}, r_t$\\
        \uIf {$a_t^a \neq a_t^h$ and $b=0$}{
         $r_t \leftarrow r_t - \lambda$}
    }
    Update $\pi_a$ using any method
}

\caption{Hard Constrained Shared Autonomy}
\label{alg:hard constrained}
\end{algorithm}
\subsection{Soft Constrained Shared Autonomy} \label{softalgo}
In hard constrained shared autonomy, we gave the agent an intuitive constraint of a maximum number of interventions. This introduces the problem of balancing utility of the agent and the preservation of user's experience manually using the budget. The optimal budget can be difficult to estimate depending on the environment. Hence, we now relax this hard constraint, and, instead penalize the agent for every intervention it takes, compared to the previous method where we start penalizing the agent only after the exhaustion of the budget. In certain use cases, it may be beneficial to not have a hard cap on the number of interventions. The magnitude of the penalty establishes a trade-off between the increased reward accumulated by intervening over the human action, and the penalty associated with doing so.
As with hard constrained shared autonomy, we introduce a modified reward function:
{\small \begin{align}
\overline{\mathcal{R}}(\overline{s},a^a, \overline{s}') &= \mathcal{R}(s,a,s') + \mathcal{R}_{\texttt{penalty}}(\overline{s},a^a)
\end{align}}
{\small
\begin{align}
\text{where}\quad\quad    \mathcal{R}_{\texttt{penalty}}(\overline{s},a^a) &=
    \begin{cases}
        0& \text{$a^h = a^a$}\\
        -\lambda& \text{$a^h \neq a^a$}
    \end{cases} 
    \nonumber
\end{align}
}
where $\lambda \geq 0$ and $\mathcal{R}_\texttt{penalty}$
penalizes the agent for intervening.
Using $\overline{\mathcal{R}}$, shared autonomy can be recast into the standard reinforcement learning setup. The MDP of this penalty method can be formalized as $\mathcal{M} = \{\overline{\mathcal{S}}, \mathcal{A}_a, \overline{\mathcal{T}}, \overline{\mathcal{R}}, \gamma \}$. The goal of the agent is to find the optimal policy $\pi_a^*$ that maximizes the expected discounted sum of rewards with the modified reward function:
{\small $\pi_a^* = \argmax \limits_{\pi_a} \mathbb{E}_{\pi_a}\bigg[\sum_{t=0}^T \gamma^t\overline{\mathcal{R}}(\overline{s}_t,a_t) \bigg]$}.

Intuitively, $\lambda$ is a hyperparameter that encodes the trade-off between performance and the
user's willingness to receive assistance from the agent.
Large $\lambda$ discourages the agent from intervening, but limits the agent's capability as a consequence. 
In contrast, small $\lambda$ encourages the agent to intervene more which might enable greater performance, but at the cost of potentially interrupting user's experience more often.
However, choosing an optimal value for $\lambda$ is not intuitive and also varies significantly according to the environment and the scale of rewards. 
Instead of leaving the user to fine-tune this sensitive hyperparameter, we propose a method to automatically optimize $\lambda$ using a dual formulation \cite{SACApps}.

Our objective is to find a optimal policy that maximizes expected return while satisfying a minimum expected intervention constraint. Note that we do not discount rewards as it would result in discounting the intervention penalty as well. We cannot justify this approach in our use case as it is unreasonable to allow for more interventions as an episode grows longer. Formally, we want to solve the following constrained optimization problem: 
{\small 
\begin{equation}
    \pi_a^*=\max_{\pi_a}\mathbb{E}_{\pi_a, \pi_h}\left [ \sum_{t=0}^{T} r(s_t, a_t) \right ] \text{s.t.} \; \mathbb{E}_{\pi_a, \pi_h}\left [ \sum_{t=0}^{T}I(a_t^a, a_t^h) \right ] \leq c 
\end{equation}
}
Written as a Lagrangian, we obtain the following constrained objective:
{\small 
\begin{equation}
\max_{\pi_a}\min_{\lambda \geq 0} 
\mathbb{E}_{\pi_a, \pi_h}
\left ( \sum_{t=0}^{T} r(s_t, a_t) +\lambda  \left ( c - \sum_{t=0}^{T}I(a_t^a, a_t^h) \right ) \right )
\end{equation}
}
According to the preceding equation, we would require an entire episode to be able to evaluate the amount of interventions that occur based on our constraint c. If we wish to optimize at every timestep, we can rewrite the constraint to be a linear function of time: $c' = \frac{c}{t}$. In the case of interventions, this can be thought of as the \textit{intervention rate}. 
{\small 
\begin{equation}
\max_{\pi_a}\min_{\lambda \geq 0} 
\mathbb{E}_{\pi_a, \pi_h}
\left ( \sum_{t=0}^{T} \left ( r(s_t, a_t) +\lambda  \left ( c' - I(a_t^a, a_t^h) \right ) \right ) \right )
\end{equation}
}
We perform the maximization of $\pi$ using a model-free RL algorithm such as DDQN \cite{DDQN} or SAC \cite{SACApps}, while we optimize $\lambda$ using the following stochastic gradient descent where the update for each time-step is written as follows. Note that the reward term does not depend on $\lambda$.
{\small 
\begin{equation}
    \lambda \gets \lambda - \alpha \nabla_{\lambda}\left ( r(s_t, a_t) +\lambda  \left ( c' - I(a_t^a, a_t^h) \right ) \right )
\end{equation}
\begin{equation}
    \lambda \gets \lambda - \alpha ( c' - I(a_t^a, a_t^h) )
\end{equation}
}
The final algorithm for this method, which we refer to as the penalty adapting method, is listed in Algorithm 2. At each timestep we update the policy, and then additionally perform a single gradient step on the dual variables. This does not give us an exact solution to the dual problem which is impractical to solve in RL. However, we find that this approach works in practice to automatically tune $\lambda$ according to the constraint on interventions. This "penalty-adapting" can be considered optional if the user desires to set the penalty manually and we provide experimental results with and without it. We refer to the method without penalty-adapting as simply, the penalty method. 

\begin{algorithm}[h!]
\SetAlgoLined
\DontPrintSemicolon
Initialize agent's policy: $\pi_a$, penalty: $\lambda$, learning rate: $\alpha_\lambda$, normalized constraint: $c'$\\
Observe $s_0$ \\
\While{\textit{not converged}}{
    \For{$t\gets0$ \dots $T$}{
        \textup{Sample}  $a_t^h \sim \pi_h(a_t^h|s_t)$\\
        $\overline{s}_t = [s_t, a_t^h]$ \\
        \textup{Sample} $a_{t}^a \sim \pi_a(a_{t}^a|\overline{s}_t)$ \\
        Execute action $a_t^a$ and observe $s_{t+1}, r_t$ \\
        
        \uIf{$I(a_t^a, a_t^h) = 1$}{
            $r_t \leftarrow r_t - \lambda$
        }
        Estimate  $L(\lambda)$ using $(\overline{s}_t, a^a_t, r_t)$ \\
        $\lambda \gets \lambda - \alpha_\lambda\nabla_{\lambda}L(\lambda)$
    }
    Update $\pi_a$ using any method
}
\label{alg:penalty adapting}
\caption{Penalty-Adapting Method }
\end{algorithm}

\begin{figure*}[h!]
  \centering
  \begin{subfigure}{\linewidth}
         \centering
         \begin{subfigure}{0.24\linewidth}
             \centering
             \includegraphics[scale=0.14]{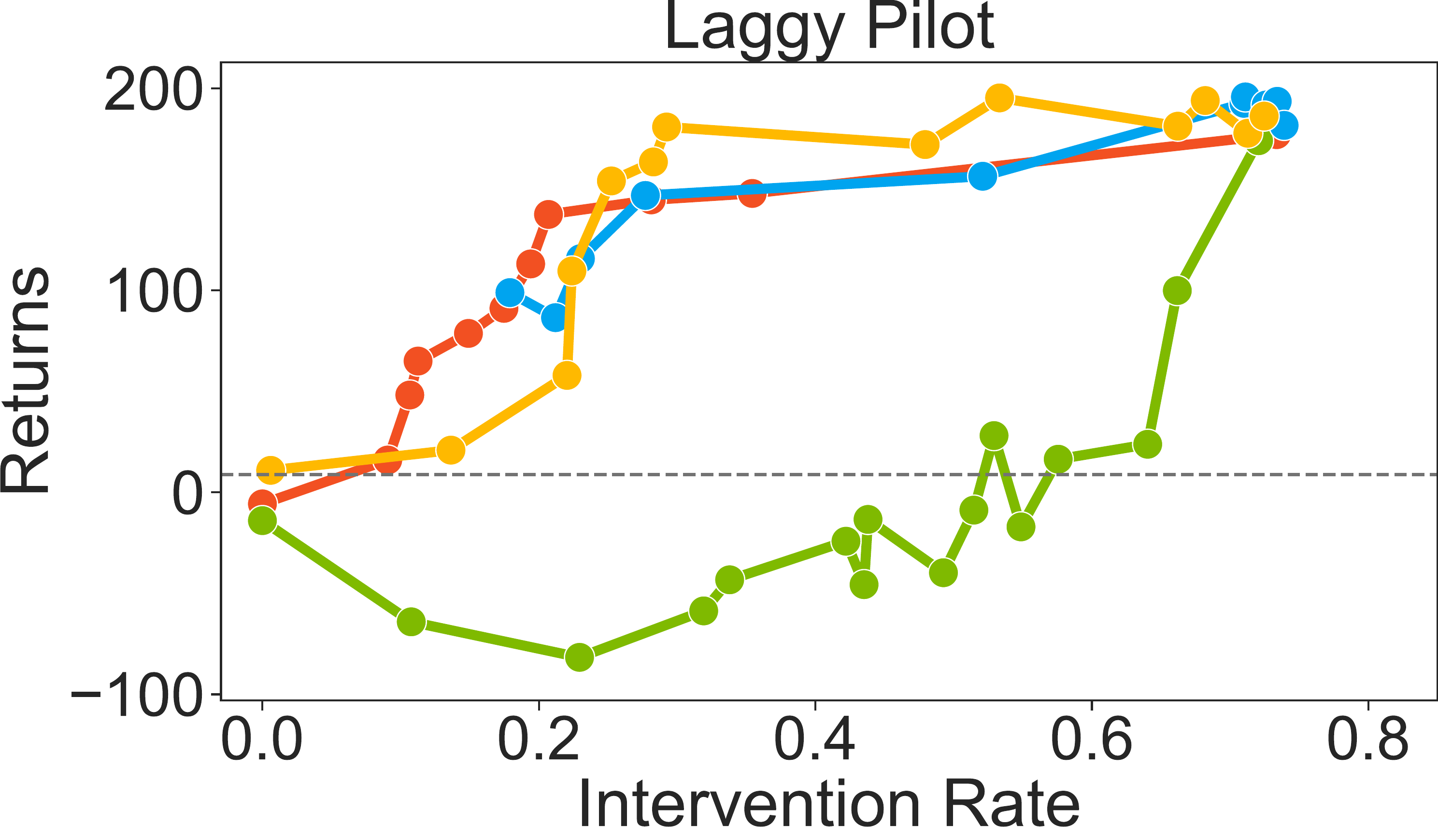}
         \end{subfigure}
         \begin{subfigure}{0.24\linewidth}
             \centering
             \includegraphics[scale=0.14]{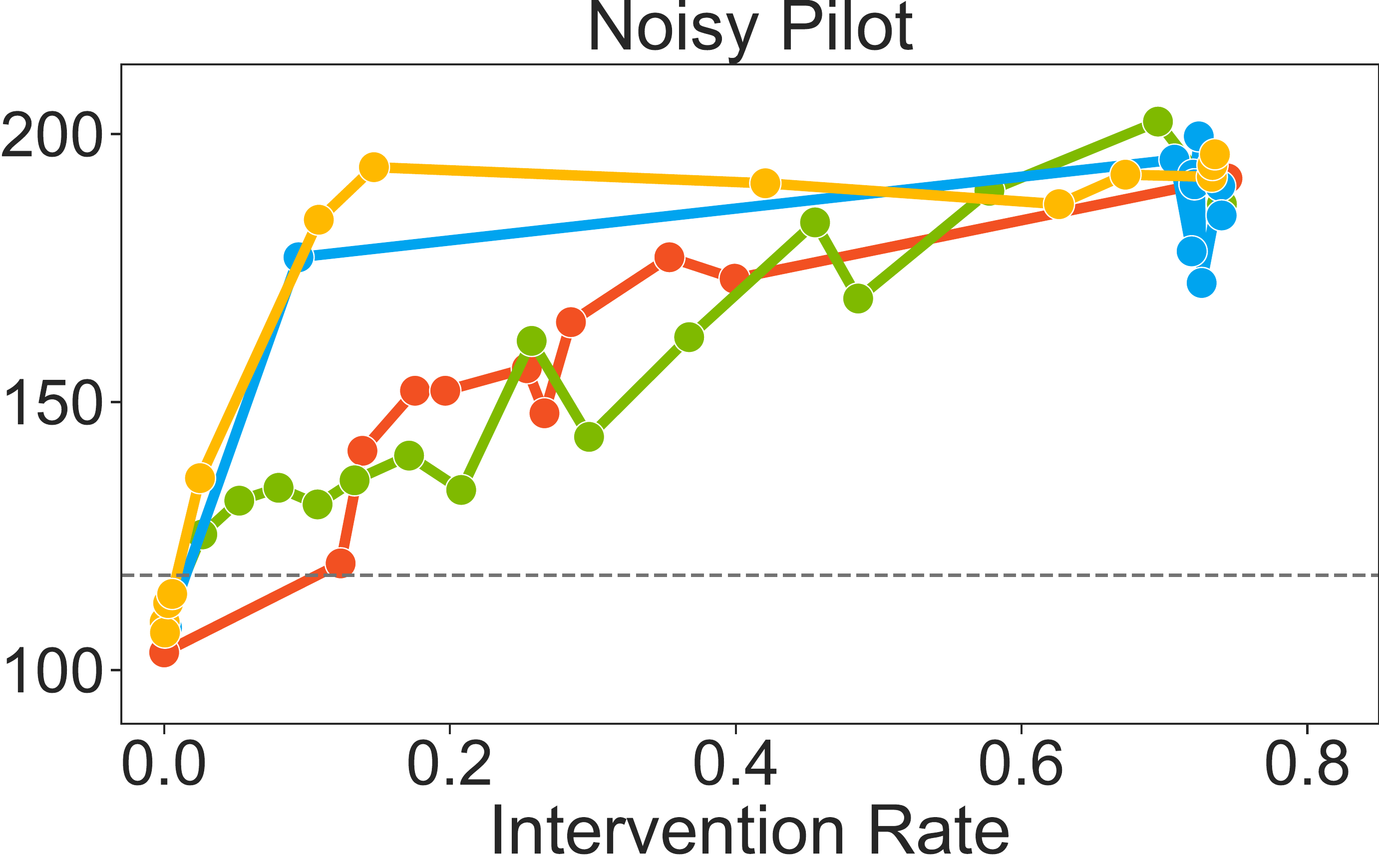}
         \end{subfigure}
         \begin{subfigure}{0.24\linewidth}
             \centering
             \includegraphics[scale=0.14]{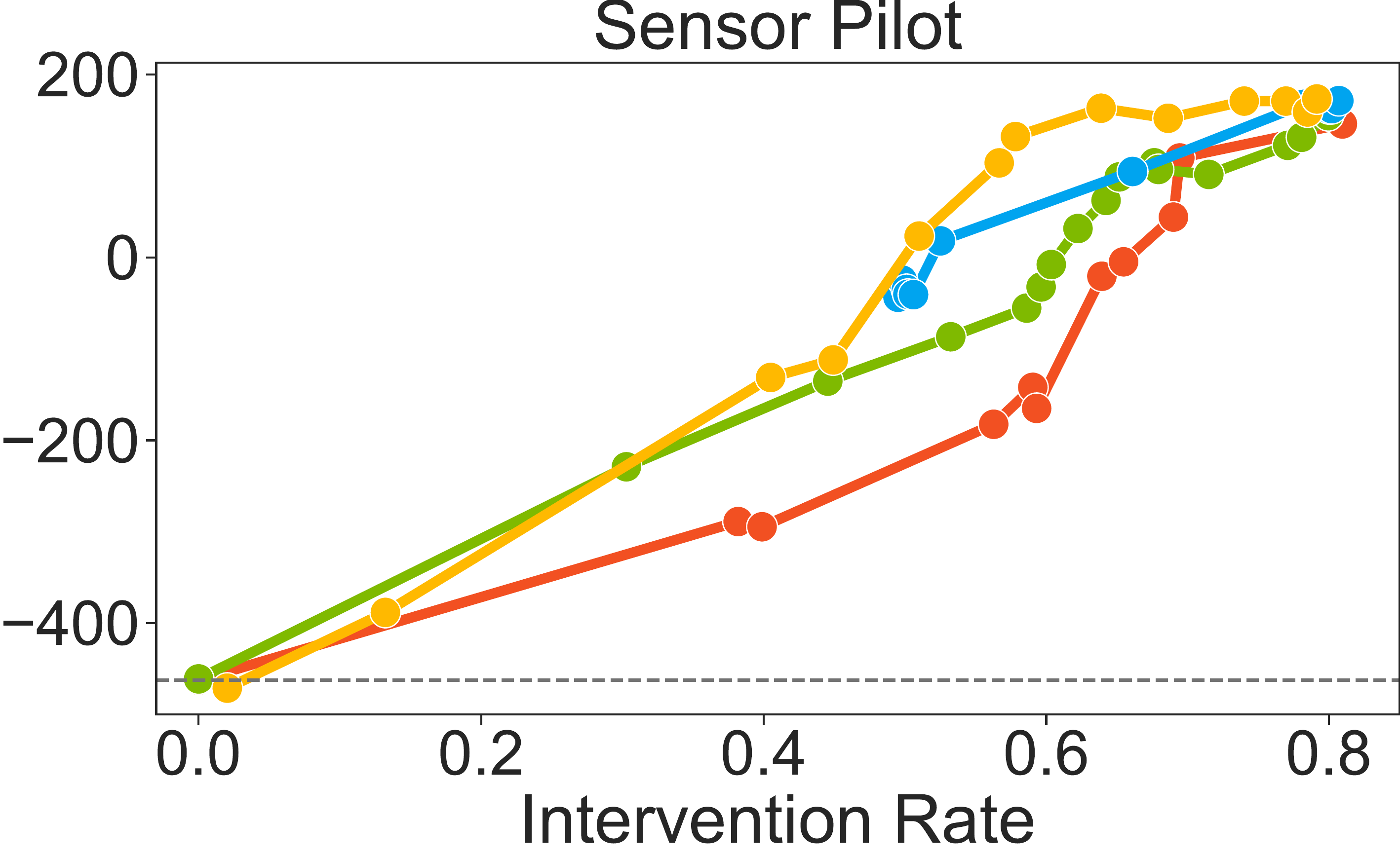}
         \end{subfigure}
         \begin{subfigure}{0.24\linewidth}
             \centering
             \includegraphics[scale=0.14]{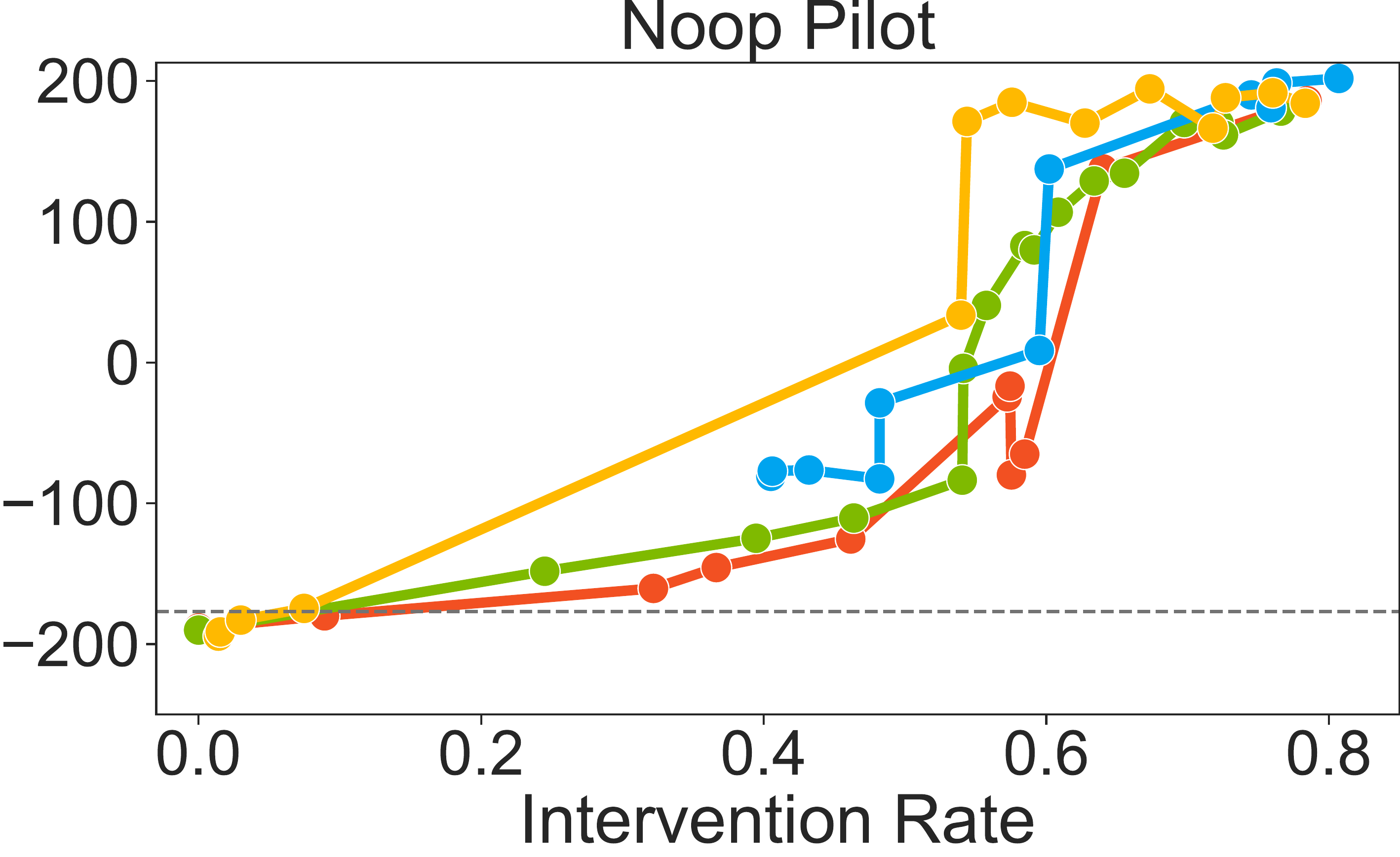}
         \end{subfigure}
     \end{subfigure}
     \begin{subfigure}{\linewidth}
         \centering
         \begin{subfigure}{\linewidth}
             \centering
             \includegraphics[scale=0.15]{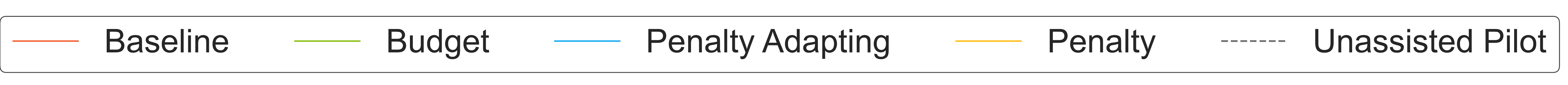}
             
         \end{subfigure}
     \end{subfigure}
  \caption{The performance of various pilots assisted by copilots trained using the baseline \cite{SharedAutonomy} method and our methods on the Lunar Lander environment. The grey dashed line represents the reward achieved by the pilot without assistance from any copilot. The purple dashed line represents the overall optimal reward that we achieved on the environment.
  }
  \label{fig:experiment1}
\end{figure*}
\section{Experiments}
\noindent
In this section we evaluate the performance of our methods on the popular Lunar Lander and \textit{Super Mario Bros.} environments using different simulated users. The goal of our experiments is to test the central hypothesis that the proposed methods can provide varying levels of assistance to improve the user's performance. 

\textbf{Lunar Lander:} A 2D simulation game from OpenAI Gym \cite{OpenAIGym}. The goal of the game is to control the engines of a lunar lander spacecraft to land at a designated landing zone without crashing the spacecraft. Each episode lasts at most 1000 steps. It provides a choice between discrete and continuous action spaces for the pilot to control. For the two-dimensional discrete action space, it consists of six discrete actions that are combinations of the main engine \{on, off\} and the lateral engines \{left, right, off\}. For the continuous action space version, the pilot can fire each engine at any power level. The state space is 8 dimensional: position, velocity, angular position, angular velocity, and whether or not each leg is in contact with the ground. The reward function penalizes speed, tilt, fuel usage, and crashing while rewarding landing at the target location.

\textbf{\textit{Super Mario Bros.}\,World 1-1:} The first stage in the first world of the classic Super Mario game is implemented by the gym-super-mario-bros library \cite{gym-smb}. The goal of the game is to control Mario and reach the target flag at the end of a map. Each episode lasts at most 1600 steps. The action space consists of 12 discrete actions made up of 7 basic actions \{up, down, right, left, jump, run, noop\}. \{left, right\} and \{run, jump\} can be combined into new actions. The state is clipped into 84X84X4 pixel image. The reward penalizes moving left, standing still, and death while rewarding moving right and reaching the target flag.

As our work is closely related to \cite{SharedAutonomy}, we use their methodology as our baseline to compare the performances of our proposed methods and adopt their terminology of referring to the human operator as the pilot and the assistive agent as the copilot. Much like our method, \cite{SharedAutonomy} can also provide different levels of assistance by tuning a hyperparameter, $\alpha$, which is the tolerance of the system to suboptimal human suggestions. If q-values of the user's action is less than the threshold of the tolerance, $\alpha q*$, the copilot will intervene. Comparing the q-values of different actions limits the baseline method to value-based method and consequently making it difficult to work on the continuous action space.

\subsection{Setup}

Similar to \cite{SharedAutonomy}, we design four agents applying different strategies to simulate real human players in the Lunar Lander environment to evaluate our methods:
\textbf{Noop Pilot:} does not apply any action (i.e. always executes a noop action),
\textbf{Laggy Pilot:} the optimal agent with 80\% possibility to repeat the previously executed action,
\textbf{Noisy Pilot:} the optimal agent with 25\% possibility to take a random action), and
\textbf{Sensor Pilot:} simply tries to move toward the landing site by firing its left or right engine. Compared to the Lunar Lander, Super Mario is much more complex and challenging. Human players commonly fail due to jumping into a pit and failing to avoid enemies or obstacles. Hence a noisy pilot that randomly takes an incorrect action (ex. will randomly fail to avoid enemies) is the most appropriate agent to simulate a human's imperfect behavior.

For each simulated pilot, we train the copilot using the baseline method \cite{SharedAutonomy} and our proposed methods: the penalty, budget, and penalty adapting methods. In order to make our method more comparable with the baseline, we use DDQN \cite{DDQN} with experience replay \cite{lin1993reinforcement} to train all the agents on Lunar Lander with discrete action space. Our Q-network is a multi-layer perceptron with two layers of 64 neurons in each layer. We use the same method to train the optimal agent for the simulated pilot. In Super Mario the state and the action representations have different dimensionality, so we cannot directly concatenate them into one vector as the input to the network. We first feed the pixel state into four convolutional layers with 32 3X3 filters and stride 2 and one fully-connected layer with the size of 1152 X 512. Then we concatenate the output with the user's action and feed it into three fully-connected layer with the size of 524X128, 128X64 and 64X12. 
Except the last layer, every layer applies a rectifier nonlinearity. 
In the appendix we show the generalization of our methods by using Soft Actor-Critic \cite{SACApps} to train the sensor pilot on Lunar Lander with a continuous action space.

For each method, we train agents with their respective hyperparameters (tolerance, budget, penalty, and intervention rate) which determine different levels of assistance. A detailed listing of the hyperparameters used is located in the appendix. We train 10 agents with different random seeds for each parameter setting and test each over 100 episodes. Finally, we take the average performance of these 10 agents for the final results. 
\begin{figure}[h!]
    \centering
    \includegraphics[scale=0.26]{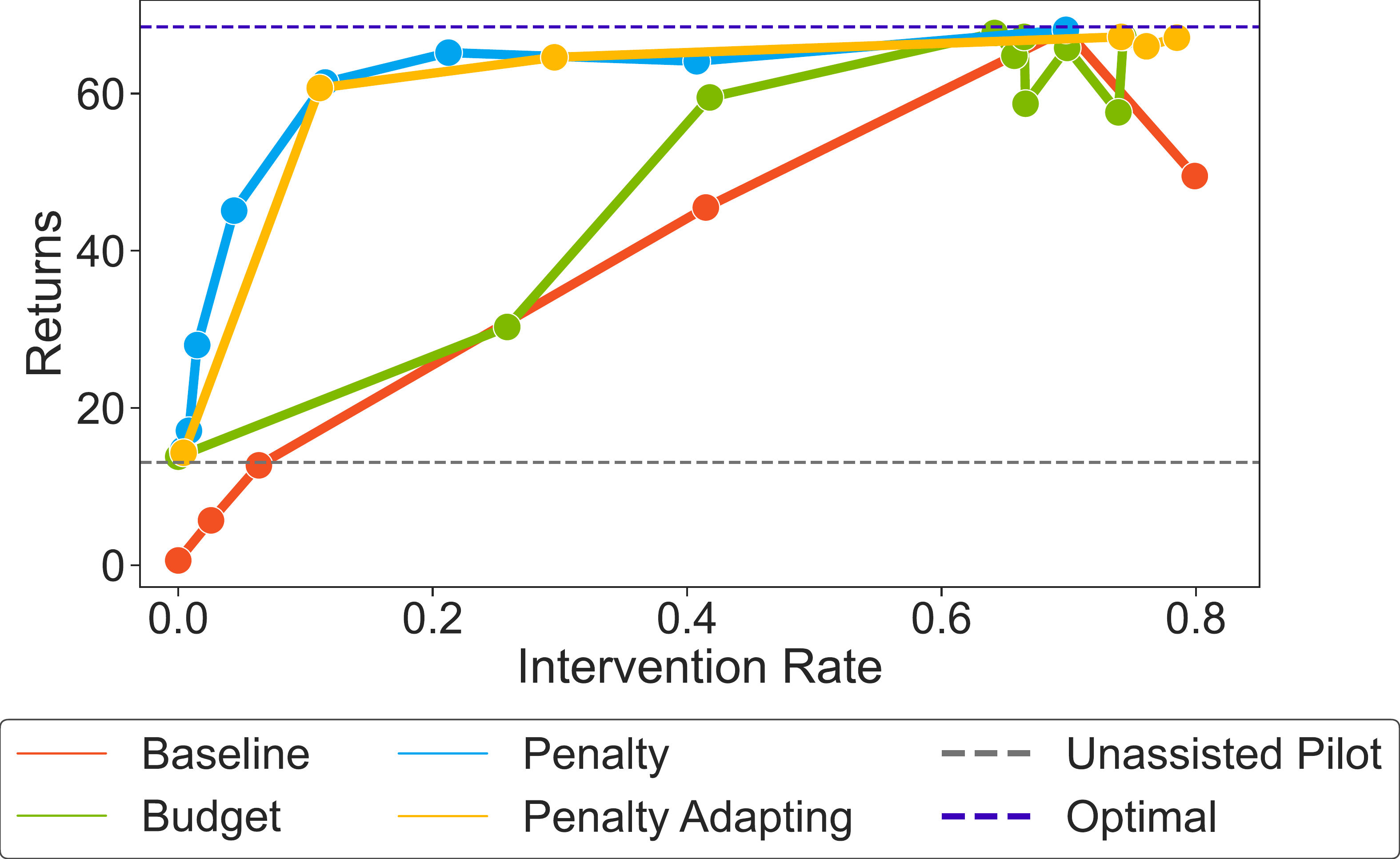}
    \caption{The performance of a noisy pilot assisted by copilots trained using the baseline \cite{SharedAutonomy} method and our methods on the Mario environment.}
    \label{fig:my_label}

\end{figure}

\begin{figure*}[t!]
  \centering
  \includegraphics[scale=0.19]{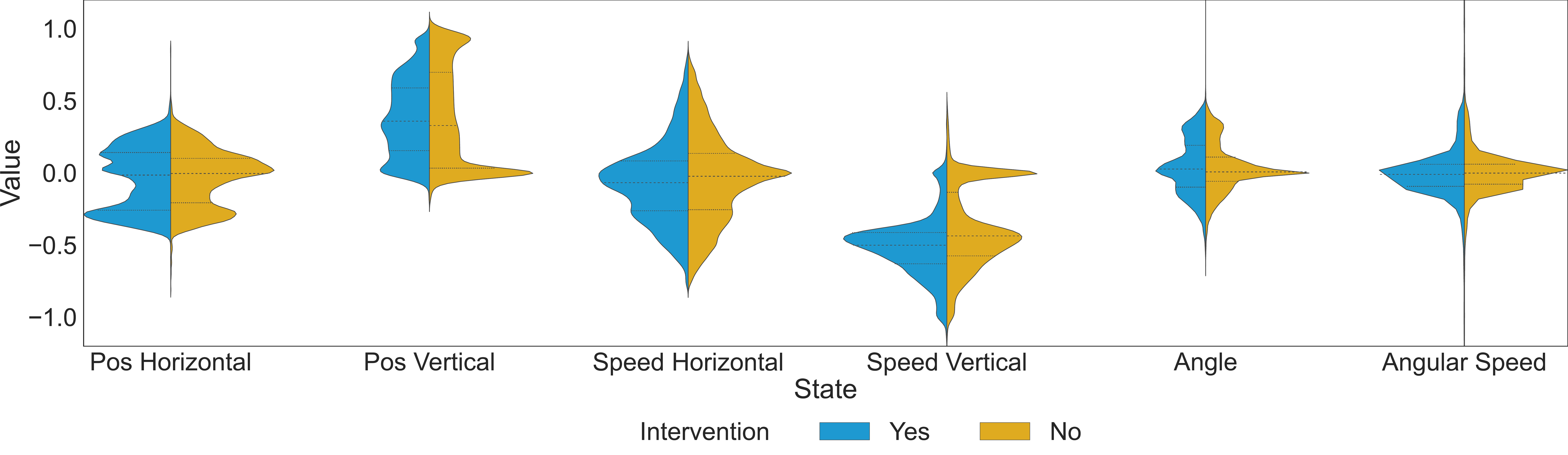}
  \caption{
    A comparison of distributions of state features using data from evaluating a sensor pilot trained with $\lambda$ = 2 on the Lunar Lander environment. Vertical position, vertical speed, and the angle of the lander are the features that show our agent is intervening increasingly at states that are likely to lead to imminent failure, such as when vertical speed is near -0.5.
      }
  \label{fig:experiment2}
\end{figure*}

\subsection{Evaluation} \label{sec:5.1}
We evaluate the performance of our methods with respect to increasing intervention rate. Figures 2 and 3 show the trend of total return at different intervention rates for individual methods on Super Mario and Lunar Lander with discrete action space, respectively. It is evident that our methods greatly improve the performance of most simulated agents. Intuitively, as tolerance threshold $\alpha$ and penalty $\lambda$ decrease, or budget $b$ and intervention rate $c'$ increase, the copilot agent should intervene more frequently and accumulate better reward. Although the hyperparameters are not shown in the figures, the appearance of the data points are consistent with the value of the parameters. For example, data points with lower intervention rate use a lower budget or use a higher penalty. 
So by tuning the hyperparameters, the copilot can assist the pilot with a different frequency of interventions, which means that both of the baseline method and our methods can provide different levels of assistance to the user.
More detailed analysis of the performance of each parameter (including standard error) can be found in the appendix. And the results of the agents trained by soft actor-critic method on lunar lander with continuous action space can be found in the appendix.

Next, we compare the performance of different methods. Since we aim to provide different levels of assistance to the user, the most practical way for comparison is setting a target return, which is the user's expectation, and assess which method achieves the target return with lower intervention rate. From the perspective of intervention rate, we can also form this comparison by setting a target intervention rate, which is the level of assistance that the user needs, and finding which method can compile higher return with this intervention rate. 

\textbf{Budget Method: } Compared to other methods, the budget method distributes more evenly across the intervention rate. The intervention rate is always less than or equal to the ratio of the budget and the total time steps. If we keep increasing the budget at a constant rate, the intervention rate will not change dramatically. This method performs extremely poorly when assisting the laggy pilot. This is due to the laggy pilot always taking significantly more time steps to play the game compared to other simulated pilots. As a result more budget is required to achieve better return. For the other three simulated pilots, the budget method shows almost similar performance to baseline method and worse than penalty and penalty adapting method.

\textbf{Penalty Method: } This method always achieves the optimal reward with lowest intervention rate among the other methods and clearly outperforms the baseline in almost all scenarios. Most notably, the penalty method outperforms the other methods by achieving near optimal returns on the Mario environment with only 0.21 intervention rate. But compared to our other methods, the hyperparameter for this method is environment dependent as we need to have a full understanding of the reward function to determine the appropriate penalty. 

\textbf{Penalty Adapting Method: } This method tends to demonstrate high intervention rate and high reward regardless of the hyperparameter configuration.
Even when the copilot is trained with low intervention rate, such as 0, 0.1 and 0.2, it still accrues significantly higher reward than the unassisted pilot. Though the intervention rate serves as a constraint to limit the behaviour of the copilot, it also guides the copilot when to intervene. It encourages the copilot to intervene when the current intervention rate is less than the constraint. This is the key difference from the budget method, which does not reward intervention. When the benefit of intervening is greater than the penalty, this method will still choose to intervene regardless of the constraint (which is why we refer to it as a soft constraint). Thus, the penalty adapting method has a comparatively high lower bound, even without fine-tuning the intervention rate for the constraint. On the other hand, it exposes the limitation of this method that it cannot handle the case where the user wants very few interventions. Overall, it still has comparable performance with the baseline method, while having the massive upside of not having to tune a black-box assistance level parameter. Effectively by setting an intervention rate of 0.1, the only hyperparameter in this method, the user is directly encoding "I do not want to lose control more than 10\% of the time".

\begin{figure*}
  \centering
  \includegraphics[scale = 0.48]{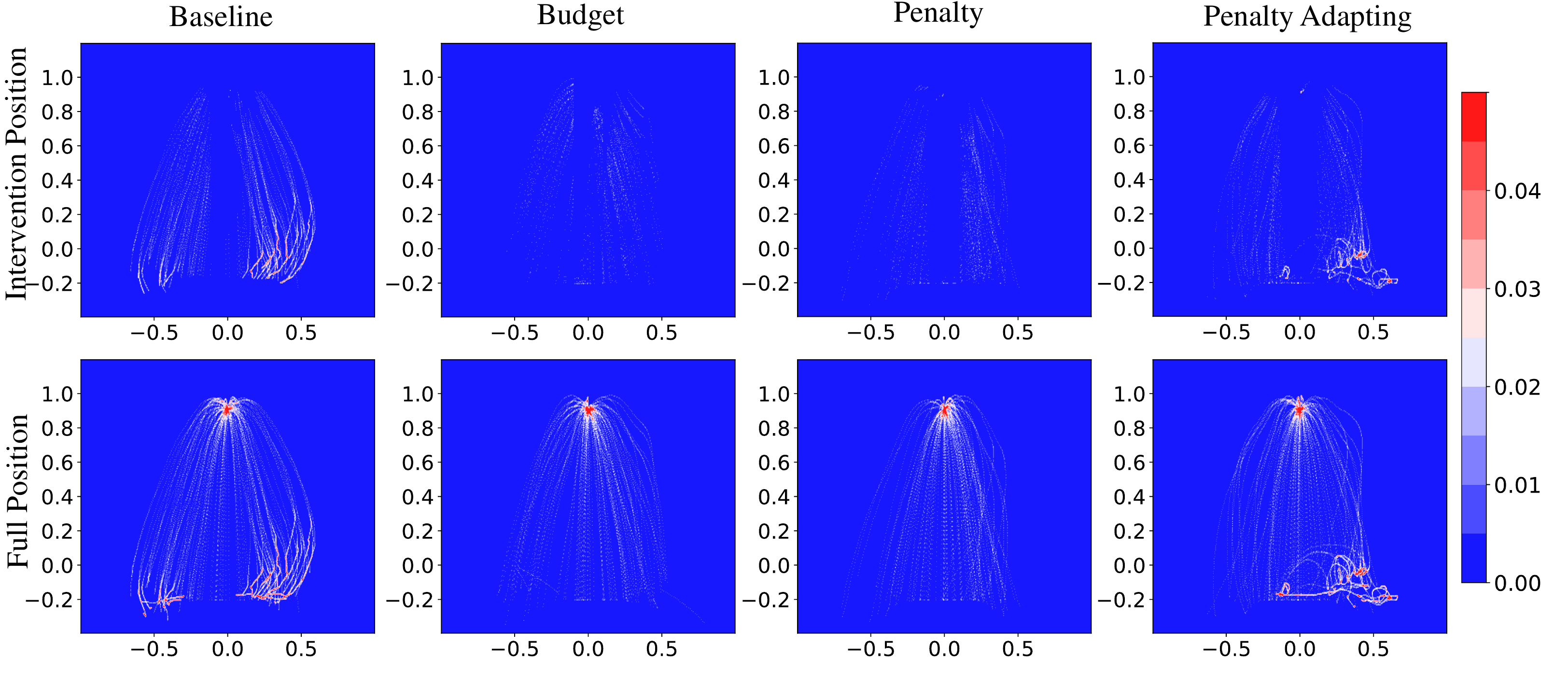}
  \caption{Heat maps visualizing the most frequently visited parts of the state space. The plots in the first row show only the positions where interventions occurred, while the second row shows the full trajectories. The intensity represents the percentage of time spent at that position and is normalized across all trials. The parameters we use to train sensor copilots are: $\alpha = 0.7$, penalty = 5, budget = 50, and intervention rate = 0.7, respectively. The realized intervention rate of each method is within a tolerance of +/-0.05 and each method is run for 100 episodes. }
  \label{fig:experiment3}
\end{figure*}

\subsection{Analysis of Interventions}
In having reviewed performances of individual proposed methods, we discussed their efficiency i.e. how intervention rate and target reward effect their ability to assist. We must also assess whether the agents are meaningfully assisting and intervening where necessary. In this contrast, we focus on the states in which our individual methods are deciding to intervene in. In the appendix we visualize a gridworld environment to show how the penalty term in our Soft Constrained Shared Autonomy method influences the behaviour of the assistive agent.

In Figure 4 we examine the distributions of each feature in the Lunar Lander state space, recorded over 100 episodes, conditioned on if an intervention occurred or not. 
We hypothesize that interventions will be more strongly distributed around states which under unassisted conditions would indicate imminent failure. These states include moments when the lander is moving too fast, its angle is too steep, or it is too far away from the objective (i.e. moving out of bounds). Conversely, distributions corresponding to where an intervention did not occur will be skewed towards values that indicate that the lander is above the goal, making steady progress, or is stable. 
Figure \ref{fig:experiment2} shows kernel density estimations (KDE) of single features of state, where one KDE corresponds to the states where interventions occurred and the corresponding one represents the states where an intervention did not occur. We observe that three of the features exhibit notably different distributions of states depending on if an intervention occurred: vertical position, vertical speed, and angle. In the most straightforward case, vertical position, relatively fewer interventions tend to occur after the lander has nearly landed at positions near 0 and at the beginning of the episode at positions near 1. We also observe that when vertical speed is higher, more interventions tend to occur. Finally, the interventions tend to occur more frequently as the angle deviates further from 0. Features such as horizontal position, horizontal speed, and angular speed exhibiting similar distributions tells us that these states either are not as critical to deciding if the agent should intervene, or the system intervenes before these features would indicate imminent failure. For example, a high angular speed would indicate the lander is spinning out of control, but the agent does not let it get out of hand in the first place. Overall, this analysis gives us confidence that our system is able to intervene when conditions are certainly about to lead to failure, but also it does not intervene more frequently when it is certainly not necessary.

Figure \ref{fig:experiment3} uses heat maps to demonstrate at which (x, y) positions in the Lunar Lander environment interventions happen most frequently. For these experiments, we fixed the landing zone, or the goal location, to be at the origin of the frame. Additionally, we executed these experiments using the sensor pilot as its behavior is the most intuitive to visualize. When the x (horizontal) position of the agent is less than -0.1, the lander is to the left of the goal and the pilot fires its left engine to move right towards the goal. It does the opposite when the x position is greater than 0.1, and the pilot does nothing when it is between -0.1 and 0.1. Both the penalty and penalty adapting methods perfectly capture the strategy of the sensor pilot. Almost all the interventions happen on both sides of the helipad when the pilot is firing its left or right engine. Few interventions happen in the area when the pilot does not fire any engine (i.e. it lands via gravity) and only when the lander is close to the ground are interventions made to maintain balance. Additionally, as Figure \ref{fig:experiment2} showed, more interventions occur when the angle is steep which correlates with why there are more interventions near the ground, but outside the goal area. In this scenario, the copilot must make major adjustments to get the lander to the goal. The baseline method almost captures the strategy, but still makes unnecessary interventions in the area where the agent does nothing.

\section{Conclusion}
\noindent We proposed two frameworks for training assistive reinforcement learning agents with a focus on being able to provide different levels of assistance by controlling the amount or the frequency of the interventions taken by the autonomous agent. Our soft-constrained method allows a user to train an agent "out of the box" with minimal and explainable hyperparameter tuning. While our hard-constrained method provides a means to train under an environment where after it violates any constraint, it is not able to continue taking actions in the environment. We analyzed these methods on the Lunar Lander and \textit{Super Mario Bros. }game environments and showed that it outperformed a previous baseline. We also analyzed where interventions occur with respect to the state space. We believe that shared autonomy with model-free RL is a promising direction to solving this problem because it is able to leverage the wealth of existing RL methods directly, while also giving flexibility to incorporate many other constraints outside of interventions. There is a large space of possible domains and different user experiences to consider. We could potentially try to leverage using real human data from video games and work on implementing our method on a real game. We also wish to test if our method is adaptable to a different distribution of human behavior than at training time and determine if means of goal inference are necessary.

\section{Acknowledgments}
This work began as part of the data science industry mentorship course at the University of Massachusetts Amherst. We thank the teaching assistants Xiang (Lorraine) Li and Rico Angell and the instructors Andrew McCallum and Tom Bernardin for their helpful feedback and mentorship. This work was performed in part using high performance computing equipment obtained under a grant from the Collaborative R\&D Fund managed by the Massachusetts Technology Collaborative. Finally, we thank Sam Devlin and the anonymous reviewers for their valuable feedback.

\bibliography{main}

\begin{thebibliography}{39}
\providecommand{\natexlab}[1]{#1}

\bibitem[{Aarno, Ekvall, and Kragic(2005)}]{aarno2005adaptive}
Aarno, D.; Ekvall, S.; and Kragic, D. 2005.
\newblock Adaptive virtual fixtures for machine-assisted teleoperation tasks.
\newblock In \emph{Proceedings of the 2005 IEEE international conference on
  robotics and automation}, 1139--1144. IEEE.

\bibitem[{Abbink et~al.(2018)Abbink, Carlson, Mulder, de~Winter, Aminravan,
  Gibo, and Boer}]{ATopology}
Abbink, D.~A.; Carlson, T.; Mulder, M.; de~Winter, J. C.~F.; Aminravan, F.;
  Gibo, T.~L.; and Boer, E.~R. 2018.
\newblock A Topology of Shared Control Systems—Finding Common Ground in
  Diversity.
\newblock \emph{IEEE Transactions on Human-Machine Systems}, 48(5): 509--525.

\bibitem[{Backman, Kulić, and Chung(2021)}]{AssistDrone}
Backman, K.; Kulić, D.; and Chung, H. 2021.
\newblock Learning to Assist Drone Landings.
\newblock \emph{IEEE Robotics and Automation Letters}, 6(2): 3192--3199.

\bibitem[{Bragg and Brunskill(2020)}]{bragg2020fake}
Bragg, J.; and Brunskill, E. 2020.
\newblock Fake it till you make it: Learning-compatible performance support.
\newblock In \emph{Uncertainty in Artificial Intelligence}, 915--924. PMLR.

\bibitem[{Broad, Murphey, and Argall(2019)}]{broad2019highly}
Broad, A.; Murphey, T.; and Argall, B. 2019.
\newblock Highly parallelized data-driven MPC for minimal intervention shared
  control.
\newblock \emph{arXiv preprint arXiv:1906.02318}.

\bibitem[{Brockman et~al.(2016)Brockman, Cheung, Pettersson, Schneider,
  Schulman, Tang, and Zaremba}]{OpenAIGym}
Brockman, G.; Cheung, V.; Pettersson, L.; Schneider, J.; Schulman, J.; Tang,
  J.; and Zaremba, W. 2016.
\newblock OpenAI Gym.
\newblock arXiv:1606.01540.

\bibitem[{Carroll et~al.(2019)Carroll, Shah, Ho, Griffiths, Seshia, Abbeel, and
  Dragan}]{OntheUtility}
Carroll, M.; Shah, R.; Ho, M.~K.; Griffiths, T.; Seshia, S.; Abbeel, P.; and
  Dragan, A. 2019.
\newblock On the Utility of Learning about Humans for Human-AI Coordination.
\newblock In \emph{Advances in Neural Information Processing Systems}.

\bibitem[{Crandall and Goodrich(2002)}]{crandall2002characterizing}
Crandall, J.~W.; and Goodrich, M.~A. 2002.
\newblock Characterizing efficiency of human robot interaction: A case study of
  shared-control teleoperation.
\newblock In \emph{IEEE/RSJ international conference on intelligent robots and
  systems}, volume~2, 1290--1295. IEEE.

\bibitem[{de~Winter and Dodou(2011)}]{cars1}
de~Winter, J.; and Dodou, D. 2011.
\newblock Preparing drivers for dangerous situations: A critical reflection on
  continuous shared control.
\newblock In \emph{2011 IEEE International Conference on Systems, Man, and
  Cybernetics}, 1050--1056.

\bibitem[{Du et~al.(2020)Du, Tiomkin, Kiciman, Polani, Abbeel, and
  Dragan}]{empowerment}
Du, Y.; Tiomkin, S.; Kiciman, E.; Polani, D.; Abbeel, P.; and Dragan, A. 2020.
\newblock Ave: Assistance via empowerment.
\newblock \emph{Advances in Neural Information Processing Systems}, 33.

\bibitem[{Gopinath, Jain, and Argall(2017)}]{HITLRobotics}
Gopinath, D.; Jain, S.; and Argall, B.~D. 2017.
\newblock Human-in-the-Loop Optimization of Shared Autonomy in Assistive
  Robotics.
\newblock \emph{IEEE Robotics and Automation Letters}, 2(1): 247--254.

\bibitem[{Haarnoja et~al.(2018)Haarnoja, Zhou, Hartikainen, Tucker, Ha, Tan,
  Kumar, Zhu, Gupta, Abbeel, and Levine}]{SACApps}
Haarnoja, T.; Zhou, A.; Hartikainen, K.; Tucker, G.; Ha, S.; Tan, J.; Kumar,
  V.; Zhu, H.; Gupta, A.; Abbeel, P.; and Levine, S. 2018.
\newblock Soft Actor-Critic Algorithms and Applications.
\newblock \emph{CoRR}, abs/1812.05905.

\bibitem[{Hasselt, Guez, and Silver(2016)}]{DDQN}
Hasselt, H.~V.; Guez, A.; and Silver, D. 2016.
\newblock Deep Reinforcement Learning with Double Q-Learning.
\newblock In \emph{Proceedings of the AAAI Conference on Artificial
  Intelligence}.

\bibitem[{Hoque et~al.(2021{\natexlab{a}})Hoque, Balakrishna, Novoseller,
  Wilcox, Brown, and Goldberg}]{hoque2021thriftydagger}
Hoque, R.; Balakrishna, A.; Novoseller, E.; Wilcox, A.; Brown, D.~S.; and
  Goldberg, K. 2021{\natexlab{a}}.
\newblock ThriftyDAgger: Budget-aware novelty and risk gating for interactive
  imitation learning.
\newblock \emph{arXiv preprint arXiv:2109.08273}.

\bibitem[{Hoque et~al.(2021{\natexlab{b}})Hoque, Balakrishna, Putterman, Luo,
  Brown, Seita, Thananjeyan, Novoseller, and Goldberg}]{hoque2021lazydagger}
Hoque, R.; Balakrishna, A.; Putterman, C.; Luo, M.; Brown, D.~S.; Seita, D.;
  Thananjeyan, B.; Novoseller, E.; and Goldberg, K. 2021{\natexlab{b}}.
\newblock LazyDAgger: Reducing Context Switching in Interactive Imitation
  Learning.
\newblock \emph{arXiv preprint arXiv:2104.00053}.

\bibitem[{Javdani et~al.(2018)Javdani, Admoni, Pellegrinelli, Srinivasa, and
  Bagnell}]{HindsightOpt}
Javdani, S.; Admoni, H.; Pellegrinelli, S.; Srinivasa, S.~S.; and Bagnell,
  J.~A. 2018.
\newblock Shared autonomy via hindsight optimization for teleoperation and
  teaming.
\newblock \emph{The International Journal of Robotics Research}, 37(7):
  717--742.

\bibitem[{Javdani, Bagnell, and Srinivasa(2016)}]{MinimizingUser}
Javdani, S.; Bagnell, J.~A.; and Srinivasa, S.~S. 2016.
\newblock Minimizing user cost for shared autonomy.
\newblock In \emph{2016 11th ACM/IEEE International Conference on Human-Robot
  Interaction (HRI)}, 621--622.

\bibitem[{Javdani, Srinivasa, and Bagnell(2015)}]{javdani2015shared}
Javdani, S.; Srinivasa, S.~S.; and Bagnell, J.~A. 2015.
\newblock Shared Autonomy via Hindsight Optimization.
\newblock \emph{Robotics science and systems: online proceedings}, 2015.

\bibitem[{Kauten(2018)}]{gym-smb}
Kauten, C. 2018.
\newblock {S}uper {M}ario {B}ros for {O}pen{AI} {G}ym.
\newblock GitHub.

\bibitem[{Kofman et~al.(2005)Kofman, Wu, Luu, and
  Verma}]{kofman2005teleoperation}
Kofman, J.; Wu, X.; Luu, T.~J.; and Verma, S. 2005.
\newblock Teleoperation of a robot manipulator using a vision-based human-robot
  interface.
\newblock \emph{IEEE transactions on industrial electronics}, 52(5):
  1206--1219.

\bibitem[{Levine and Koltun(2012)}]{levine2012continuous}
Levine, S.; and Koltun, V. 2012.
\newblock Continuous inverse optimal control with locally optimal examples.
\newblock \emph{arXiv preprint arXiv:1206.4617}.

\bibitem[{Lin(1993)}]{lin1993reinforcement}
Lin, L.-J. 1993.
\newblock Reinforcement learning for robots using neural networks.
\newblock Technical report, Carnegie-Mellon Univ Pittsburgh PA School of
  Computer Science.

\bibitem[{Matni and Oishi(2008)}]{airplane1}
Matni, N.; and Oishi, M. 2008.
\newblock Reachability-based abstraction for an aircraft landing under shared
  control.
\newblock In \emph{2008 American Control Conference}, 2278--2284.

\bibitem[{Muelling et~al.(2017)Muelling, Venkatraman, Valois, Downey, Weiss,
  Javdani, Hebert, Schwartz, Collinger, and Bagnell}]{muelling2017autonomy}
Muelling, K.; Venkatraman, A.; Valois, J.-S.; Downey, J.~E.; Weiss, J.;
  Javdani, S.; Hebert, M.; Schwartz, A.~B.; Collinger, J.~L.; and Bagnell,
  J.~A. 2017.
\newblock Autonomy Infused Teleoperation with Application to Brain Computer
  Interface Controlled Manipulation.
\newblock \emph{Autonomous Robots}, 41(6): 1401--1422.

\bibitem[{Ng, Russell et~al.(2000)}]{ng2000algorithms}
Ng, A.~Y.; Russell, S.~J.; et~al. 2000.
\newblock Algorithms for inverse reinforcement learning.
\newblock In \emph{Icml}, volume~1, 2.

\bibitem[{Palmer, Thursfield, and Judge(2005)}]{palmer2005evaluation}
Palmer, P.; Thursfield, C.; and Judge, S. 2005.
\newblock An evaluation of the psychosocial impact of assistive devices scale.
\newblock In \emph{Assistive technology: from virtuality to reality}, 16,
  740--744. IOS Press.

\bibitem[{Puterman(2014)}]{puterman2014markov}
Puterman, M.~L. 2014.
\newblock \emph{Markov decision processes: discrete stochastic dynamic
  programming}.
\newblock John Wiley \& Sons.

\bibitem[{Raffin et~al.(2019)Raffin, Hill, Ernestus, Gleave, Kanervisto, and
  Dormann}]{stable-baselines3}
Raffin, A.; Hill, A.; Ernestus, M.; Gleave, A.; Kanervisto, A.; and Dormann, N.
  2019.
\newblock Stable Baselines3.
\newblock \url{https://github.com/DLR-RM/stable-baselines3}.

\bibitem[{Ratliff, Bagnell, and Zinkevich(2006)}]{ratliff2006maximum}
Ratliff, N.~D.; Bagnell, J.~A.; and Zinkevich, M.~A. 2006.
\newblock Maximum margin planning.
\newblock In \emph{Proceedings of the 23rd international conference on Machine
  learning}, 729--736.

\bibitem[{Reddy, Dragan, and Levine(2018{\natexlab{a}})}]{SharedAutonomy}
Reddy, S.; Dragan, A.; and Levine, S. 2018{\natexlab{a}}.
\newblock Shared Autonomy via Deep Reinforcement Learning.
\newblock In \emph{Proceedings of Robotics: Science and Systems}. Pittsburgh,
  Pennsylvania.

\bibitem[{Reddy, Dragan, and Levine(2018{\natexlab{b}})}]{WhereDo}
Reddy, S.; Dragan, A.; and Levine, S. 2018{\natexlab{b}}.
\newblock Where Do You Think You\textquotesingle re Going?: Inferring Beliefs
  about Dynamics from Behavior.
\newblock In \emph{Advances in Neural Information Processing Systems}.

\bibitem[{Reddy et~al.(2020)Reddy, Dragan, Levine, Legg, and
  Leike}]{HumanObjectives}
Reddy, S.; Dragan, A.~D.; Levine, S.; Legg, S.; and Leike, J. 2020.
\newblock Learning Human Objectives by Evaluating Hypothetical Behavior.
\newblock In \emph{International Conference on Learning Representations}.

\bibitem[{Sadigh et~al.(2018)Sadigh, Landolfi, Sastry, Seshia, and
  Dragan}]{sadigh2018planning}
Sadigh, D.; Landolfi, N.; Sastry, S.~S.; Seshia, S.~A.; and Dragan, A.~D. 2018.
\newblock Planning for cars that coordinate with people: leveraging effects on
  human actions for planning and active information gathering over human
  internal state.
\newblock \emph{Autonomous Robots}, 42(7): 1405--1426.

\bibitem[{Sadigh et~al.(2016)Sadigh, Sastry, Seshia, and
  Dragan}]{sadigh2016information}
Sadigh, D.; Sastry, S.~S.; Seshia, S.~A.; and Dragan, A. 2016.
\newblock Information gathering actions over human internal state.
\newblock In \emph{2016 IEEE/RSJ International Conference on Intelligent Robots
  and Systems (IROS)}, 66--73. IEEE.

\bibitem[{Schaff and Walter(2020)}]{ResidualPolicy}
Schaff, C.; and Walter, M. 2020.
\newblock {Residual Policy Learning for Shared Autonomy}.
\newblock In \emph{Proceedings of Robotics: Science and Systems}. Corvalis,
  Oregon, USA.

\bibitem[{Seo et~al.(2021)Seo, Kennedy{-}Metz, Zenati, Shah, Dias, and
  Unhelkar}]{seo2021}
Seo, S.; Kennedy{-}Metz, L.~R.; Zenati, M.~A.; Shah, J.~A.; Dias, R.~D.; and
  Unhelkar, V.~V. 2021.
\newblock Towards an {AI} Coach to Infer Team Mental Model Alignment in
  Healthcare.
\newblock \emph{CoRR}, abs/2102.08507.

\bibitem[{Verdonck, McCormack, and Chard(2011)}]{verdonck2011irish}
Verdonck, M.; McCormack, C.; and Chard, G. 2011.
\newblock Irish occupational therapists' views of electronic assistive
  technology.
\newblock \emph{British Journal of Occupational Therapy}, 74(4): 185--190.

\bibitem[{XBOX(2019)}]{XBOX}
XBOX. 2019.
\newblock Xbox Adaptive Controller: Xbox.

\bibitem[{You and Hauser(2012)}]{you2012assisted}
You, E.; and Hauser, K. 2012.
\newblock Assisted teleoperation strategies for aggressively controlling a
  robot arm with 2d input.
\newblock In \emph{Robotics: science and systems}, volume~7, 354.

\end{thebibliography}

\newpage

\onecolumn

\section{Appendix}
In this appendix, we first derive transition function dynamics for our custom MDPs and then provide the results trained with Soft Actor-Critic \citep{SACApps} on Lunar Lander with continuous action space to show that different from our baseline \citep{SharedAutonomy} , our method can work with both discrete and continuous action space. Finally, we demonstrate more experiment results about the performance with different hyperparameters of different methods to show how our methods provide different levels of assistance by tuning the hyperparameters on Lunar Lander with discrete action space and Super Mario environments.

\subsection{Transition Function Derivations}

In accordance with reinforcement learning literature, we have:

\begin{equation*}
    \mathcal{T}(s,a,s') = \mbox{Pr}(s_{t+1}=s'|s_t = s, a_t = a)
\end{equation*}
As a shorthand, we write this as:
\begin{equation*}
    \mathcal{T}(s,a,s') = \mbox{Pr}(s'|s, a)
\end{equation*}

In Section 3, we represent the transition function $\overline{\mathcal{T}}$ of our MDP $\mathcal{M}$ in terms of a generic transition function $\mathcal{T}$ corresponding to any valid environment in reinforcement learning. This is done to ensure that $\mathcal{M}$ indeed is a valid MDP with its transition function corresponding to a valid conditional probability distribution. We derive it as follows:
\begin{equation}
    \overline{\mathcal{T}}(\overline{s},  a^a, \overline{s}') = \mbox{Pr}(\overline{s}'| \overline{s},  a^a)
\end{equation}

\begin{equation}
    \therefore \overline{\mathcal{T}}(\overline{s},  a^a, \overline{s}') = \mbox{Pr}(s', a^{h'}| s, a^h,  a^a)
    \label{2}
\end{equation}
Here, \eqref{2} comes from the definition of $\overline{s}$.

\begin{equation}
    \therefore \overline{\mathcal{T}}(\overline{s},  a^a, \overline{s}') = \mbox{Pr}(a^{h'}|s',  s, a^h,  a^a) \times \mbox{Pr}(s'|  s, a^h,  a^a)
    \label{3}
\end{equation}

\begin{equation}
    \therefore \overline{\mathcal{T}}(\overline{s},  a^a, \overline{s}') = \mbox{Pr}(a^{h'}|s') \times \mbox{Pr}(s'|  s,  a^a)
    \label{4}
\end{equation}
We get to \eqref{4} from \eqref{3} by noting two crucial observations from the definition of our MDP: The human action $a^{h'}$ at time $t+1$ does not depend on state $s$, human action $a^h$, and agent action $a^a$ at time $t$, given the state $s'$ at time $t+1$. Similarly, for the generic MDP which $\mathcal{T}$ belongs to, we note that state $s'$ at time $t+1$ does not depend on the human action $a^h$ at time $t$ given the state $s$, and the agent action $a^a$ at time $t$. Both MDPs are structured around the agent's action space. 
\begin{equation}
    \therefore \overline{\mathcal{T}}(\overline{s},  a^a, \overline{s}') = \pi_h(s', a^{h'}) \times \mathcal{T}(s, a^a, s')
    \label{5}
\end{equation}

Similarly, for our MDP $\mathcal{M}_b$ for Hard Constrained Shared Autonomy, we have:

\begin{equation}
    \mathcal{T}_b(s^b,  a^a, s^{b'}) = \mbox{Pr}(s^{b'}| s^b,  a^a)
\end{equation}

\begin{equation}
    \mathcal{T}_b(s^b,  a^a, s^{b'}) = \mbox{Pr}(s', a^{h'}, b'| s, a^h, b, a^a)
    \label{7}
\end{equation}
Here \eqref{7} comes from the definition of $s^b$.

\begin{equation}
    \therefore \mathcal{T}_b(s^b,  a^a, s^{b'}) = \mbox{Pr}(a^{h'}|s', b', s, a^h, b, a^a) \times \mbox{Pr}(s', b'| s, a^h, b, a^a)
    \label{8}
\end{equation}

\begin{equation}
    \therefore \mathcal{T}_b(s^b,  a^a, s^{b'}) = \mbox{Pr}(a^{h'}|s', b', s, a^h, b, a^a) \times \mbox{Pr}(s' | b', s, a^h, b, a^a) \times \mbox{Pr}(b'| s, a^h, b, a^a)
    \label{9}
\end{equation}

\begin{equation}
    \therefore \mathcal{T}_b(s^b,  a^a, s^{b'}) = \mbox{Pr}(a^{h'}|s') \times \mbox{Pr}(s' | s, a^a) \times \mbox{Pr}(b'|b,  a^h, a^a)
    \label{10}
\end{equation}

Here, \eqref{10} comes from similar observations as discussed above. There is an addition term $\mbox{Pr}(b'|b,  a^h, a^a)$. This term is derived as the budget $b'$ at time $t+1$ does not depend on the state $s$ at time $t$, given the budget $b$, human action $a^h$, and agent action $a^a$ at time $t$. A further note on this probability distribution: This probability function is deterministic. If $b=0$, then $b'=0 \forall a^h, a^a$. This is because the budget is exhausted. On the other hand, if $b \neq 0$, then $b'=b$ if $a^h=a^a$, and $b'= b-1$ otherwise. 
\begin{equation}
    \therefore \mathcal{T}_b(s^b,  a^a, s^{b'}) = \pi_h(s', a^{h'}) \times \mathcal{T}(s,a,s') \times \mbox{Pr}(b'|b,  a^h, a^a)
    \label{11}
\end{equation}

\subsection{Lunar Lander with Continuous Action Space}
We provide additional results showcasing the ability of our methods to be easily extendable to a continuous domain, in this case continuous lunar lander \cite{OpenAIGym}. Continuous lunar lander is similar to the discrete version we described in the Experiments section. The only difference is instead of six discrete actions, there are two continuous actions. One action controls main engine and is in the range [-1.0, +1.0] where -1 corresponds to fully off and +1 is on. The second action is for firing the left and right engines and for [-0.5, +0.5] the engine is off.

\begin{figure}[h!]
  \centering
  \includegraphics[scale=0.26]{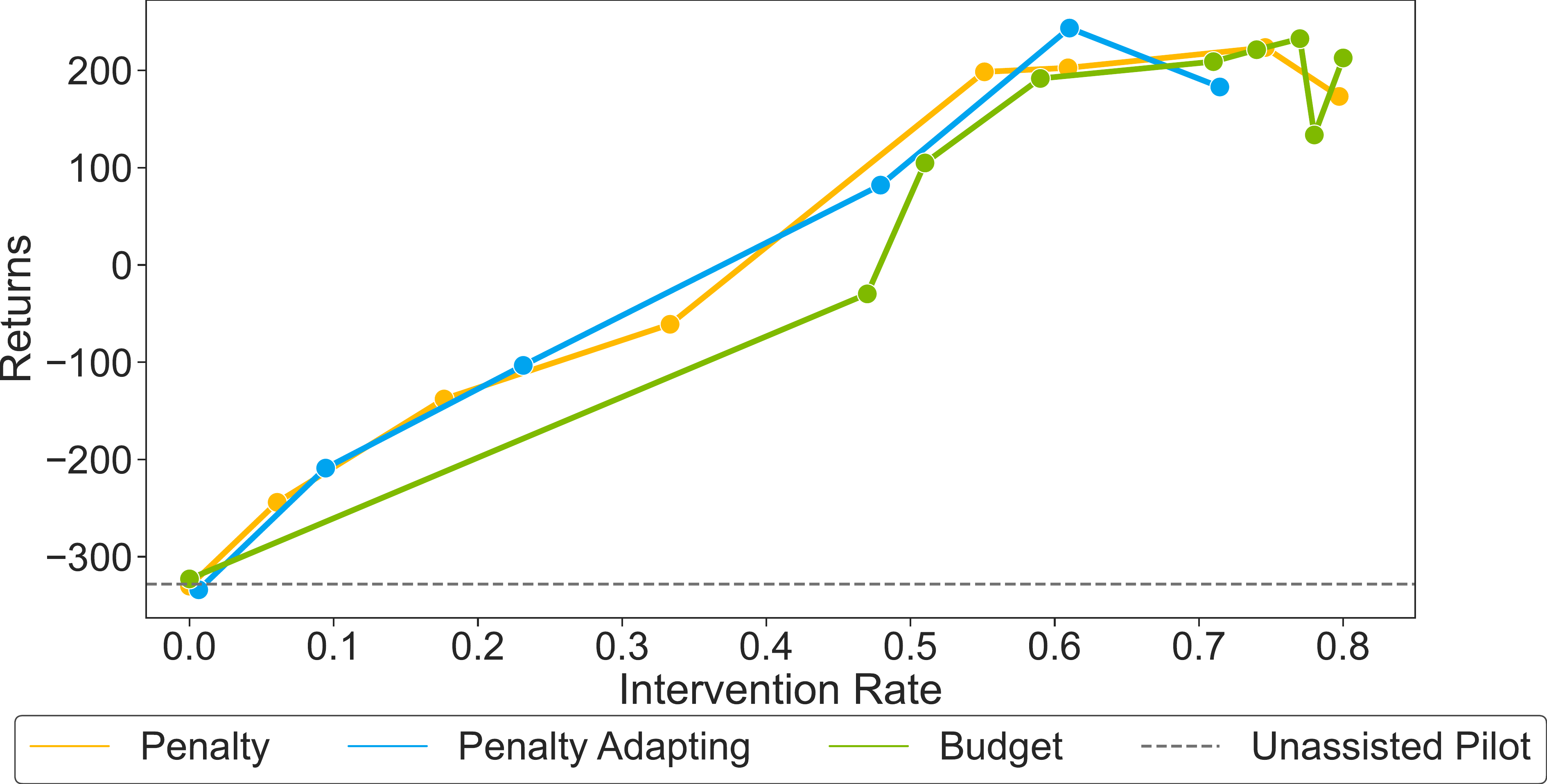}
  \caption{The performance of sensor pilot for various copilots on Lunar Lander with continuous action space.}
  \label{fig:main_sac}
\end{figure}

\begin{figure*}[h!]
     \centering
     \begin{subfigure}[t]{\textwidth}
         \centering
         \begin{subfigure}[t]{0.28\textwidth}
             \centering
             \vspace{0.01in}
             \includegraphics[width=\textwidth]{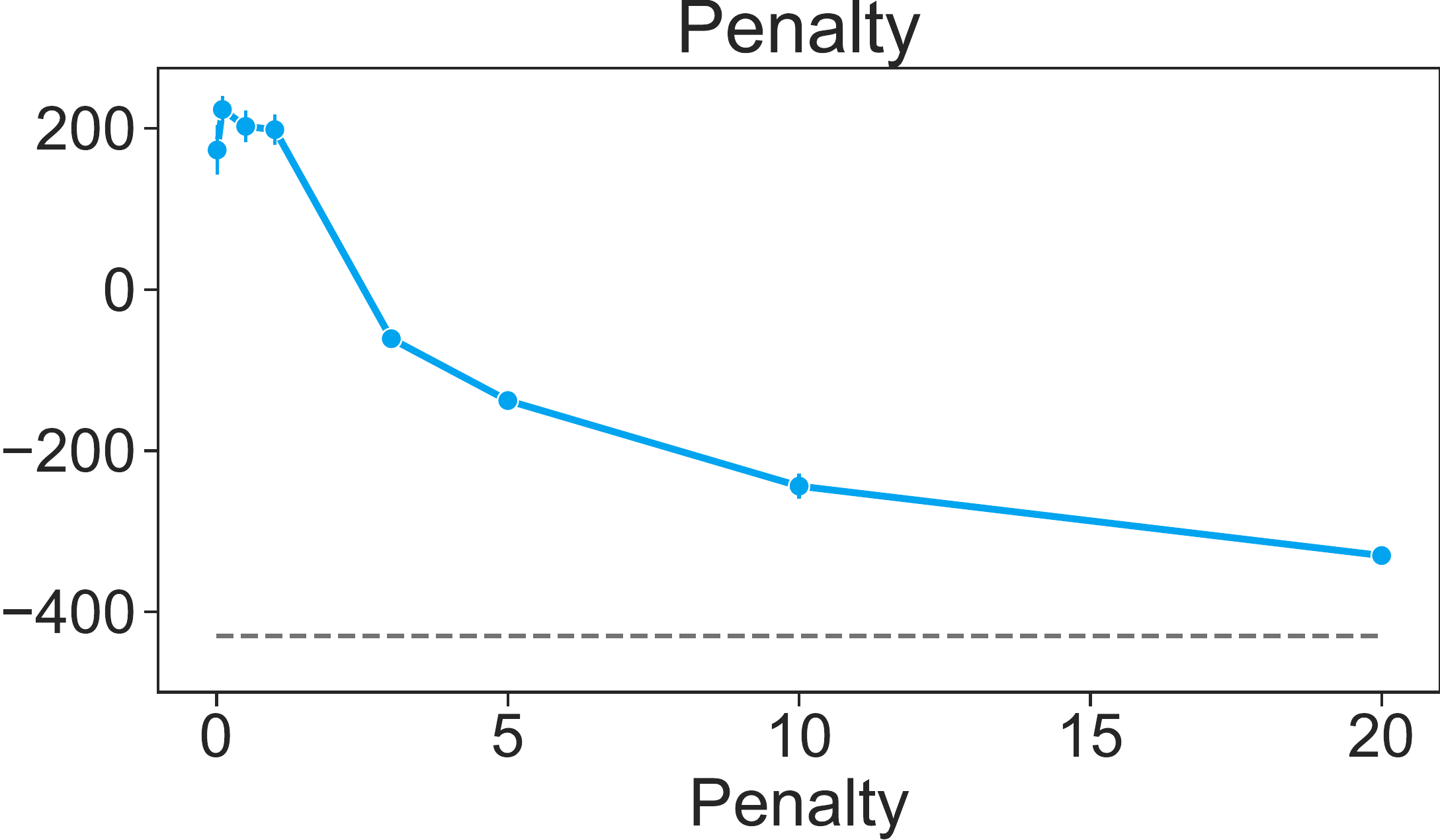}
         \end{subfigure}
         \begin{subfigure}[t]{0.28\textwidth}
             \centering
             \vspace{0.01in}
             \includegraphics[width=\textwidth]{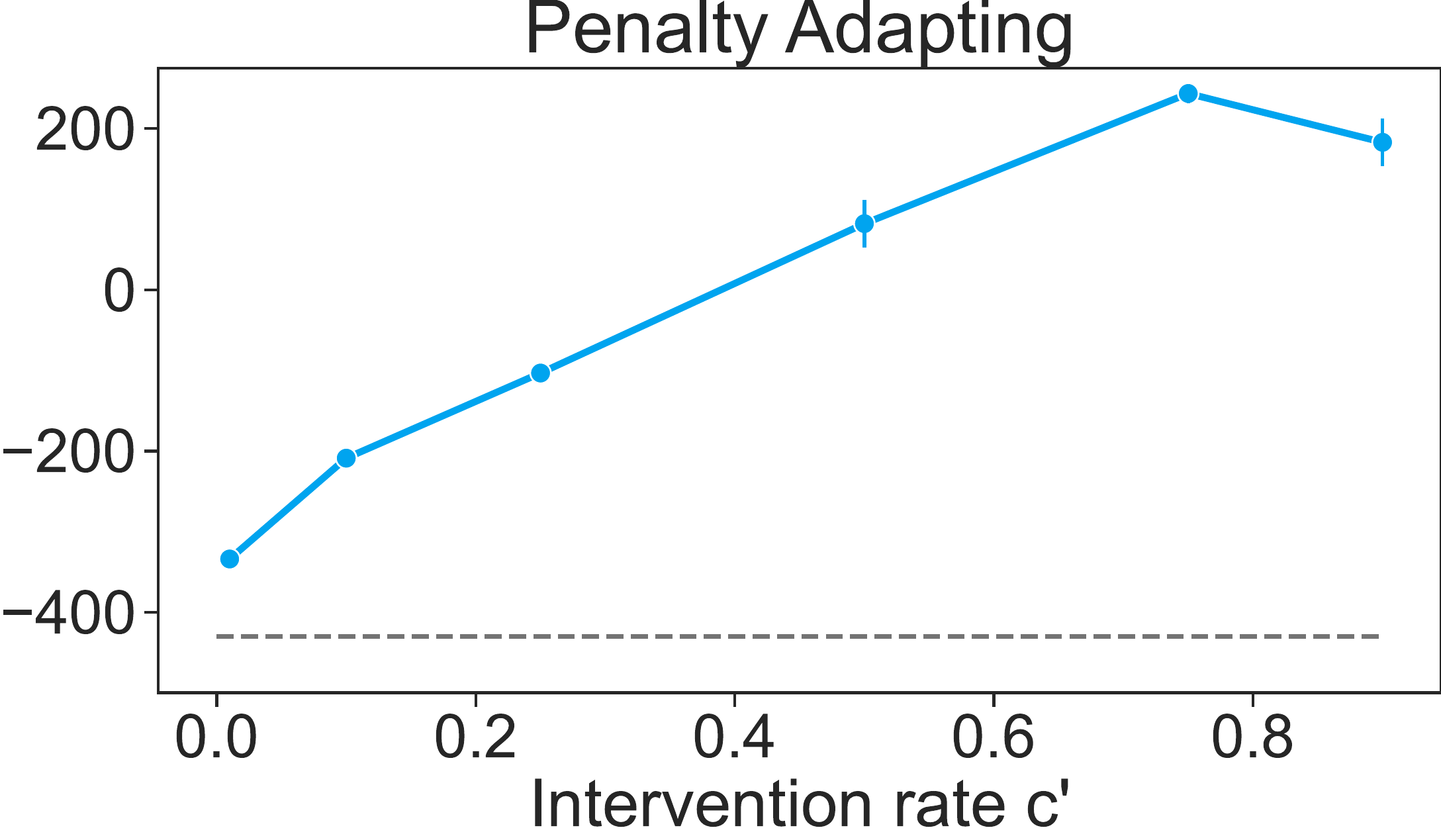}
         \end{subfigure}
         \begin{subfigure}[t]{0.41\textwidth}
             \centering
             \vspace{0.01in}
             \includegraphics[width=\textwidth]{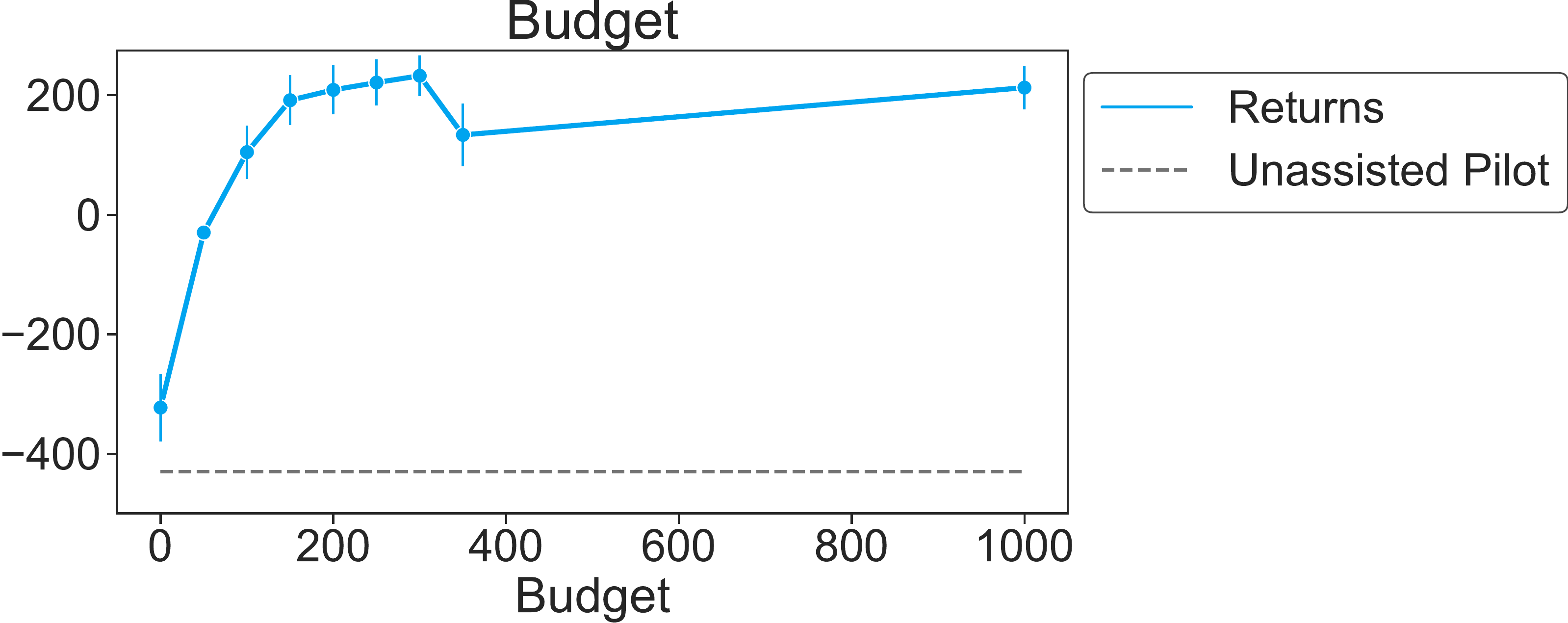}
         \end{subfigure}
        \begin{subfigure}[t]{\textwidth}
         \centering
         \begin{subfigure}[t]{0.28\textwidth}
             \centering
             \vspace{0.01in}
             \includegraphics[width=\textwidth]{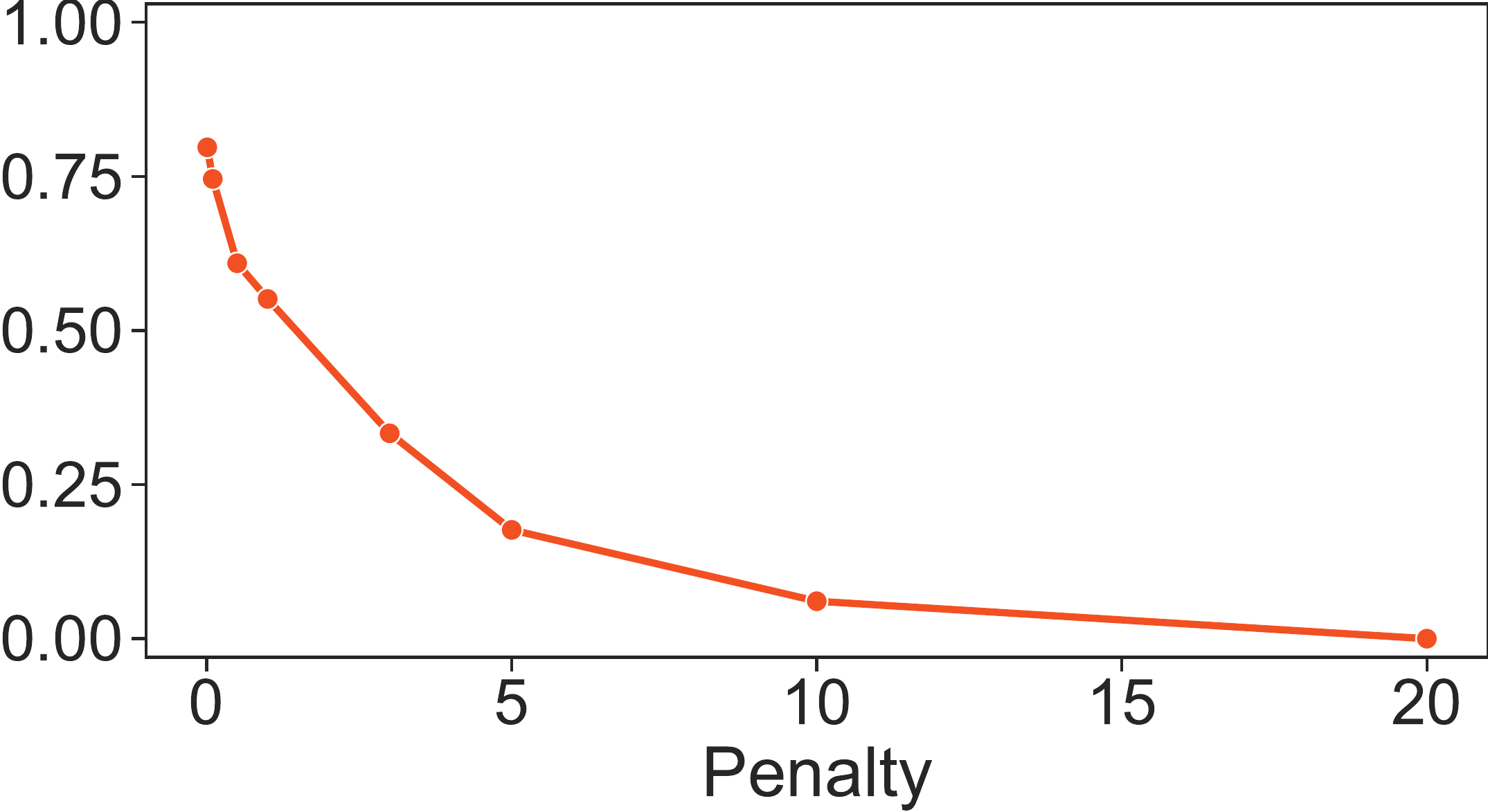}
         \end{subfigure}
         \begin{subfigure}[t]{0.28\textwidth}
             \centering
             \vspace{0.01in}
             \includegraphics[width=\textwidth]{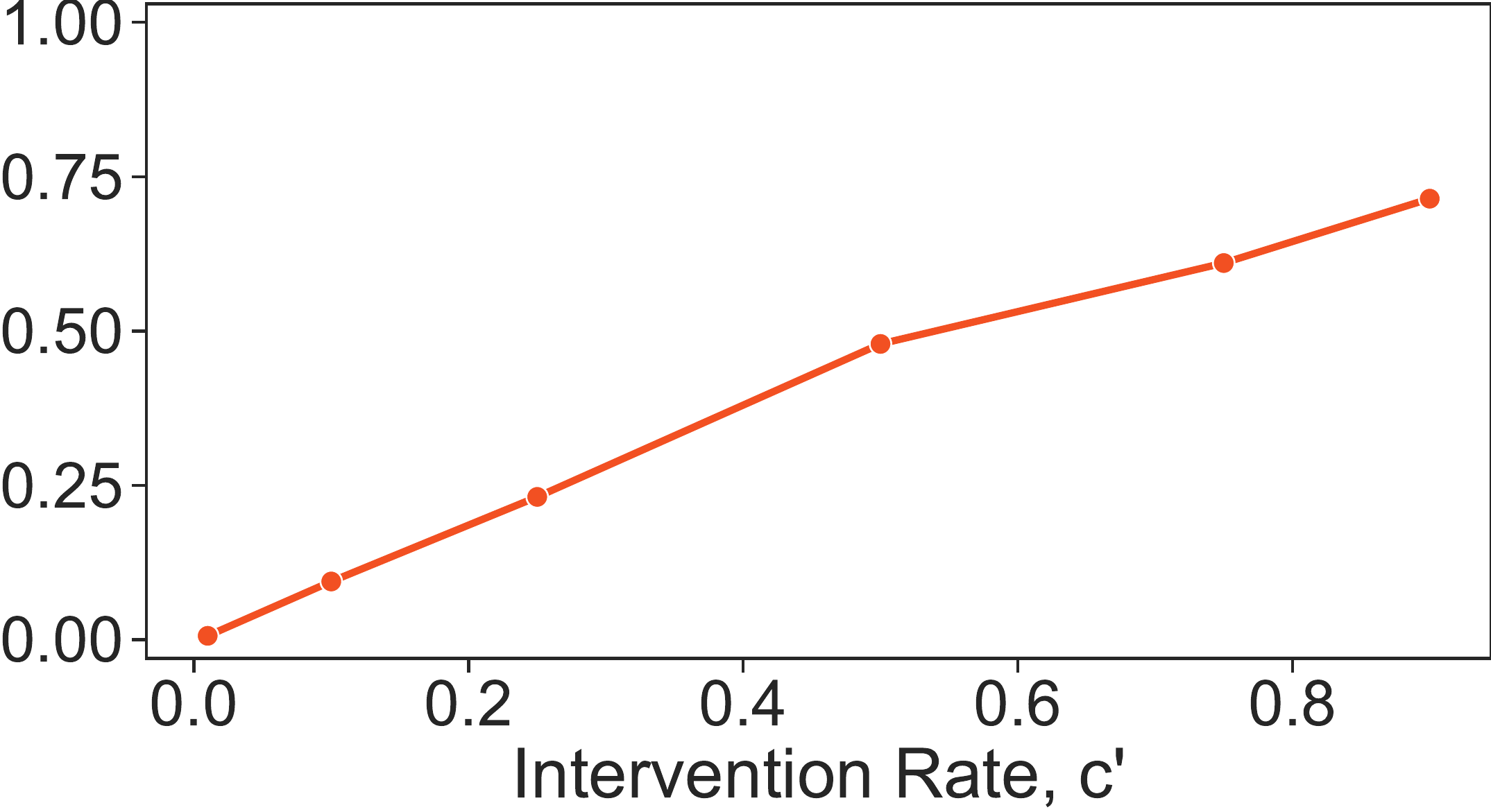}
         \end{subfigure}
         \begin{subfigure}[t]{0.41\textwidth}
             \centering
             \vspace{0.01in}
             \includegraphics[width=\textwidth]{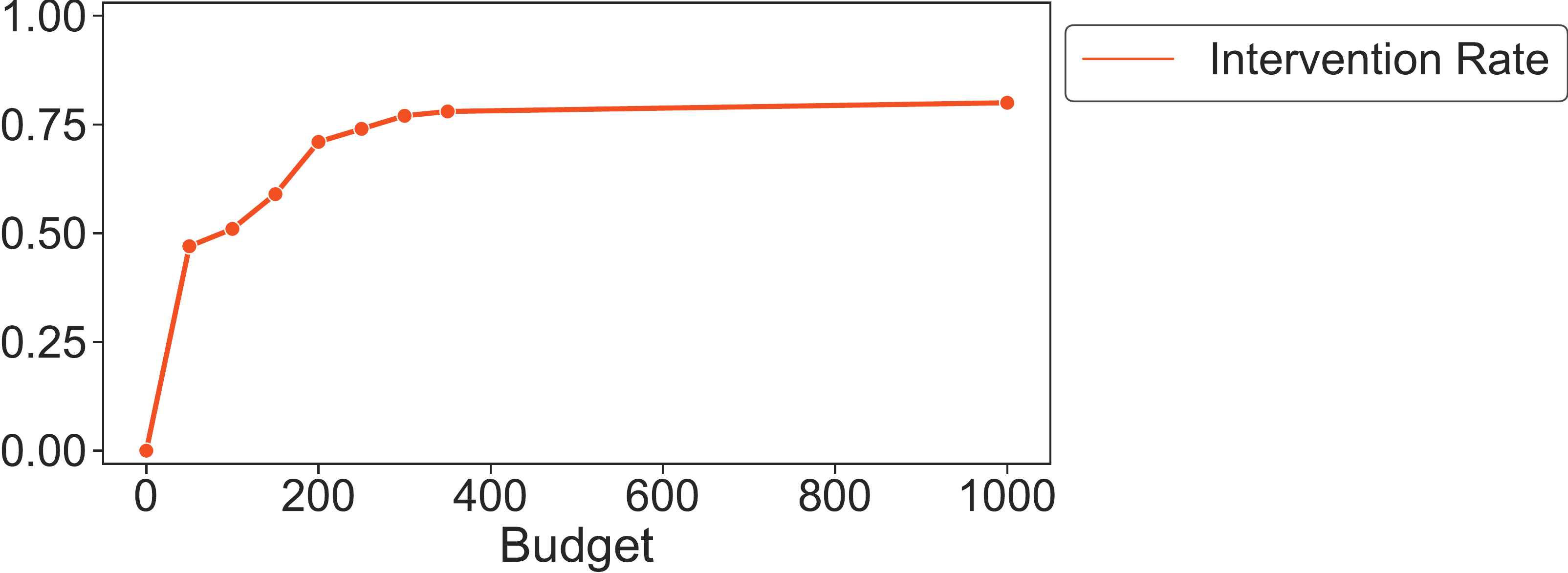}
         \end{subfigure}
     \end{subfigure}
         
     \end{subfigure}
     
     \caption{The returns and intervention rate of sensor pilot assisted by the copilot trained with different parameters for different methods on Lunar Lander with continuous action space.}
     \label{fig:sac_all}
\end{figure*}

We train copilots using each of our proposed methods. The baseline method is not included as it is unclear how to apply it to continuous action space domains. We focus this experimentation on sensor pilot as its policy is not impacted by the change in action space. We use Soft Actor Critic \cite{SACApps} to train each agent under our framework using the default implementation by \citet{stable-baselines3}. Similar to Figures 2 and 3 in the paper, Figure \ref{fig:main_sac} shows the trend of total return at different intervention rates for individual methods. Figure \ref{fig:sac_all} shows the returns and intervention rate of sensor pilot assisted by the copilot trained with different parameters for different methods. The results here are as expected in terms of the trade off between intervention rate and returns achieved. The budget method underperforms the other methods slightly similarly to sensor pilot in the discrete case due to the hard constraint being imposed.

\subsection{How Hyperparameter Influences the Performance}
In the main paper, figures 2 and 3 show the relation between the intervention rate and the return of our methods. Here, we provide more details of how the hyperparameter of different methods influence the performance.

\subsubsection{Grid World}
\begin{figure*}[]
  \begin{floatrow}
    \ffigbox[\FBwidth]{
      \includegraphics[width=0.45\textwidth]{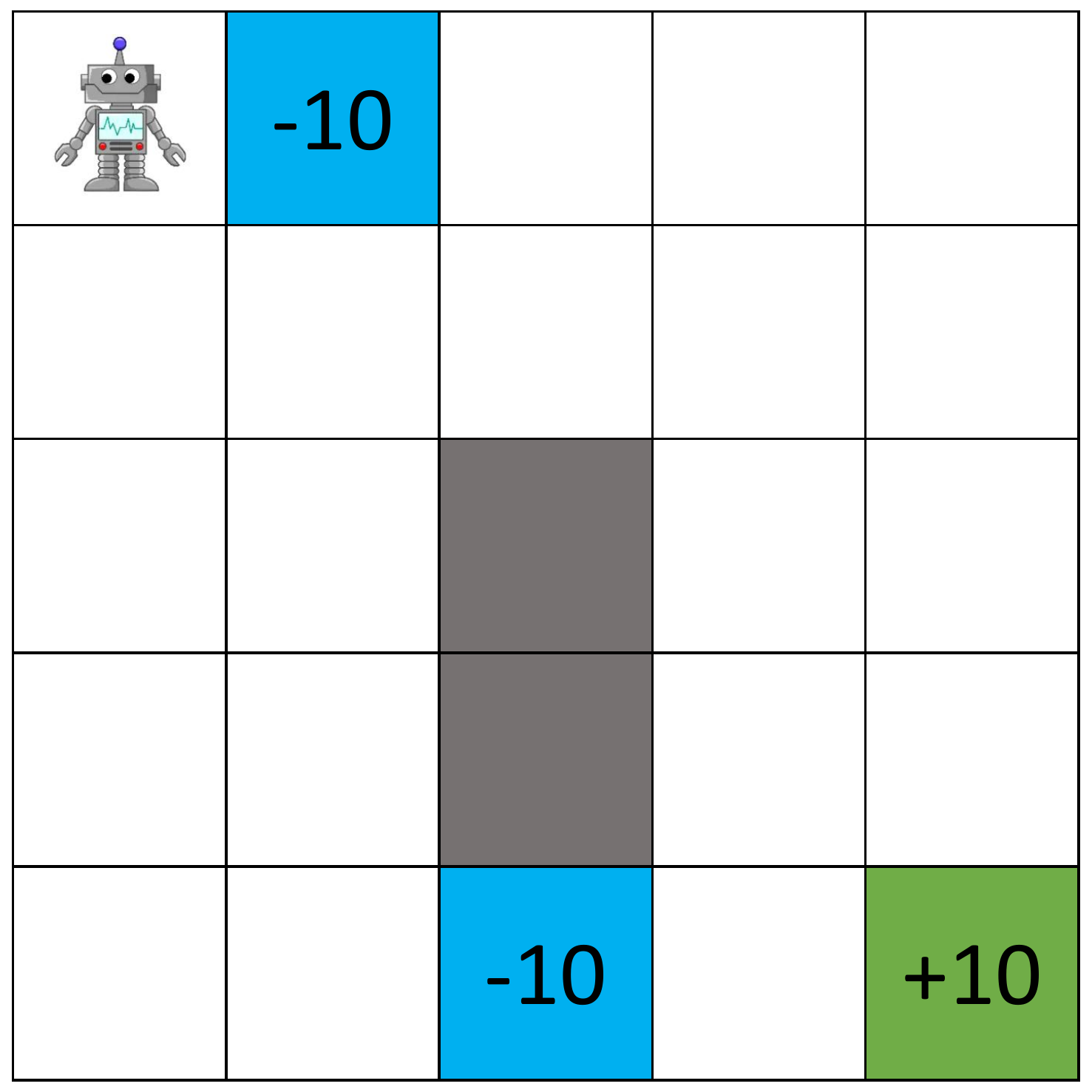}
    }{\caption{The map of the gridworld.}\label{fig:gridworld_demo}}
    \ffigbox[\FBwidth]{
      \includegraphics[width=0.45\textwidth]{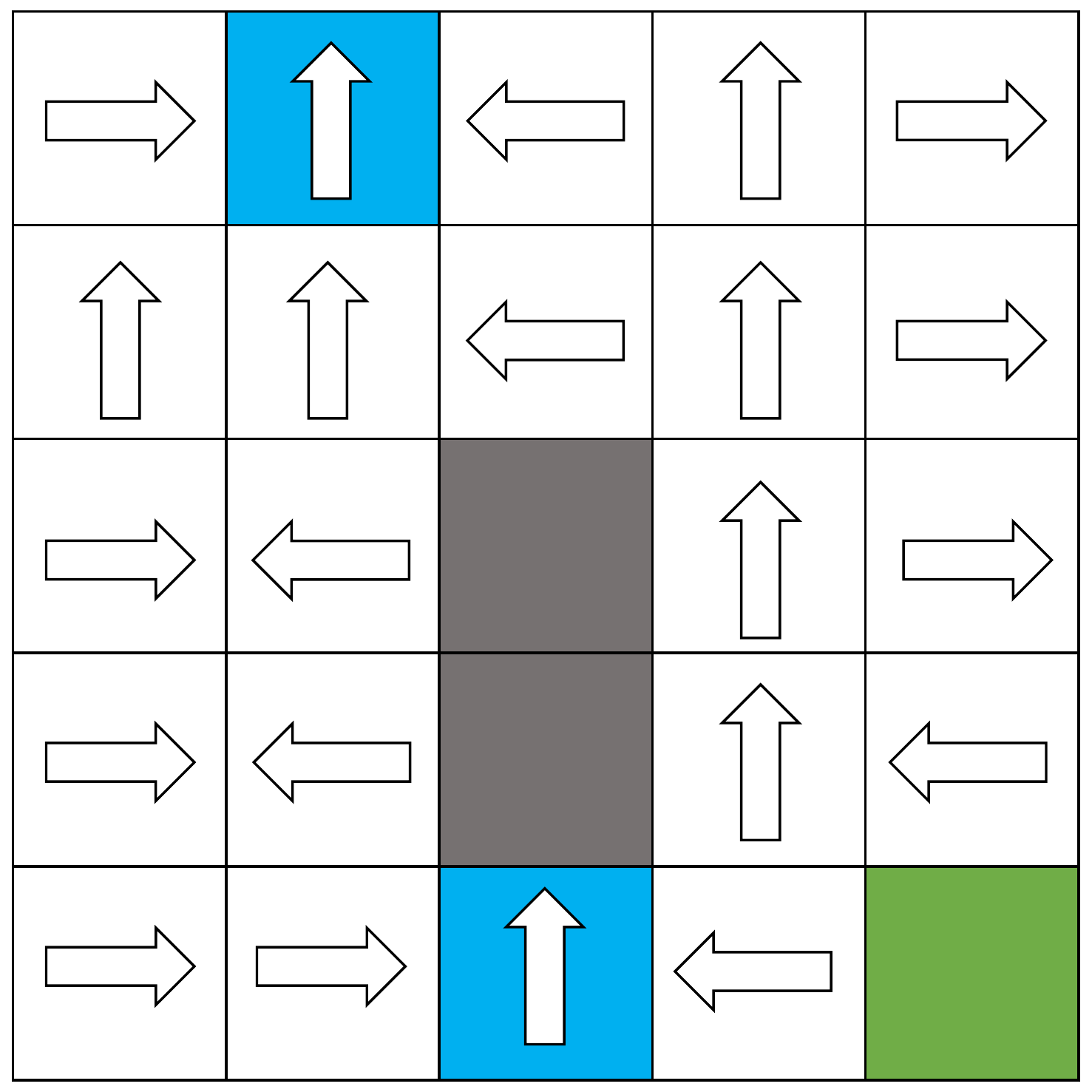}
    }{\caption{Human player's policy. The human player will take the action corresponding to the arrow in the cell.}\label{fig:gridworld_human_policy}}
  \end{floatrow}
\end{figure*}

\begin{figure*}[]
  \ffigbox[\textwidth]%
  {%
    \begin{subfloatrow}[3]
      \ffigbox[\FBwidth]{
        \includegraphics[width=0.28\textwidth]{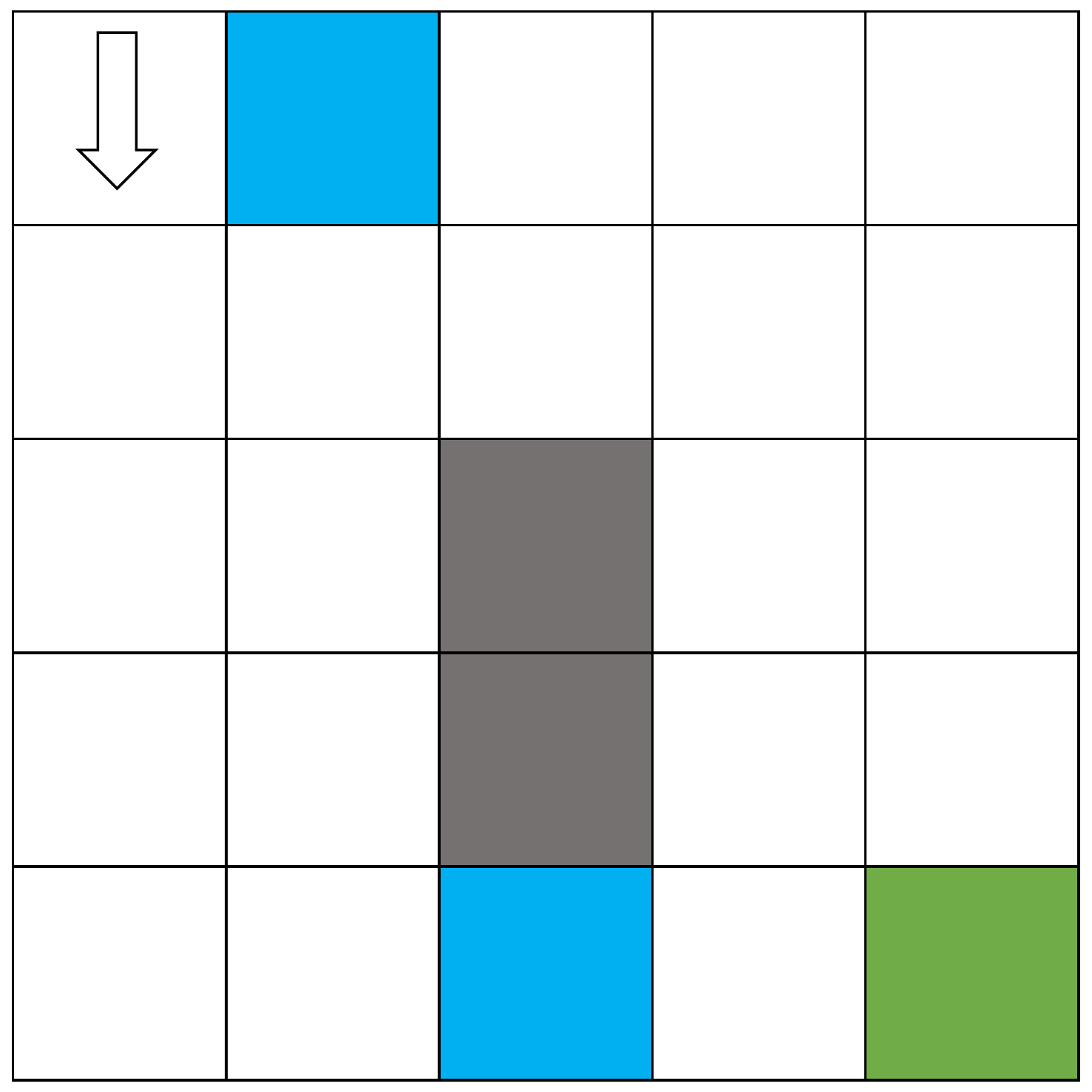}
      }{\caption{penalty = 3}\label{fig:penalty3}}
      \ffigbox[\FBwidth]{
        \includegraphics[width=0.28\textwidth]{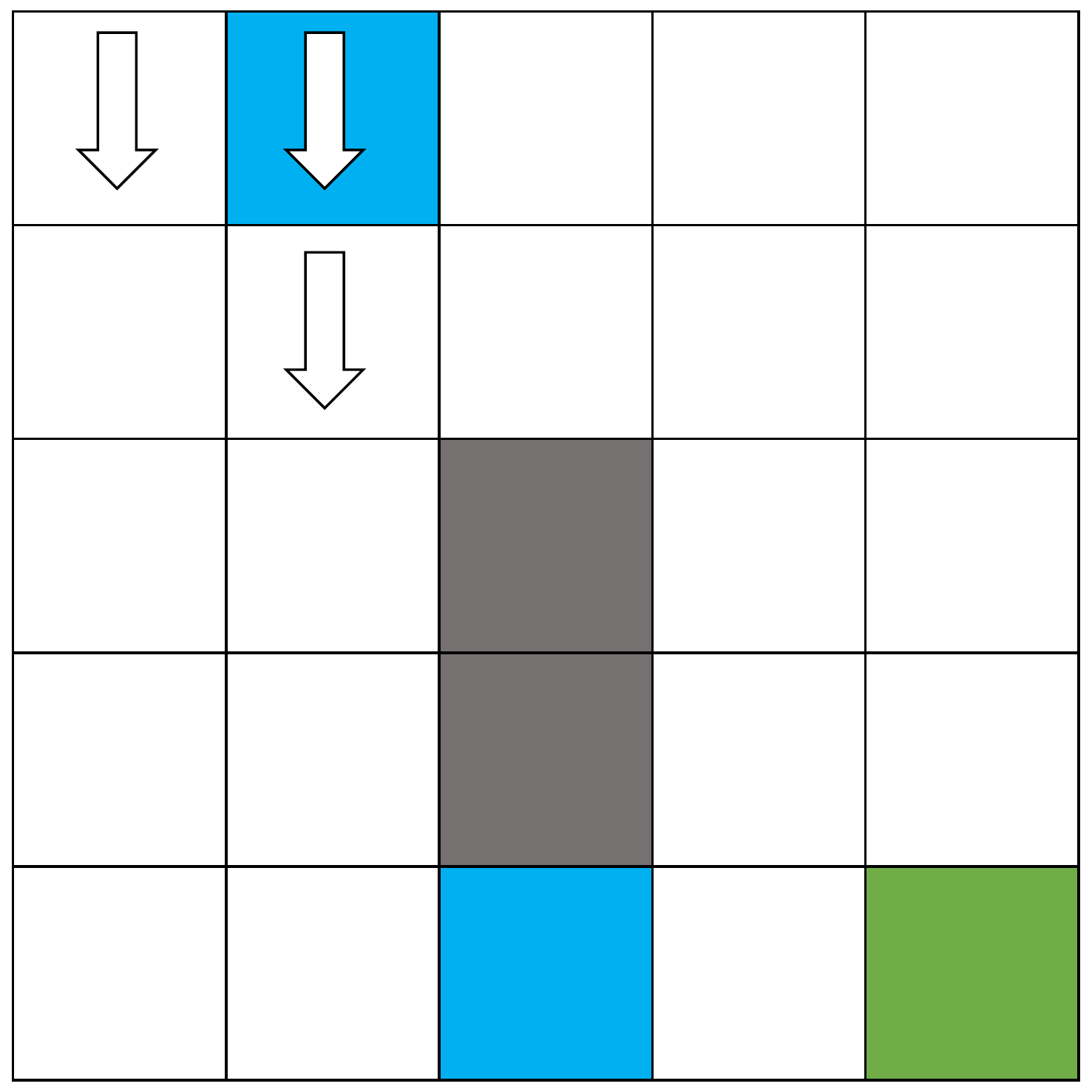}
      }{\caption{penalty = 2}\label{fig:penalty2}}
      \ffigbox[\FBwidth]{
        \includegraphics[width=0.28\textwidth]{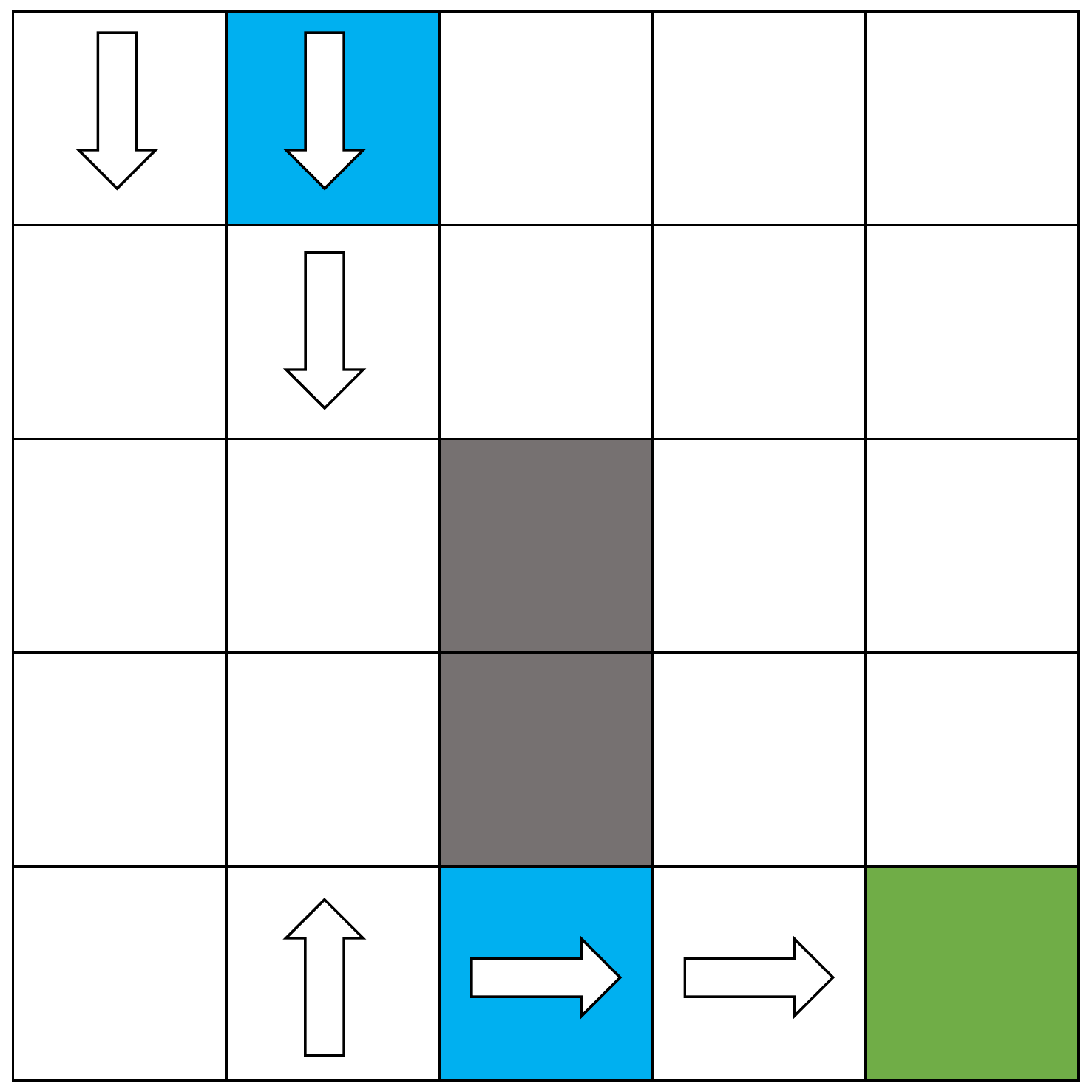}
      }{\caption{penalty = 1}\label{fig:penalty1}}
    \end{subfloatrow}    
    \begin{subfloatrow}[3]
      \ffigbox[\FBwidth]{
        \includegraphics[width=0.28\textwidth]{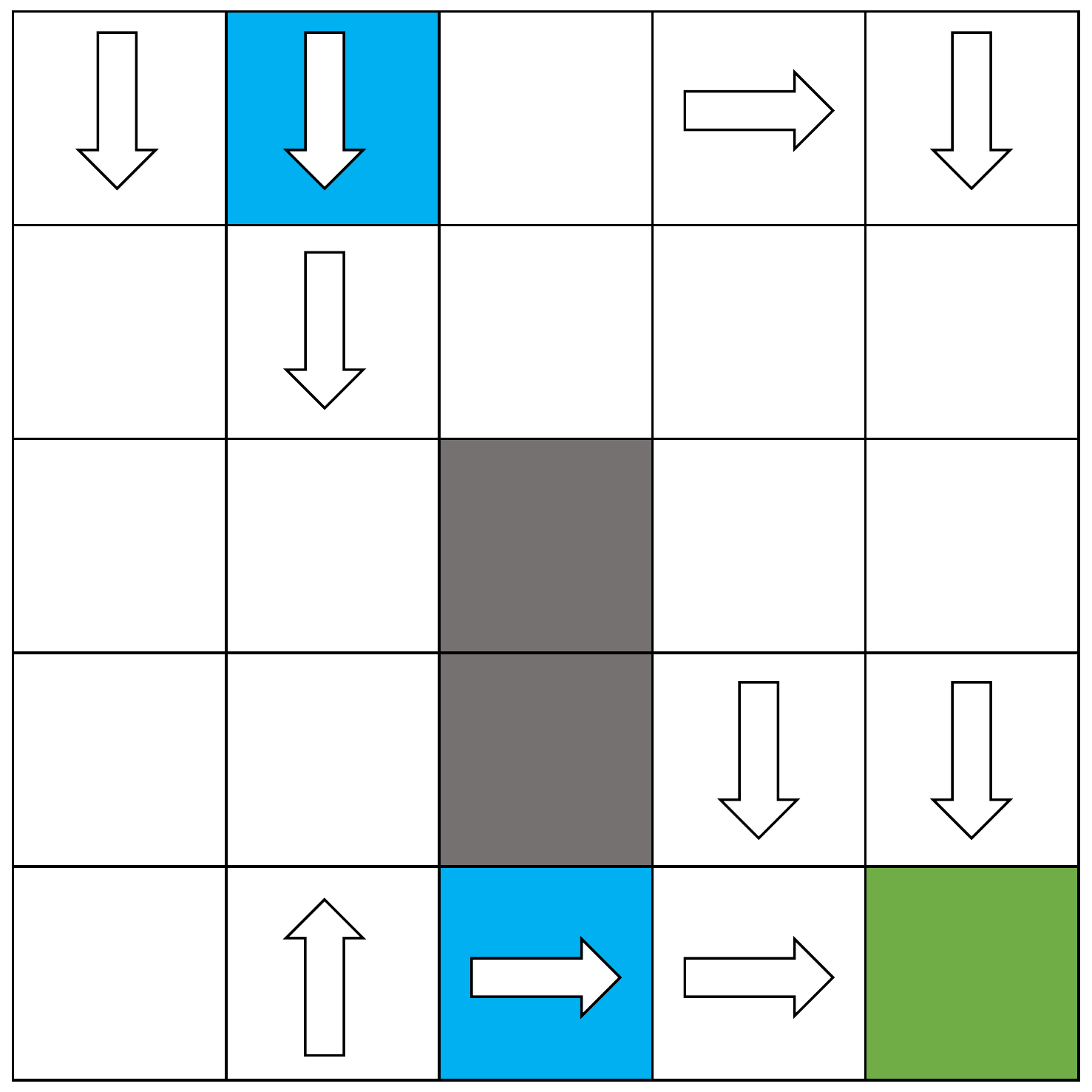}
      }{\caption{penalty = 0.5}\label{fig:penalty0.5}}
      \ffigbox[\FBwidth]{
        \includegraphics[width=0.28\textwidth]{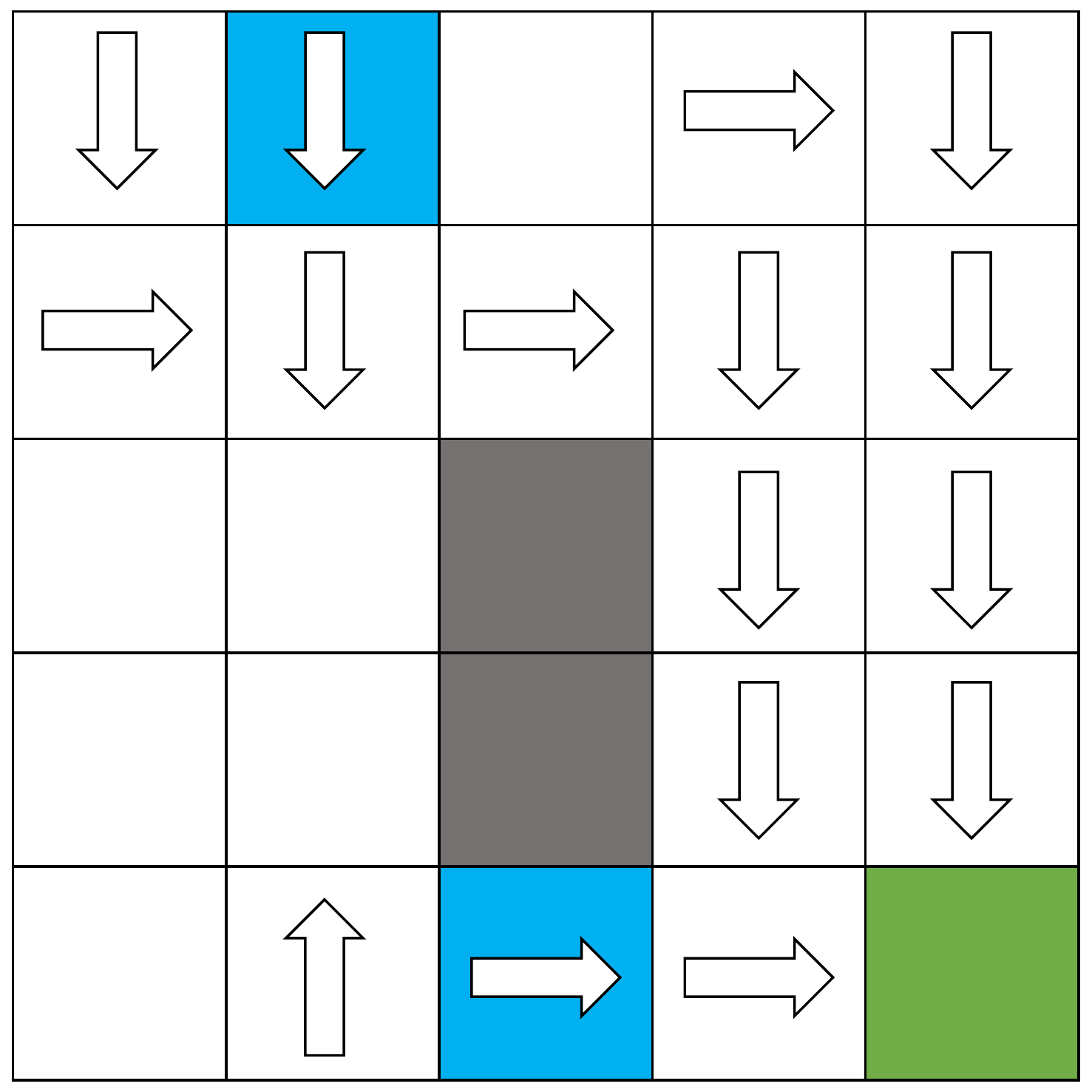}
      }{\caption{penalty = 0.1}\label{fig:penalty0.1}}
      \ffigbox[\FBwidth]{
        \includegraphics[width=0.28\textwidth]{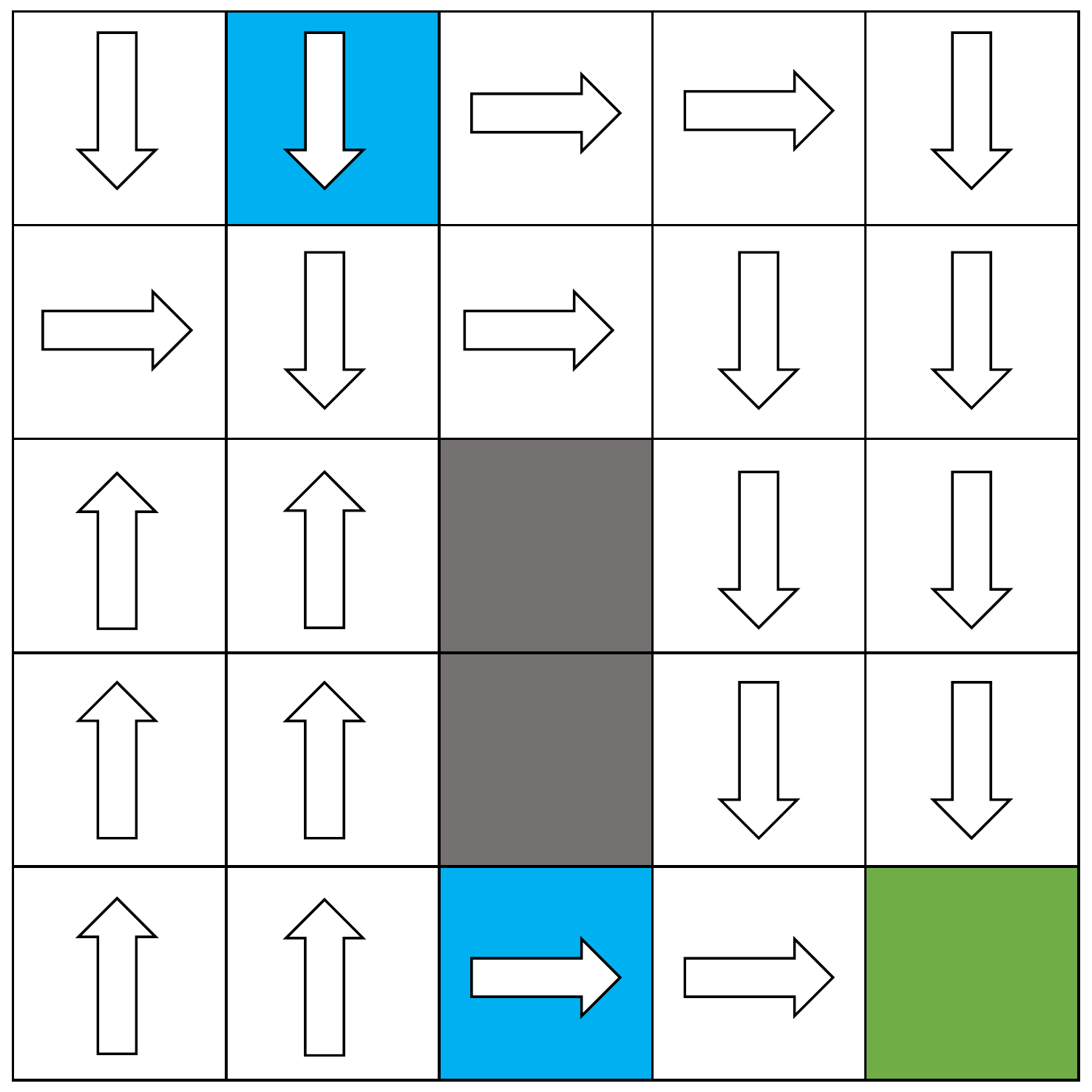}
      }{\caption{penalty = 0}\label{fig:penalty0}}
    \end{subfloatrow}
  }{\caption{Visualization of the states that the assistive agent trained with different penalty will intervene. The direction of the arrow in the cell means that the assistive agent intervenes in this cell and chooses to take the corresponding action. The other empty cells mean that the agent does not intervene and still use the human policy in these cells.}\label{fig:gridworld_penalty}}
\end{figure*}

In this section, we visualize a gridworld environment to show how the penalty term in our method influences the behaviour of the assistive agent. Figure \ref{fig:gridworld_demo} shows the layout of the gridworld. An agent is placed in the top right corner. The green cell is the terminal state, where the agent will receive +10 reward when entering this cell. The blue cells contain water where the agent will receive a -10 reward when entering this cell. The grey 
cells are stones that the agent cannot enter. The other cells are the states where the agent can enter. The agent has 5\% possibility to veer to the right -- it moves +90 degrees from where it attempted to move(E.g. From moving up to moving right). And the agent has 5\% possibility to veer to the left -- it moves -90 degrees from where it attempted to move (E.g. From moving up to moving left). And it also has a 10\% possibility of doing nothing and staying. \\

A sub-optimal human policy in figure is shown in \ref{fig:gridworld_human_policy}. We use this policy to simulate real human's action. In Figure \ref{fig:gridworld_penalty}, the arrows indicate when the human player will be intervened by 6 different assistive agents. These agents are trained with tabular Q-learning method with $penalty = [3, 2, 1, 0.5, 0.1, 0]$ respectively.  \\

As the penalty decreases, the agent will intervene in more states. When the penalty is high, the agent only intervenes when the human is going to enter the water cell resulting in receiving a large negative reward. As penalty decreases, more interventions are used to guide the human to the target cell.


\subsection{Lunar Lander with Discrete Action Space}
\begin{figure*}[!htb]
     \centering
     \begin{subfigure}[t]{\textwidth}
         \centering
         \begin{subfigure}[t]{0.24\textwidth}
             \centering
             \vspace{0.01in}
             \includegraphics[width=\textwidth]{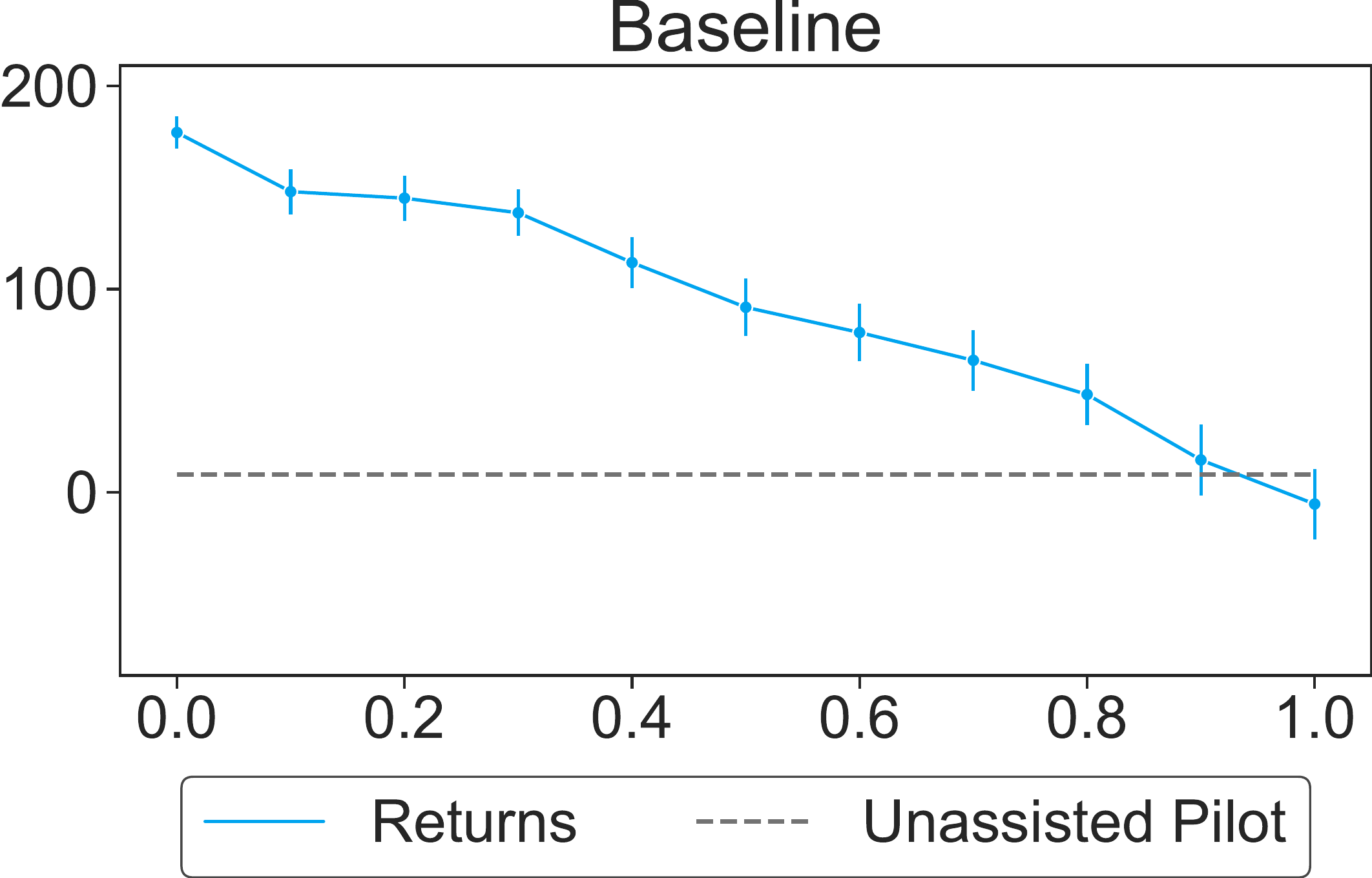}
         \end{subfigure}
         \begin{subfigure}[t]{0.24\textwidth}
             \centering
             \vspace{0.01in}
             \includegraphics[width=\textwidth]{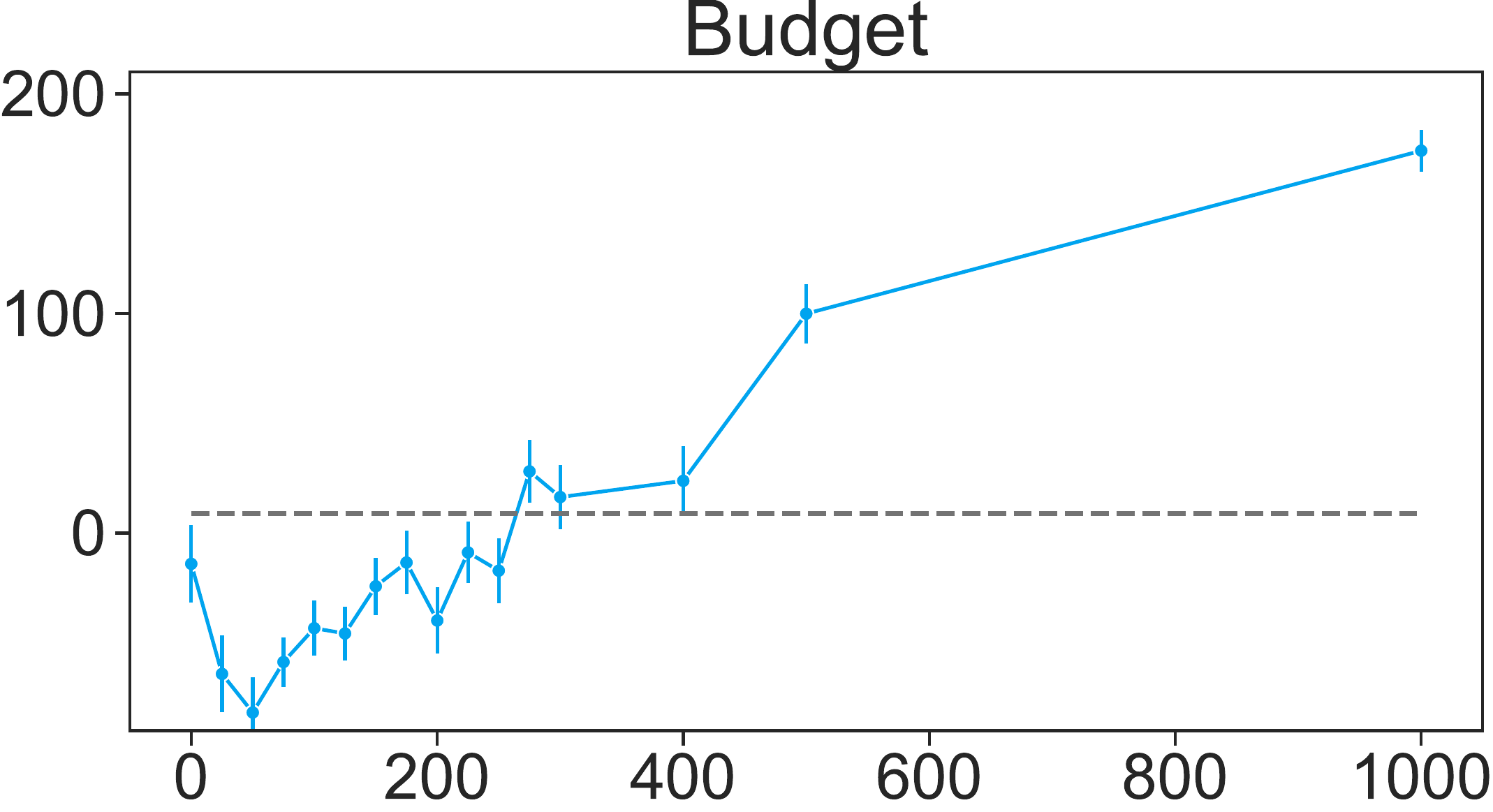}
         \end{subfigure}
         \begin{subfigure}[t]{0.24\textwidth}
             \centering
             \vspace{0.01in}
             \includegraphics[width=\textwidth]{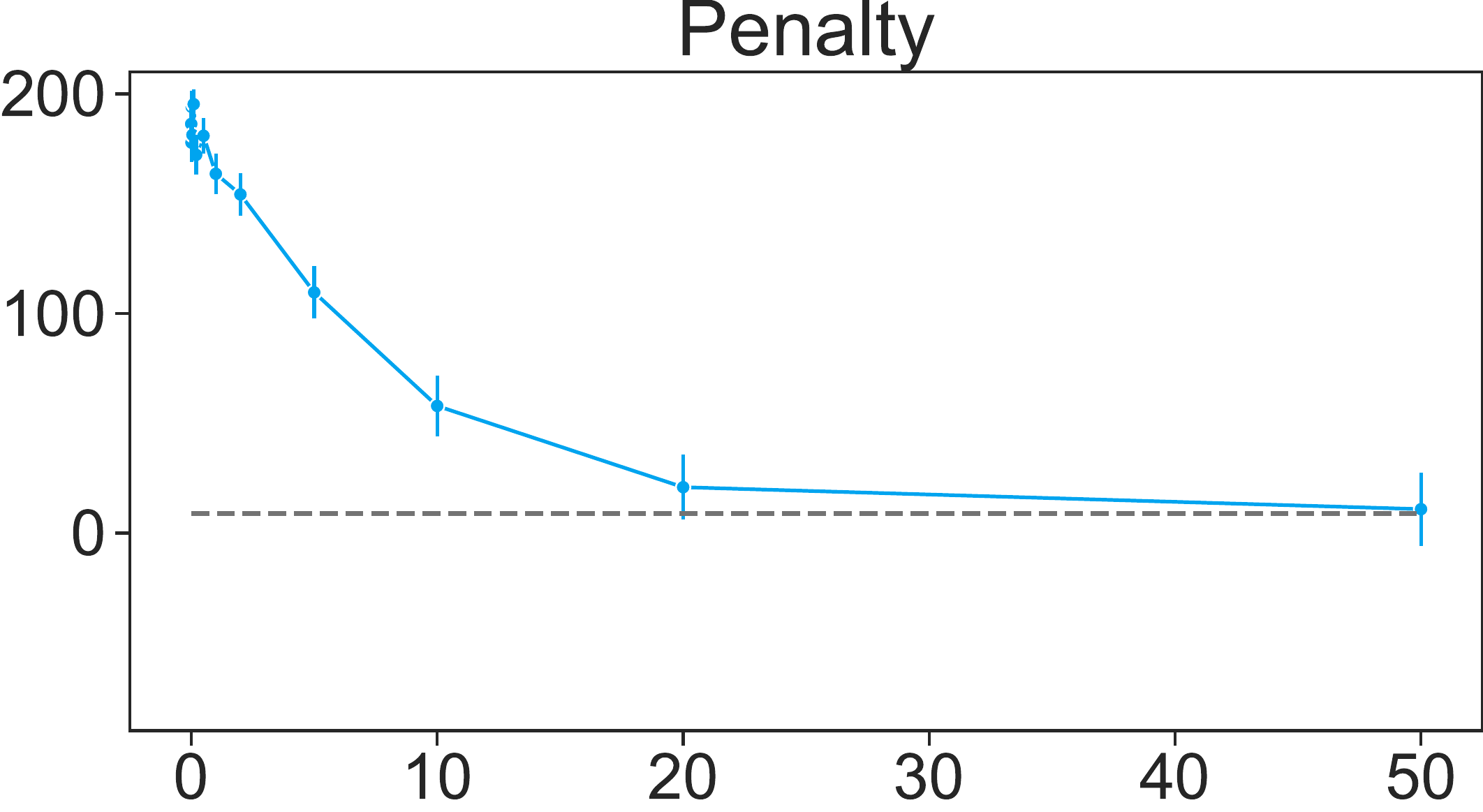}
         \end{subfigure}
         \begin{subfigure}[t]{0.24\textwidth}
             \centering
             \vspace{0.01in}
             \includegraphics[width=\textwidth]{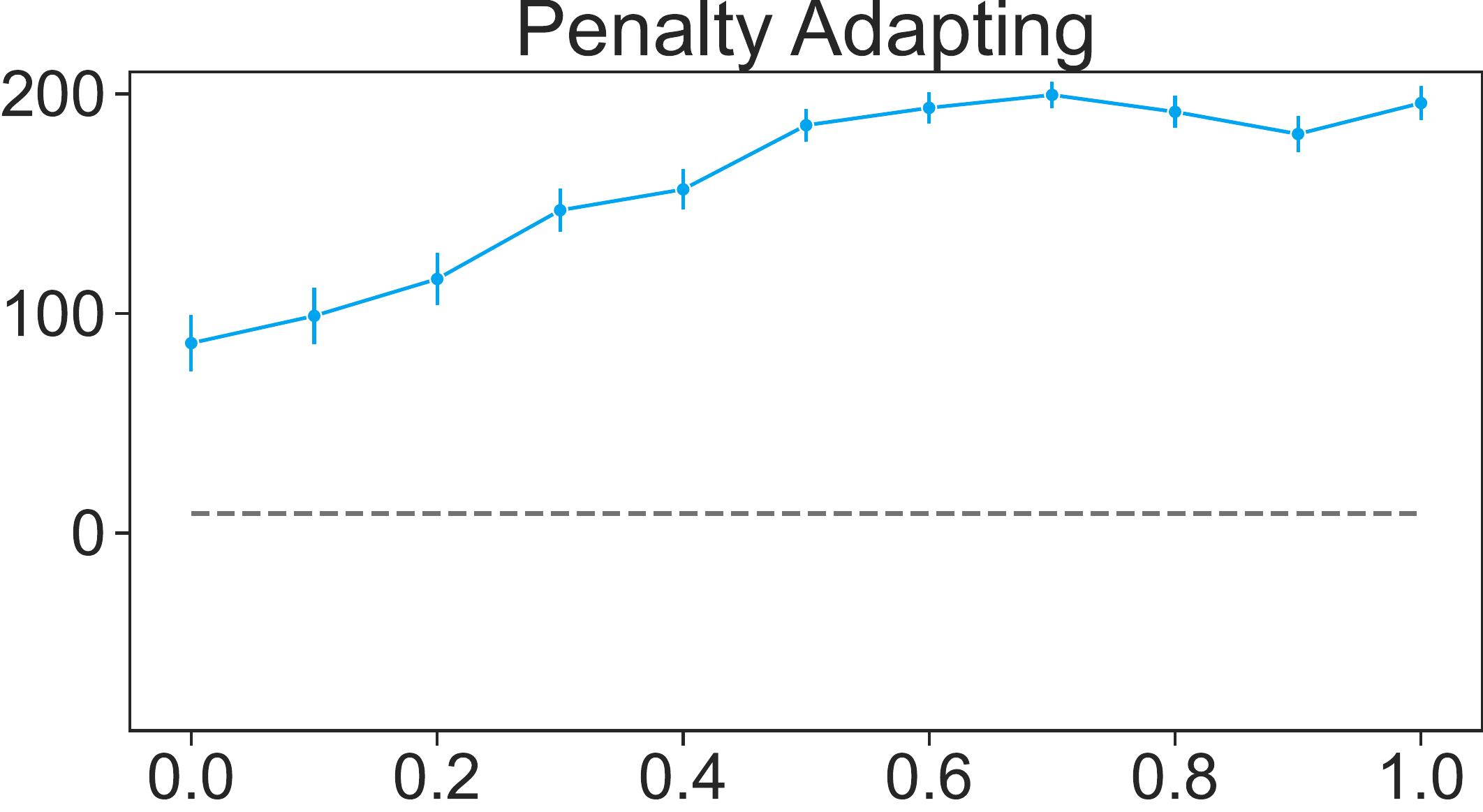}
         \end{subfigure}
     \end{subfigure}
     \begin{subfigure}[t]{\textwidth}
         \centering
         \begin{subfigure}[t]{0.24\textwidth}
             \centering
             \vspace{0.01in}
             \includegraphics[width=\textwidth]{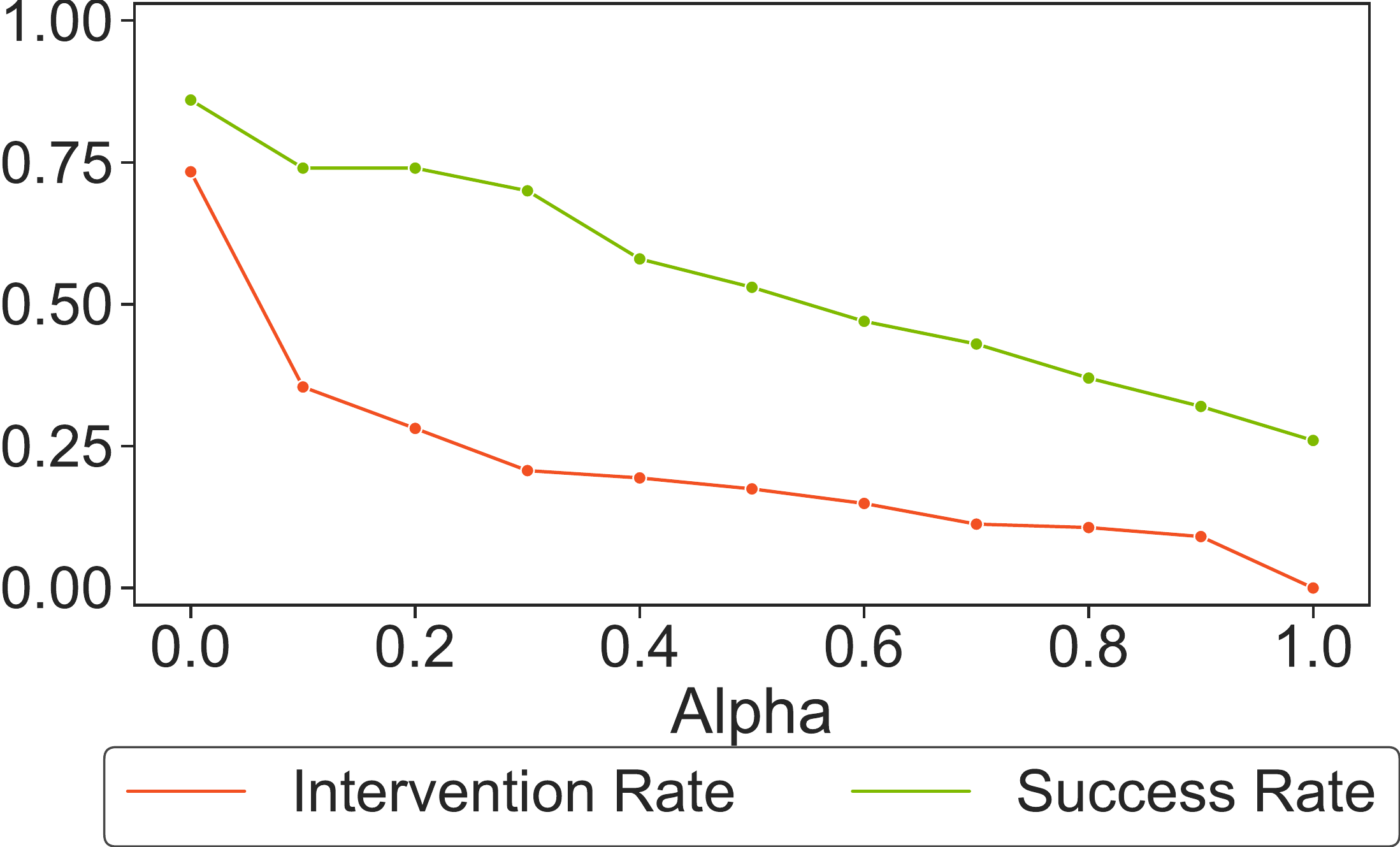}
         \end{subfigure}
         \begin{subfigure}[t]{0.24\textwidth}
             \centering
             \vspace{0.01in}
             \includegraphics[width=\textwidth]{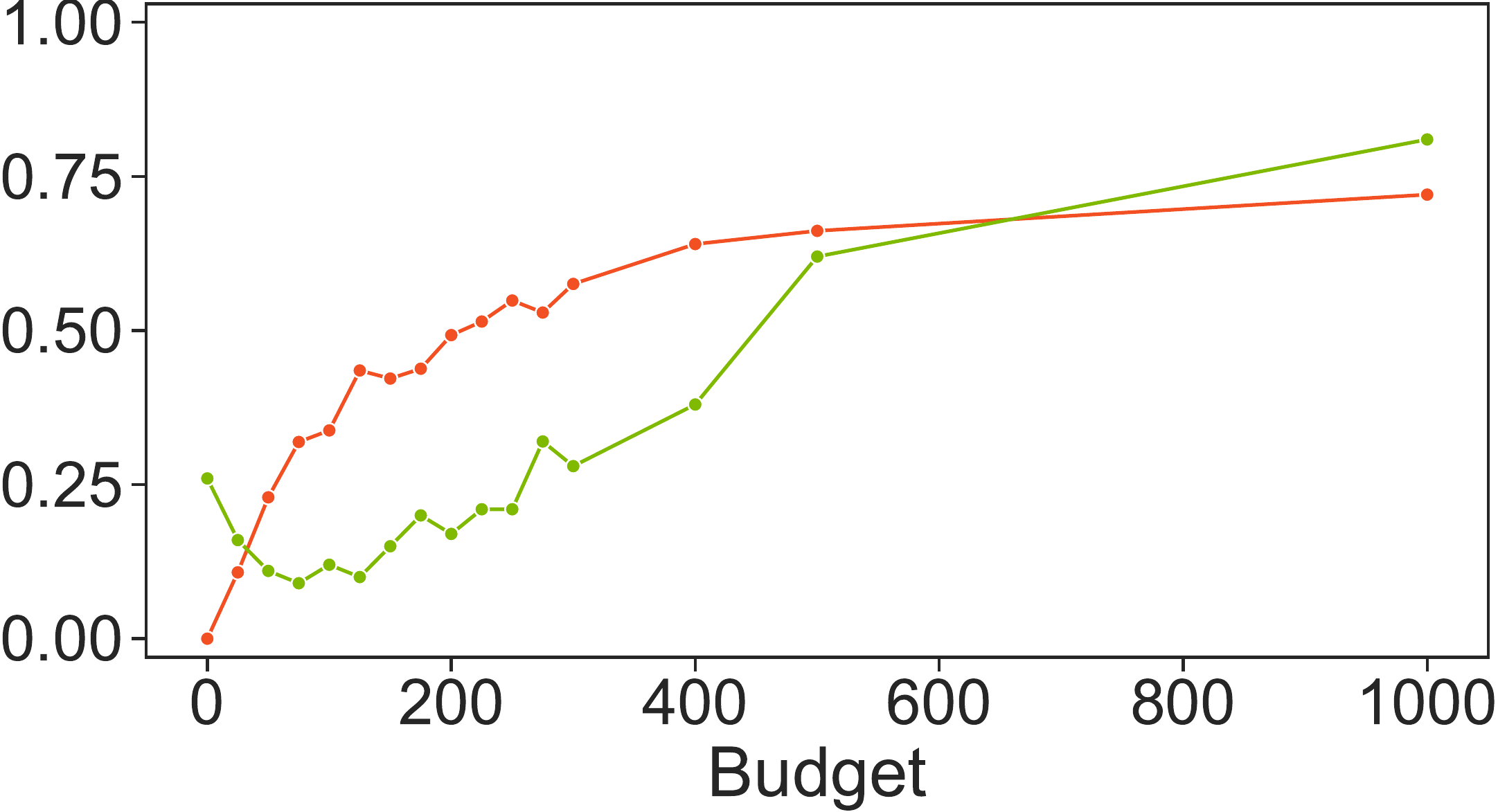}
         \end{subfigure}
         \begin{subfigure}[t]{0.24\textwidth}
             \centering
             \vspace{0.01in}
             \includegraphics[width=\textwidth]{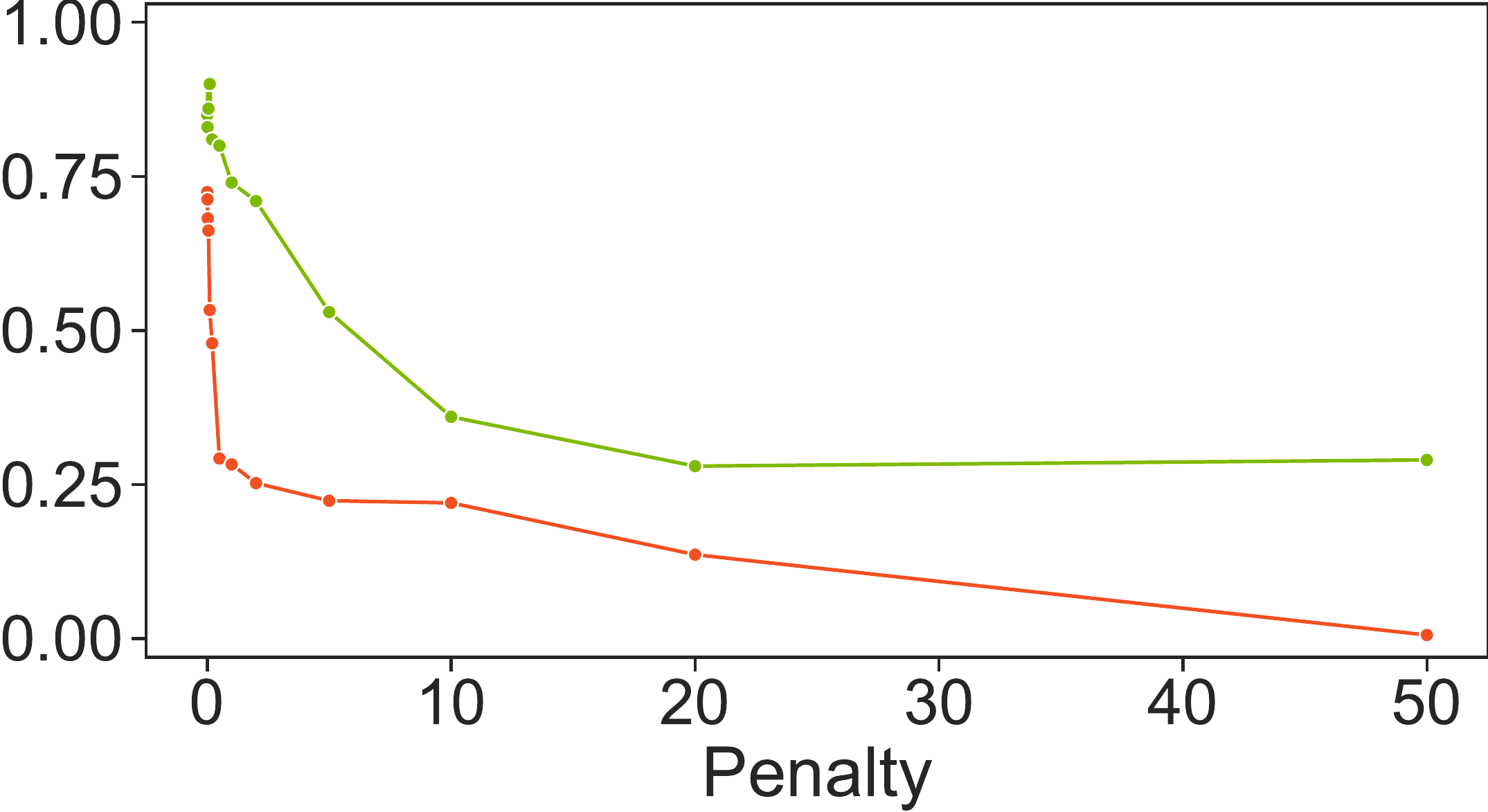}
         \end{subfigure}
         \begin{subfigure}[t]{0.24\textwidth}
             \centering
             \vspace{0.01in}
             \includegraphics[width=\textwidth]{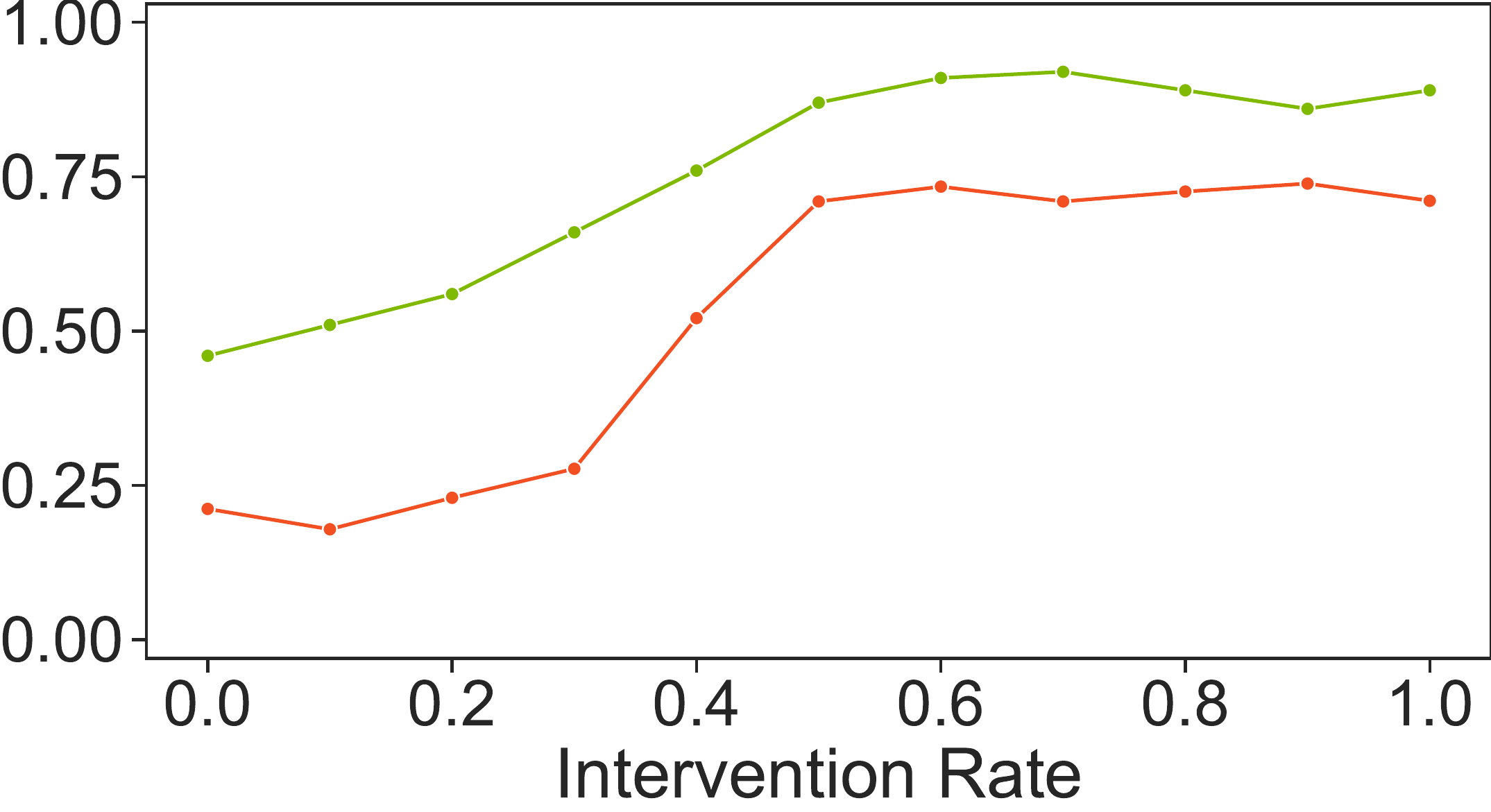}
         \end{subfigure}
     \end{subfigure}
     \caption{The return, intervention rate and success rate of the \textbf{laggy} pilot assisted by the copilot trained with different parameters for different methods on Lunar Lander with discrete action space.}
     \label{fig:lander laggy}
\end{figure*}

\begin{figure*}[!htb]
     \centering
     \begin{subfigure}[t]{\textwidth}
         \centering
         \begin{subfigure}[t]{0.24\textwidth}
             \centering
             \vspace{0.05in}
             \includegraphics[width=\textwidth]{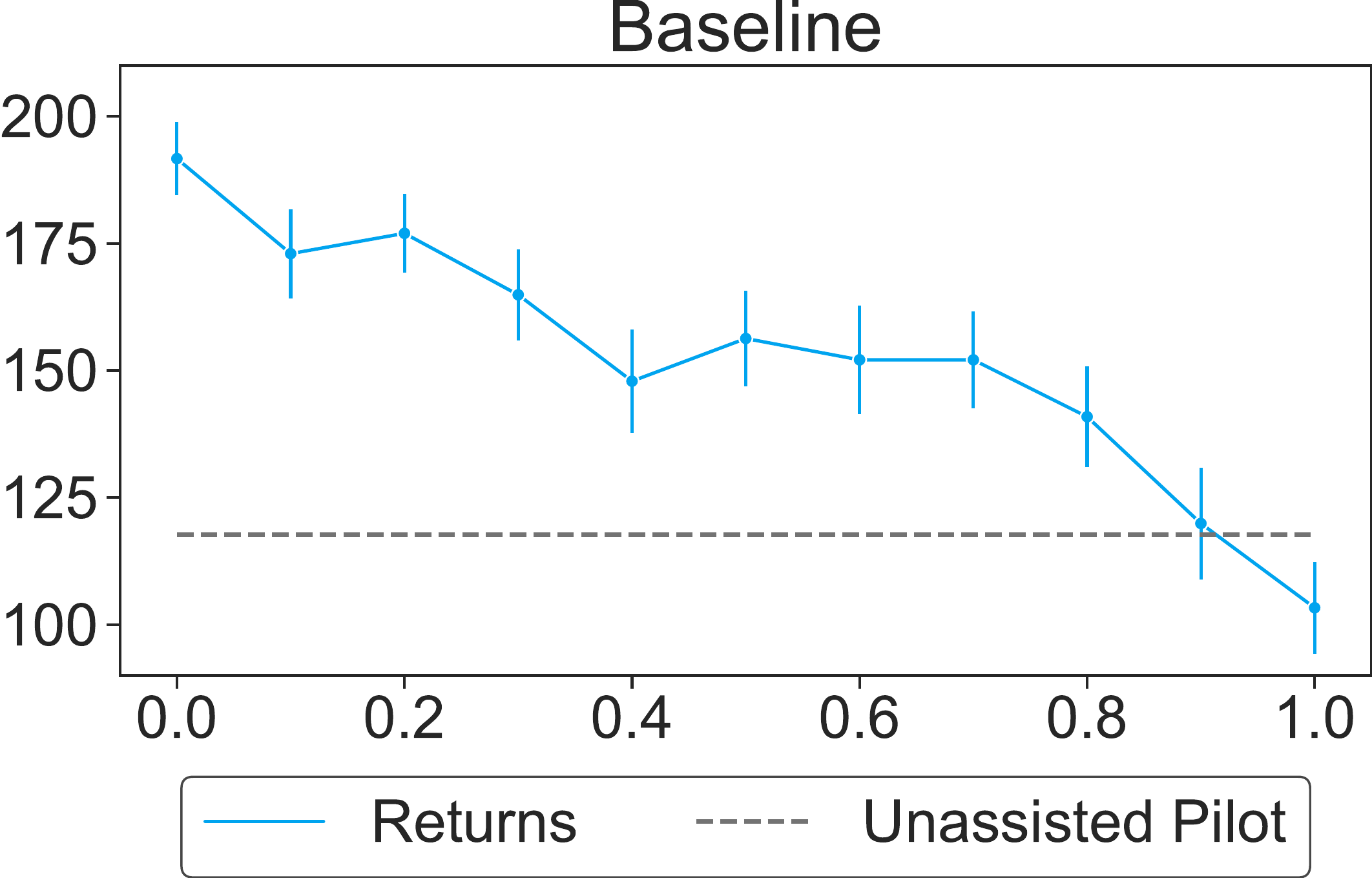}
         \end{subfigure}
         \begin{subfigure}[t]{0.24\textwidth}
             \centering
             \vspace{0.05in}
             \includegraphics[width=\textwidth]{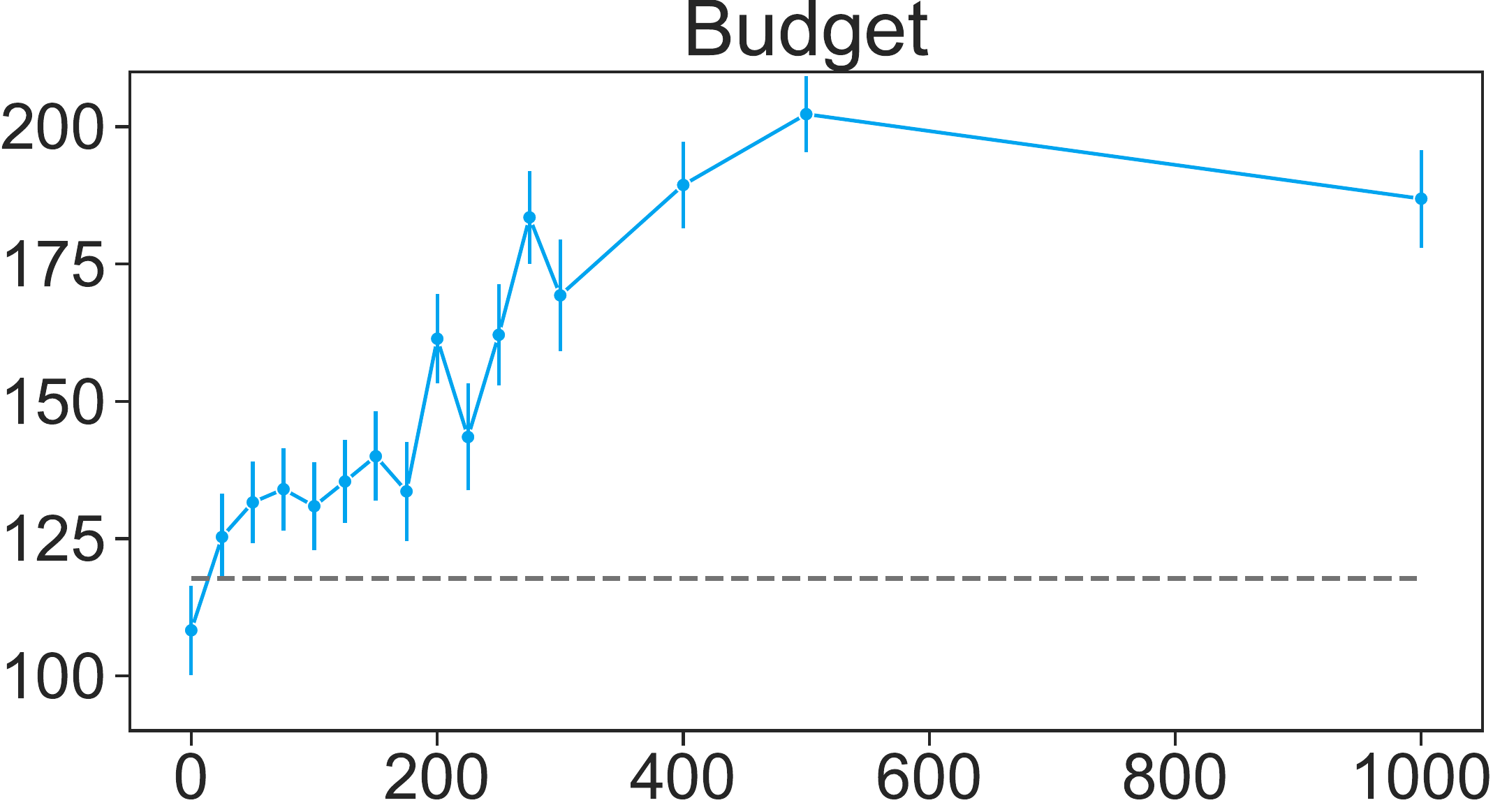}
         \end{subfigure}
         \begin{subfigure}[t]{0.24\textwidth}
             \centering
             \vspace{0.05in}
             \includegraphics[width=\textwidth]{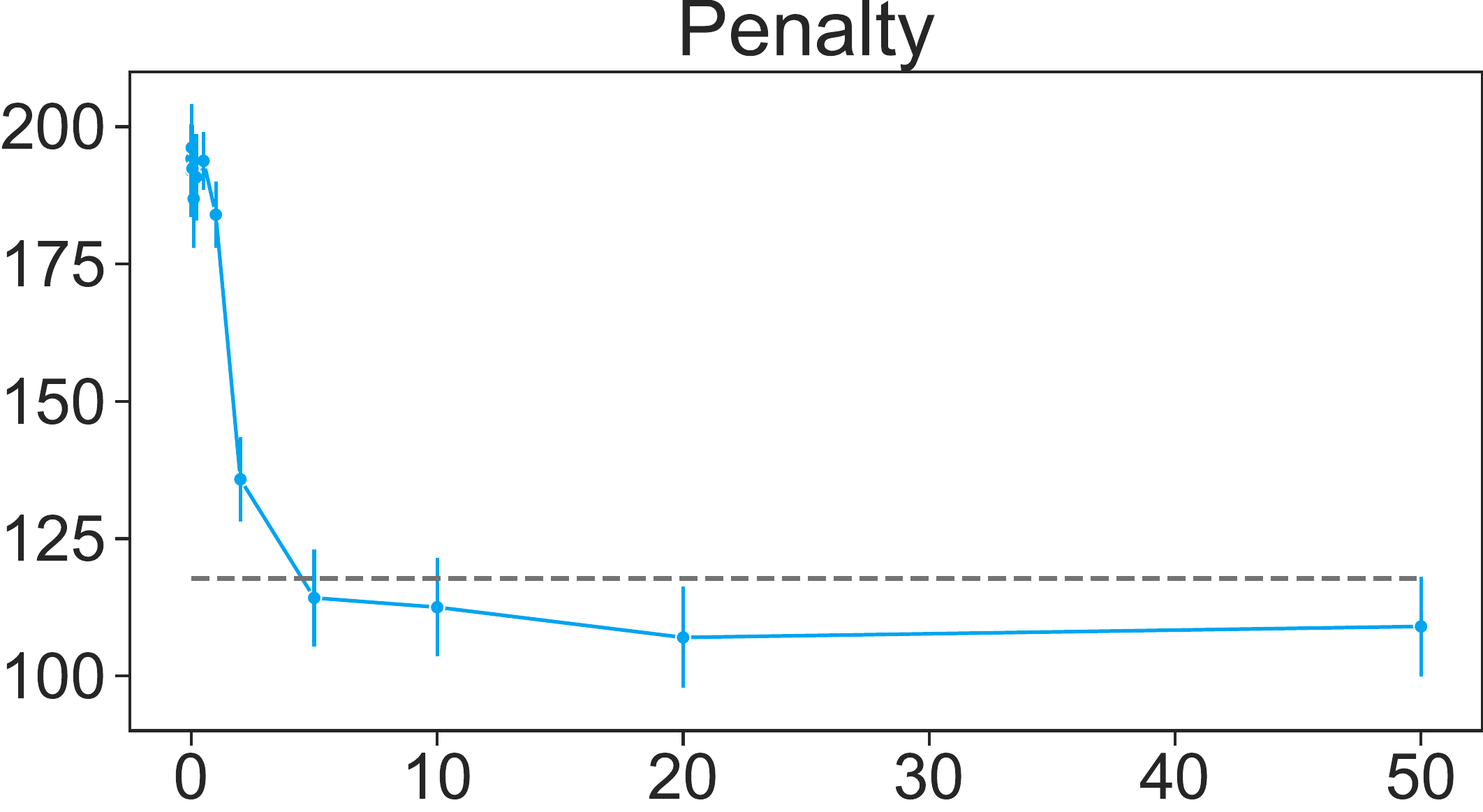}
         \end{subfigure}
         \begin{subfigure}[t]{0.24\textwidth}
             \centering
             \vspace{0.05in}
             \includegraphics[width=\textwidth]{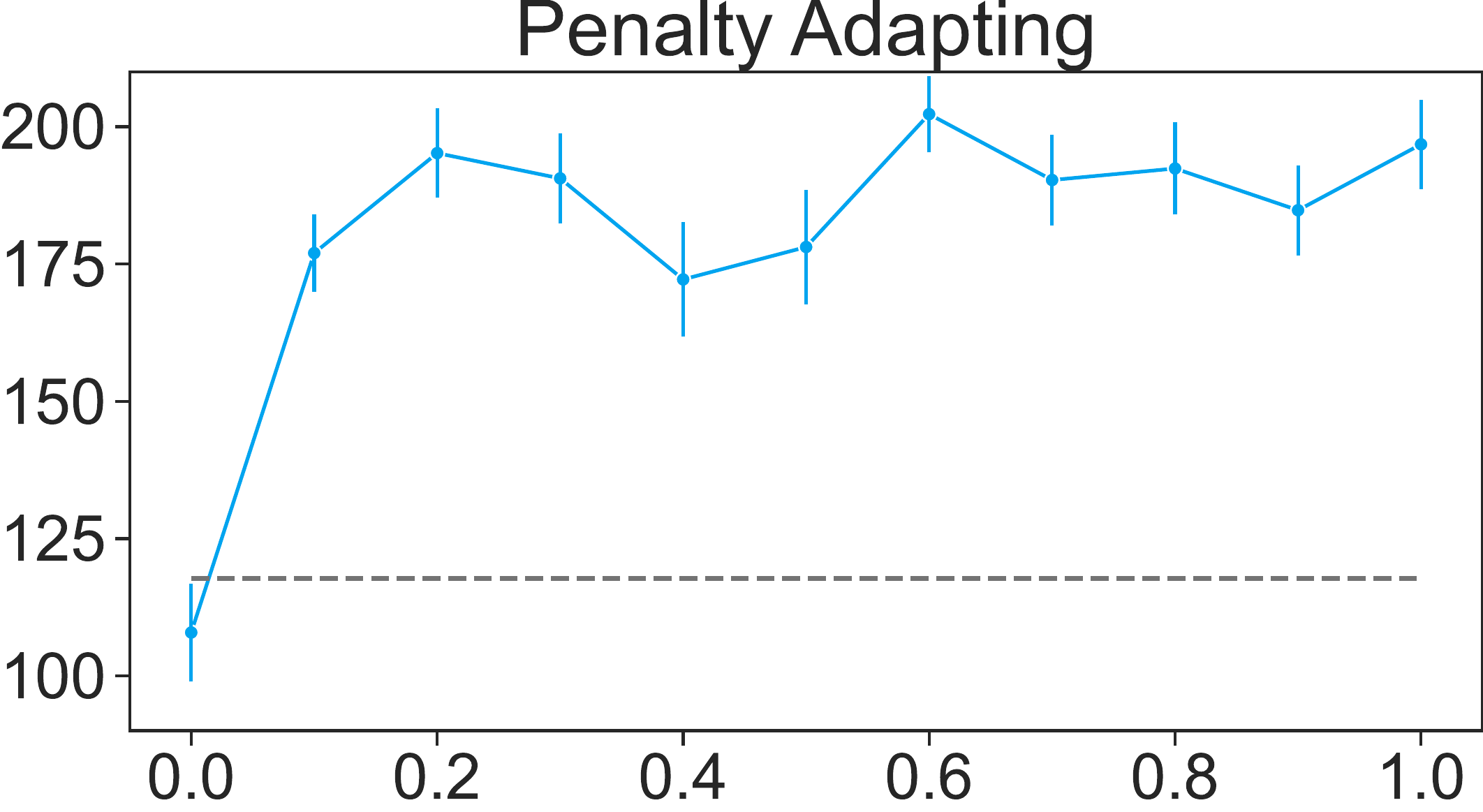}
         \end{subfigure}
     \end{subfigure}
     \begin{subfigure}[t]{\textwidth}
         \centering
         \begin{subfigure}[t]{0.24\textwidth}
             \centering
             \vspace{0.05in}
             \includegraphics[width=\textwidth]{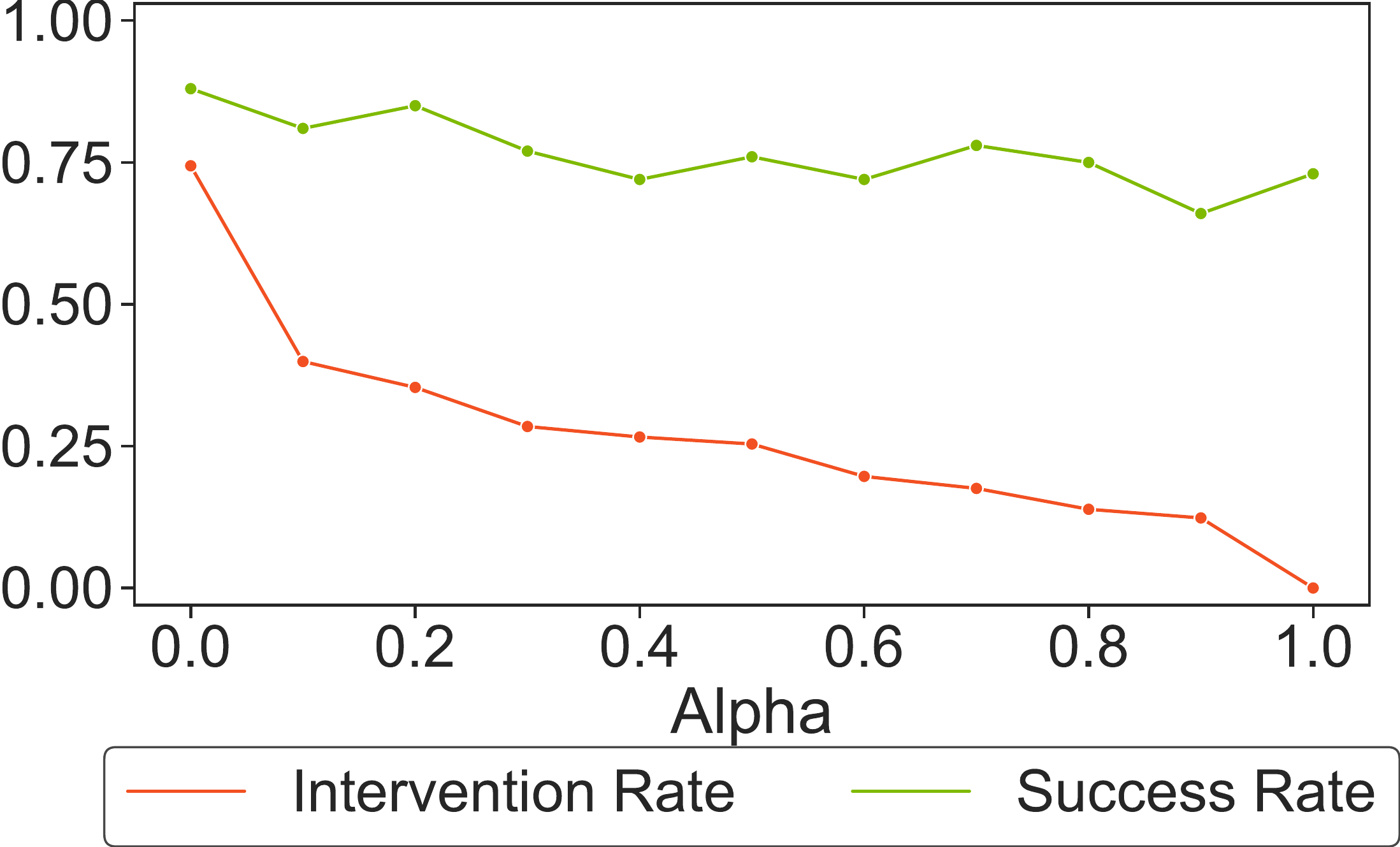}
         \end{subfigure}
         \begin{subfigure}[t]{0.24\textwidth}
             \centering
             \vspace{0.05in}
             \includegraphics[width=\textwidth]{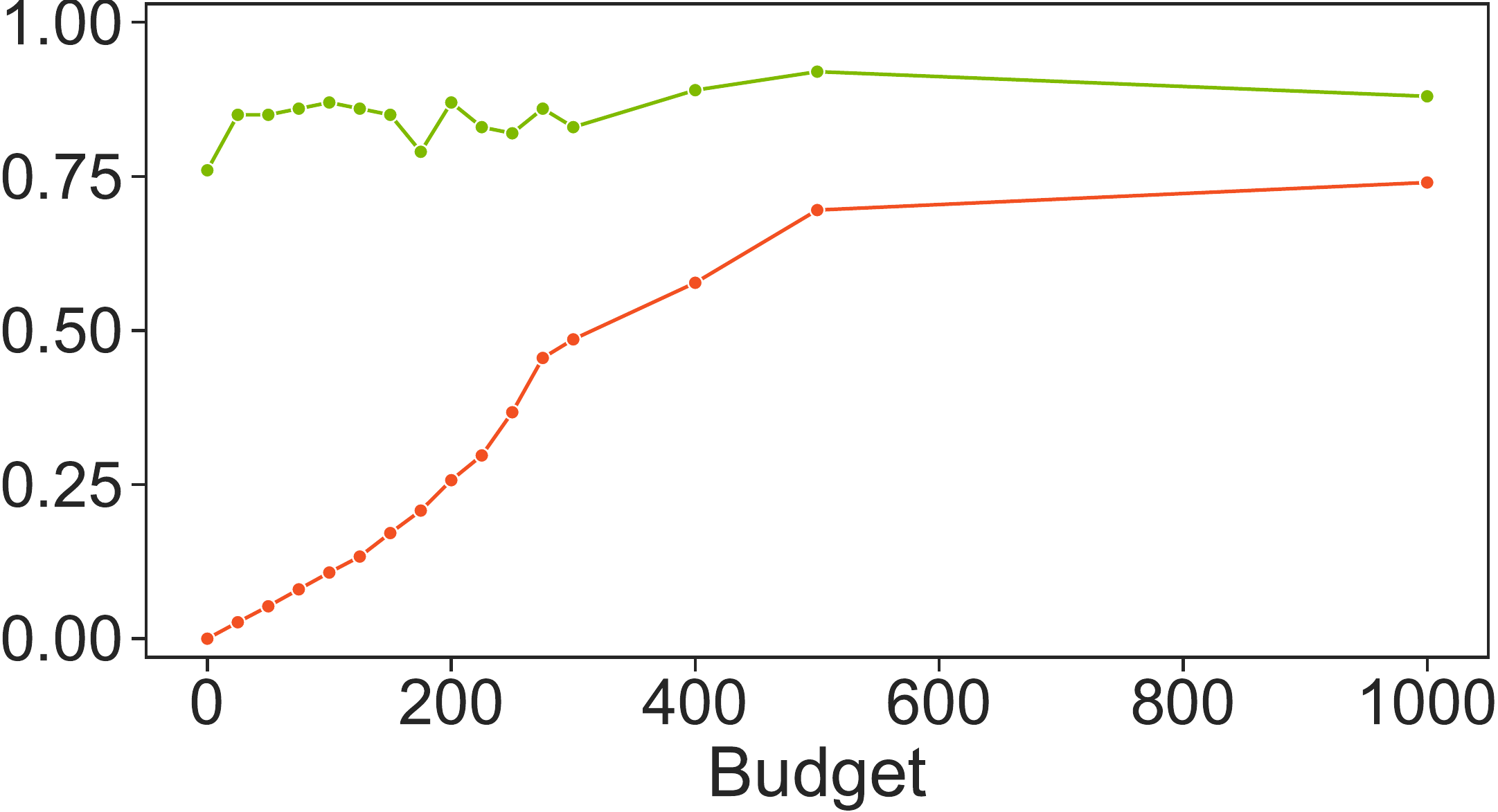}
         \end{subfigure}
         \begin{subfigure}[t]{0.24\textwidth}
             \centering
             \vspace{0.05in}
             \includegraphics[width=\textwidth]{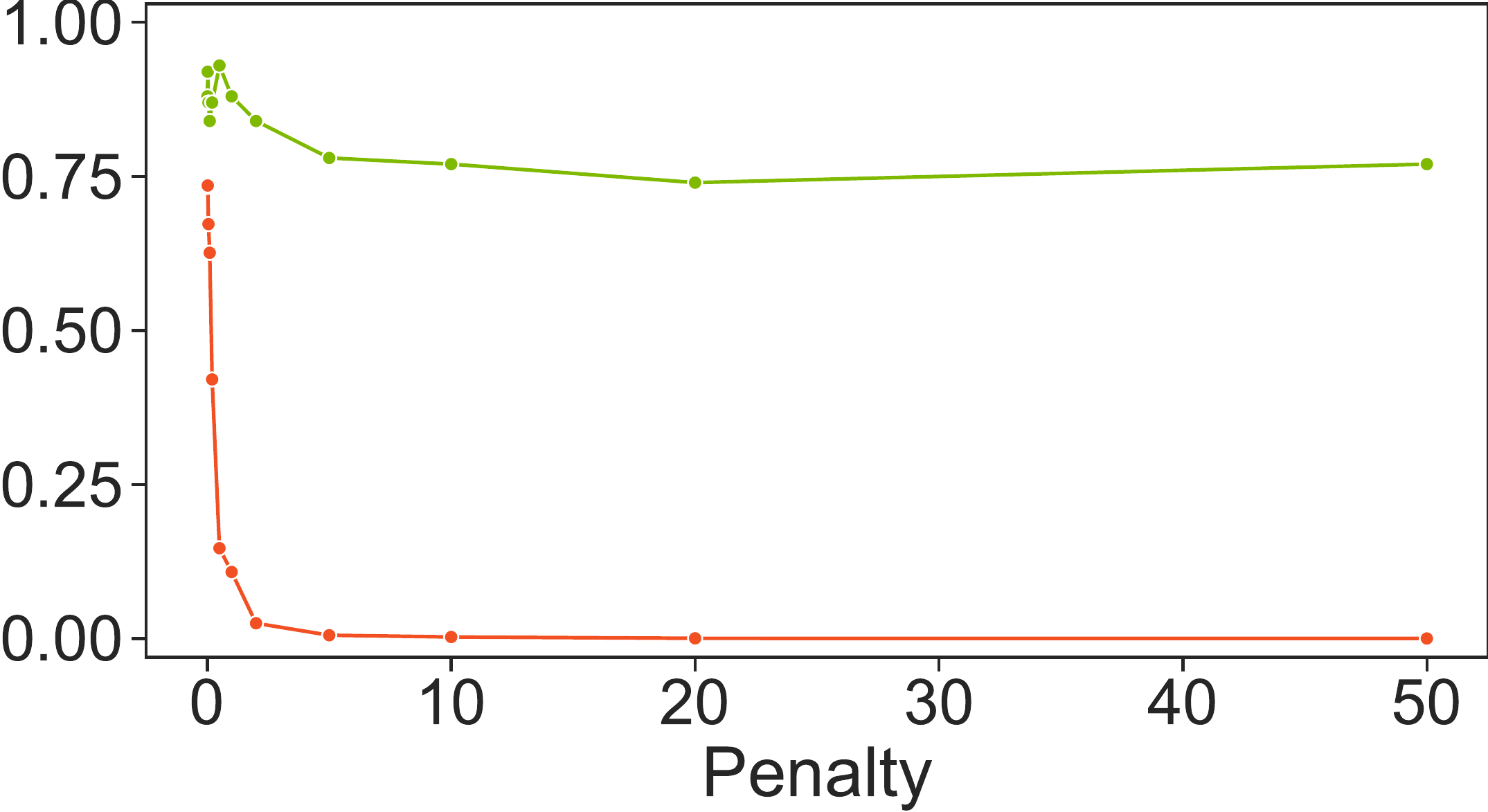}
         \end{subfigure}
         \begin{subfigure}[t]{0.24\textwidth}
             \centering
             \vspace{0.05in}
             \includegraphics[width=\textwidth]{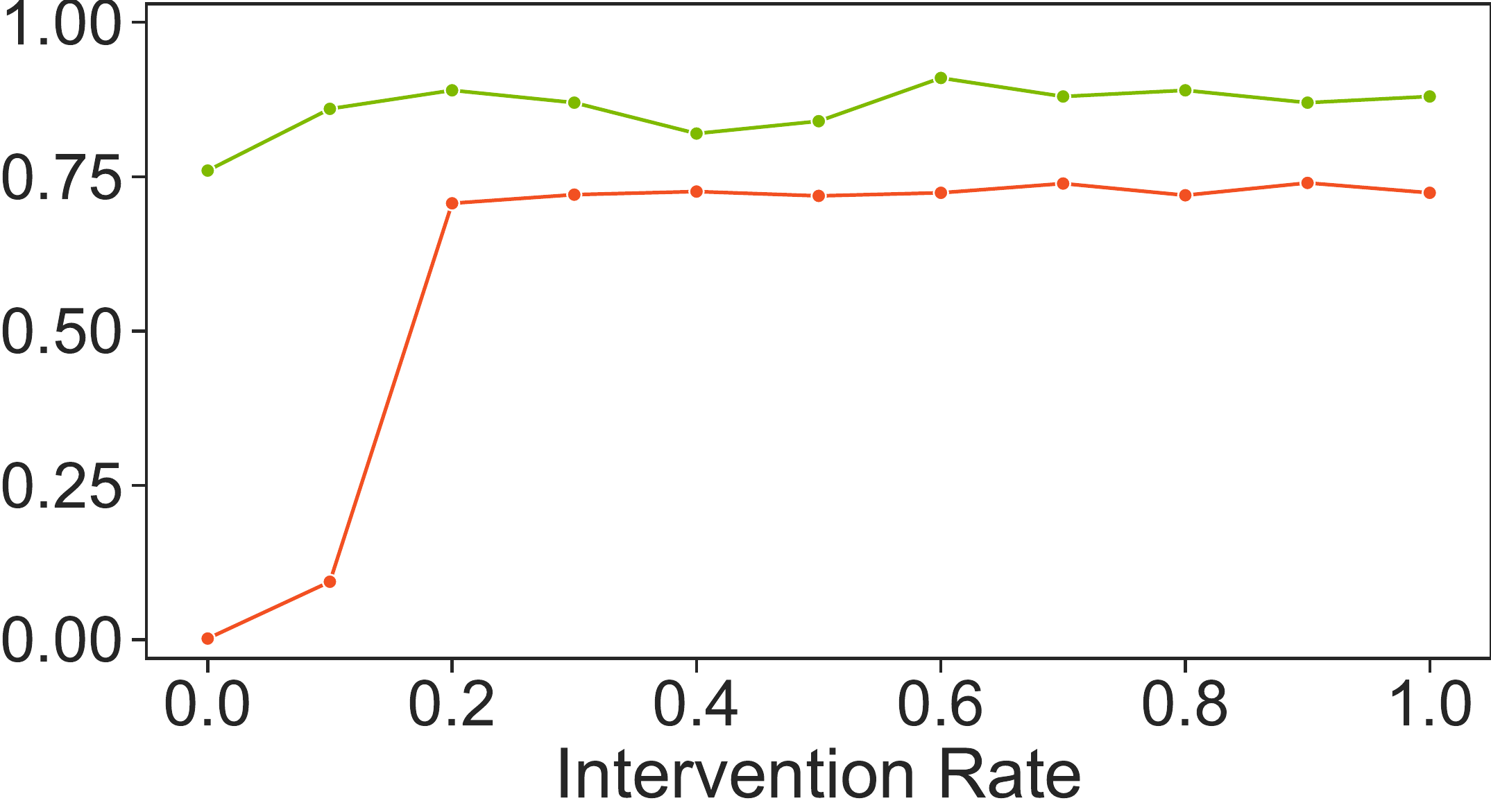}
         \end{subfigure}
     \end{subfigure}
     \caption{The return, intervention rate and success rate of the \textbf{noisy} pilot assisted by the copilot trained with different parameters for different methods on Lunar Lander with discrete action space.}
     \label{fig:lander noisy}
\end{figure*}

\begin{figure*}[!htb]
     \centering
     \begin{subfigure}[t]{\textwidth}
         \centering
         \begin{subfigure}[t]{0.24\textwidth}
             \centering
             \vspace{0.05in}
             \includegraphics[width=\textwidth]{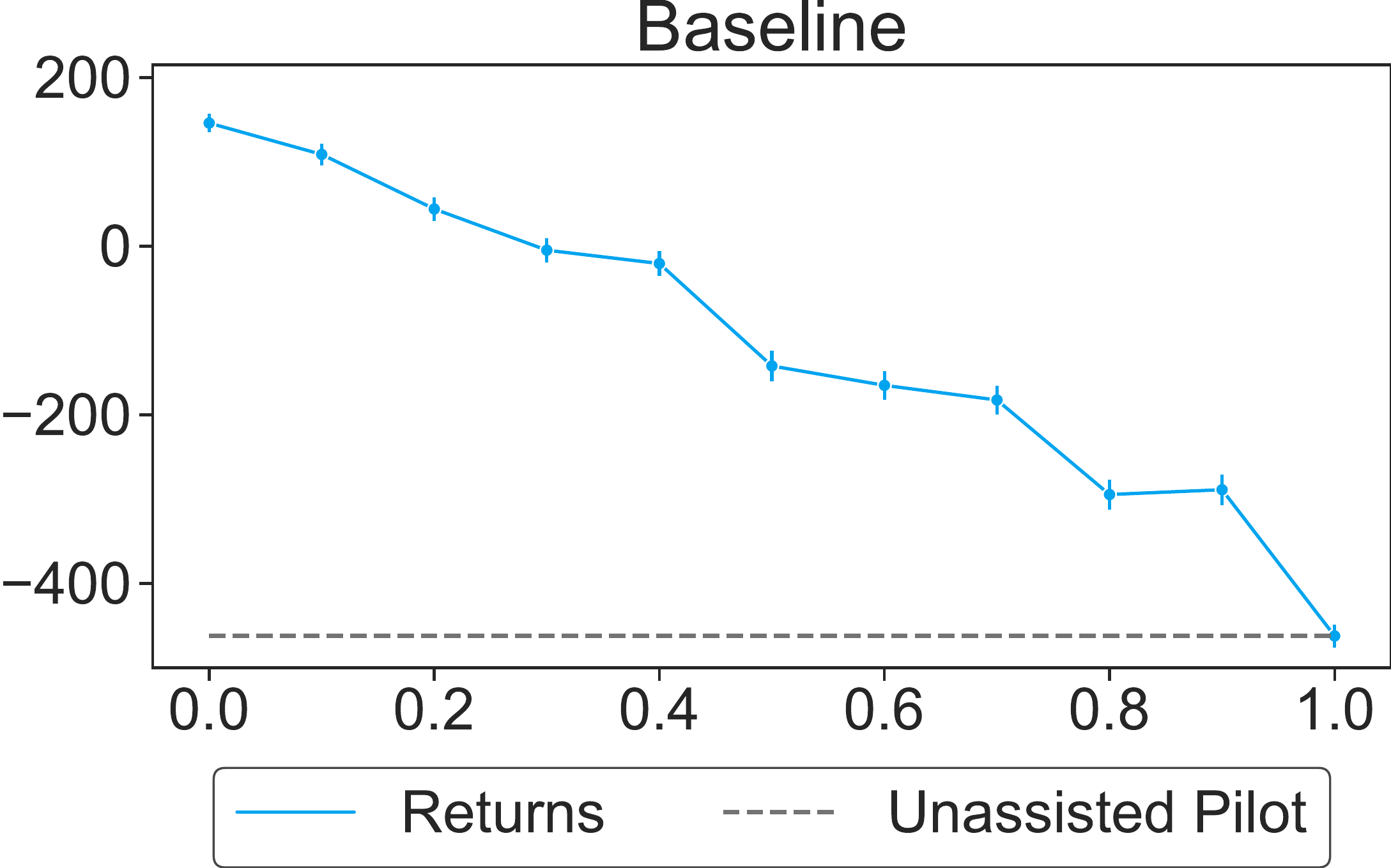}
         \end{subfigure}
         \begin{subfigure}[t]{0.24\textwidth}
             \centering
             \vspace{0.05in}
             \includegraphics[width=\textwidth]{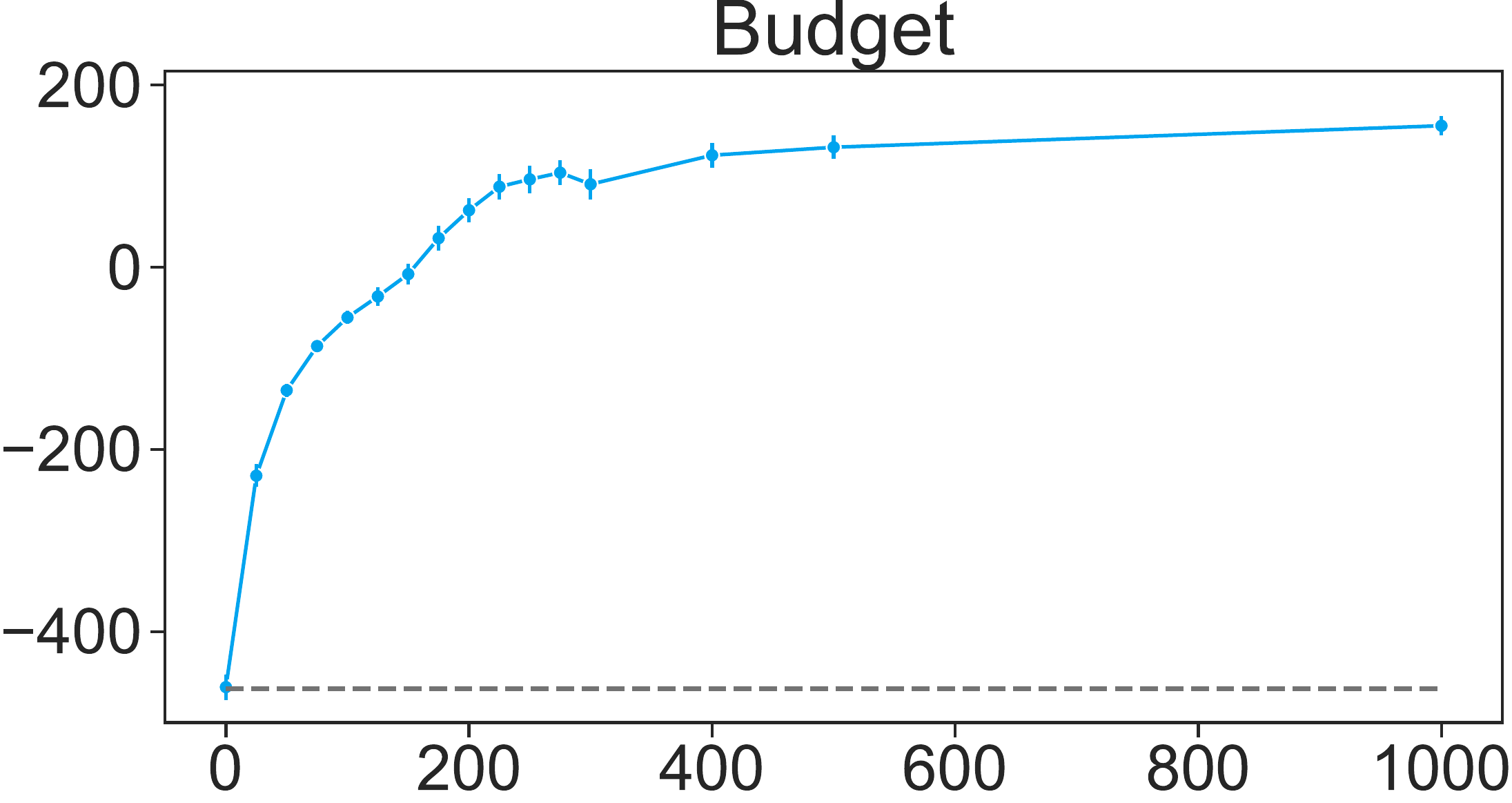}
         \end{subfigure}
         \begin{subfigure}[t]{0.24\textwidth}
             \centering
             \vspace{0.05in}
             \includegraphics[width=\textwidth]{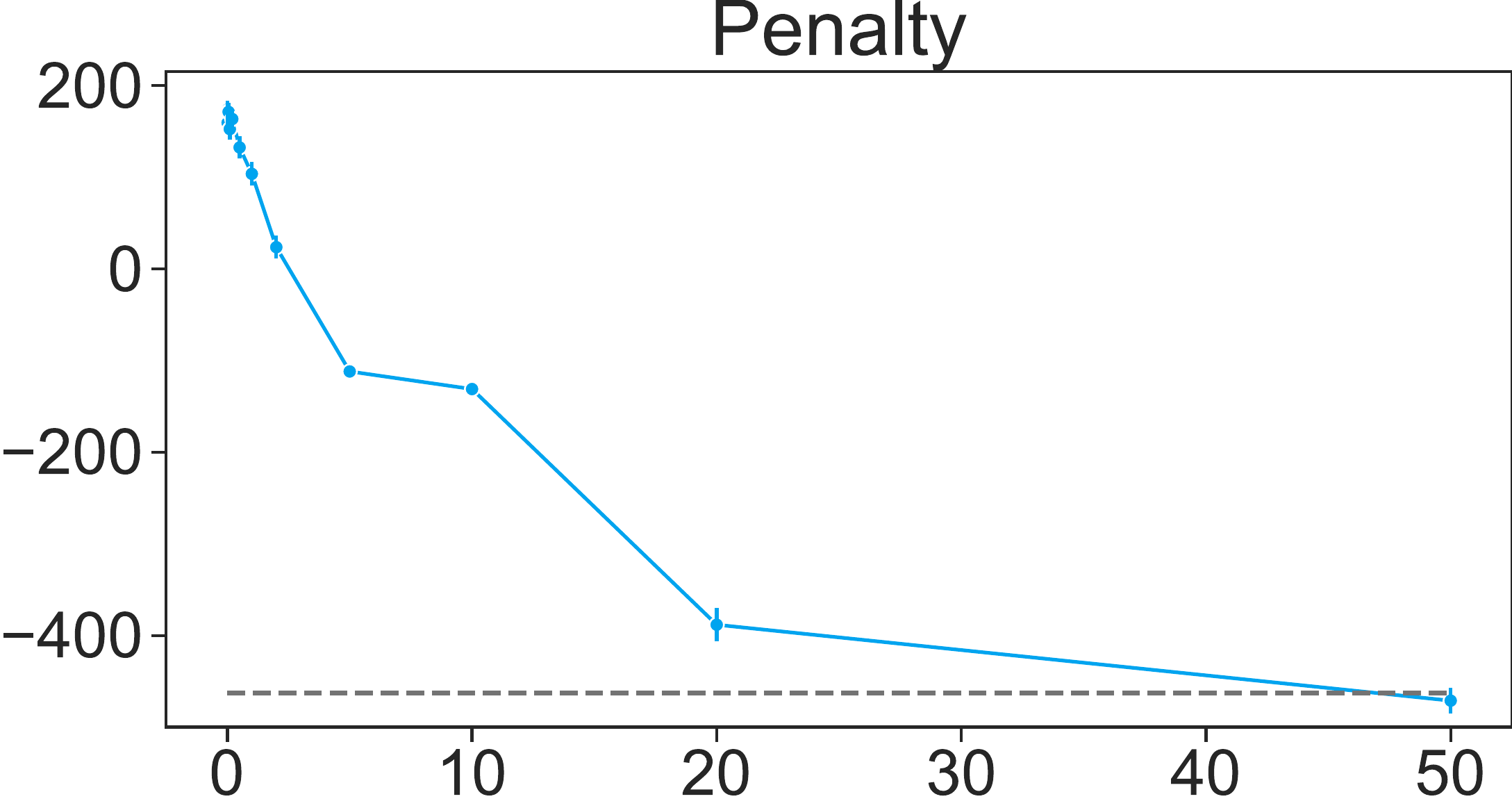}
         \end{subfigure}
         \begin{subfigure}[t]{0.24\textwidth}
             \centering
             \vspace{0.05in}
             \includegraphics[width=\textwidth]{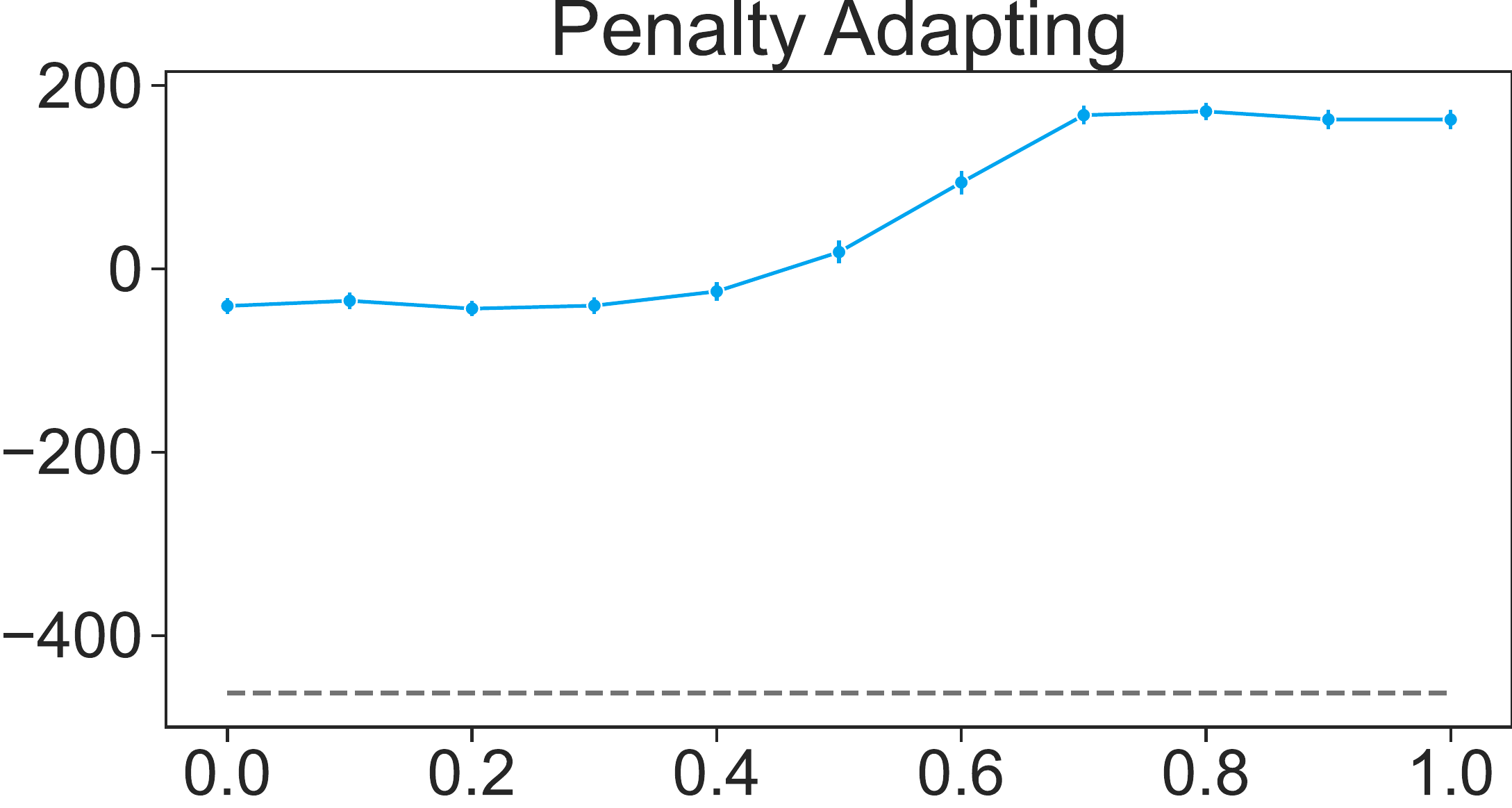}
         \end{subfigure}
     \end{subfigure}
     \begin{subfigure}[t]{\textwidth}
         \centering
         \begin{subfigure}[t]{0.24\textwidth}
             \centering
             \vspace{0.05in}
             \includegraphics[width=\textwidth]{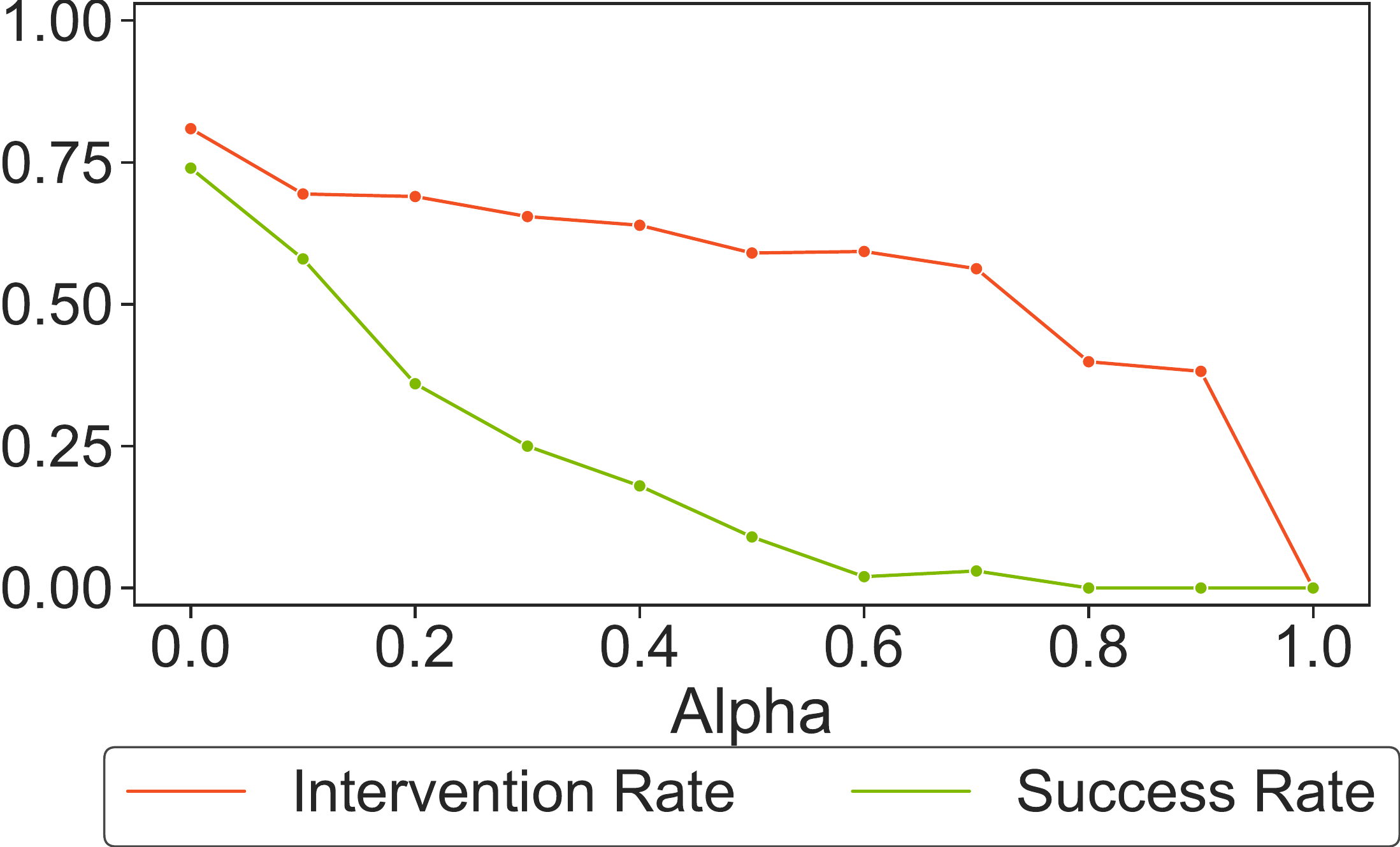}
         \end{subfigure}
         \begin{subfigure}[t]{0.24\textwidth}
             \centering
             \vspace{0.05in}
             \includegraphics[width=\textwidth]{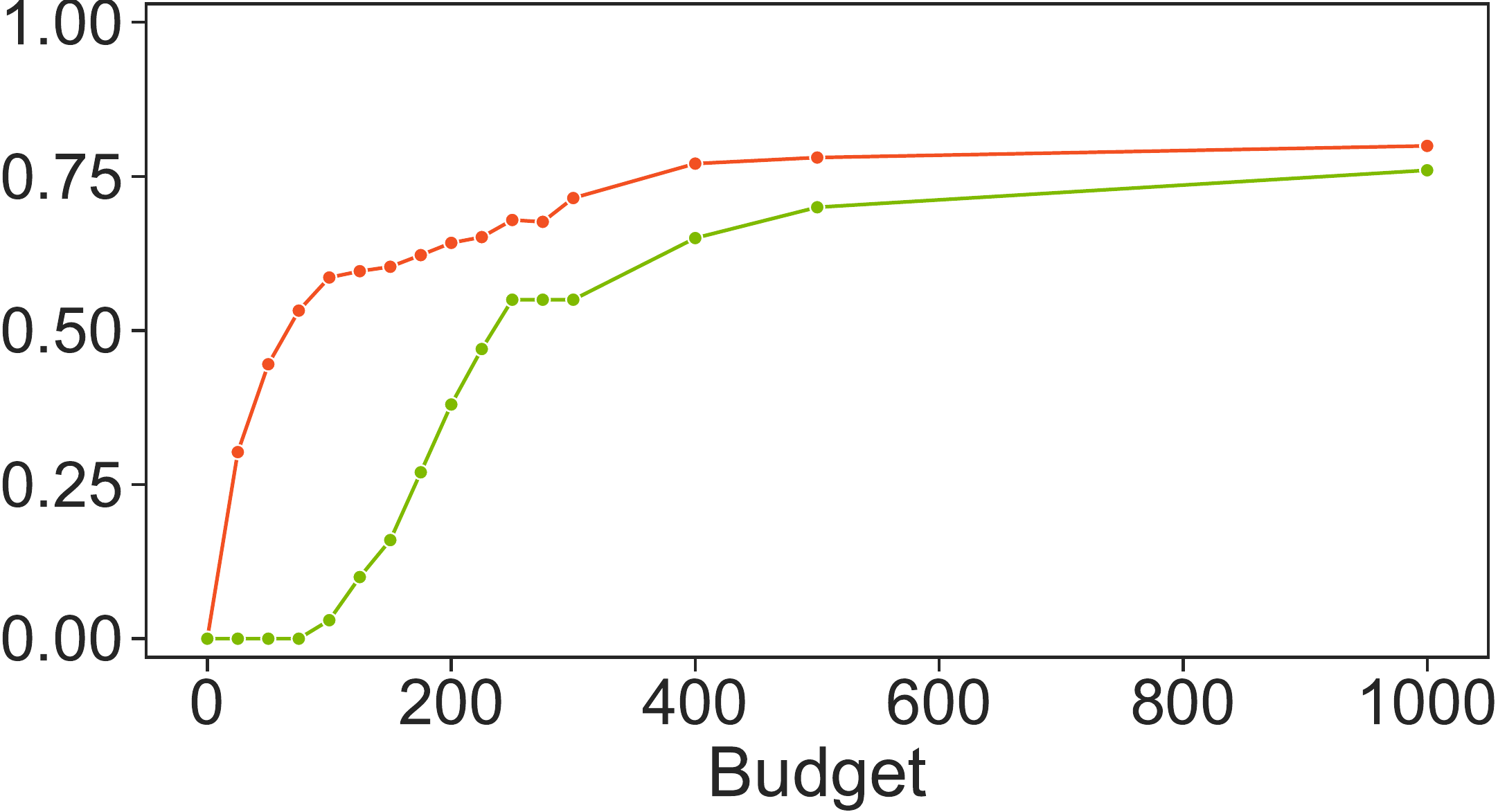}
         \end{subfigure}
         \begin{subfigure}[t]{0.24\textwidth}
             \centering
             \vspace{0.05in}
             \includegraphics[width=\textwidth]{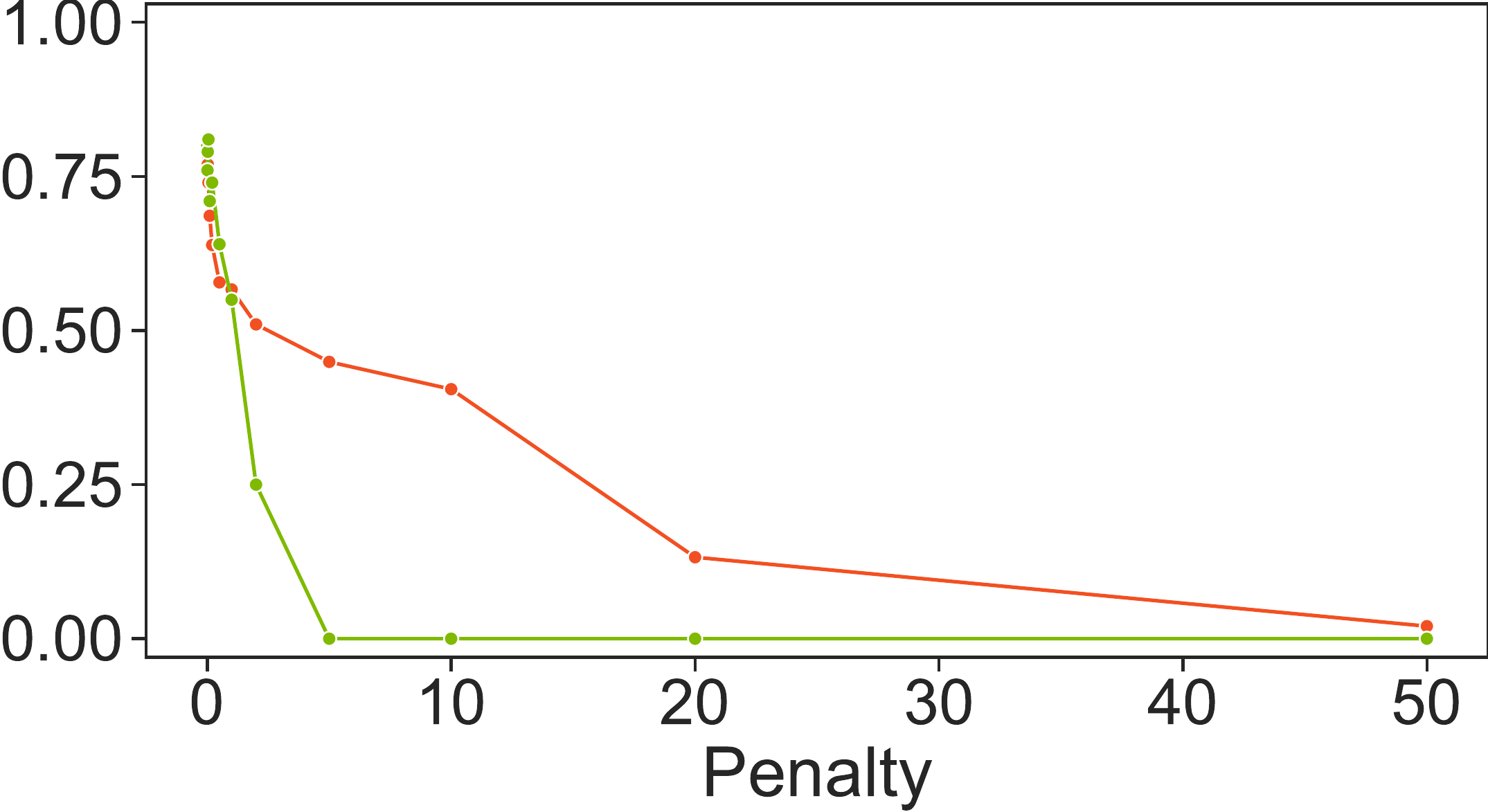}
         \end{subfigure}
         \begin{subfigure}[t]{0.24\textwidth}
             \centering
             \vspace{0.05in}
             \includegraphics[width=\textwidth]{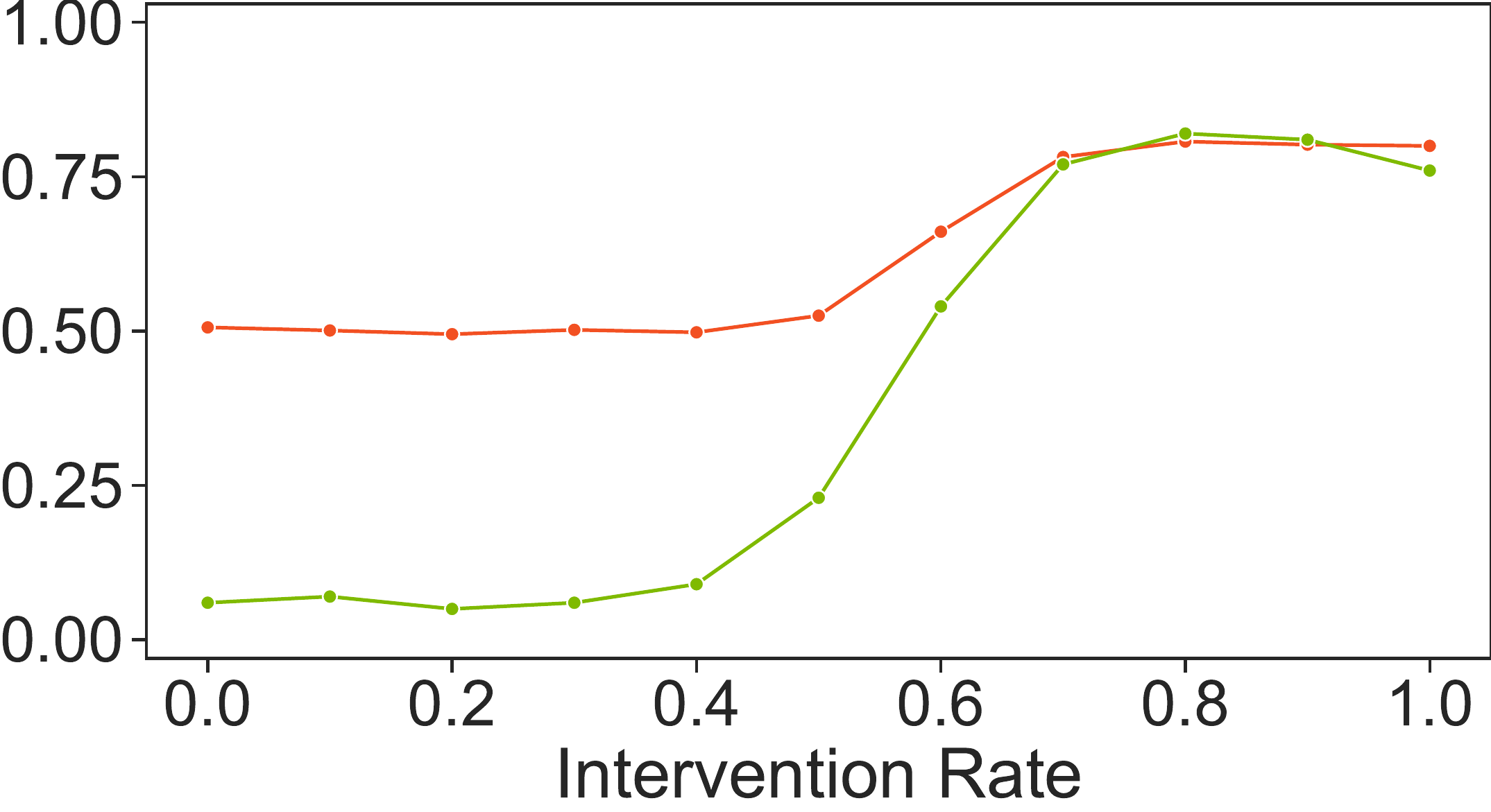}
         \end{subfigure}
     \end{subfigure}
     \caption{The return, intervention rate and success rate of the \textbf{sensor} pilot assisted by the copilot trained with different parameters for different methods on Lunar Lander with discrete action space.}
     \label{fig:lander sensor}
\end{figure*}

\begin{figure*}[!htb]
     \centering
     \begin{subfigure}[t]{\textwidth}
         \centering
         \begin{subfigure}[t]{0.24\textwidth}
             \centering
             \vspace{0.05in}
             \includegraphics[width=\textwidth]{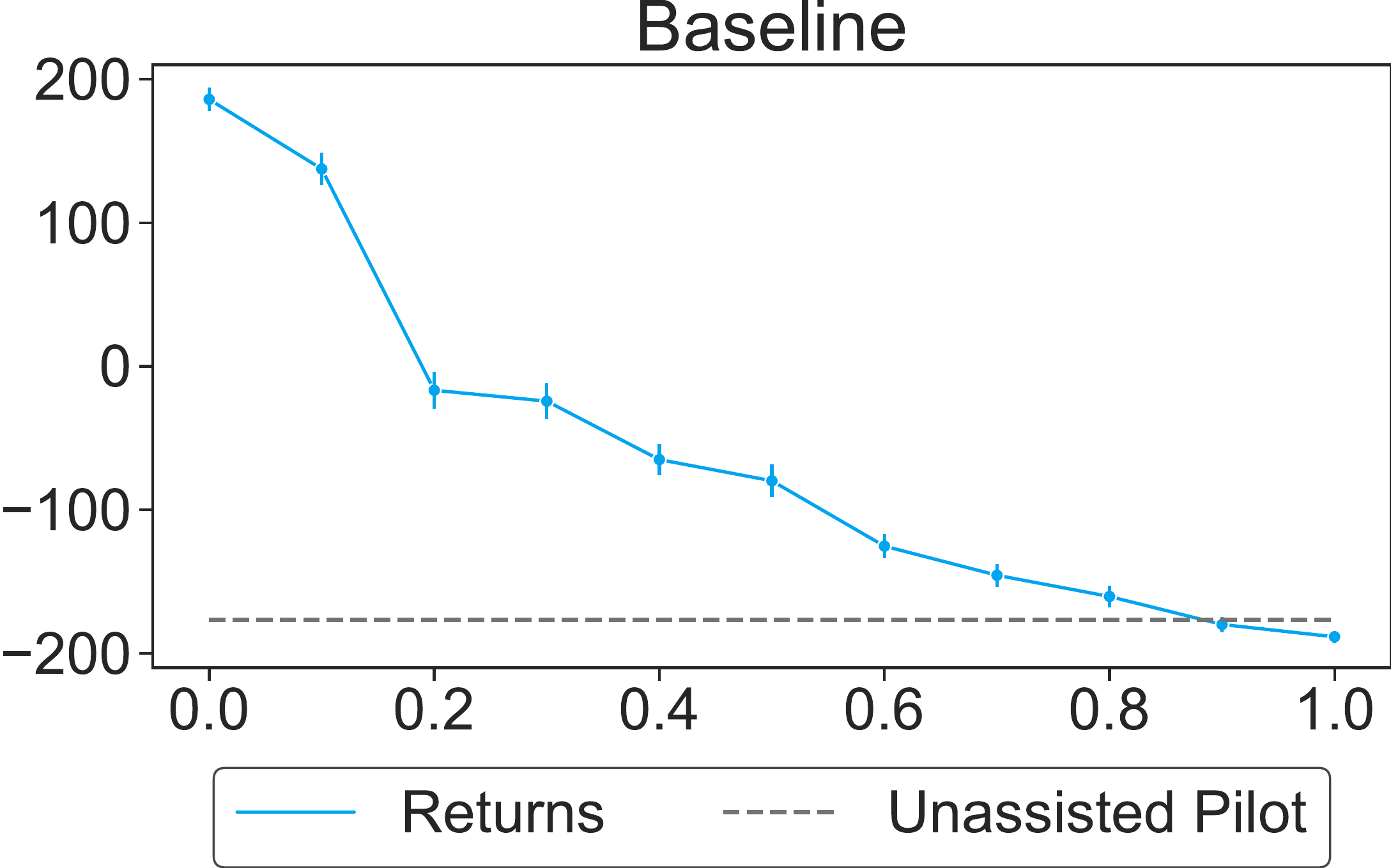}
         \end{subfigure}
         \begin{subfigure}[t]{0.24\textwidth}
             \centering
             \vspace{0.05in}
             \includegraphics[width=\textwidth]{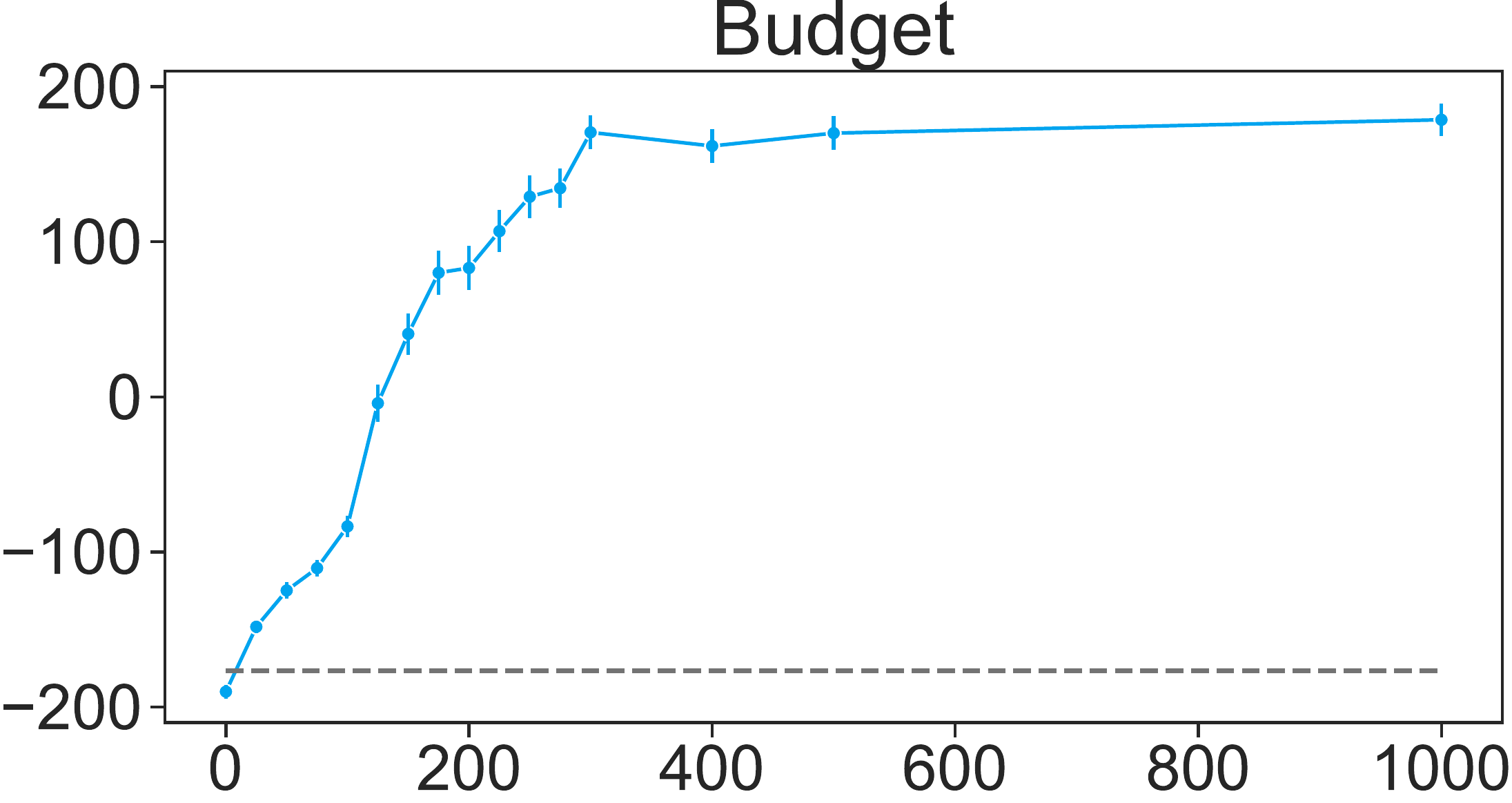}
         \end{subfigure}
         \begin{subfigure}[t]{0.24\textwidth}
             \centering
             \vspace{0.05in}
             \includegraphics[width=\textwidth]{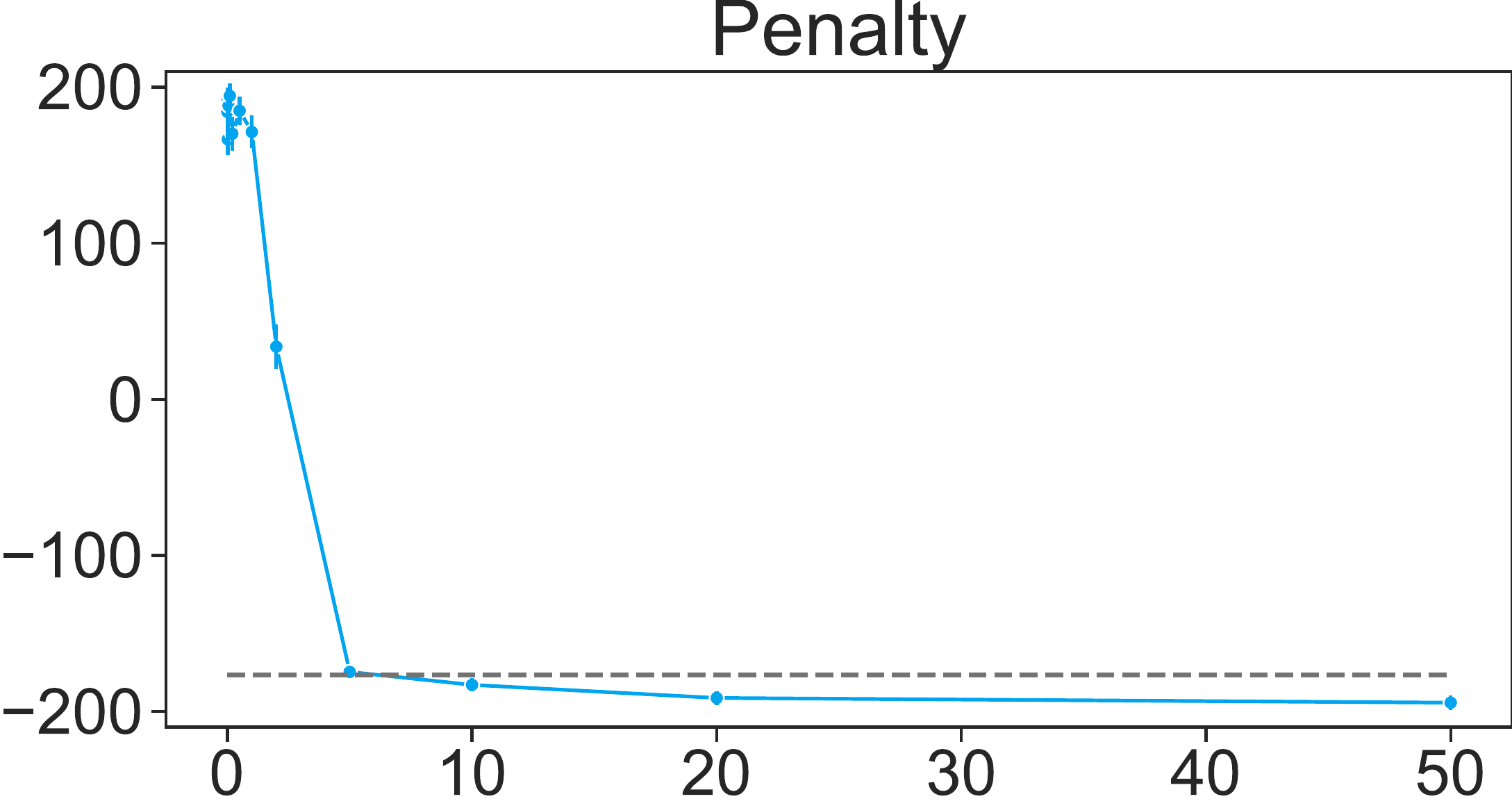}
         \end{subfigure}
         \begin{subfigure}[t]{0.24\textwidth}
             \centering
             \vspace{0.05in}
             \includegraphics[width=\textwidth]{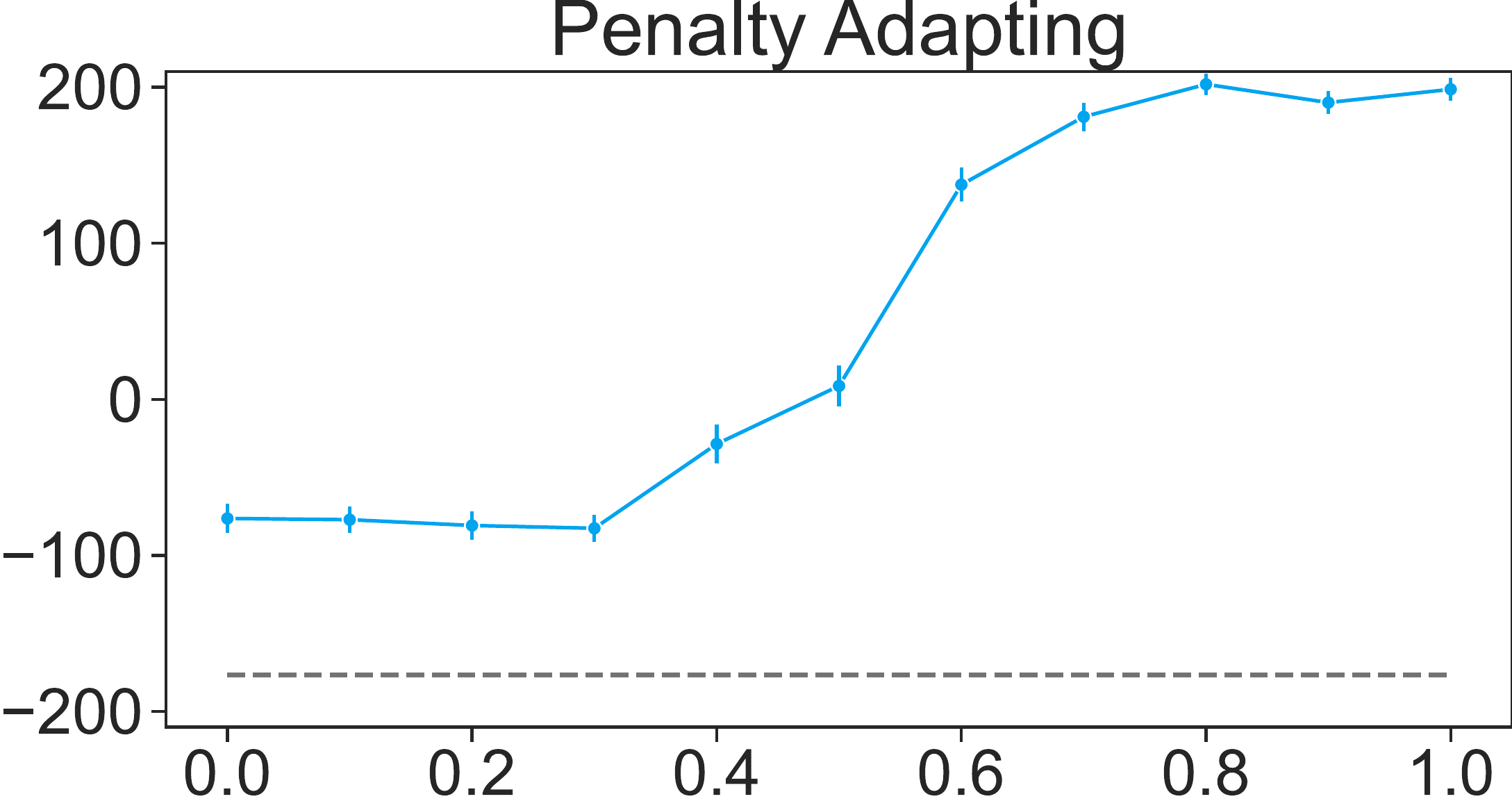}
         \end{subfigure}
     \end{subfigure}
     \begin{subfigure}[t]{\textwidth}
         \centering
         \begin{subfigure}[t]{0.24\textwidth}
             \centering
             \vspace{0.05in}
             \includegraphics[width=\textwidth]{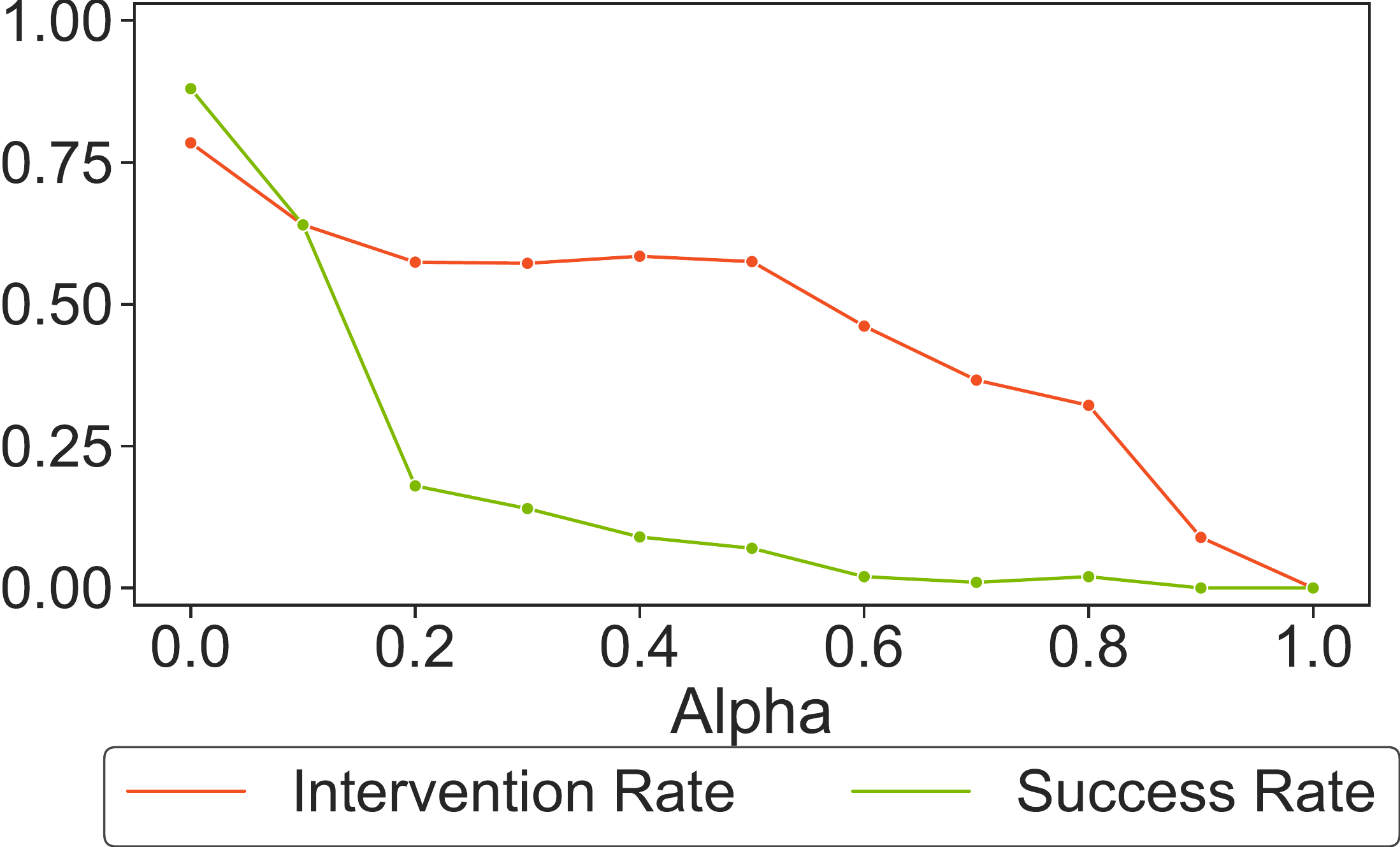}
         \end{subfigure}
         \begin{subfigure}[t]{0.24\textwidth}
             \centering
             \vspace{0.05in}
             \includegraphics[width=\textwidth]{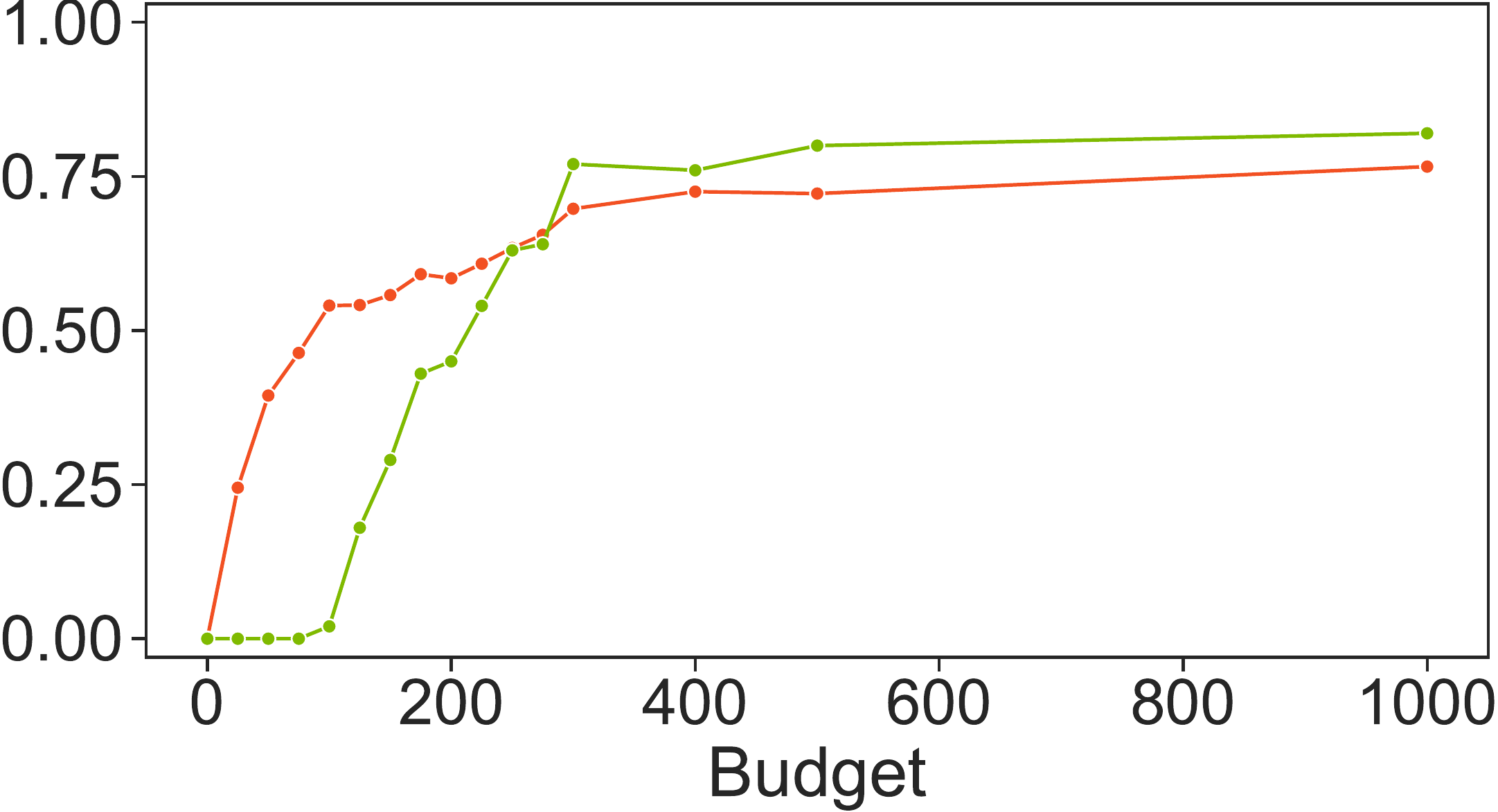}
         \end{subfigure}
         \begin{subfigure}[t]{0.24\textwidth}
             \centering
             \vspace{0.05in}
             \includegraphics[width=\textwidth]{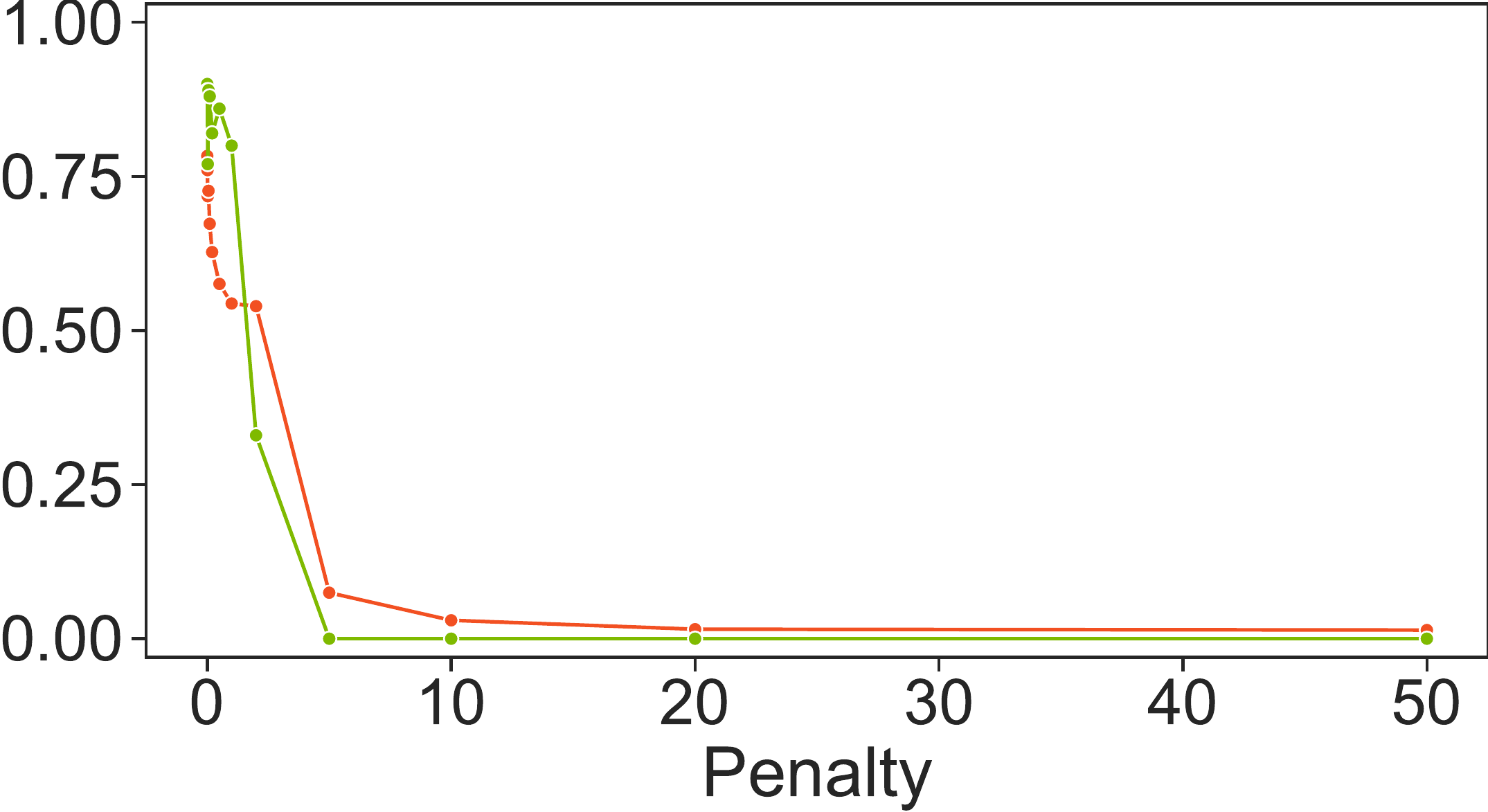}
         \end{subfigure}
         \begin{subfigure}[t]{0.24\textwidth}
             \centering
             \vspace{0.05in}
             \includegraphics[width=\textwidth]{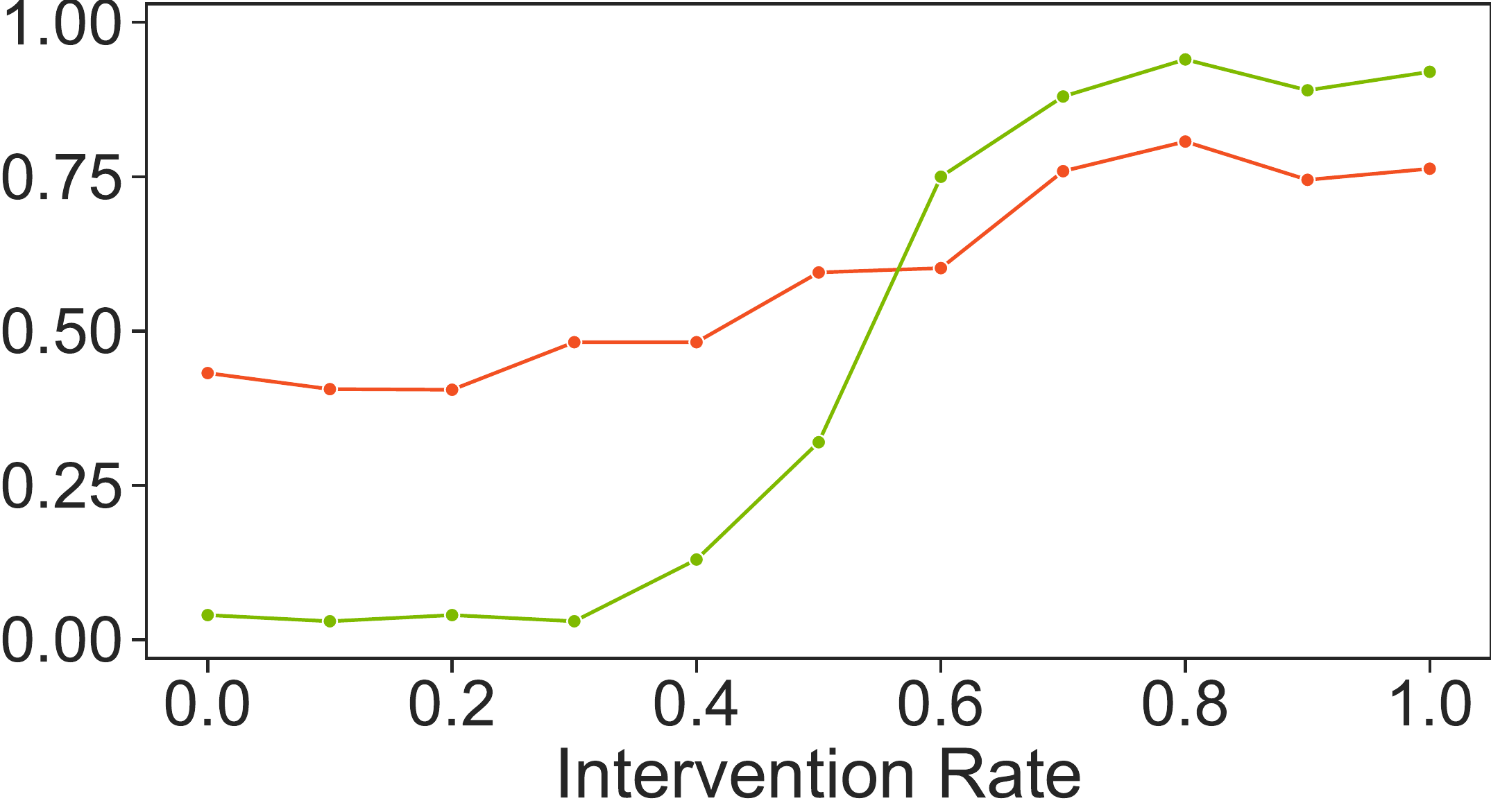}
         \end{subfigure}
     \end{subfigure}
     \caption{The return, intervention rate and success rate of the \textbf{no-op} pilot assisted by the copilot trained with different parameters for different methods on Lunar Lander with discrete action space.}
     \label{fig:lander noop}
\end{figure*}

Here we present the results of our experiments at the most granular level, which shows the performance of our system in terms of returns, intervention rate, and success rate. For Lunar Lander with discrete action space, we use four simulated agents as pilot, laggy pilot, noisy pilot, sensor pilot and noop pilot, which correspond to Figure \ref{fig:lander laggy}, \ref{fig:lander noisy}, \ref{fig:lander sensor}, \ref{fig:lander noop}, respectively.

\subsection{Super Mario}
\begin{figure*}[!htb]
     \centering
     \begin{subfigure}[t]{\textwidth}
         \centering
         \begin{subfigure}[t]{0.24\textwidth}
             \centering
             \vspace{0.01in}
             \includegraphics[width=\textwidth]{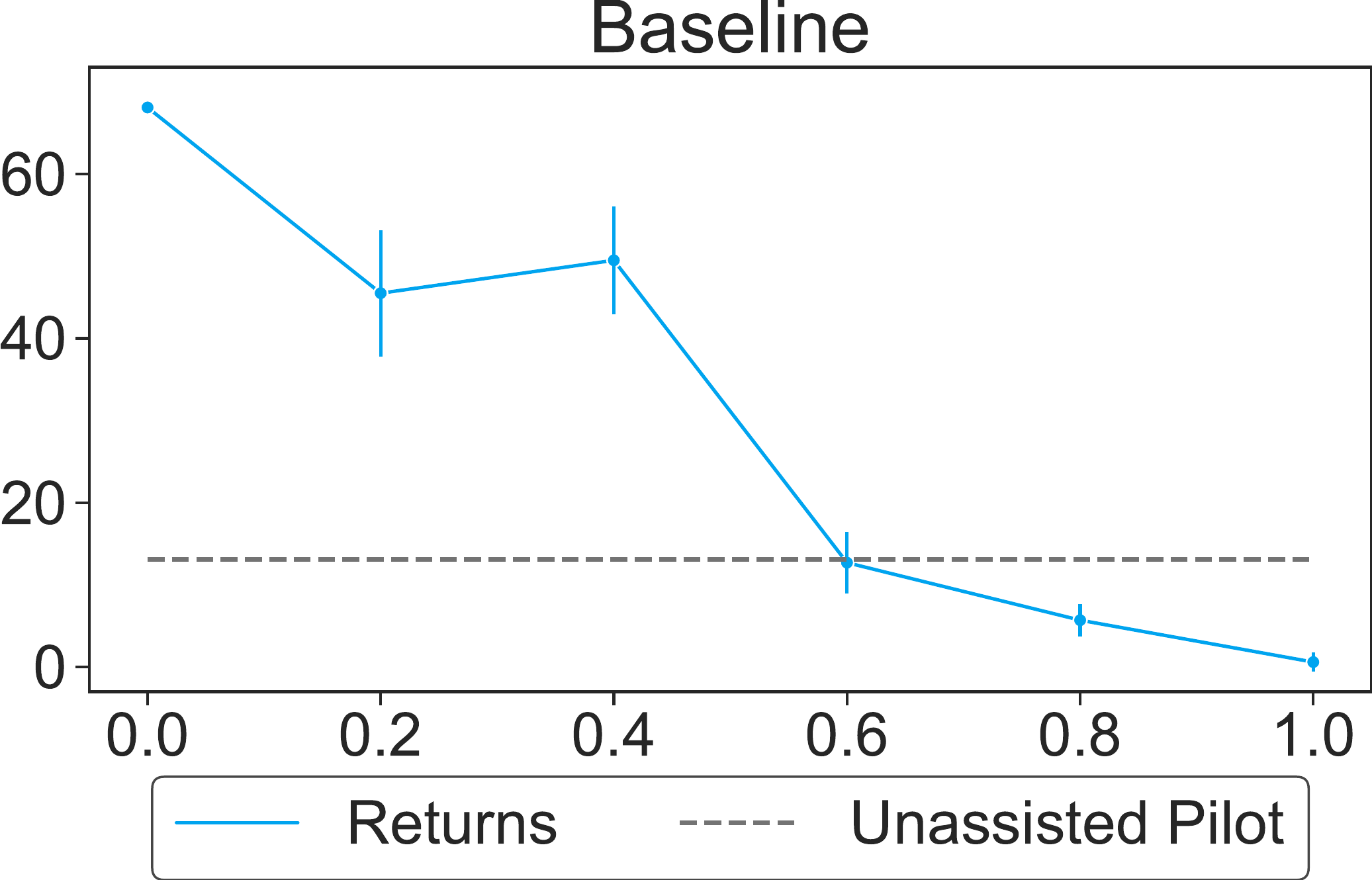}
         \end{subfigure}
         \begin{subfigure}[t]{0.24\textwidth}
             \centering
             \vspace{0.01in}
             \includegraphics[width=\textwidth]{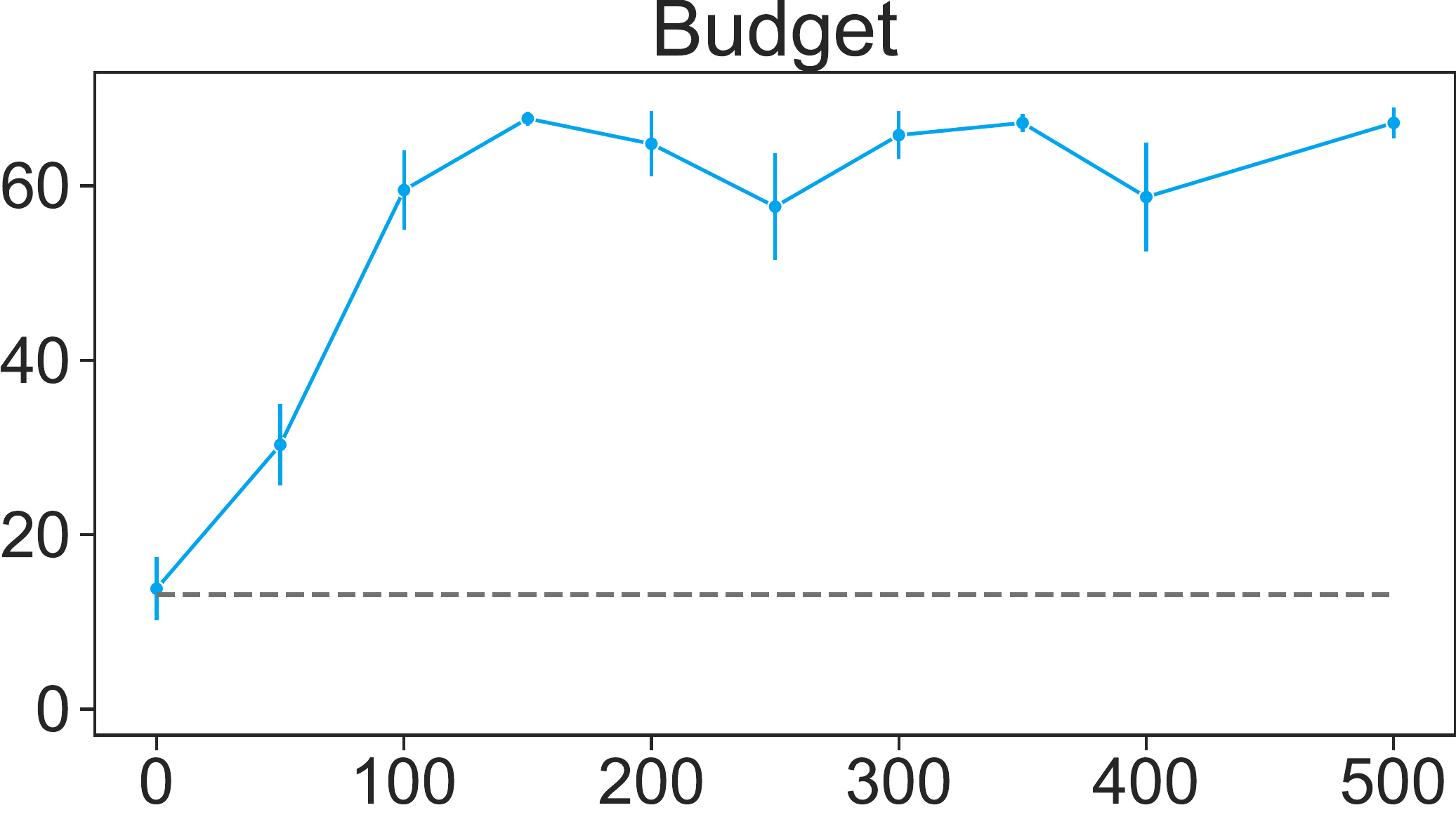}
         \end{subfigure}
         \begin{subfigure}[t]{0.24\textwidth}
             \centering
             \vspace{0.01in}
             \includegraphics[width=\textwidth]{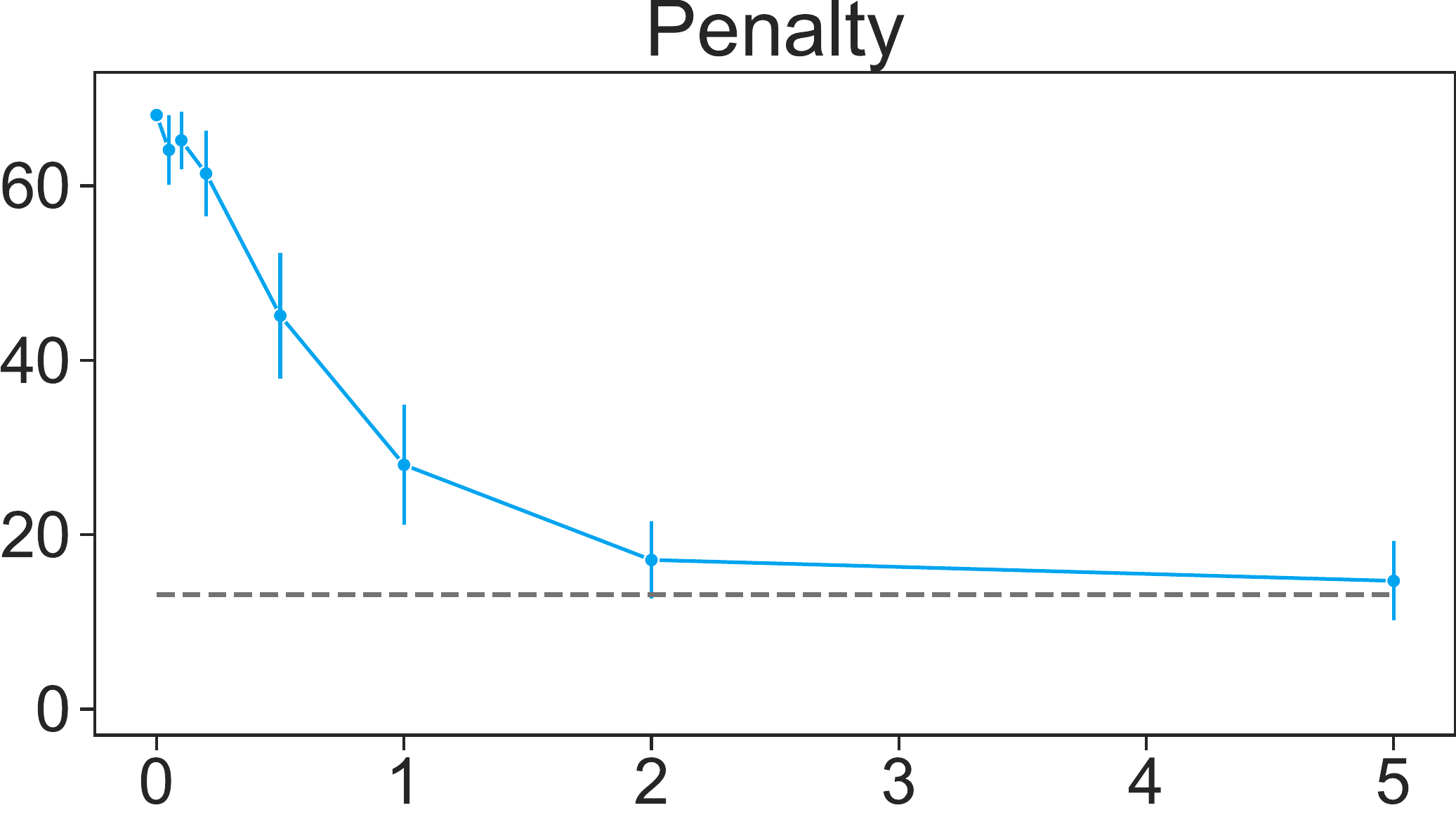}
         \end{subfigure}
         \begin{subfigure}[t]{0.24\textwidth}
             \centering
             \vspace{0.01in}
             \includegraphics[width=\textwidth]{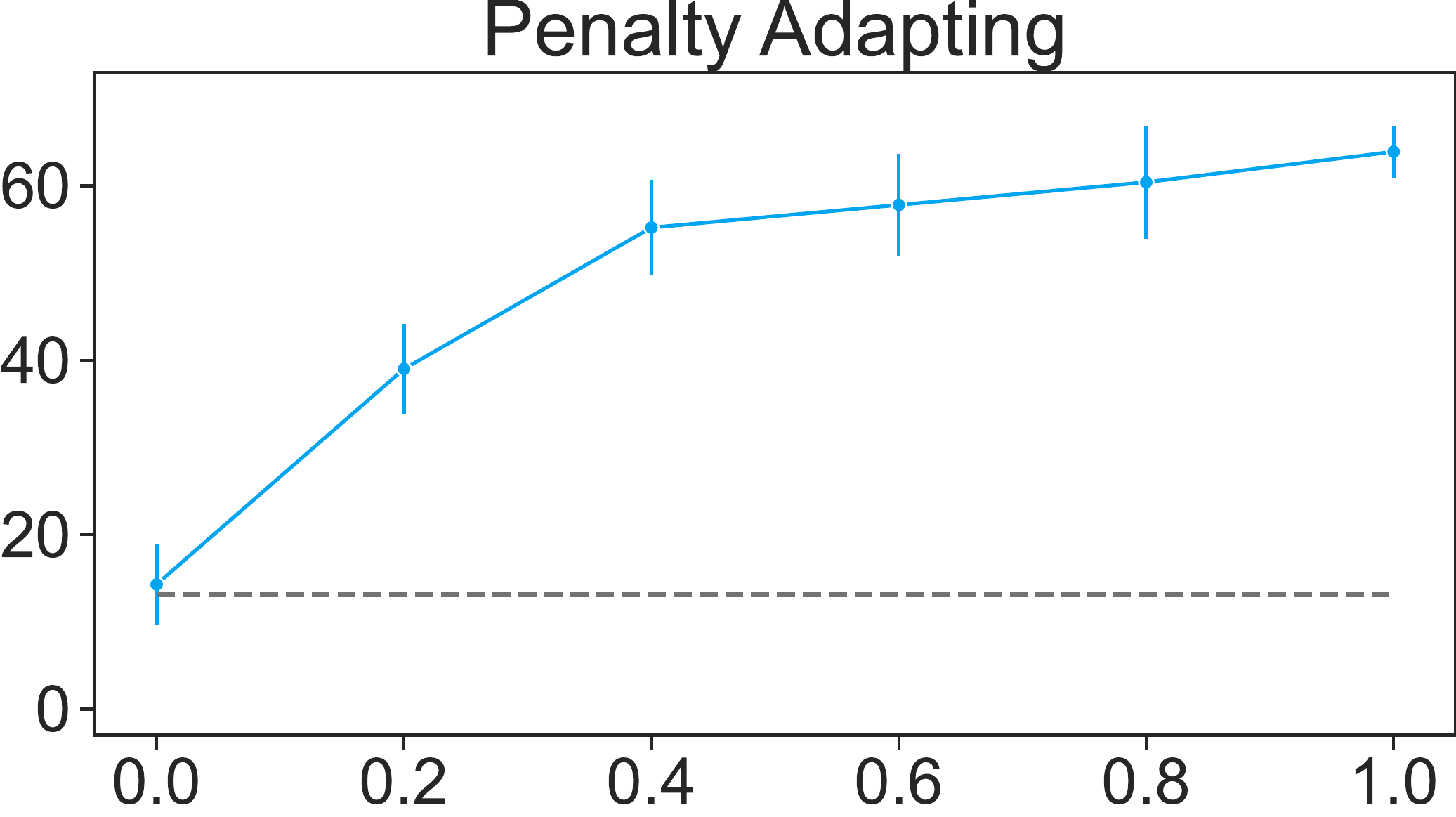}
         \end{subfigure}
     \end{subfigure}
     \begin{subfigure}[t]{\textwidth}
         \centering
         \begin{subfigure}[t]{0.24\textwidth}
             \centering
             \vspace{0.01in}
             \includegraphics[width=\textwidth]{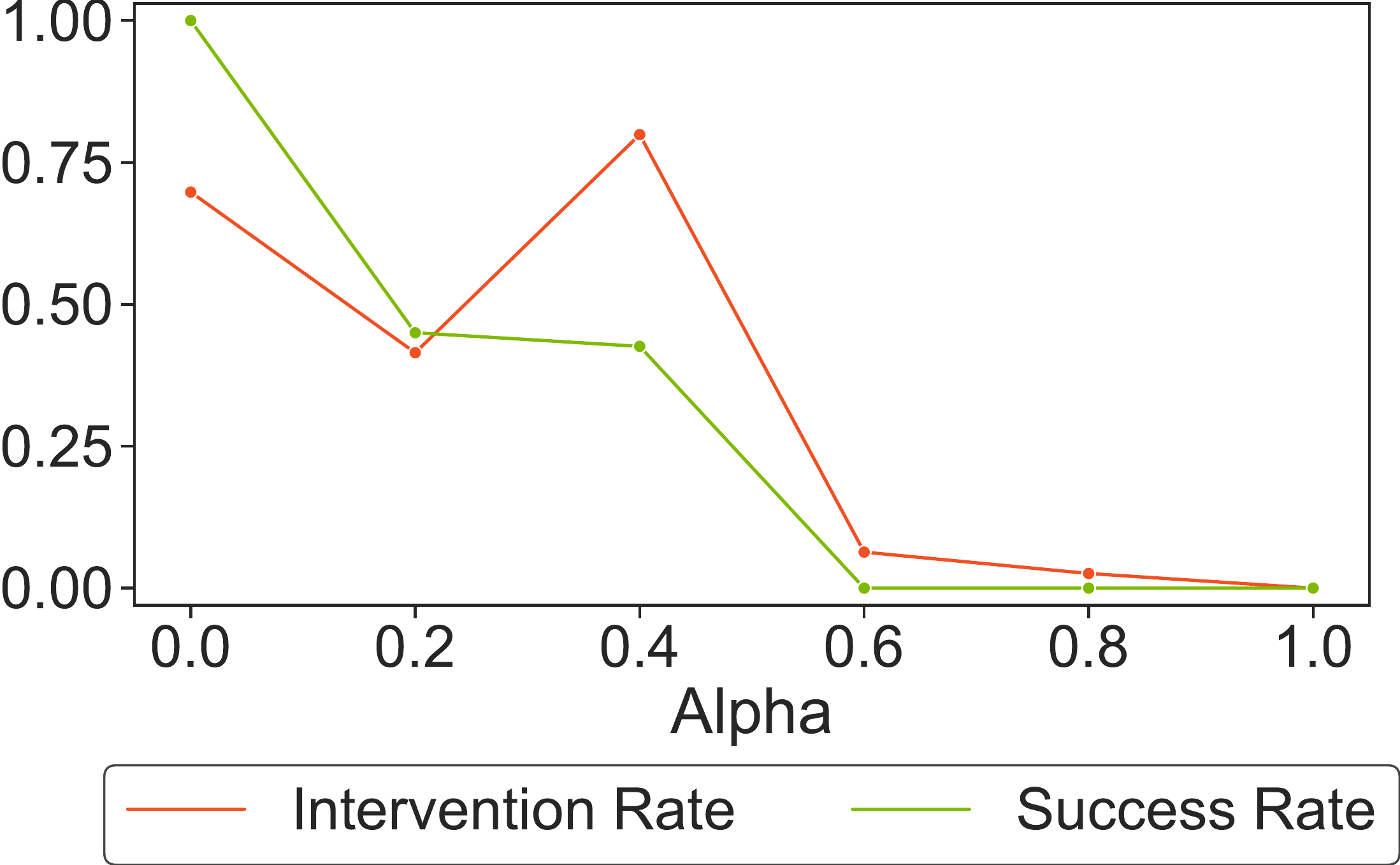}
         \end{subfigure}
         \begin{subfigure}[t]{0.24\textwidth}
             \centering
             \vspace{0.01in}
             \includegraphics[width=\textwidth]{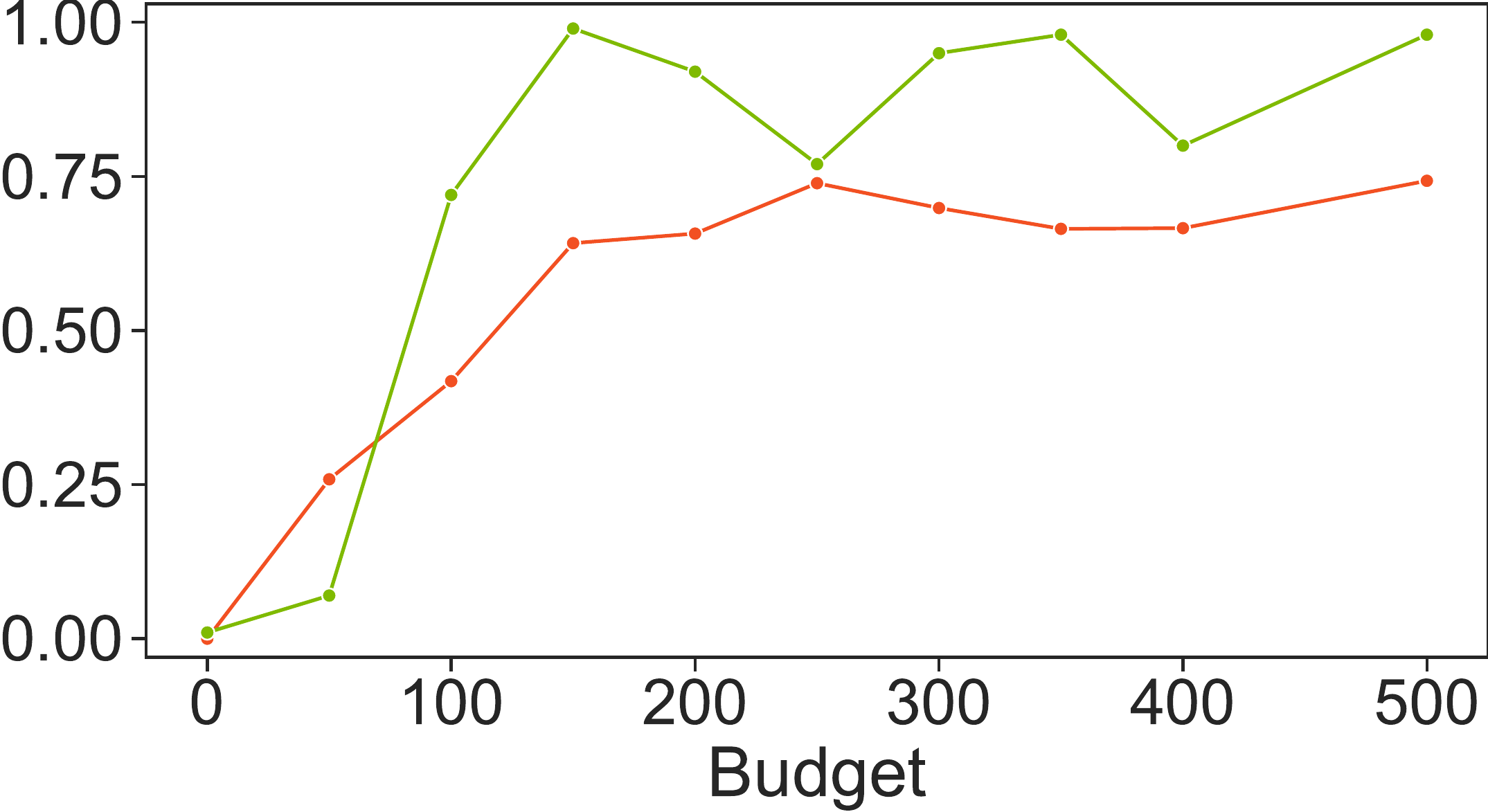}
         \end{subfigure}
         \begin{subfigure}[t]{0.24\textwidth}
             \centering
             \vspace{0.01in}
             \includegraphics[width=\textwidth]{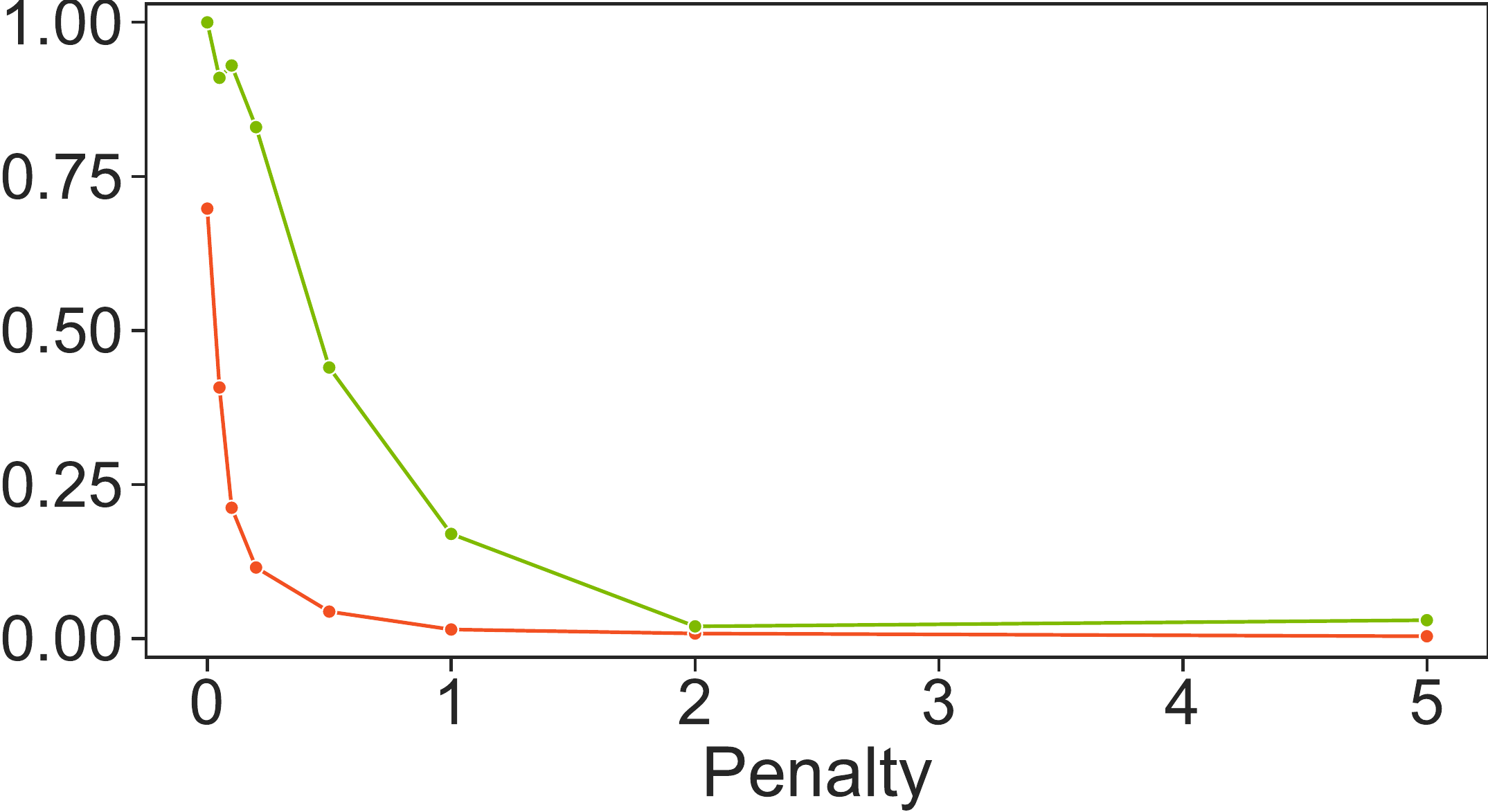}
         \end{subfigure}
         \begin{subfigure}[t]{0.24\textwidth}
             \centering
             \vspace{0.01in}
             \includegraphics[width=\textwidth]{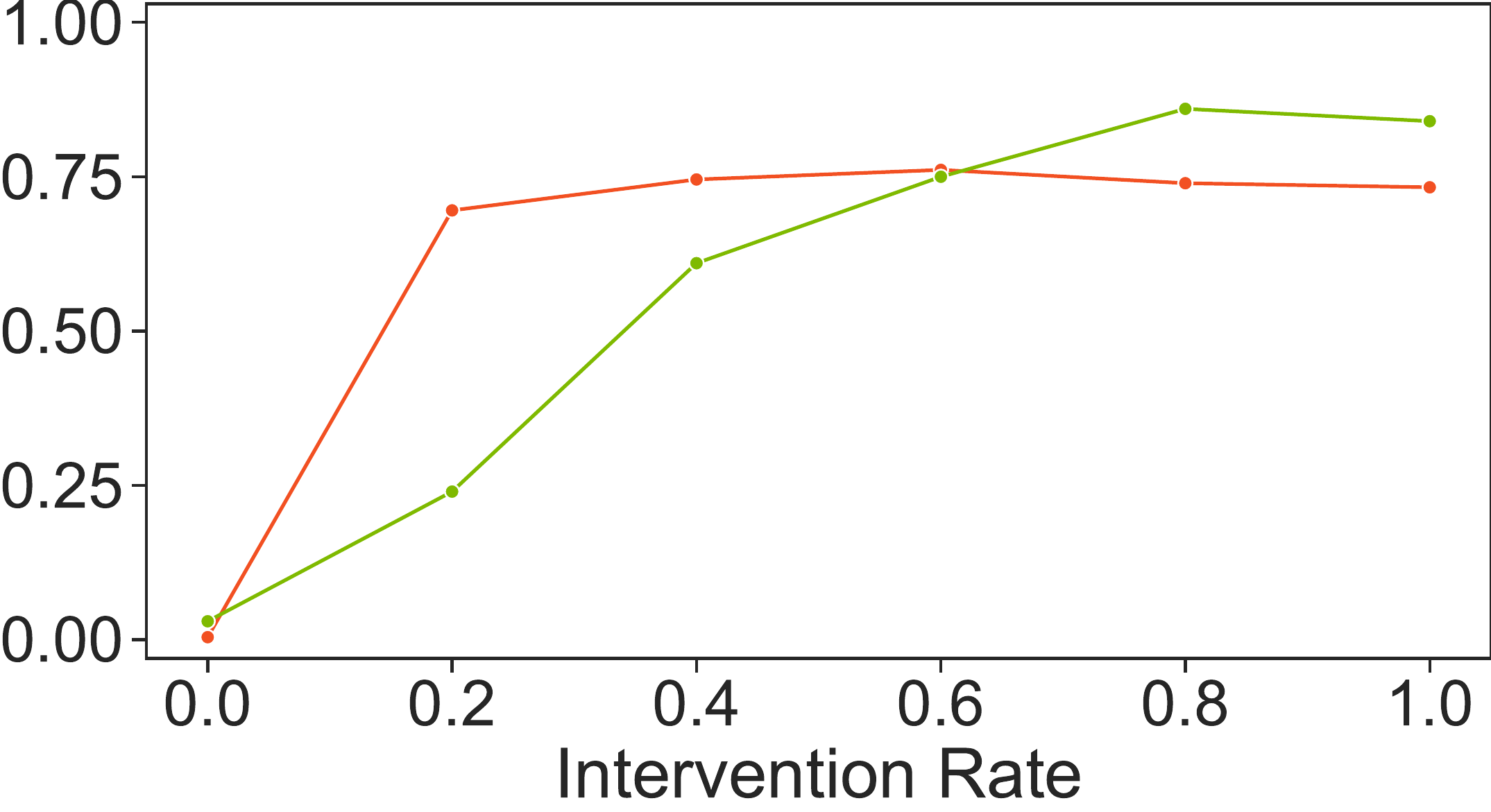}
         \end{subfigure}
     \end{subfigure}
     \caption{The return, intervention rate and success rate of the noisy pilot assisted by the copilot trained with different parameters for different methods on Super Mario.}
     \label{fig:mario noisy}
\end{figure*}

For Super Mario, we use the noisy pilot, which corresponds to Figure \ref{fig:mario noisy}.

\clearpage

\subsection{Learning Curve of Different Methods in Lunar Lander}
\begin{figure*}[h!]
     \centering
     \begin{subfigure}[t]{\textwidth}
         \centering
         \begin{subfigure}[t]{0.24\textwidth}
             \centering
             \includegraphics[width=\textwidth]{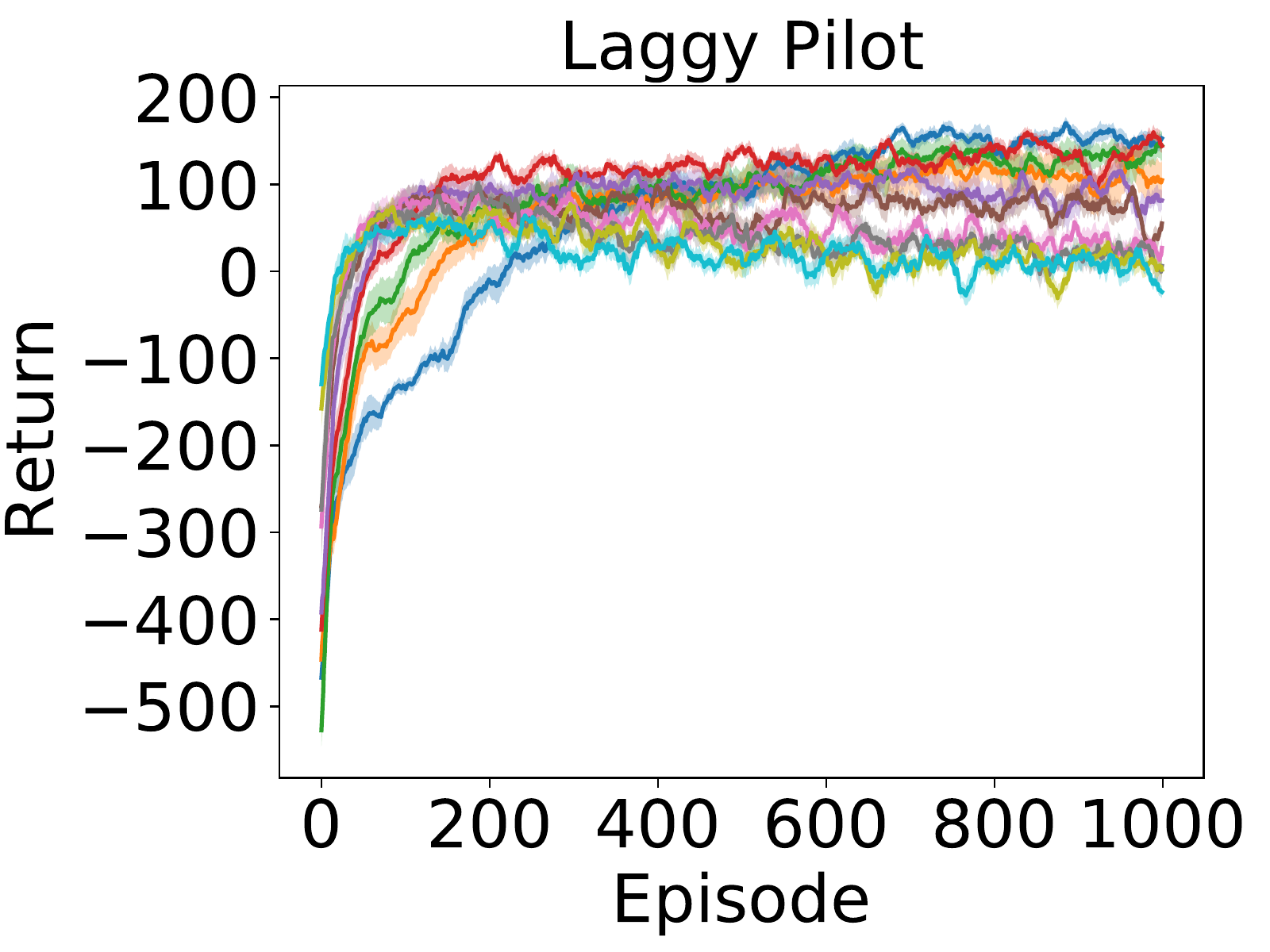}
         \end{subfigure}
         \begin{subfigure}[t]{0.24\textwidth}
             \centering
             \includegraphics[width=\textwidth]{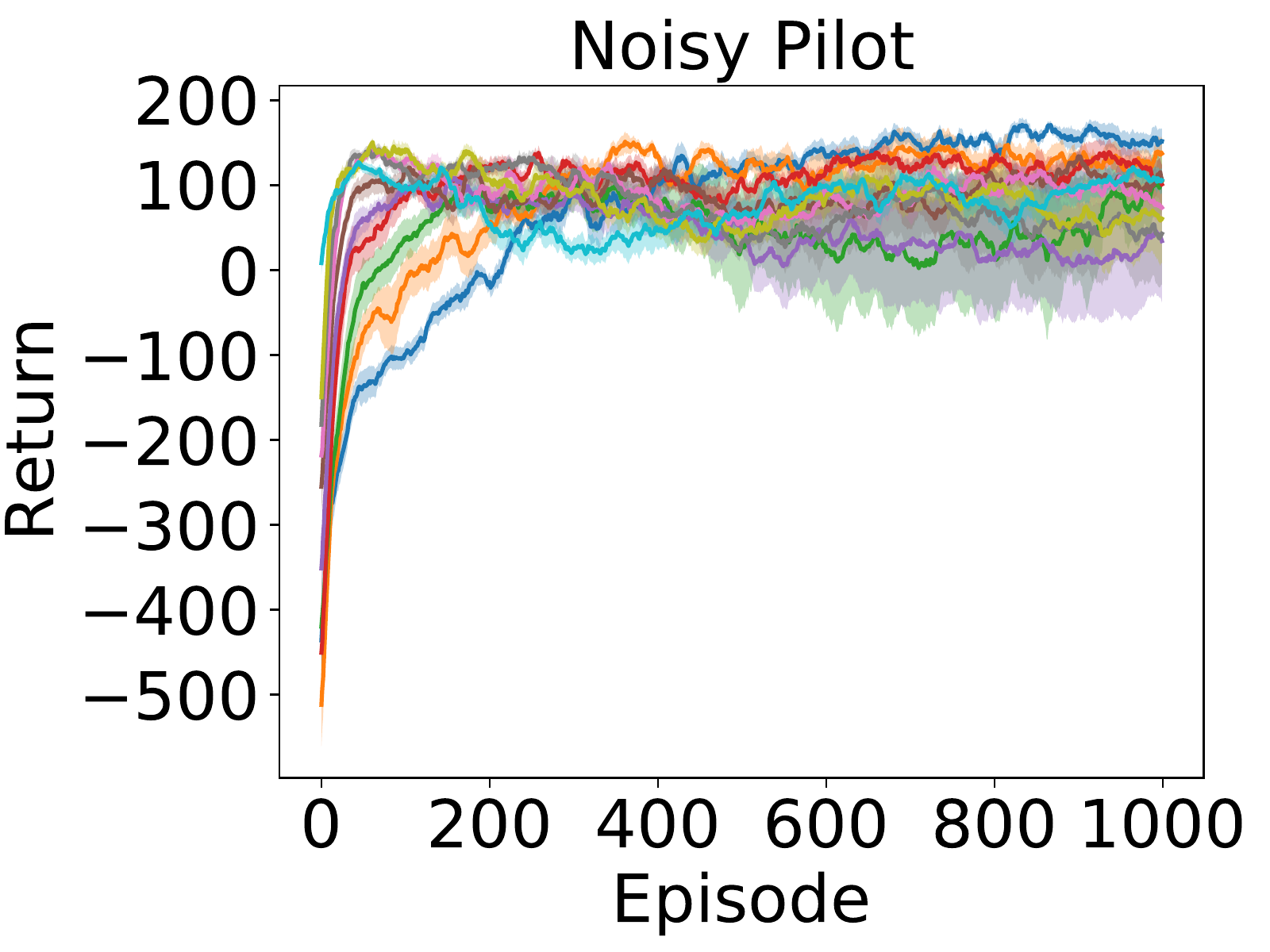}
         \end{subfigure}
         \begin{subfigure}[t]{0.24\textwidth}
             \centering
             \includegraphics[width=\textwidth]{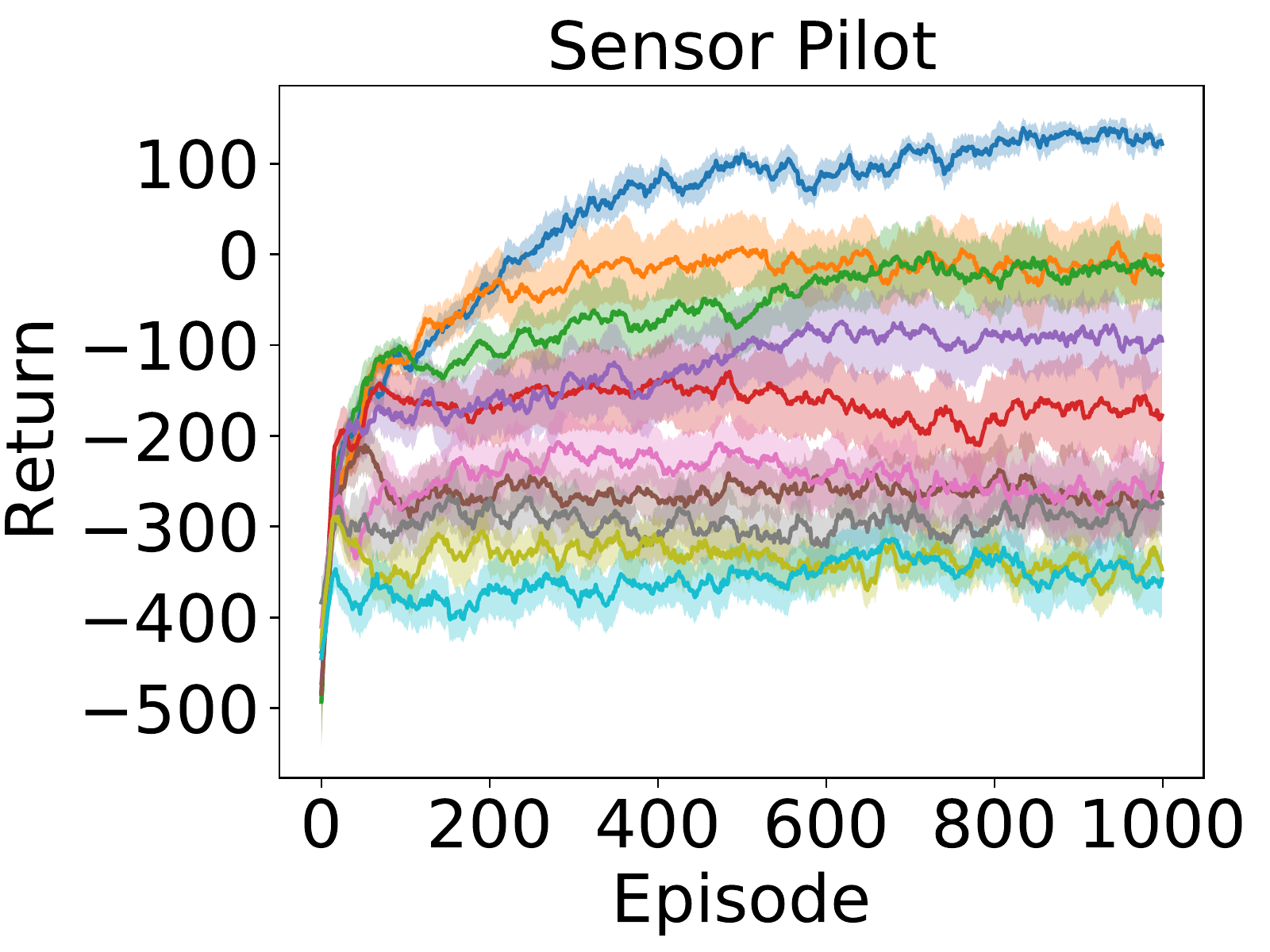}
         \end{subfigure}
         \begin{subfigure}[t]{0.24\textwidth}
             \centering
             \includegraphics[width=\textwidth]{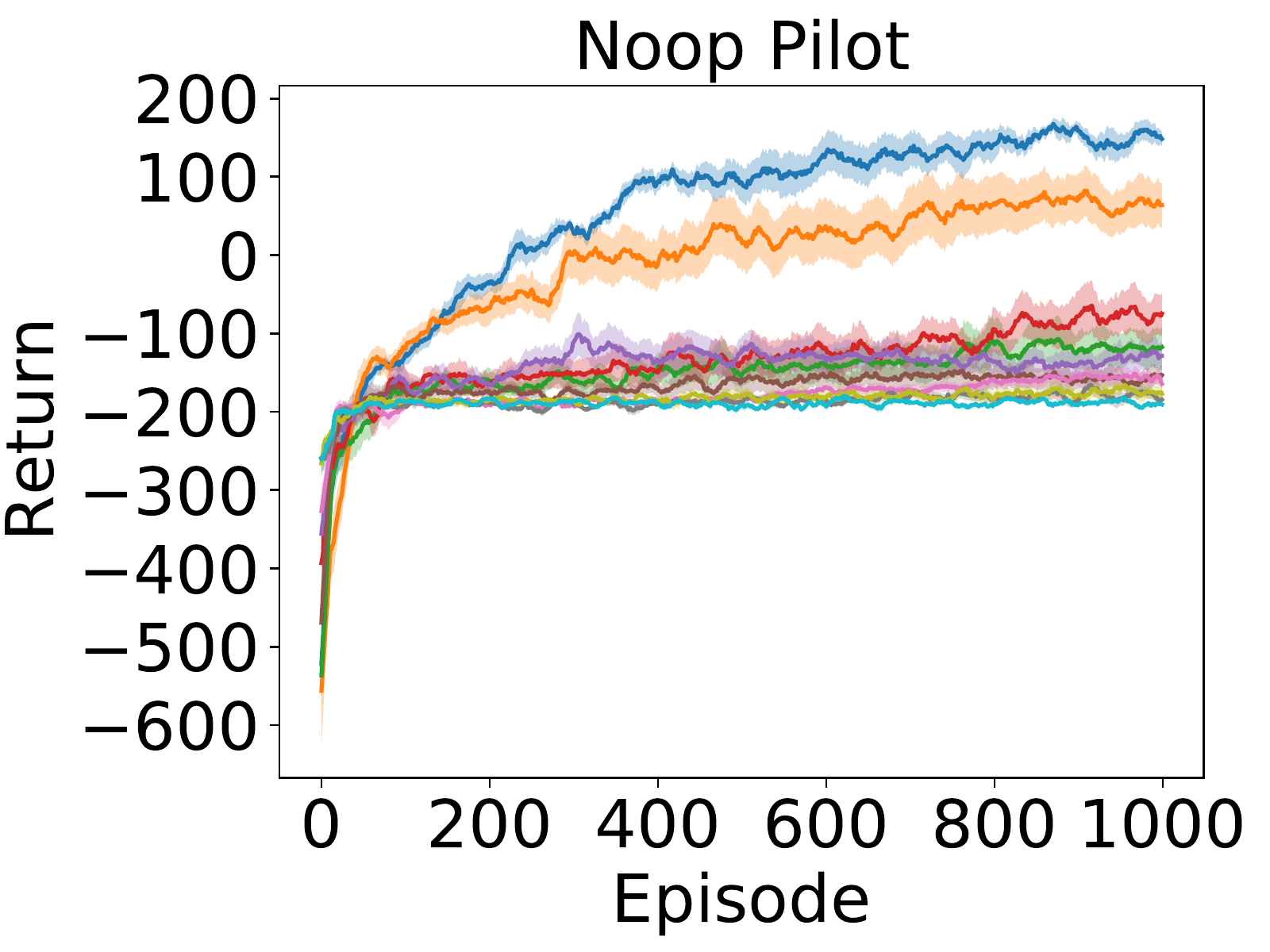}
         \end{subfigure}
         \begin{subfigure}[t]{0.24\textwidth}
             \centering
             \includegraphics[width=\textwidth]{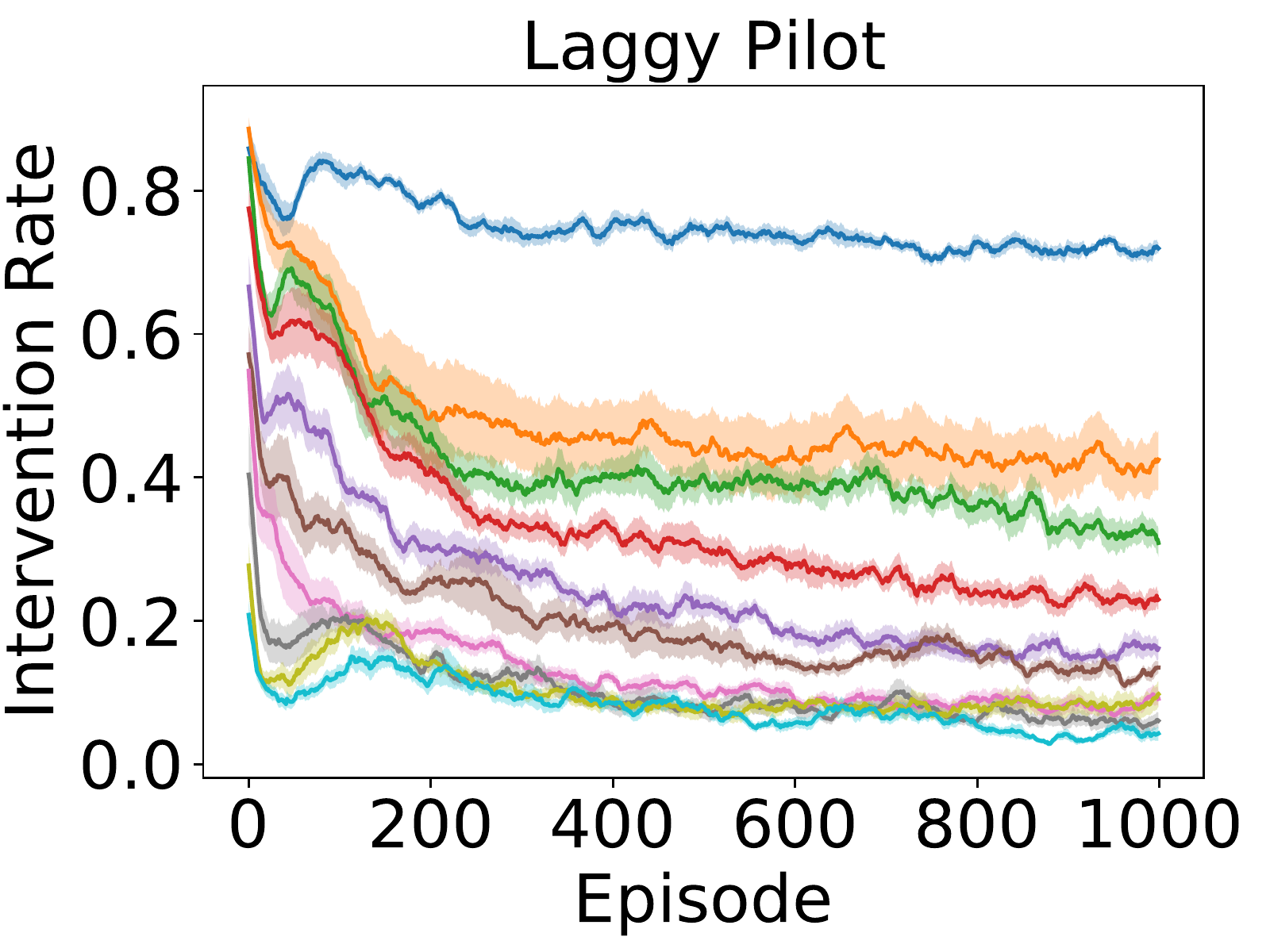}
         \end{subfigure}
         \begin{subfigure}[t]{0.24\textwidth}
             \centering
             \includegraphics[width=\textwidth]{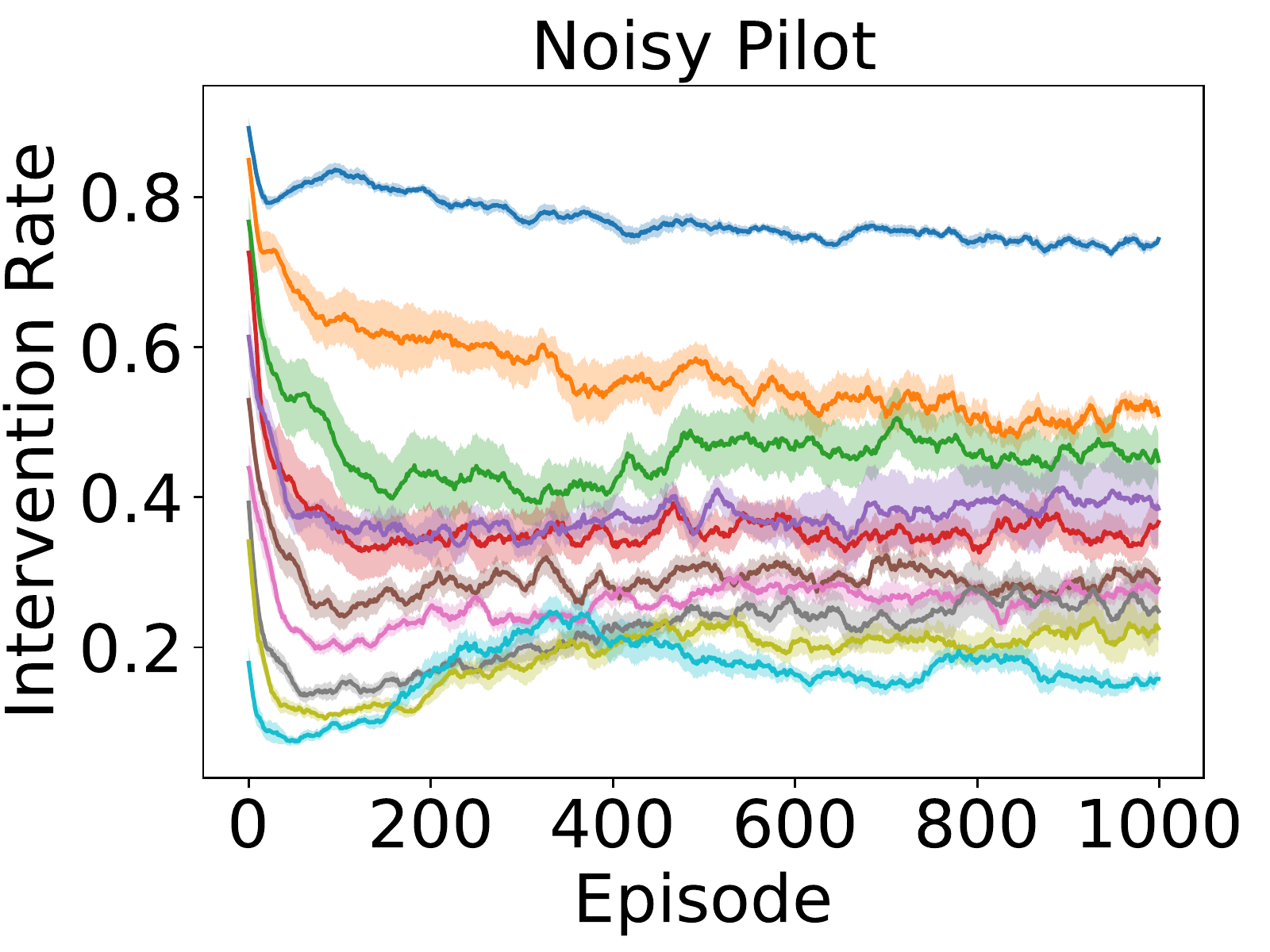}
         \end{subfigure}
         \begin{subfigure}[t]{0.24\textwidth}
             \centering
             \includegraphics[width=\textwidth]{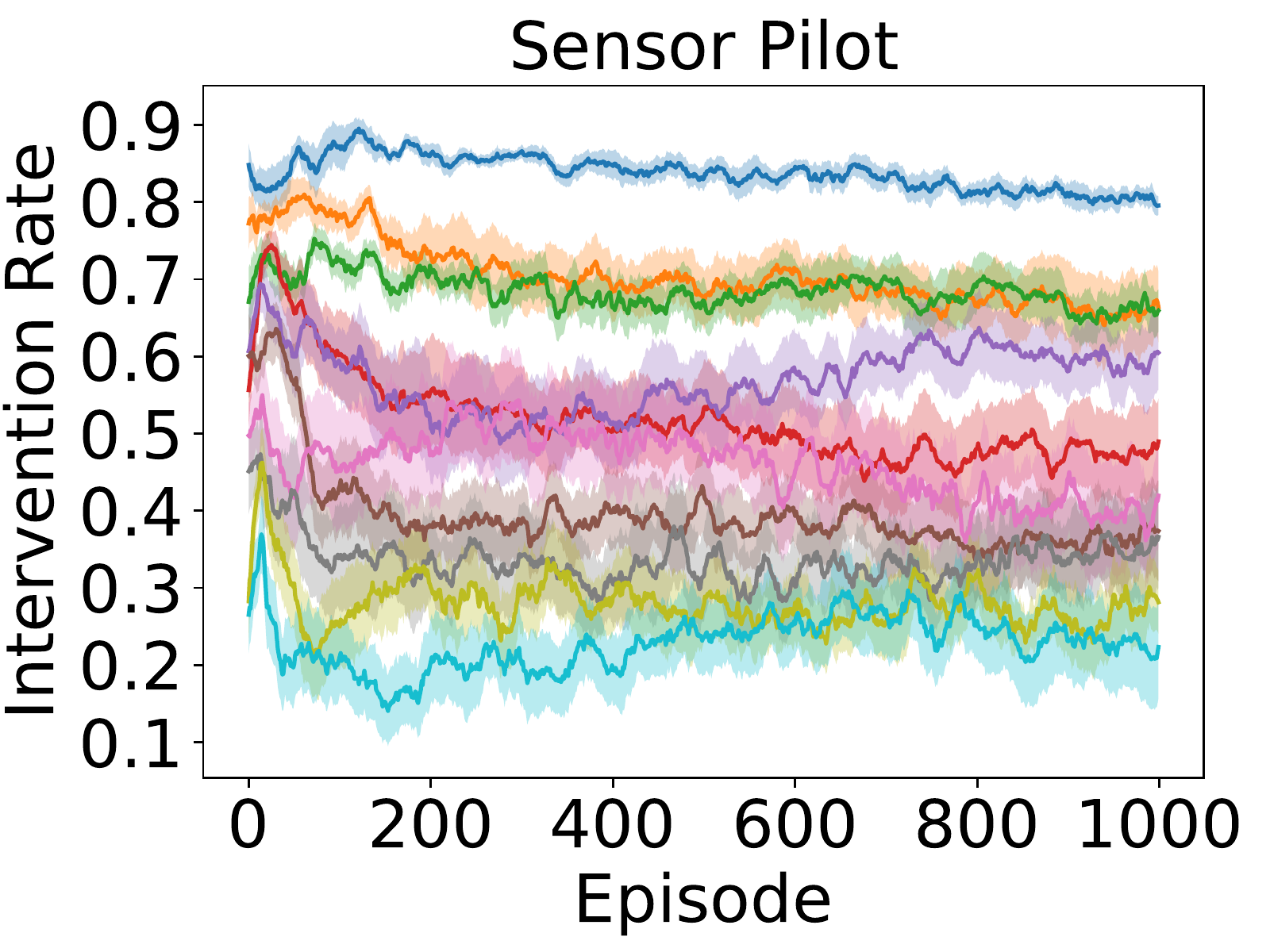}
         \end{subfigure}
         \begin{subfigure}[t]{0.24\textwidth}
             \centering
             \includegraphics[width=\textwidth]{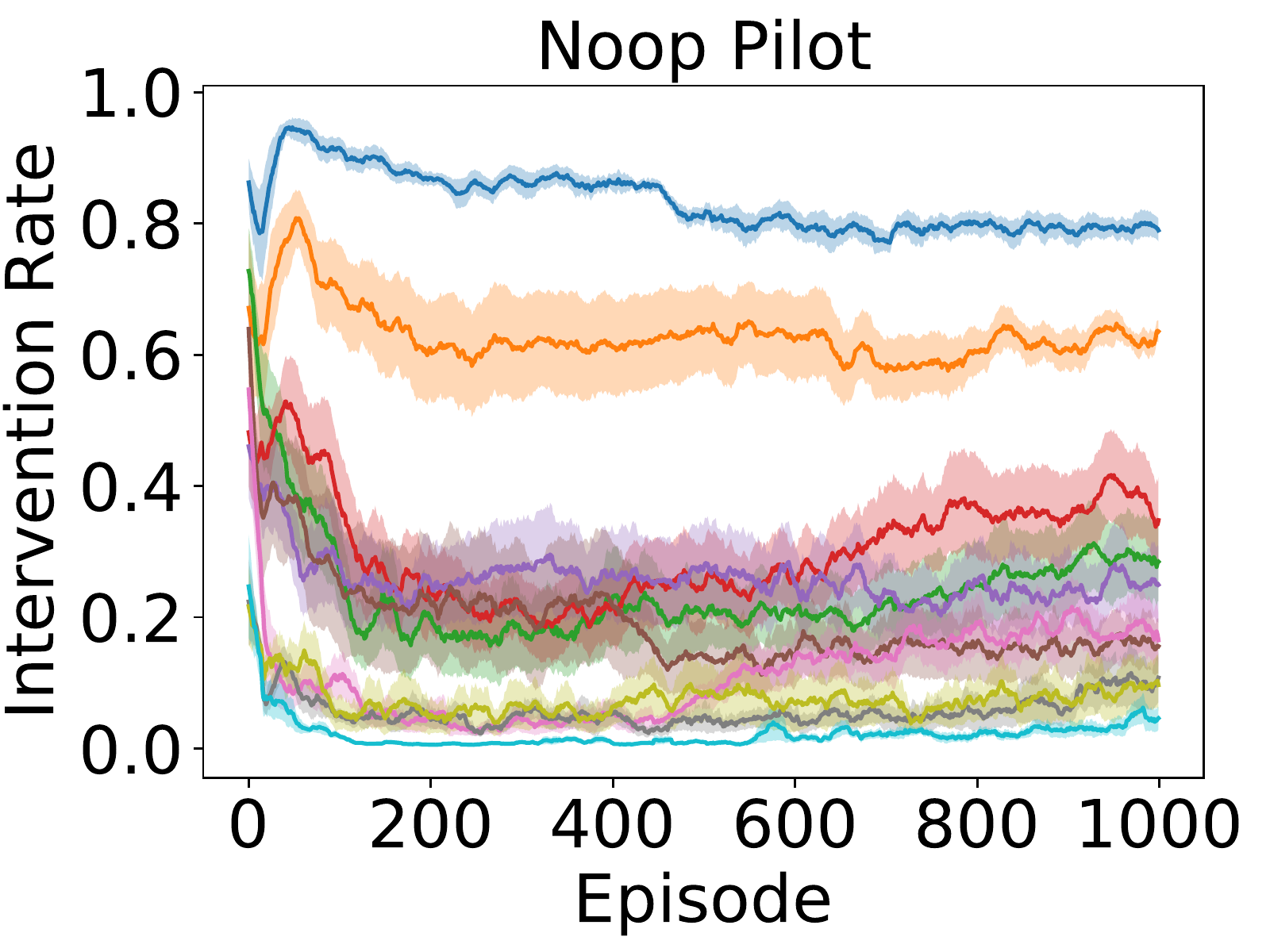}
         \end{subfigure}
         \begin{subfigure}[t]{0.24\textwidth}
             \centering
             \includegraphics[width=\textwidth]{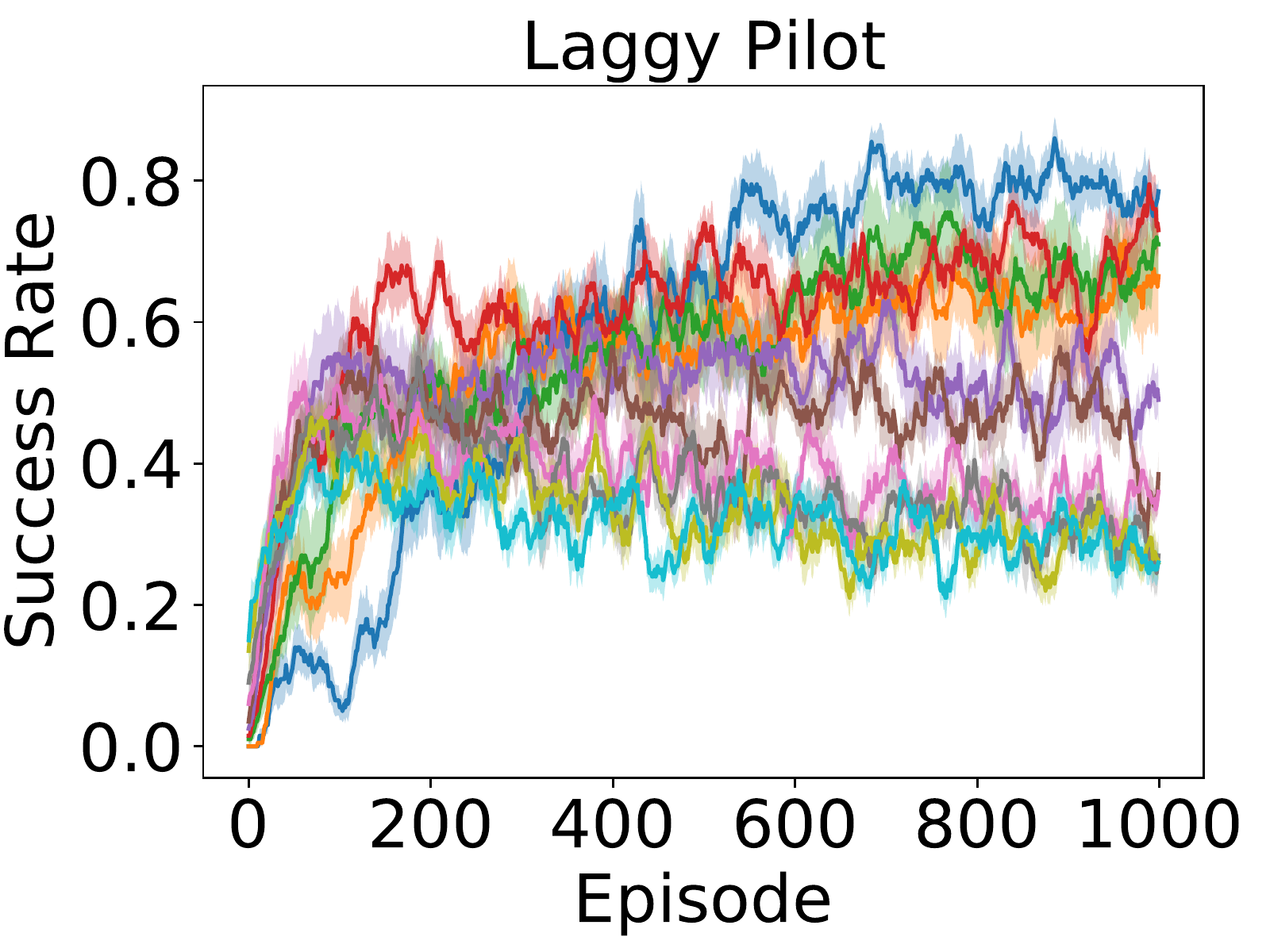}
         \end{subfigure}
         \begin{subfigure}[t]{0.24\textwidth}
             \centering
             \includegraphics[width=\textwidth]{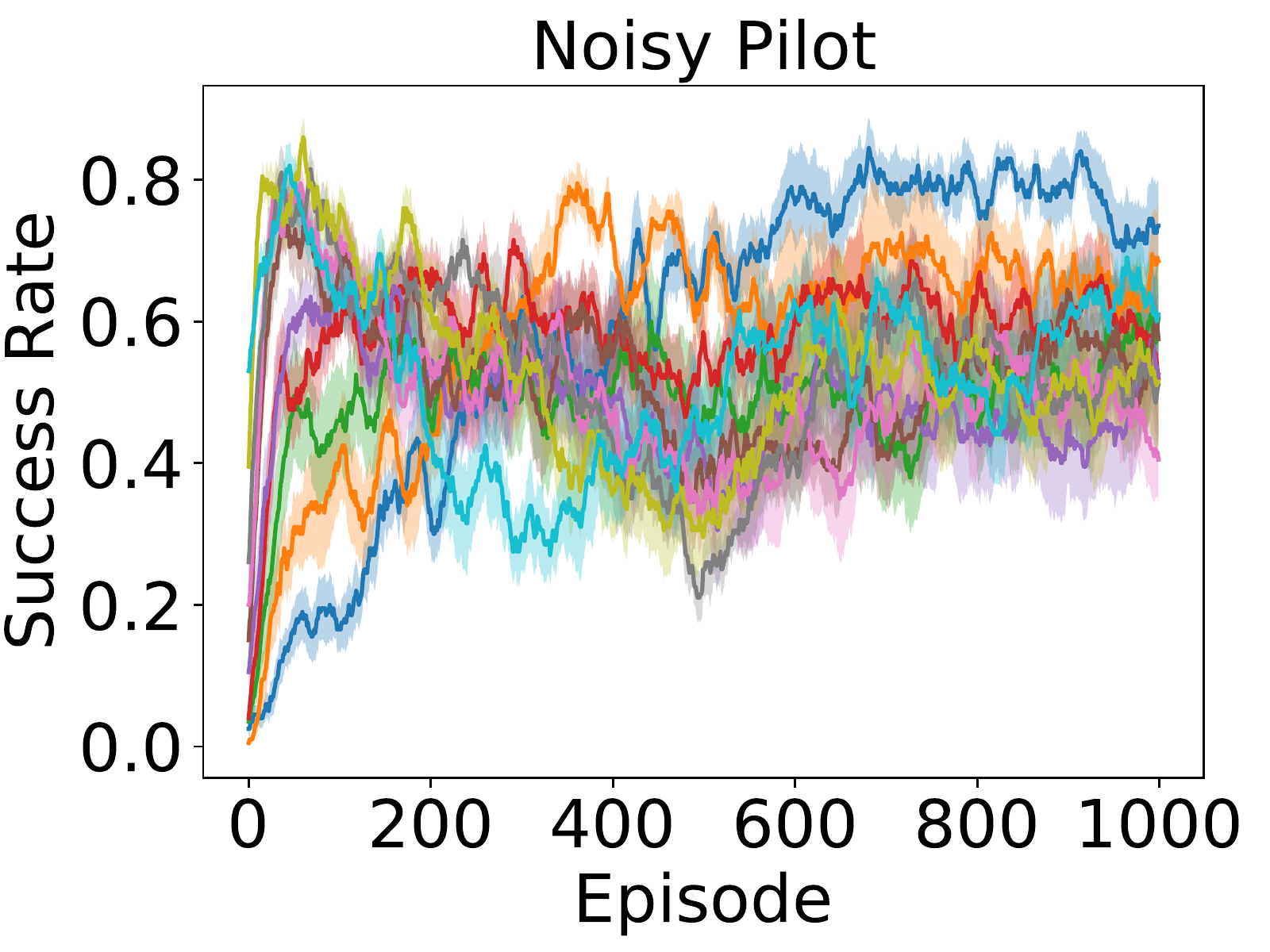}
         \end{subfigure}
         \begin{subfigure}[t]{0.24\textwidth}
             \centering
             \includegraphics[width=\textwidth]{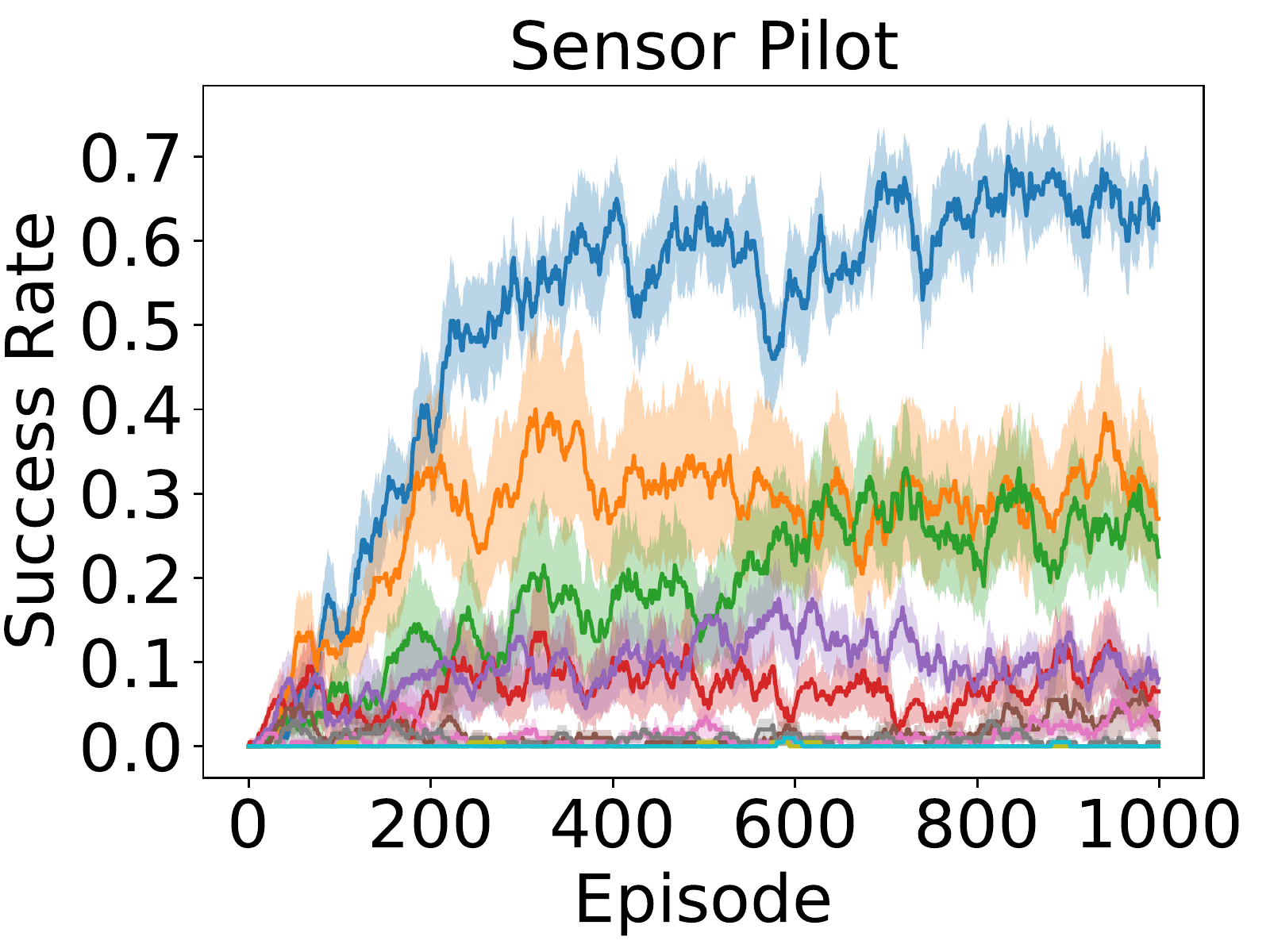}
         \end{subfigure}
         \begin{subfigure}[t]{0.24\textwidth}
             \centering
             \includegraphics[width=\textwidth]{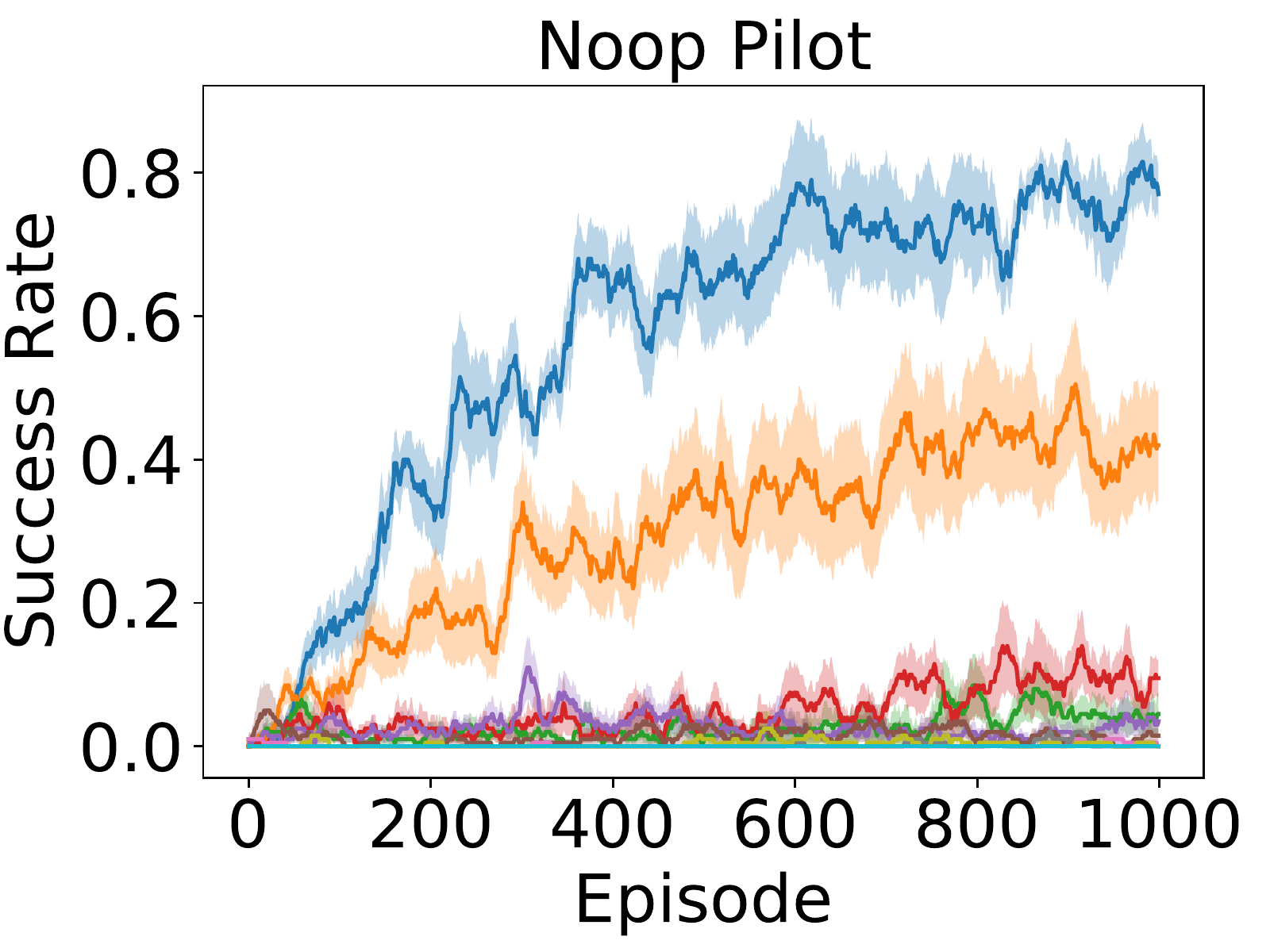}
         \end{subfigure}
         
         \begin{subfigure}[t]{\textwidth}
             \centering
             
             \includegraphics[width=\textwidth]{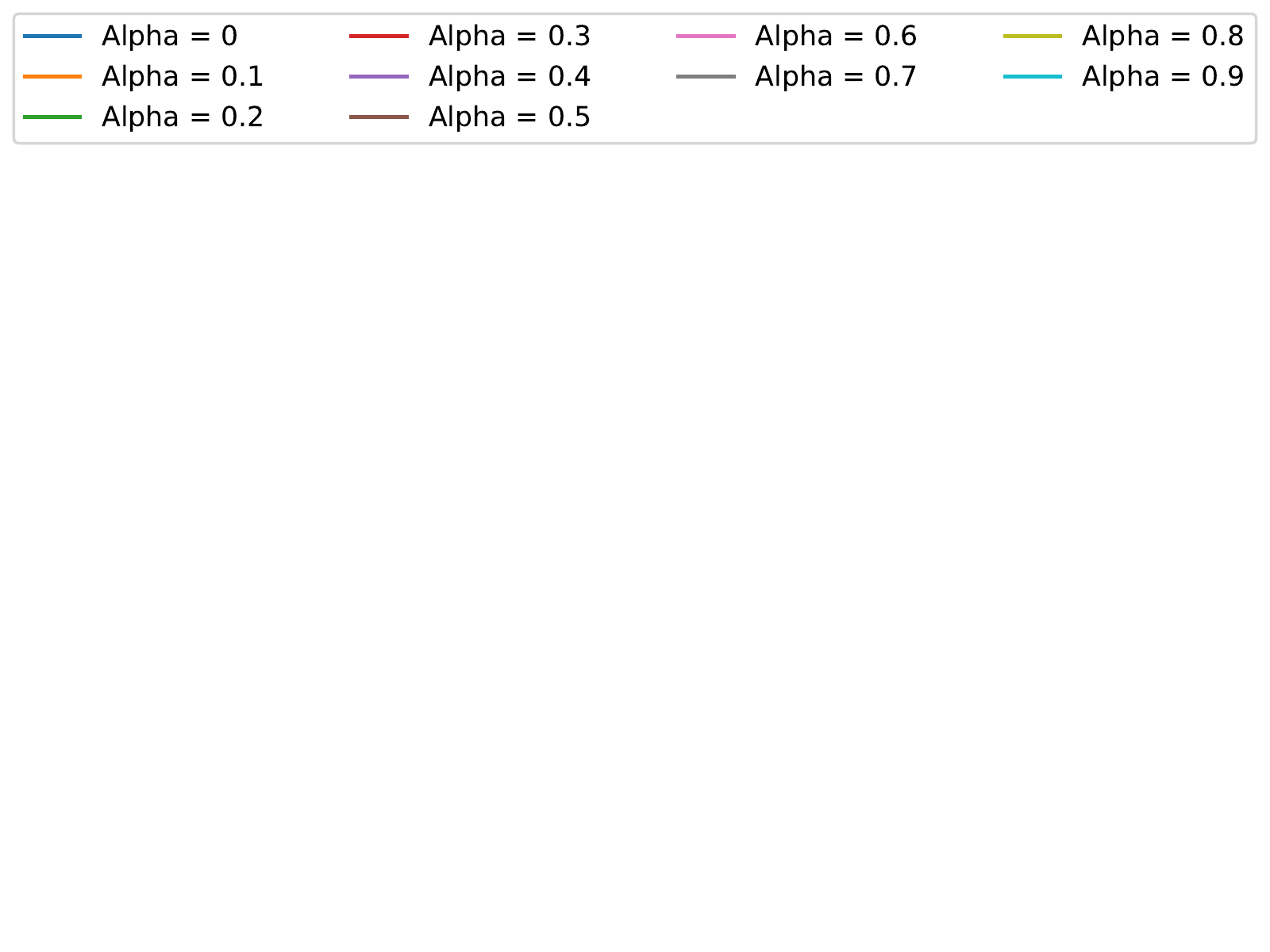}
          \end{subfigure}
     \end{subfigure}
     \vspace{-9cm}
     \caption{The learning curves for return, intervention rate, and success rate of the four types of pilots assisted by \textbf{baseline} copilot during training. Return, intervention rate, and success rate are smoothed using a moving average with a window size of 20 episodes.}
     \label{fig:learning_curve_alpha}
\end{figure*}

\begin{figure*}[h!]
     \centering
     \begin{subfigure}[t]{\textwidth}
         \centering
         \begin{subfigure}[t]{0.24\textwidth}
             \centering
             \includegraphics[width=\textwidth]{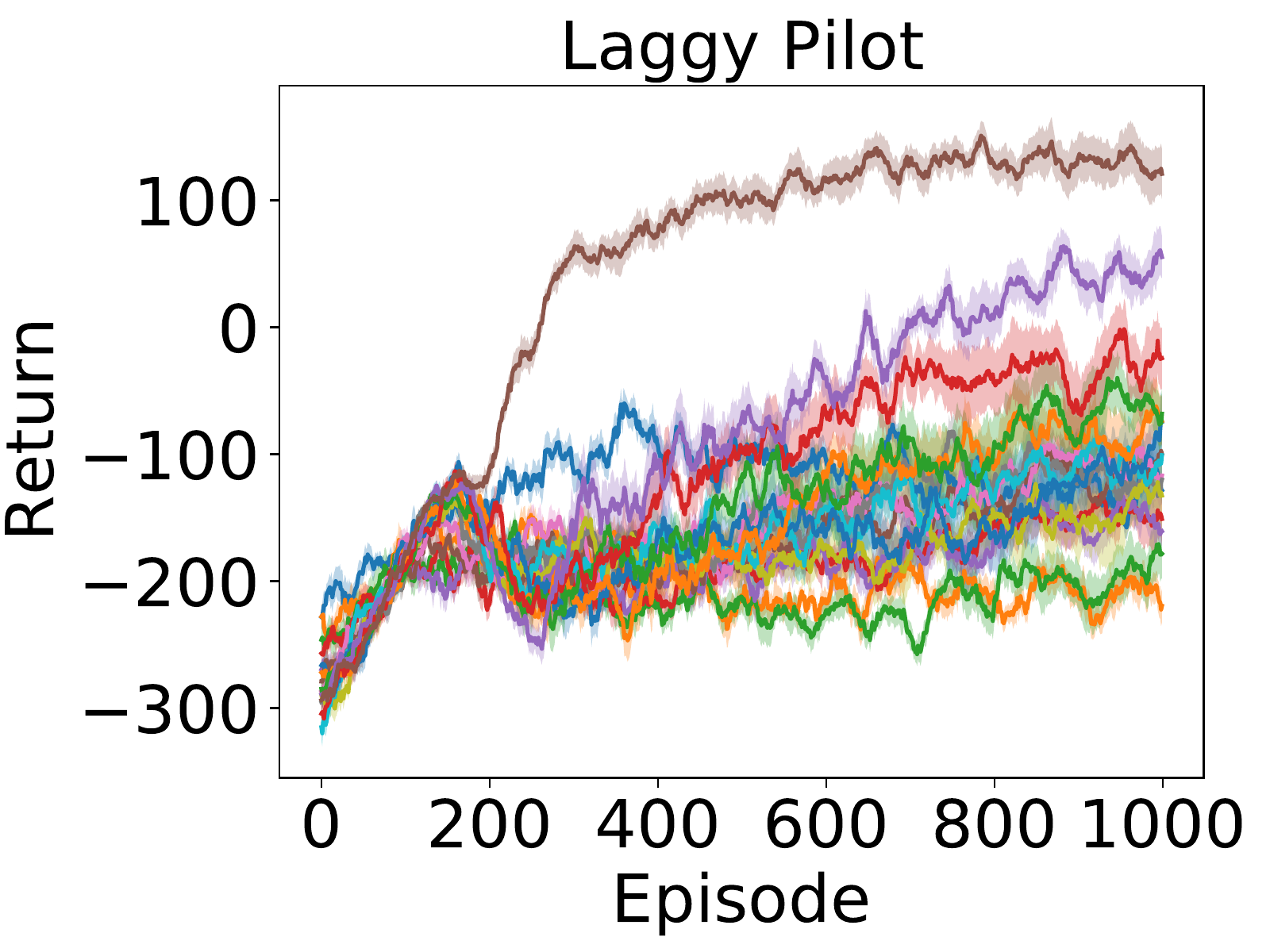}
         \end{subfigure}
         \begin{subfigure}[t]{0.24\textwidth}
             \centering
             \includegraphics[width=\textwidth]{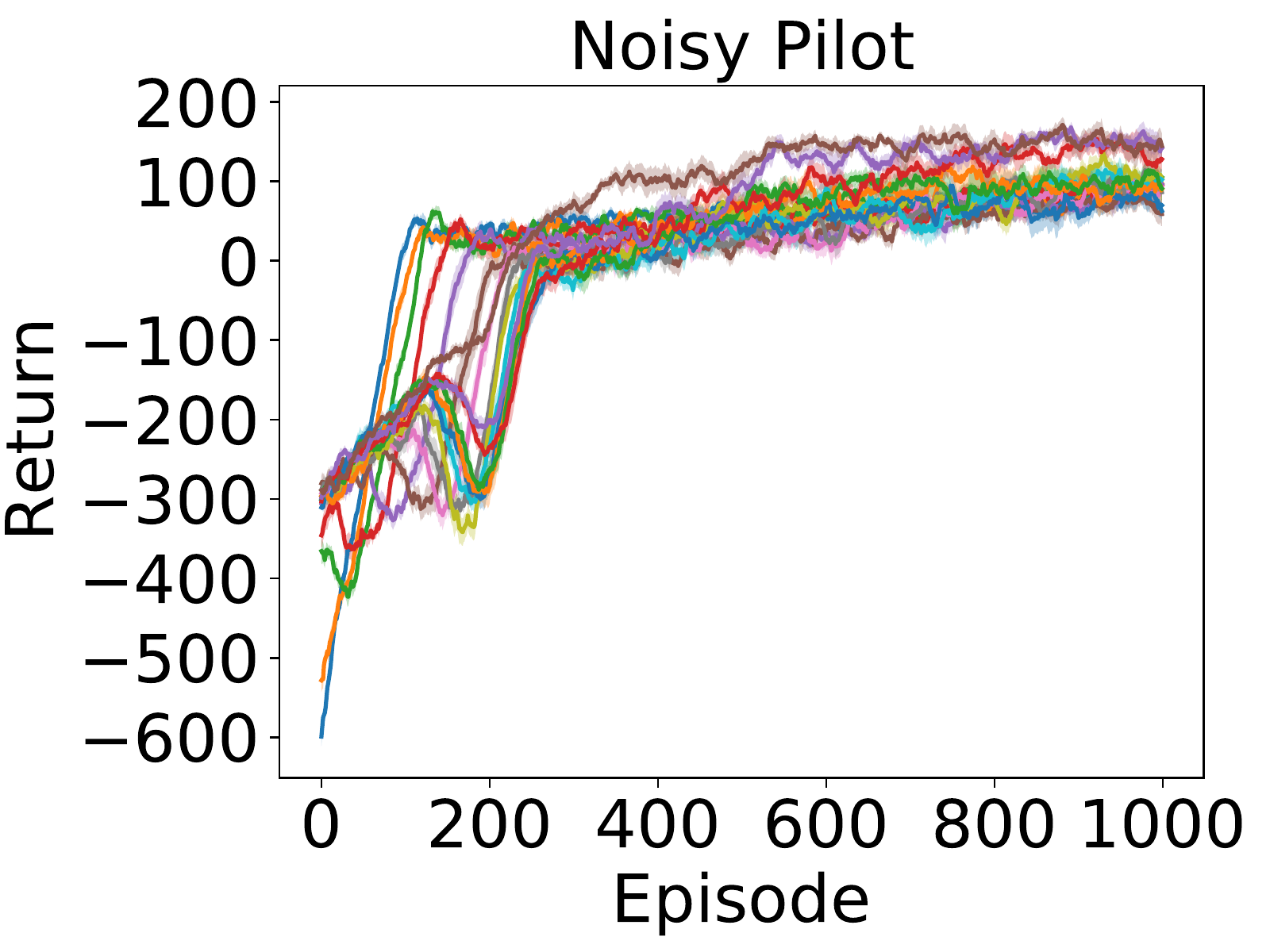}
         \end{subfigure}
         \begin{subfigure}[t]{0.24\textwidth}
             \centering
             \includegraphics[width=\textwidth]{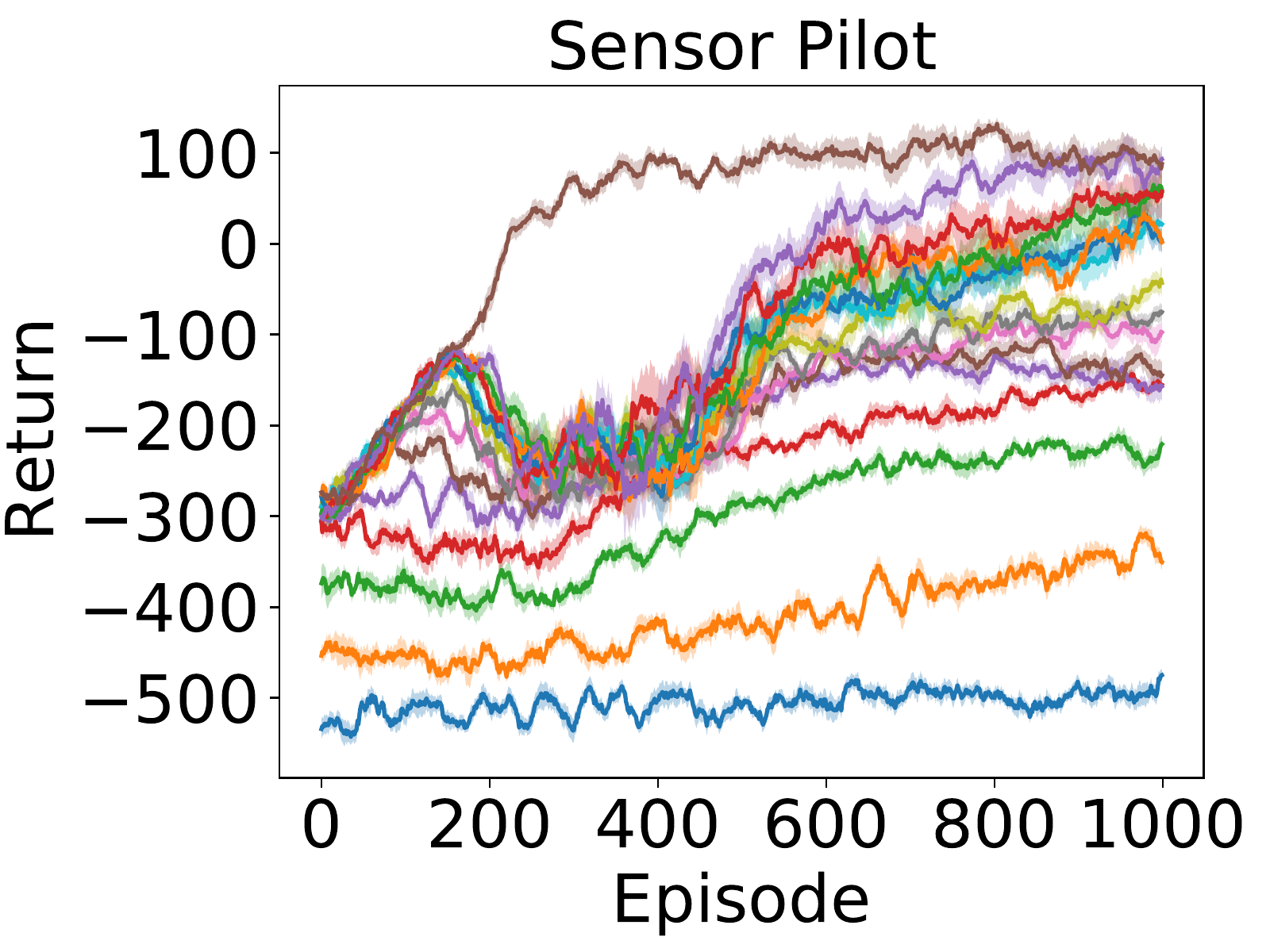}
         \end{subfigure}
         \begin{subfigure}[t]{0.24\textwidth}
             \centering
             \includegraphics[width=\textwidth]{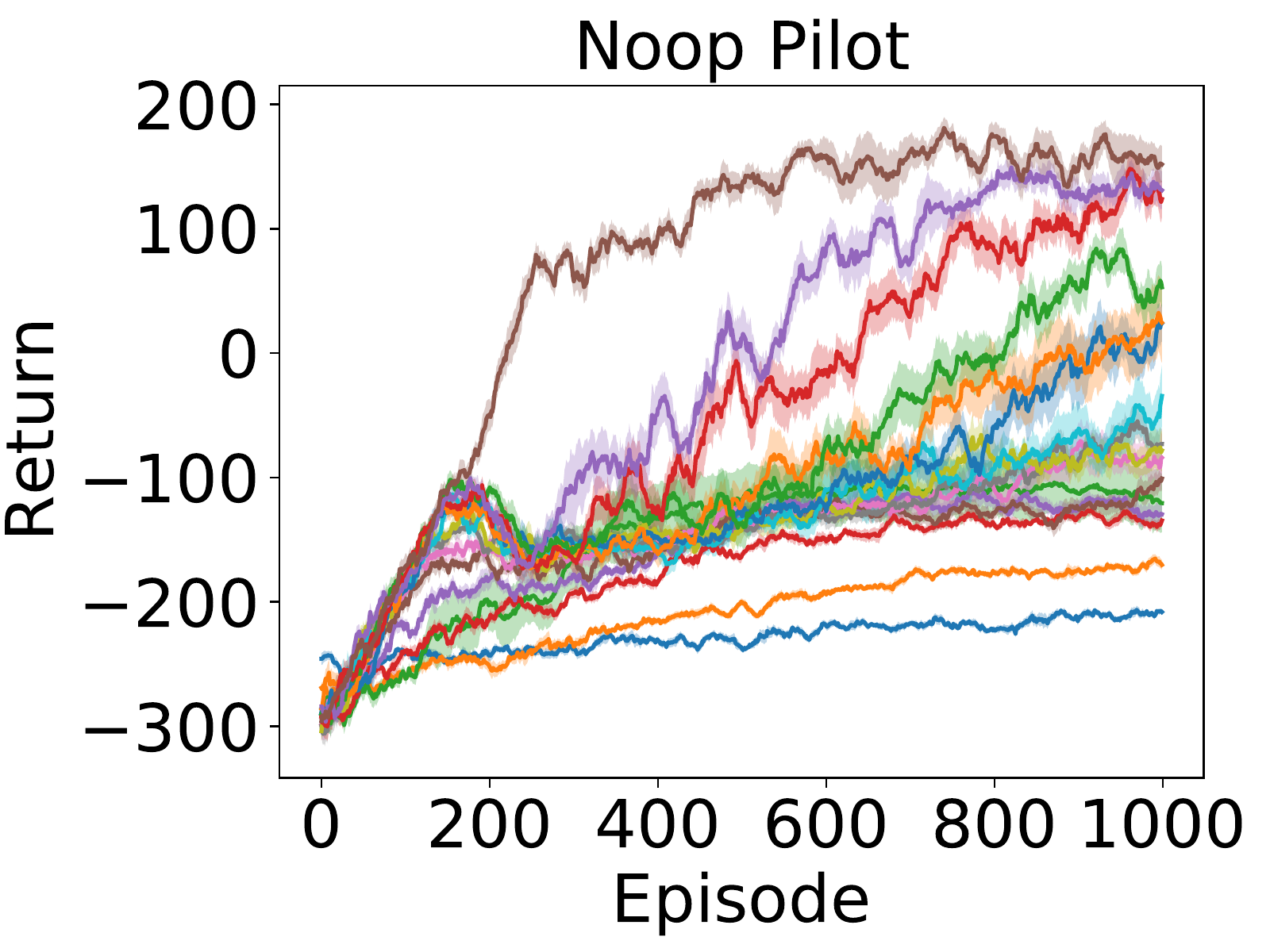}
         \end{subfigure}
         \begin{subfigure}[t]{0.24\textwidth}
             \centering
             \includegraphics[width=\textwidth]{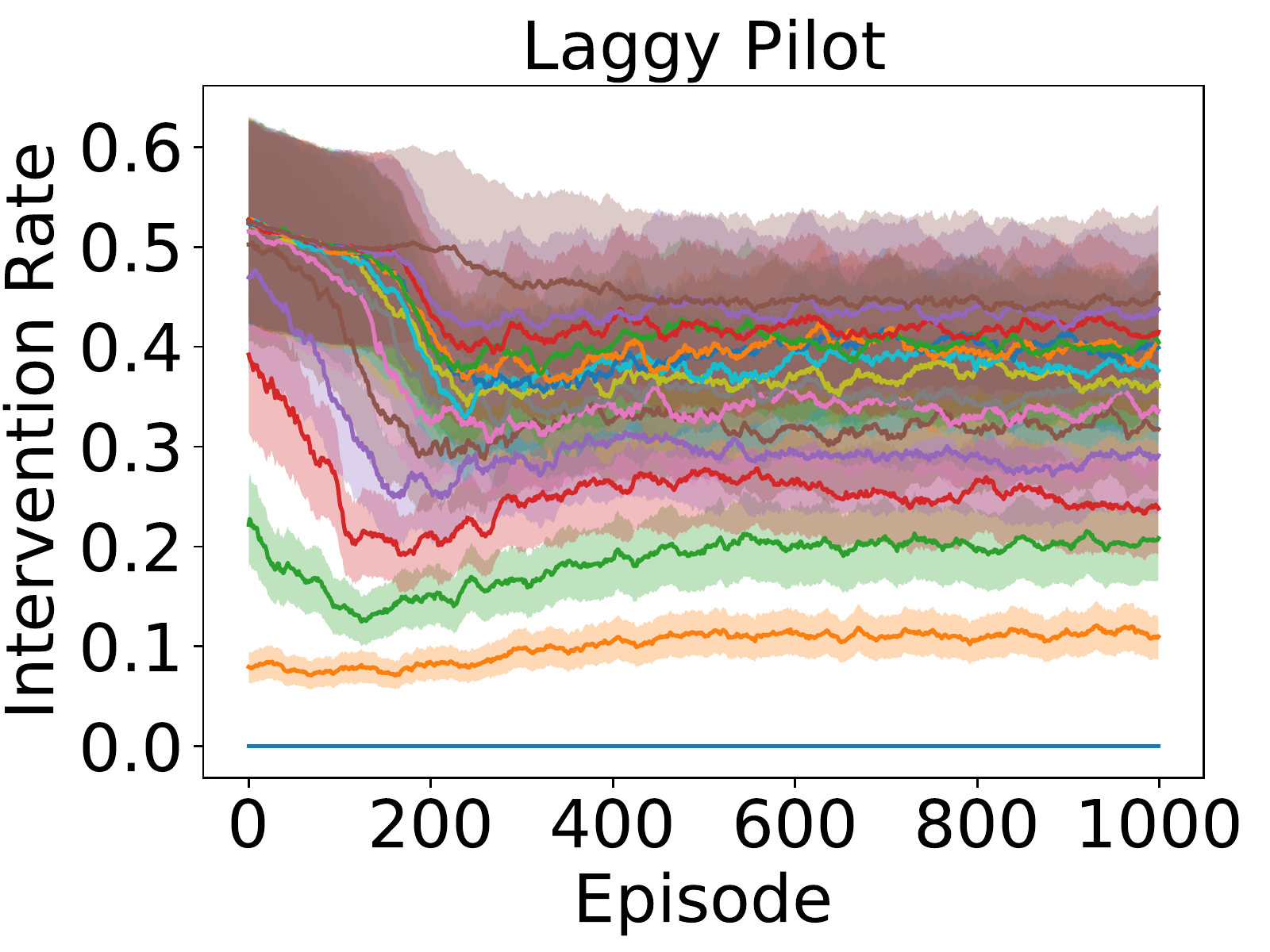}
         \end{subfigure}
         \begin{subfigure}[t]{0.24\textwidth}
             \centering
             \includegraphics[width=\textwidth]{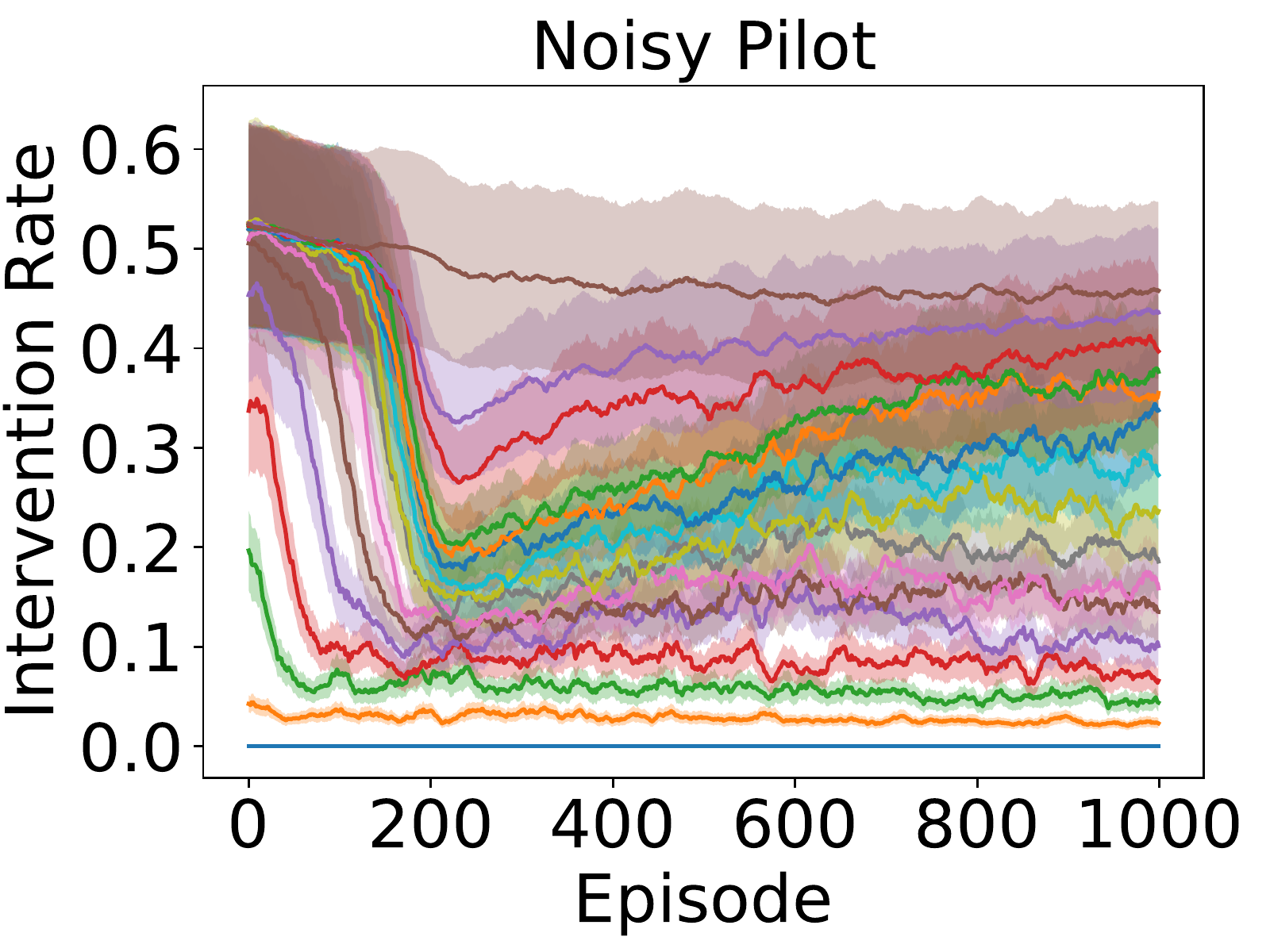}
         \end{subfigure}
         \begin{subfigure}[t]{0.24\textwidth}
             \centering
             \includegraphics[width=\textwidth]{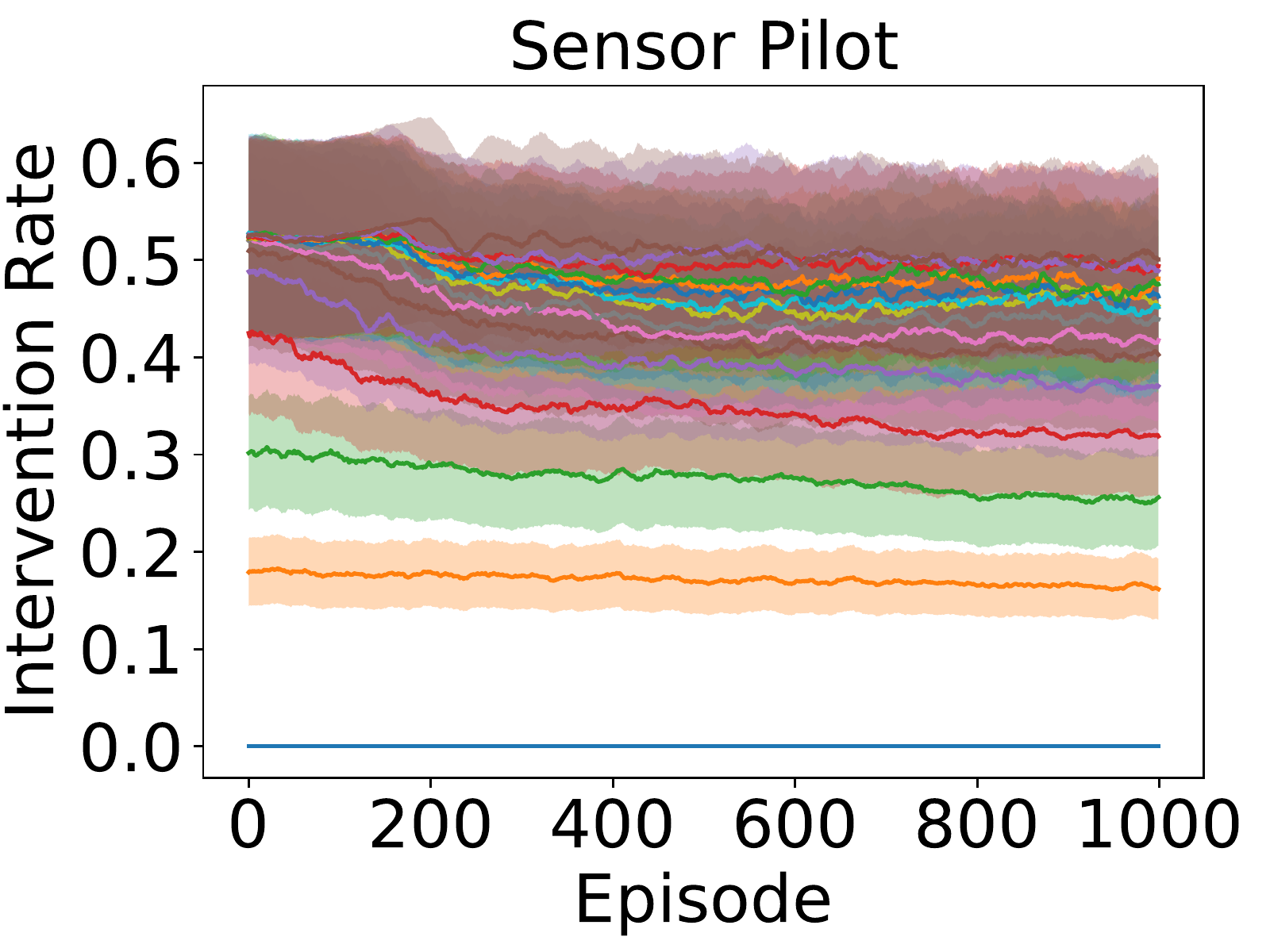}
         \end{subfigure}
         \begin{subfigure}[t]{0.24\textwidth}
             \centering
             \includegraphics[width=\textwidth]{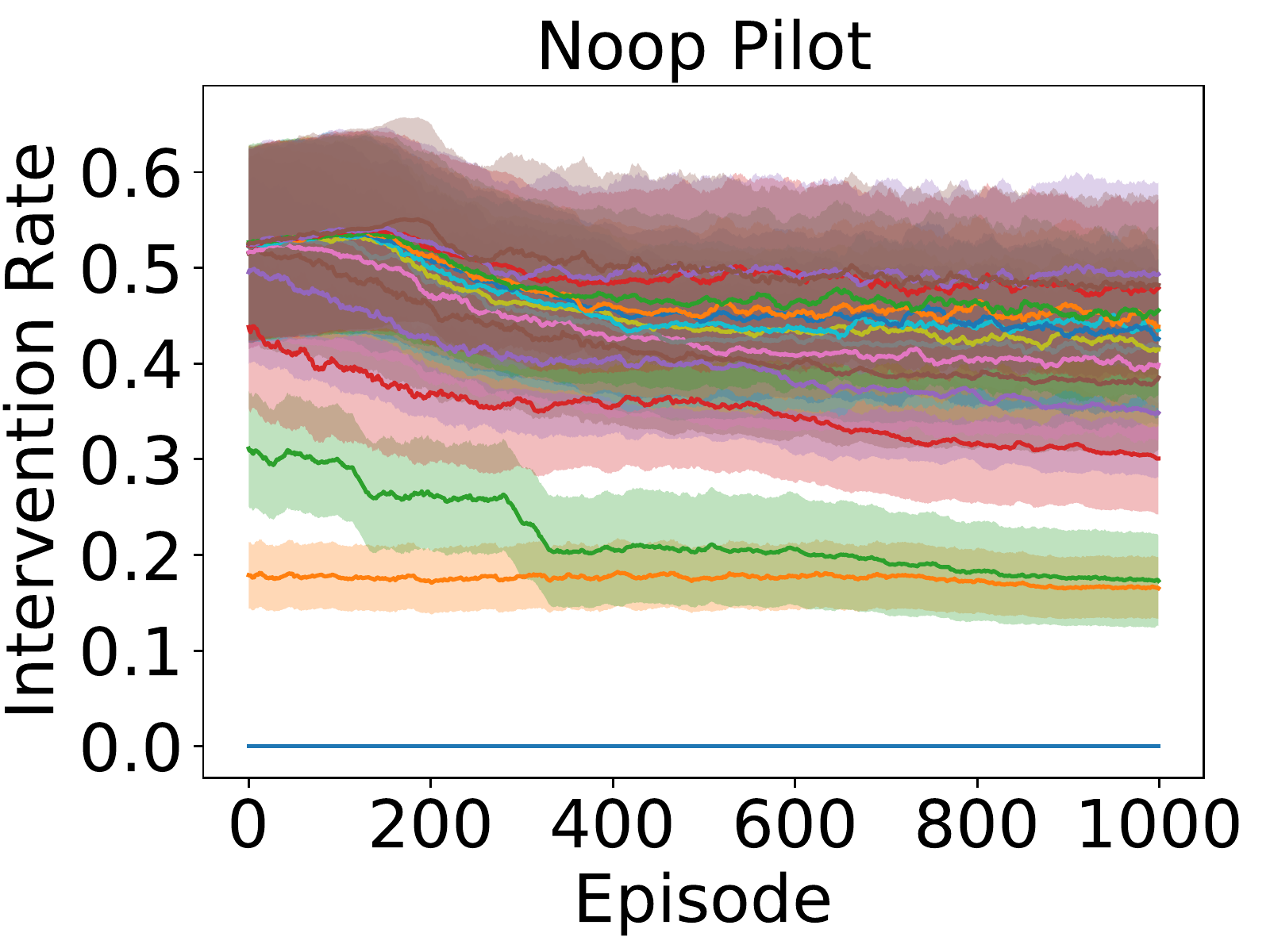}
         \end{subfigure}
         \begin{subfigure}[t]{0.24\textwidth}
             \centering
             \includegraphics[width=\textwidth]{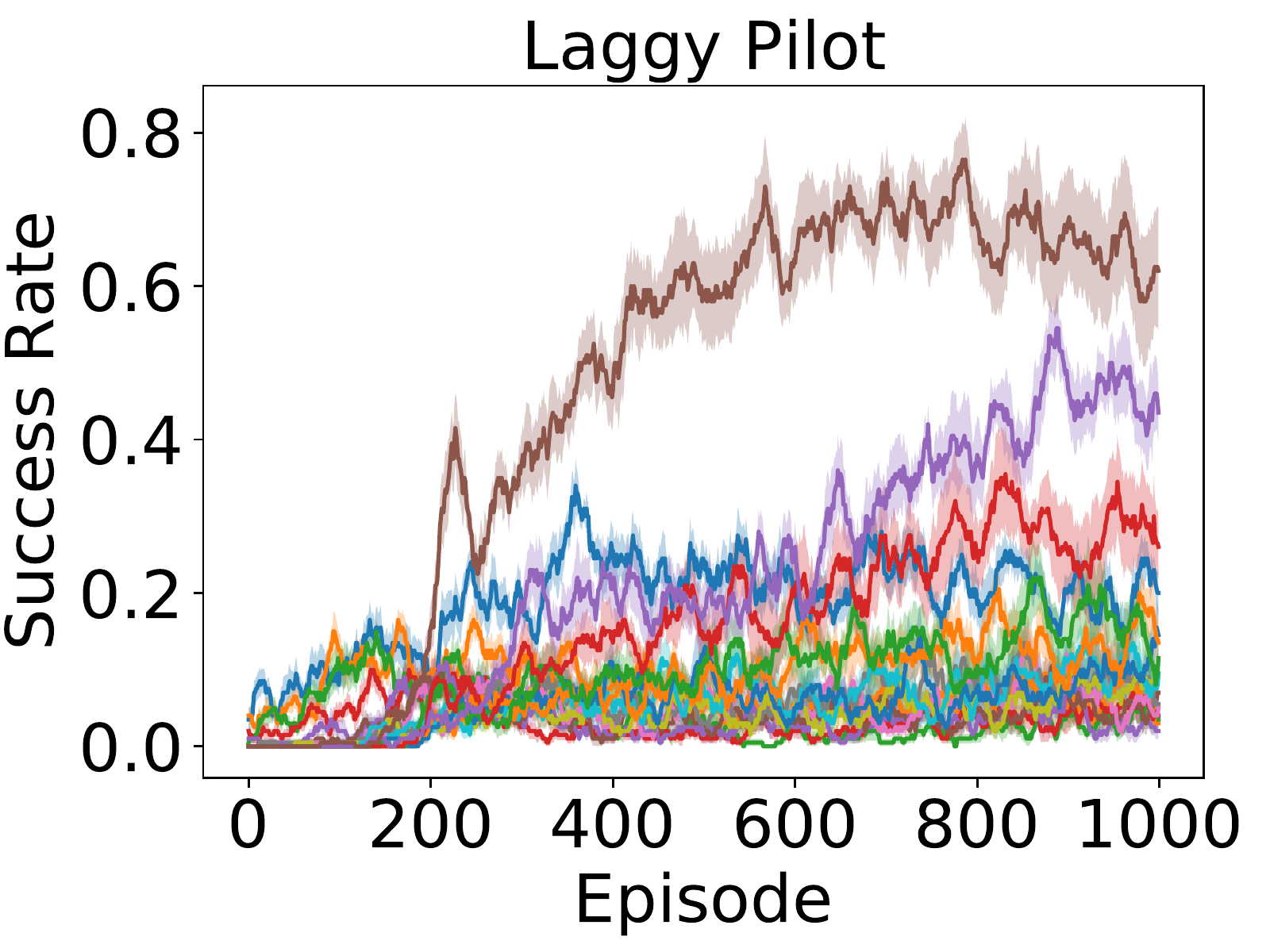}
         \end{subfigure}
         \begin{subfigure}[t]{0.24\textwidth}
             \centering
             \includegraphics[width=\textwidth]{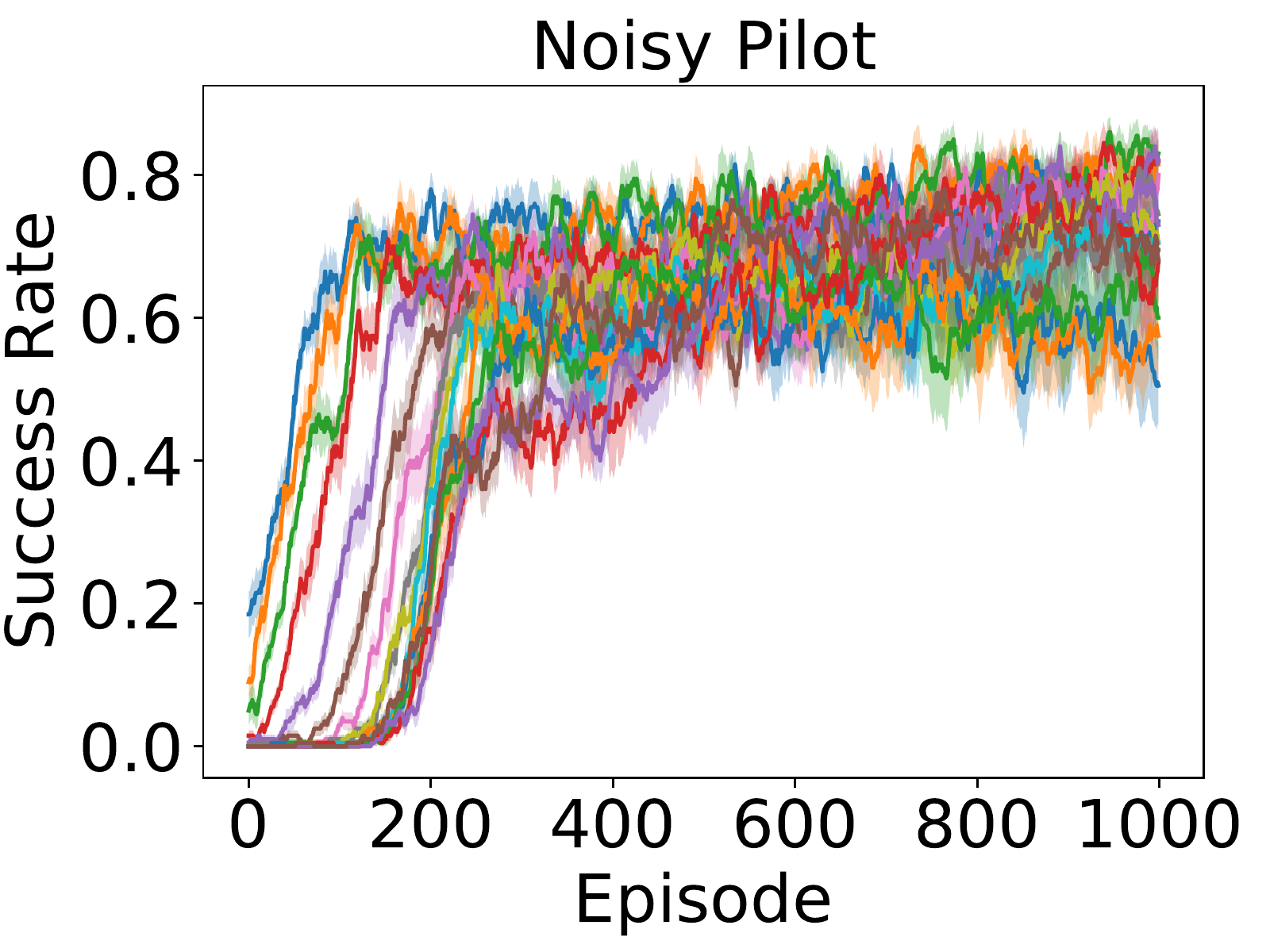}
         \end{subfigure}
         \begin{subfigure}[t]{0.24\textwidth}
             \centering
             \includegraphics[width=\textwidth]{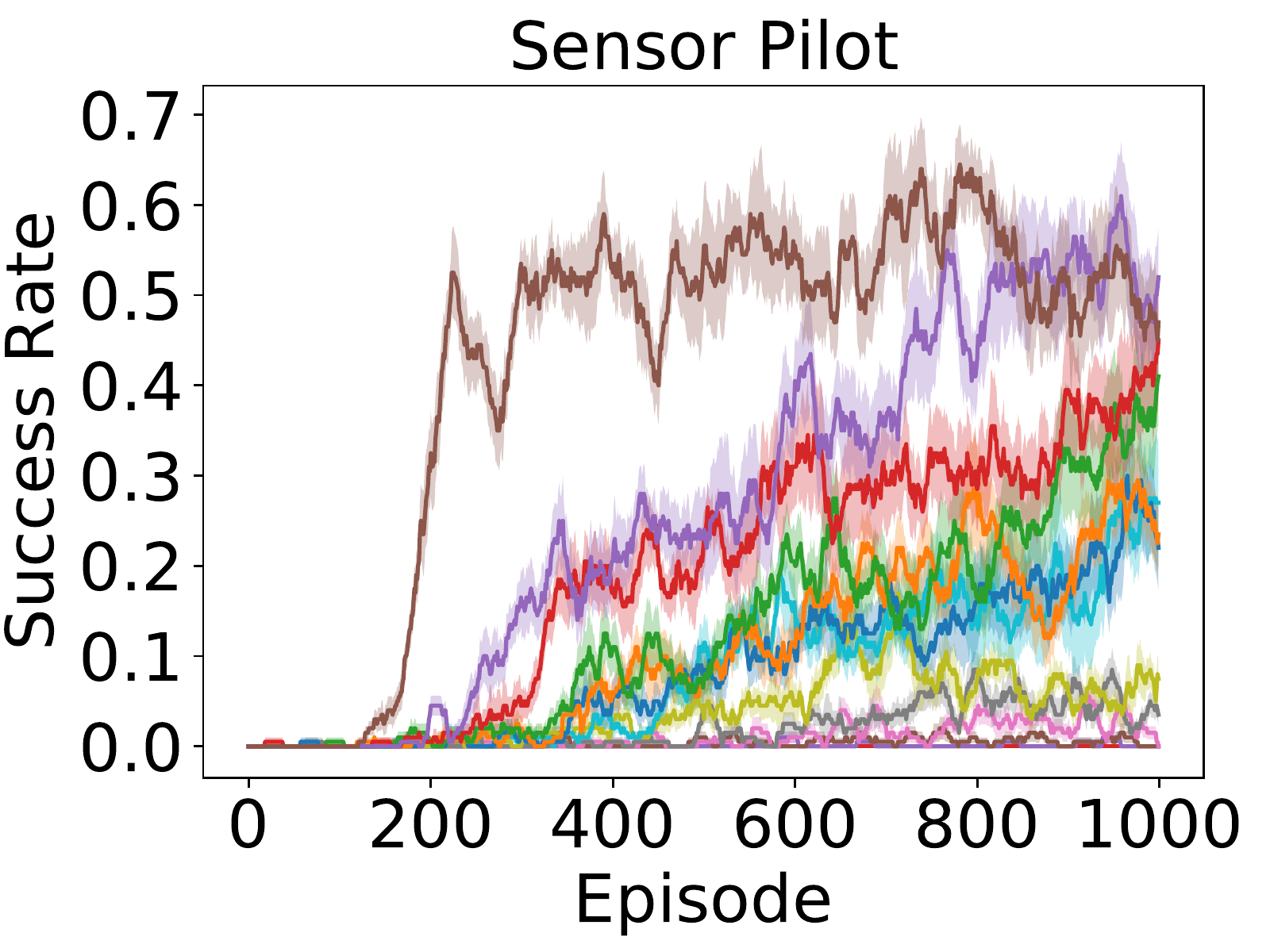}
         \end{subfigure}
         \begin{subfigure}[t]{0.24\textwidth}
             \centering
             \includegraphics[width=\textwidth]{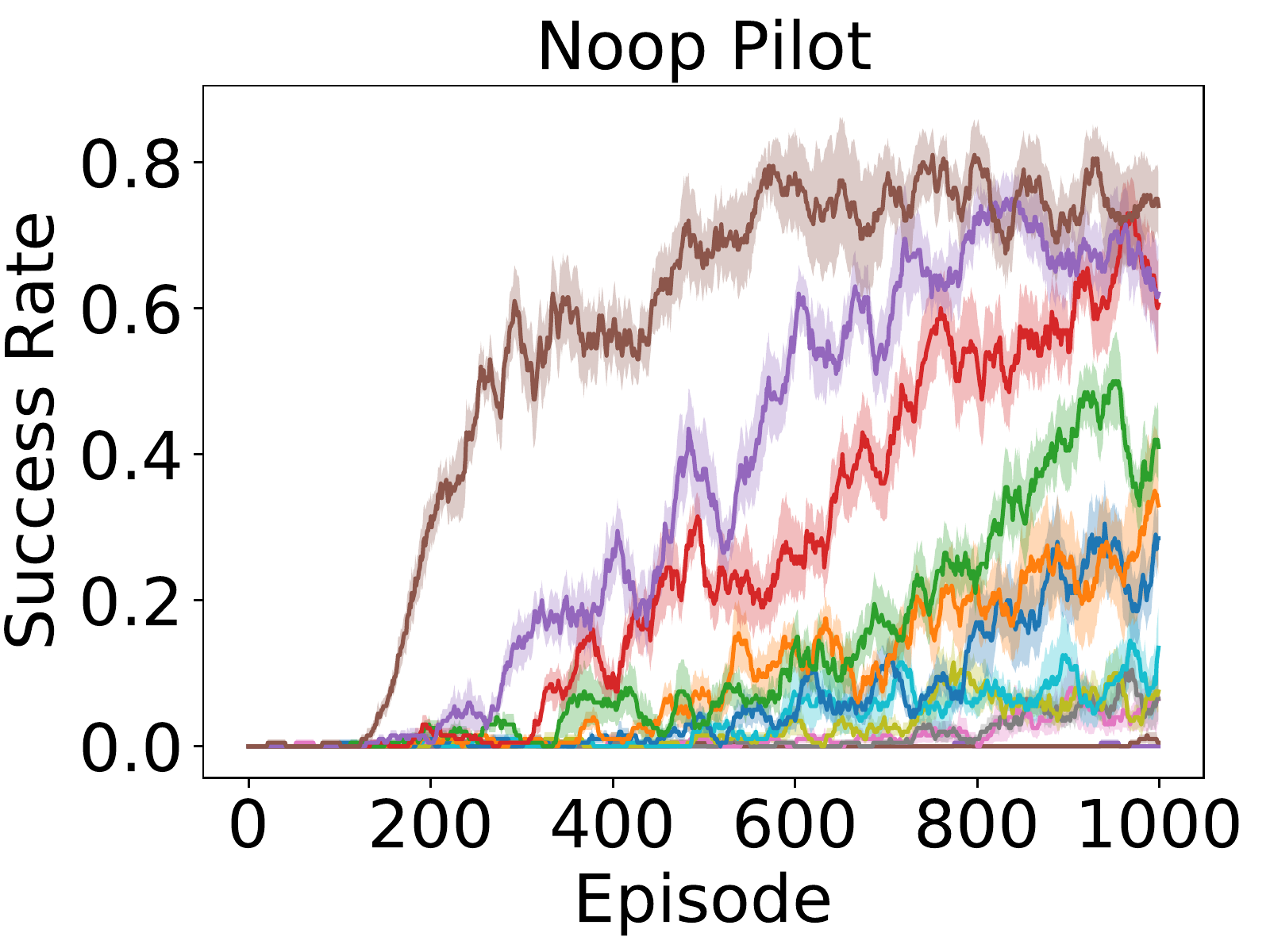}
         \end{subfigure}
         
         \begin{subfigure}[t]{\textwidth}
             \centering
             
             \includegraphics[width=\textwidth]{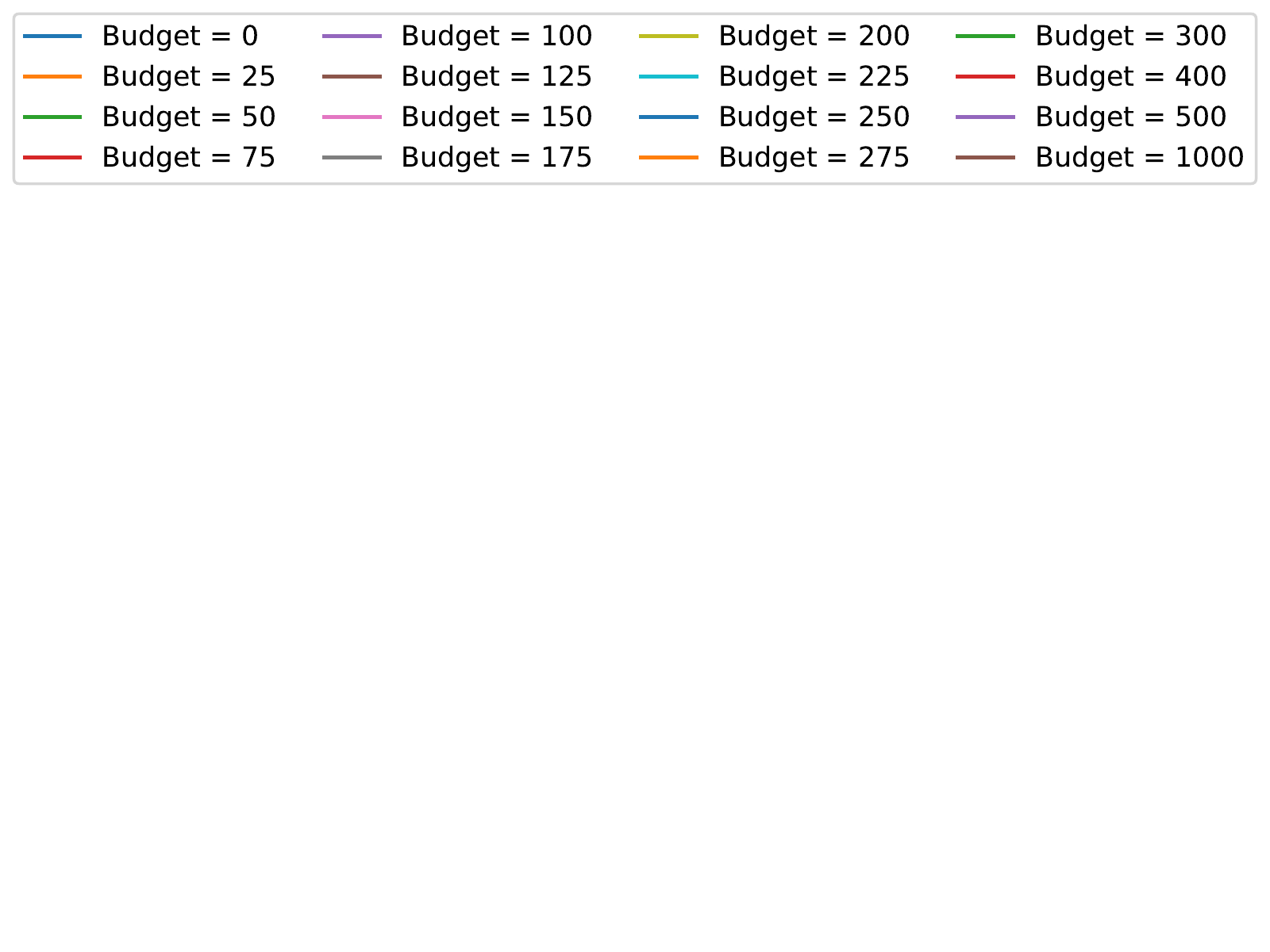}
          \end{subfigure}
     \end{subfigure}
     \vspace{-8.5cm}
     \caption{The learning curves for return, intervention rate, and success rate of the four types of pilots assisted by \textbf{budget} method copilot during training. Return, intervention rate, and success rate are smoothed using a moving average with a window size of 20 episodes.}
     \label{fig:learning_curve_budget}
\end{figure*}

\begin{figure*}[h!]
     \centering
     \begin{subfigure}[t]{\textwidth}
         \centering
         \begin{subfigure}[t]{0.24\textwidth}
             \centering
             \includegraphics[width=\textwidth]{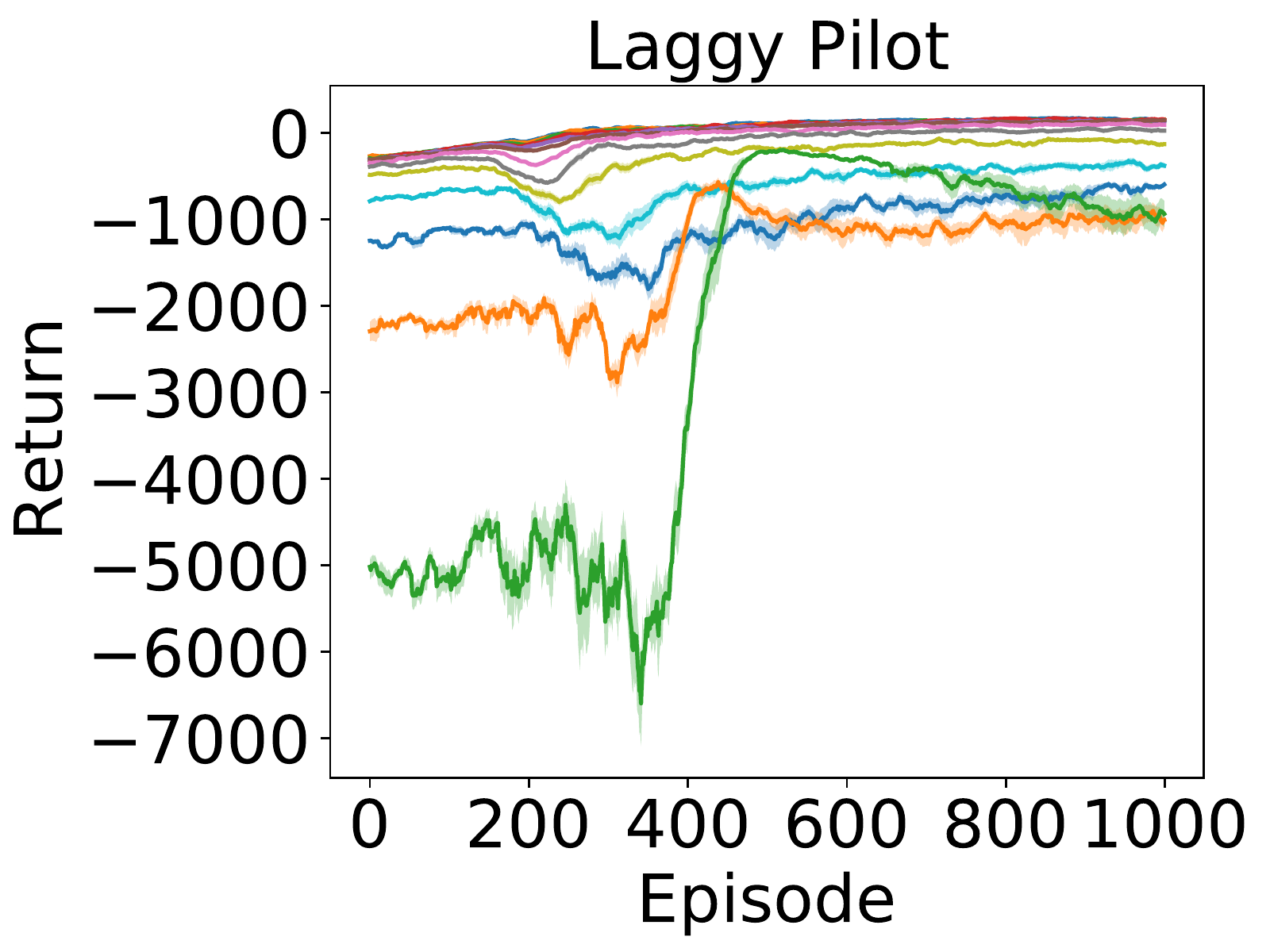}
         \end{subfigure}
         \begin{subfigure}[t]{0.24\textwidth}
             \centering
             \includegraphics[width=\textwidth]{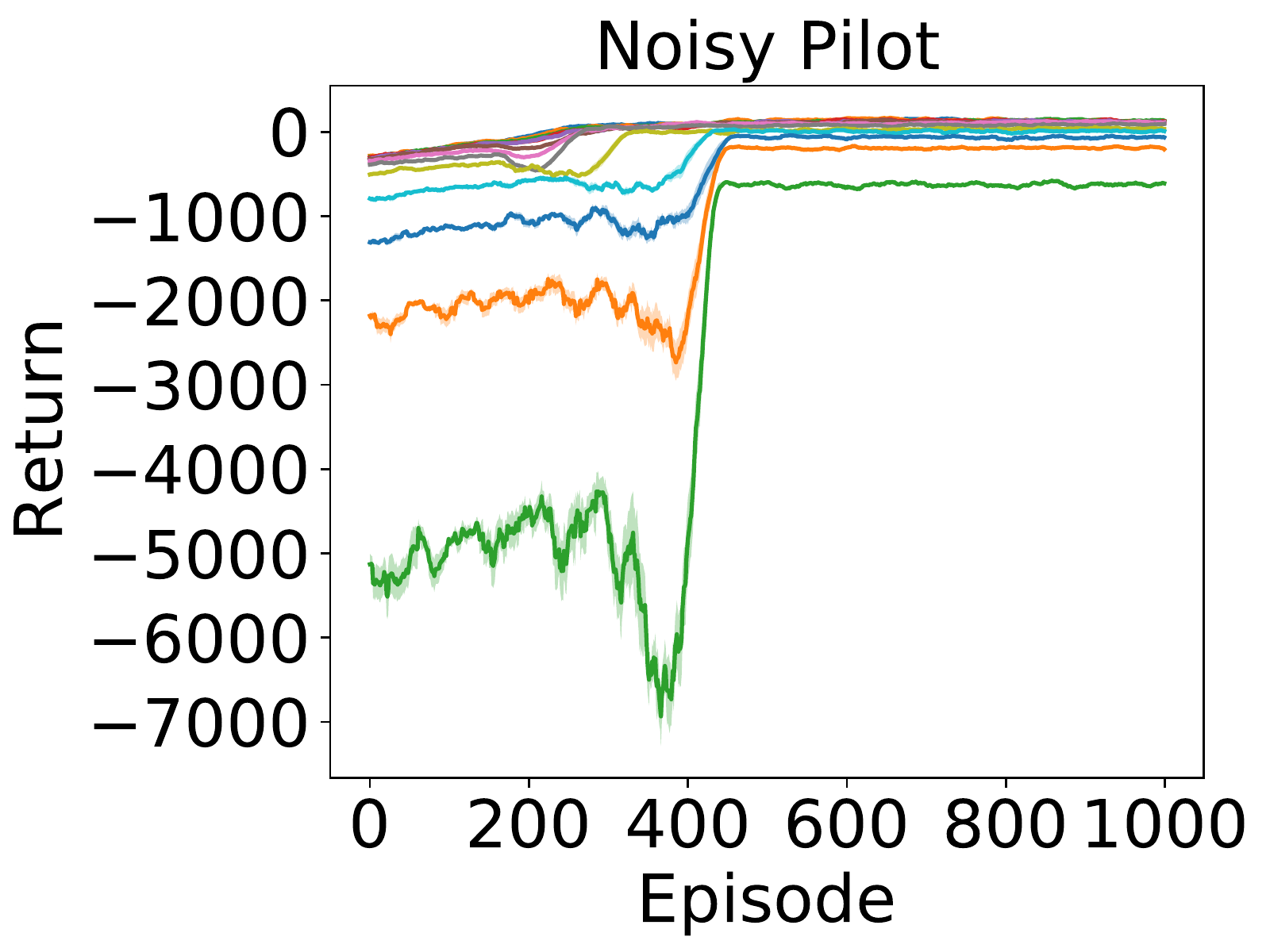}
         \end{subfigure}
         \begin{subfigure}[t]{0.24\textwidth}
             \centering
             \includegraphics[width=\textwidth]{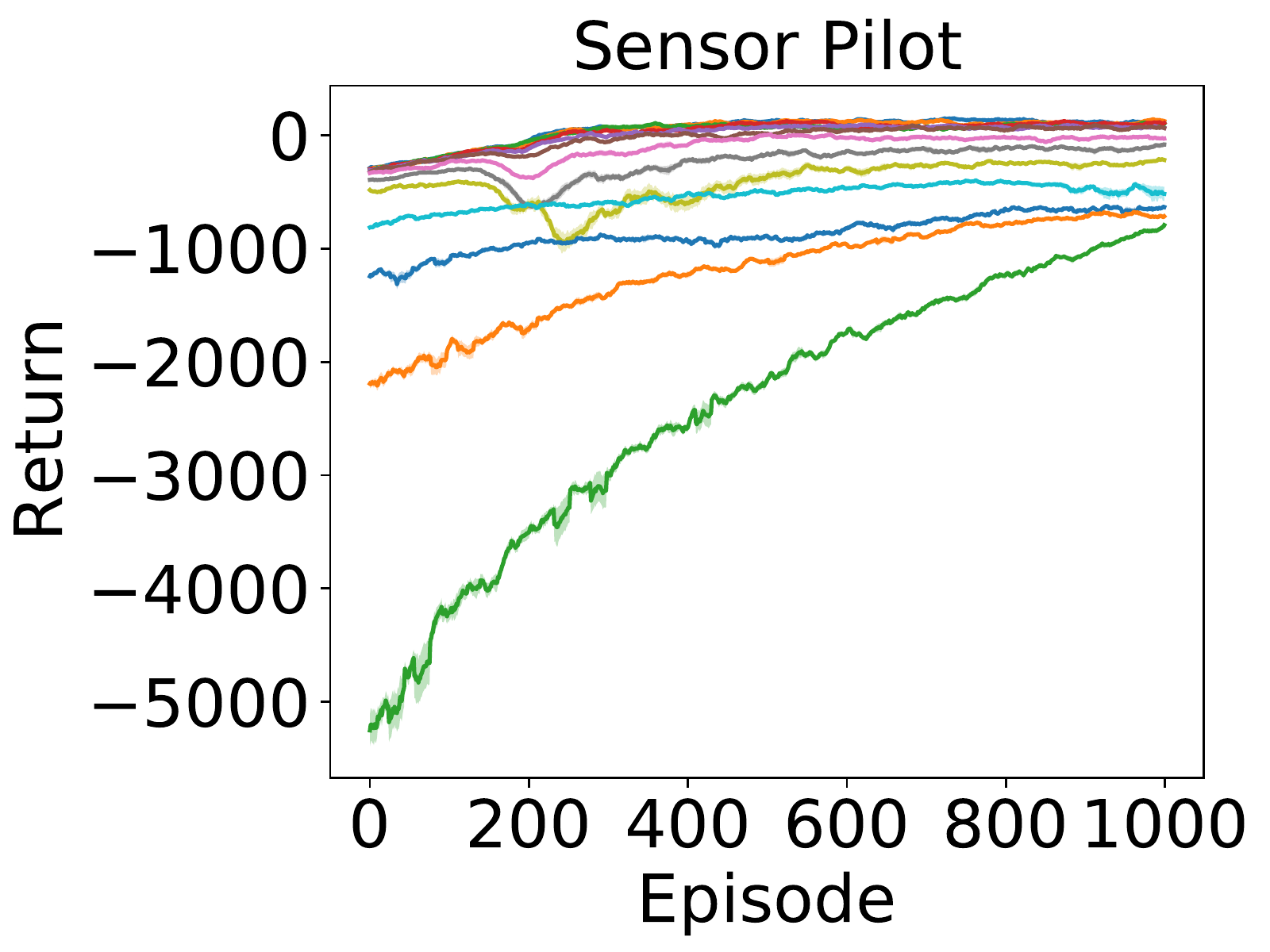}
         \end{subfigure}
         \begin{subfigure}[t]{0.24\textwidth}
             \centering
             \includegraphics[width=\textwidth]{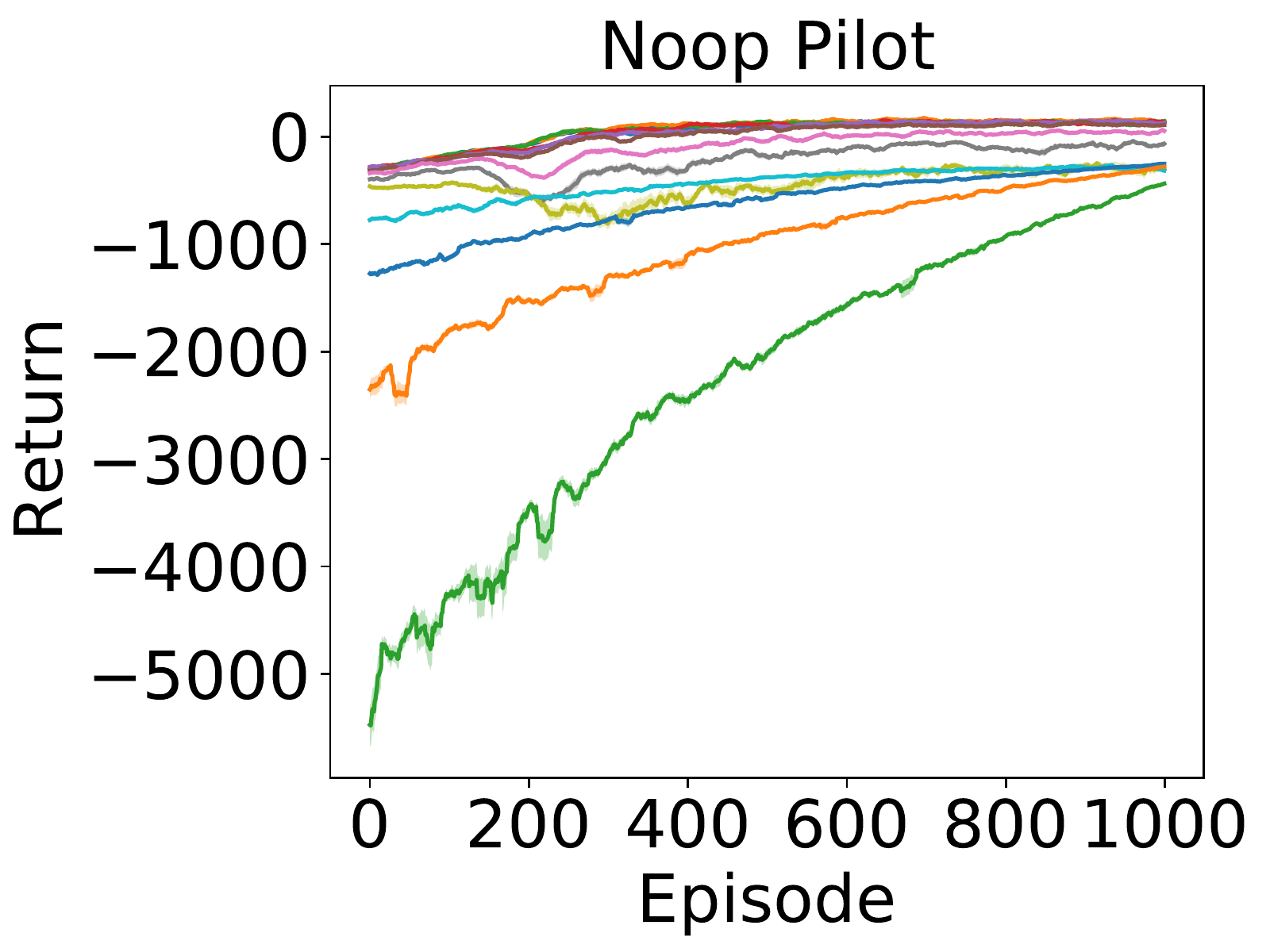}
         \end{subfigure}
         \begin{subfigure}[t]{0.24\textwidth}
             \centering
             \includegraphics[width=\textwidth]{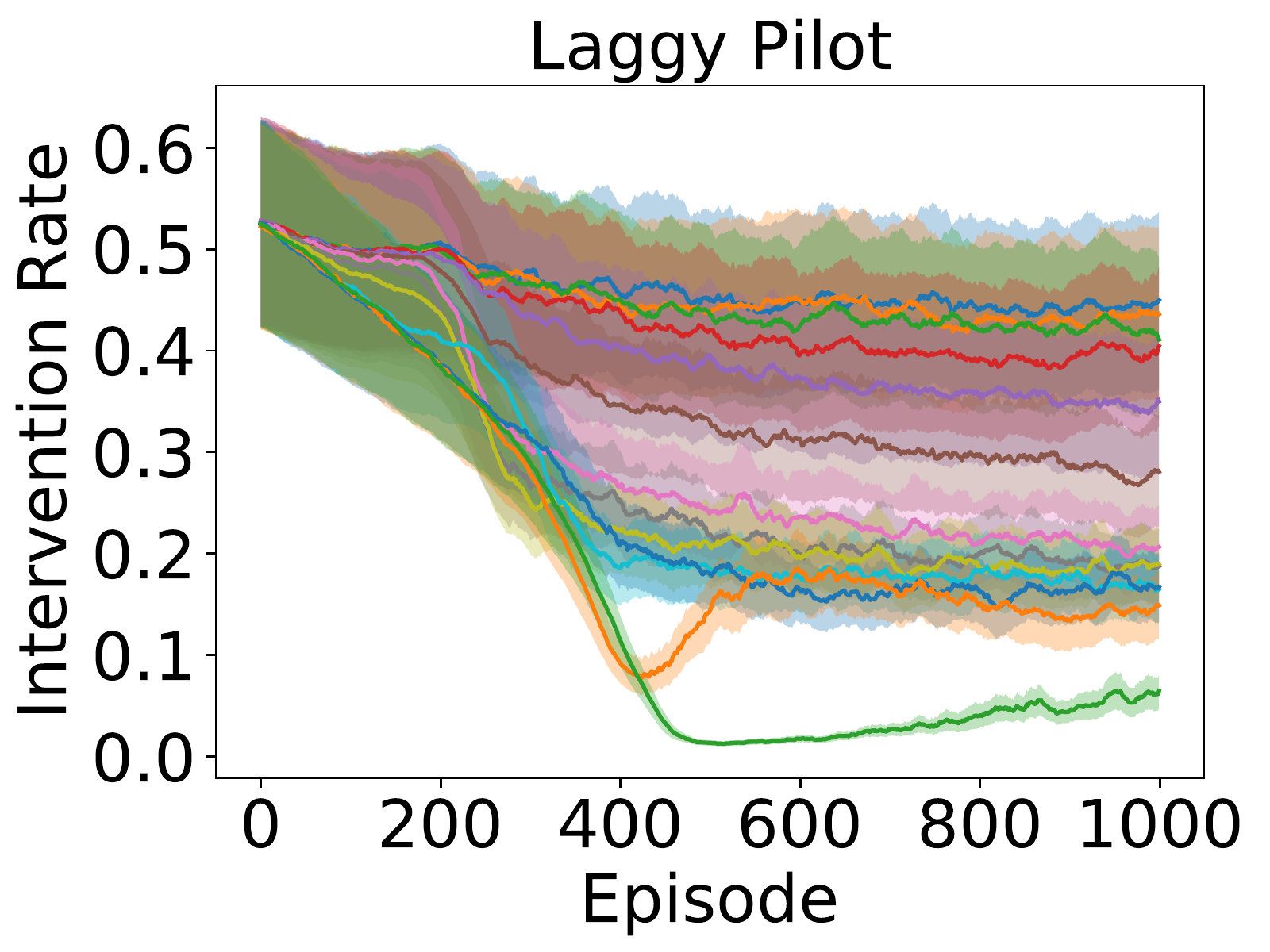}
         \end{subfigure}
         \begin{subfigure}[t]{0.24\textwidth}
             \centering
             \includegraphics[width=\textwidth]{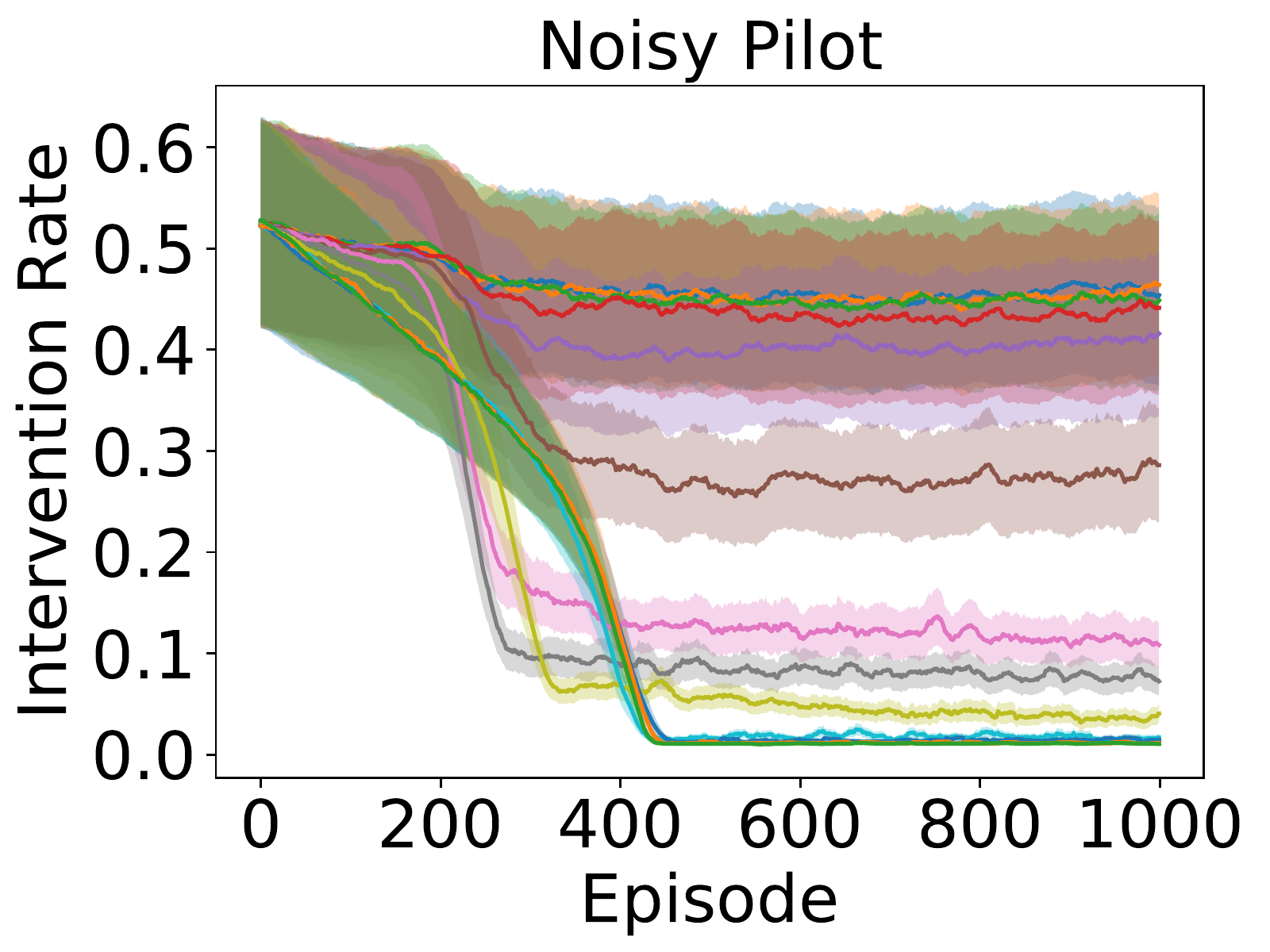}
         \end{subfigure}
         \begin{subfigure}[t]{0.24\textwidth}
             \centering
             \includegraphics[width=\textwidth]{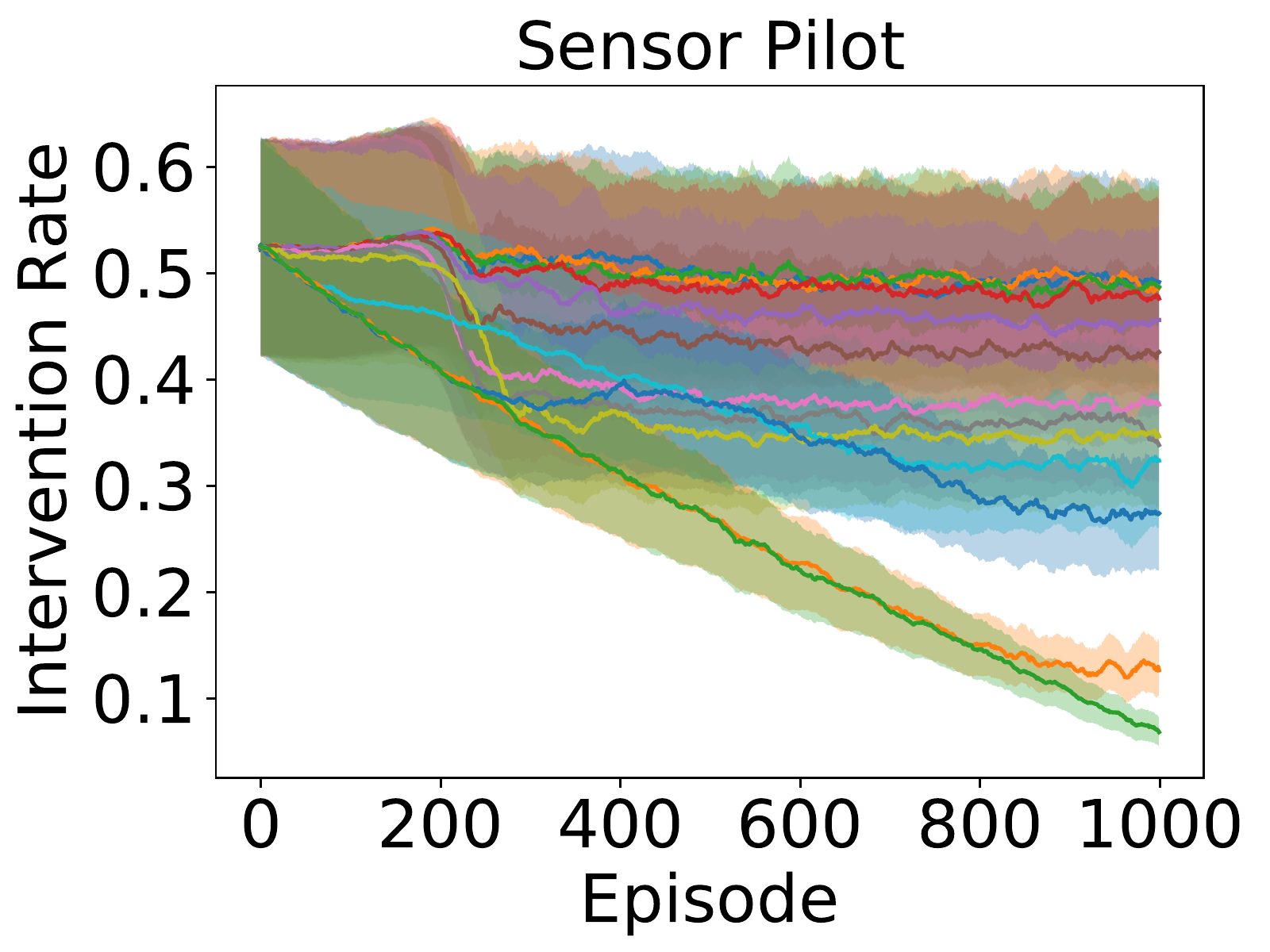}
         \end{subfigure}
         \begin{subfigure}[t]{0.24\textwidth}
             \centering
             \includegraphics[width=\textwidth]{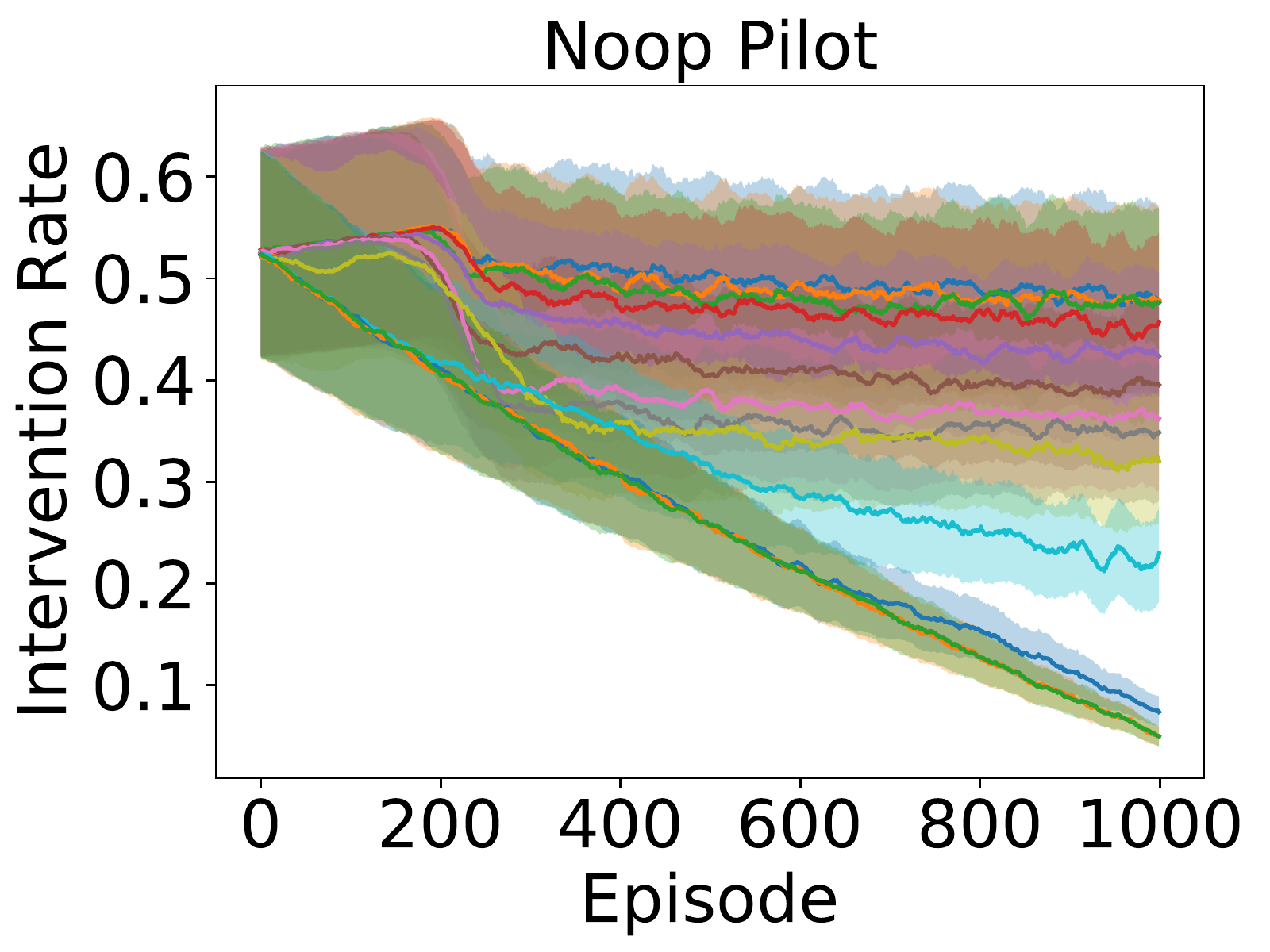}
         \end{subfigure}
         \begin{subfigure}[t]{0.24\textwidth}
             \centering
             \includegraphics[width=\textwidth]{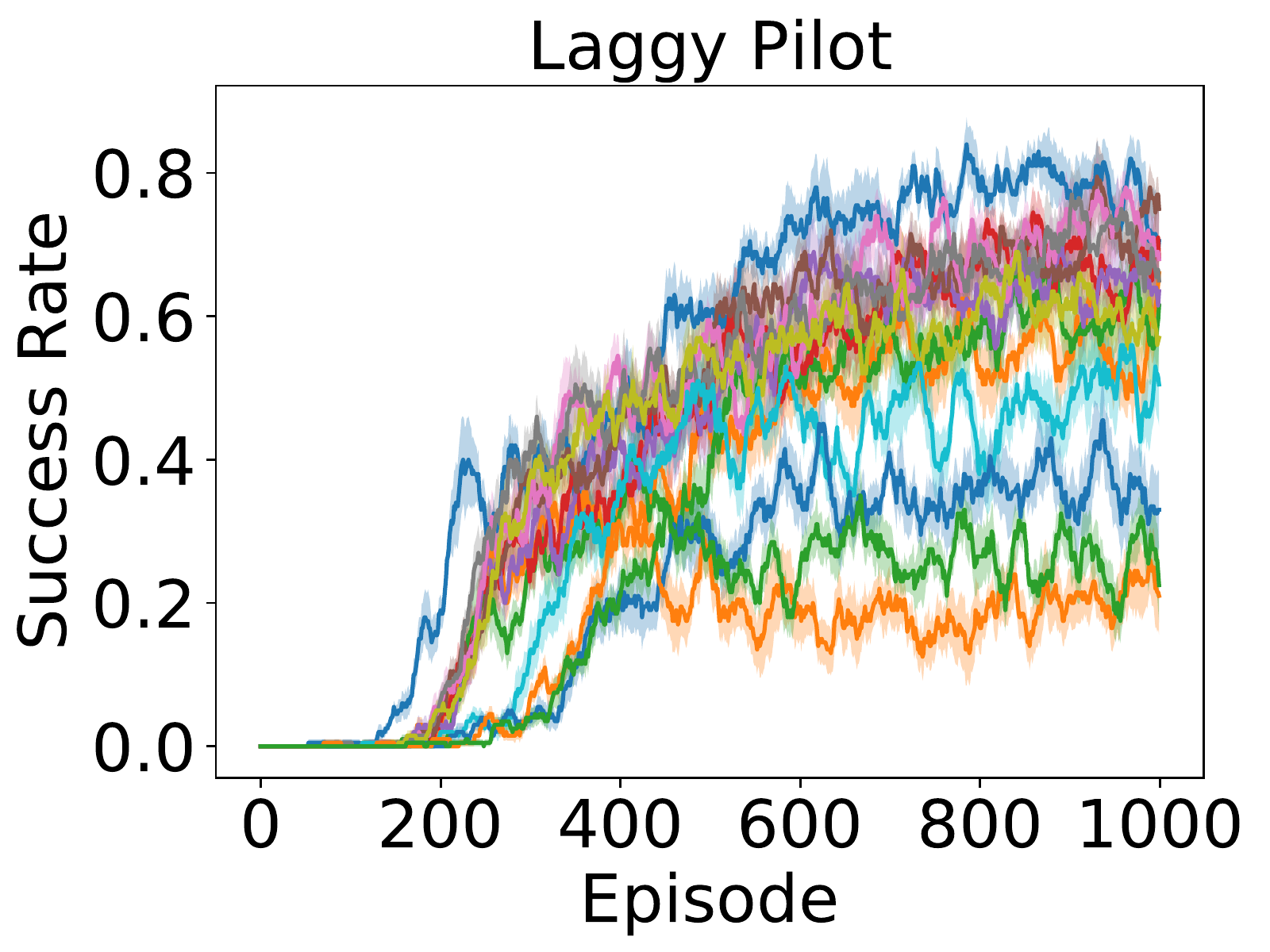}
         \end{subfigure}
         \begin{subfigure}[t]{0.24\textwidth}
             \centering
             \includegraphics[width=\textwidth]{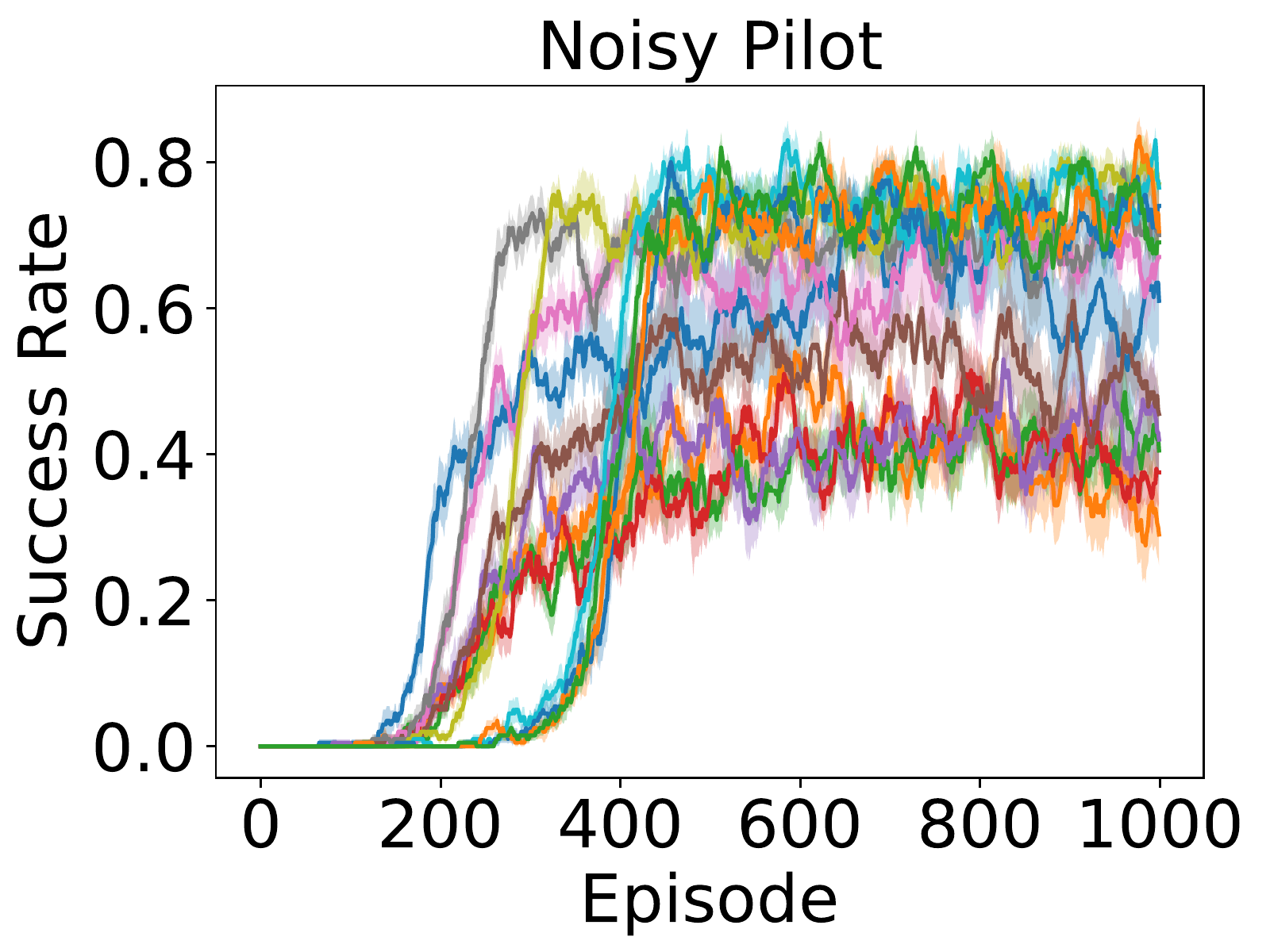}
         \end{subfigure}
         \begin{subfigure}[t]{0.24\textwidth}
             \centering
             \includegraphics[width=\textwidth]{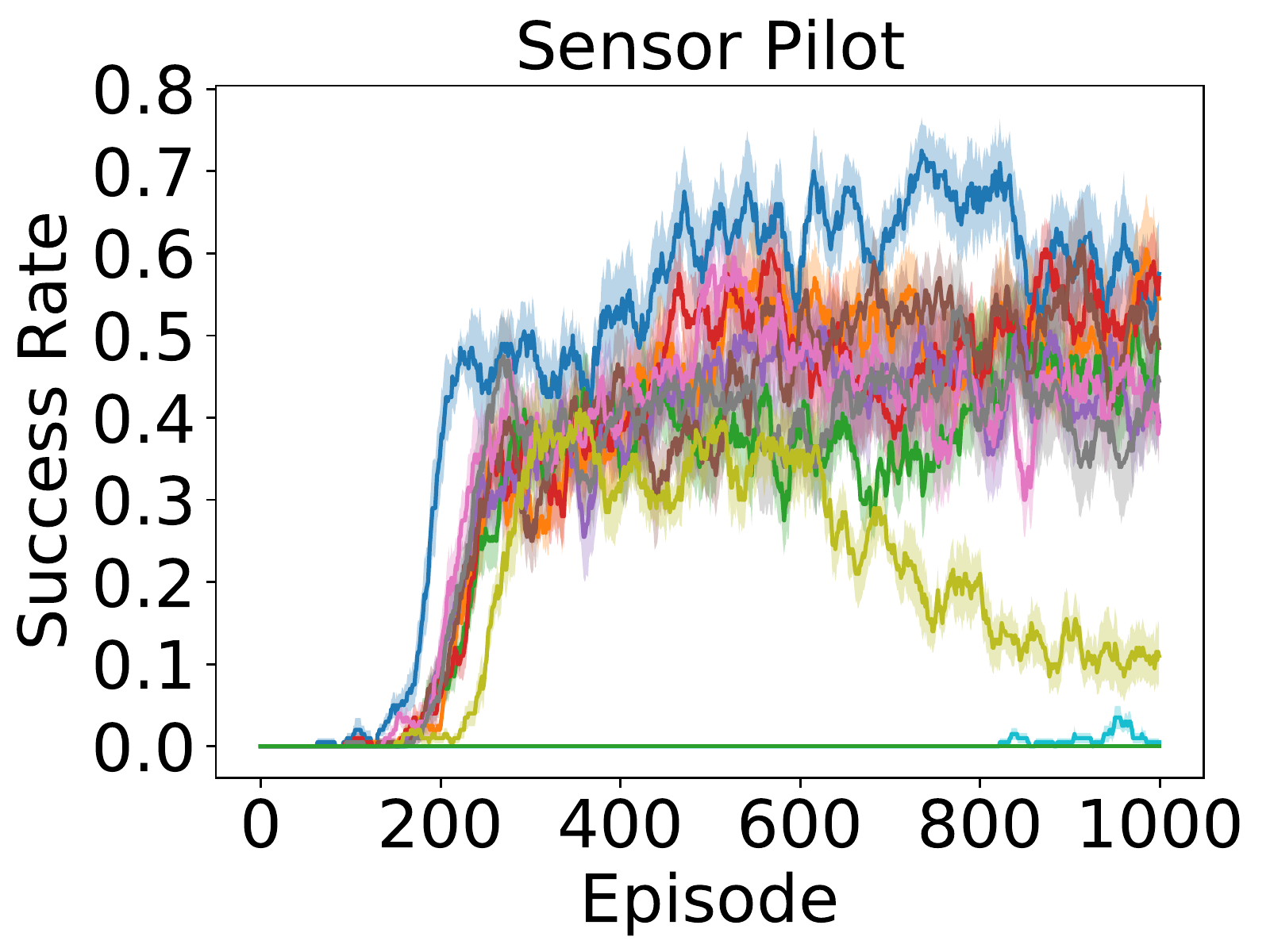}
         \end{subfigure}
         \begin{subfigure}[t]{0.24\textwidth}
             \centering
             \includegraphics[width=\textwidth]{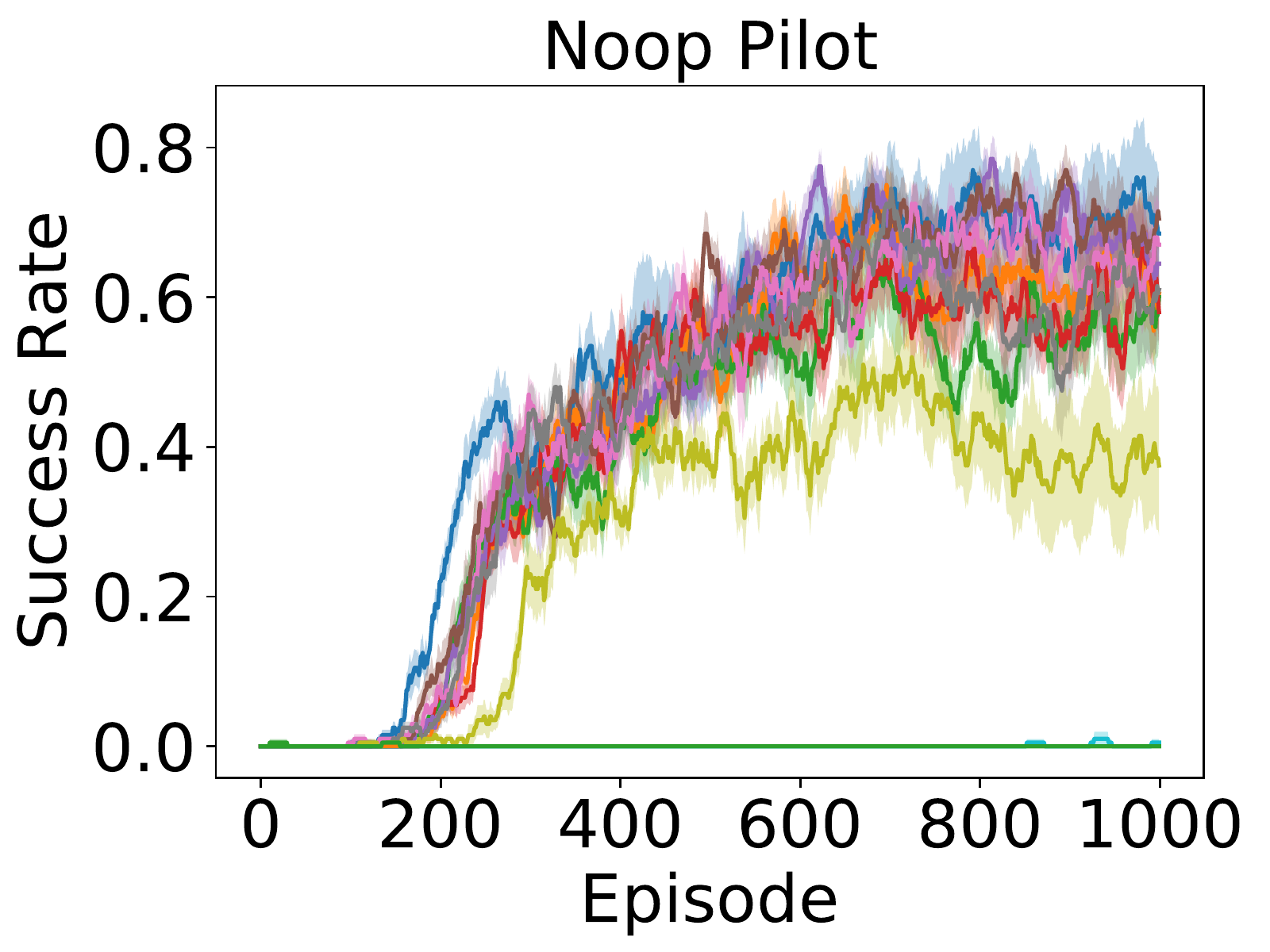}
         \end{subfigure}
         
         \begin{subfigure}[t]{\textwidth}
             \centering
             
             \includegraphics[width=\textwidth]{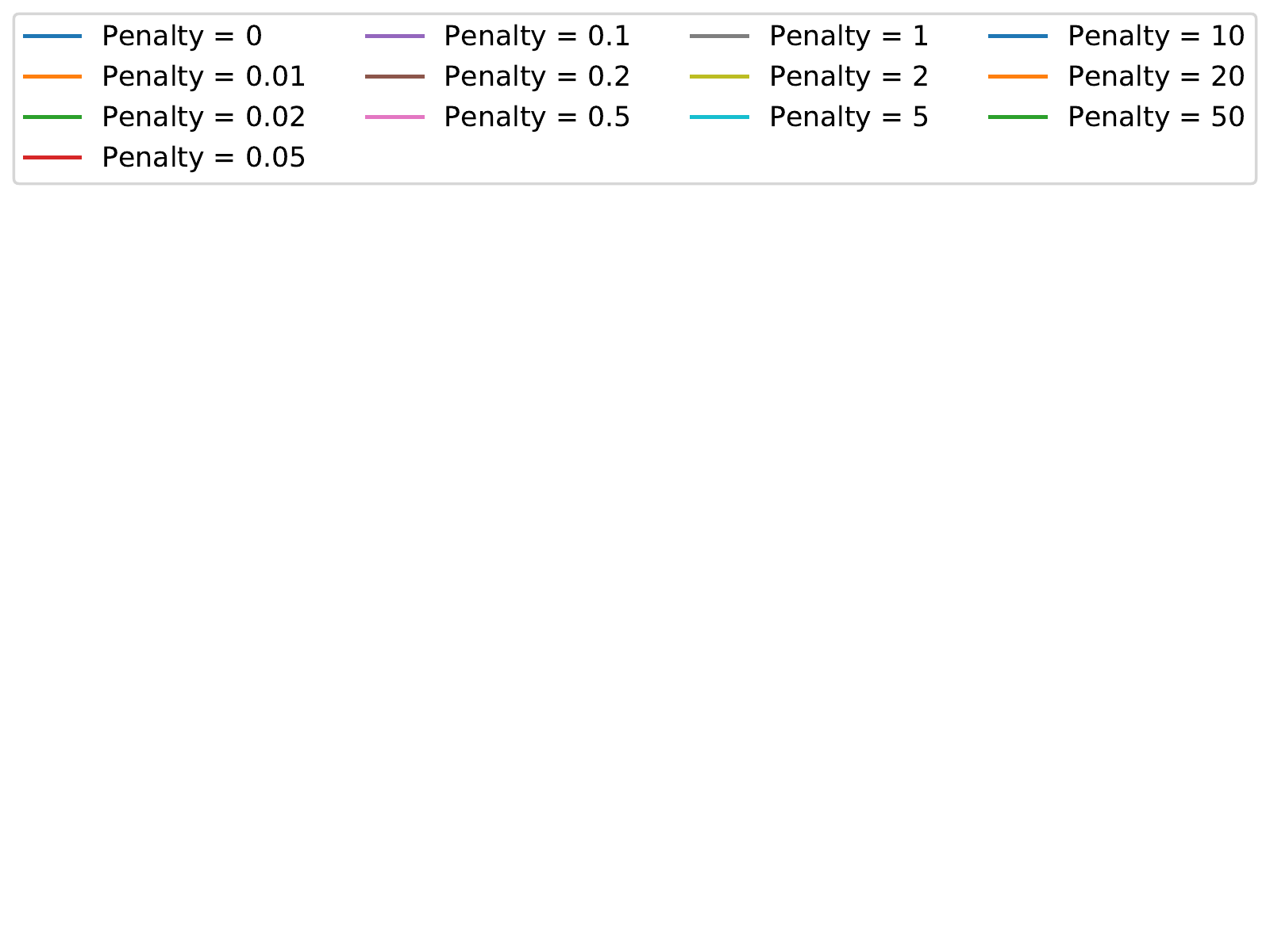}
          \end{subfigure}
     \end{subfigure}
     \vspace{-8.5cm}
     \caption{The learning curves for return, intervention rate, and success rate of the four types of pilots assisted by \textbf{penalty} method copilot during training. Return, intervention rate, and success rate are smoothed using a moving average with a window size of 20 episodes.}
     \label{fig:learning_curve_penalty}
\end{figure*}

\begin{figure*}[h!]
     \centering
     \begin{subfigure}[t]{\textwidth}
         \centering
         \begin{subfigure}[t]{0.24\textwidth}
             \centering
             \includegraphics[width=\textwidth]{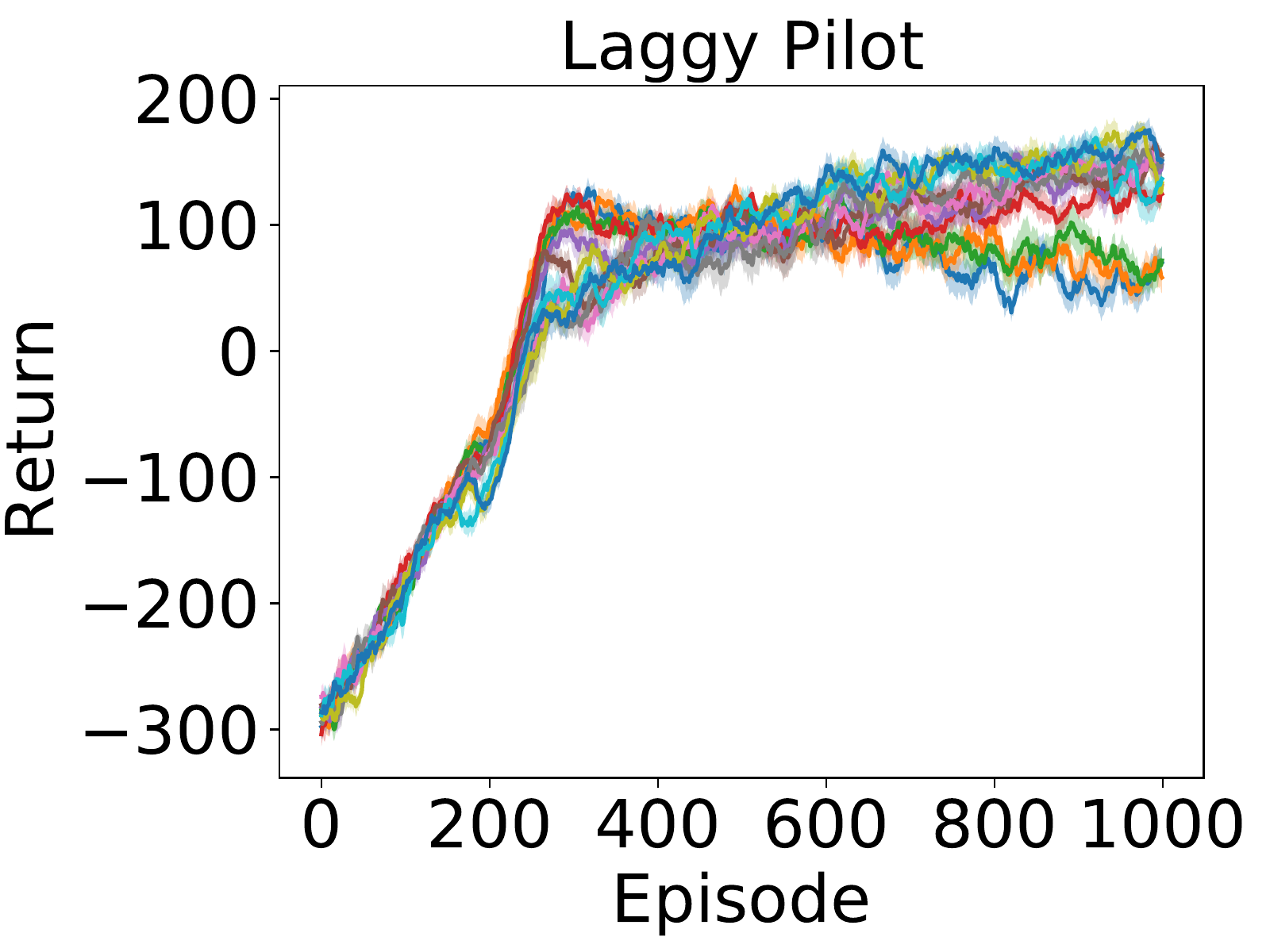}
         \end{subfigure}
         \begin{subfigure}[t]{0.24\textwidth}
             \centering
             \includegraphics[width=\textwidth]{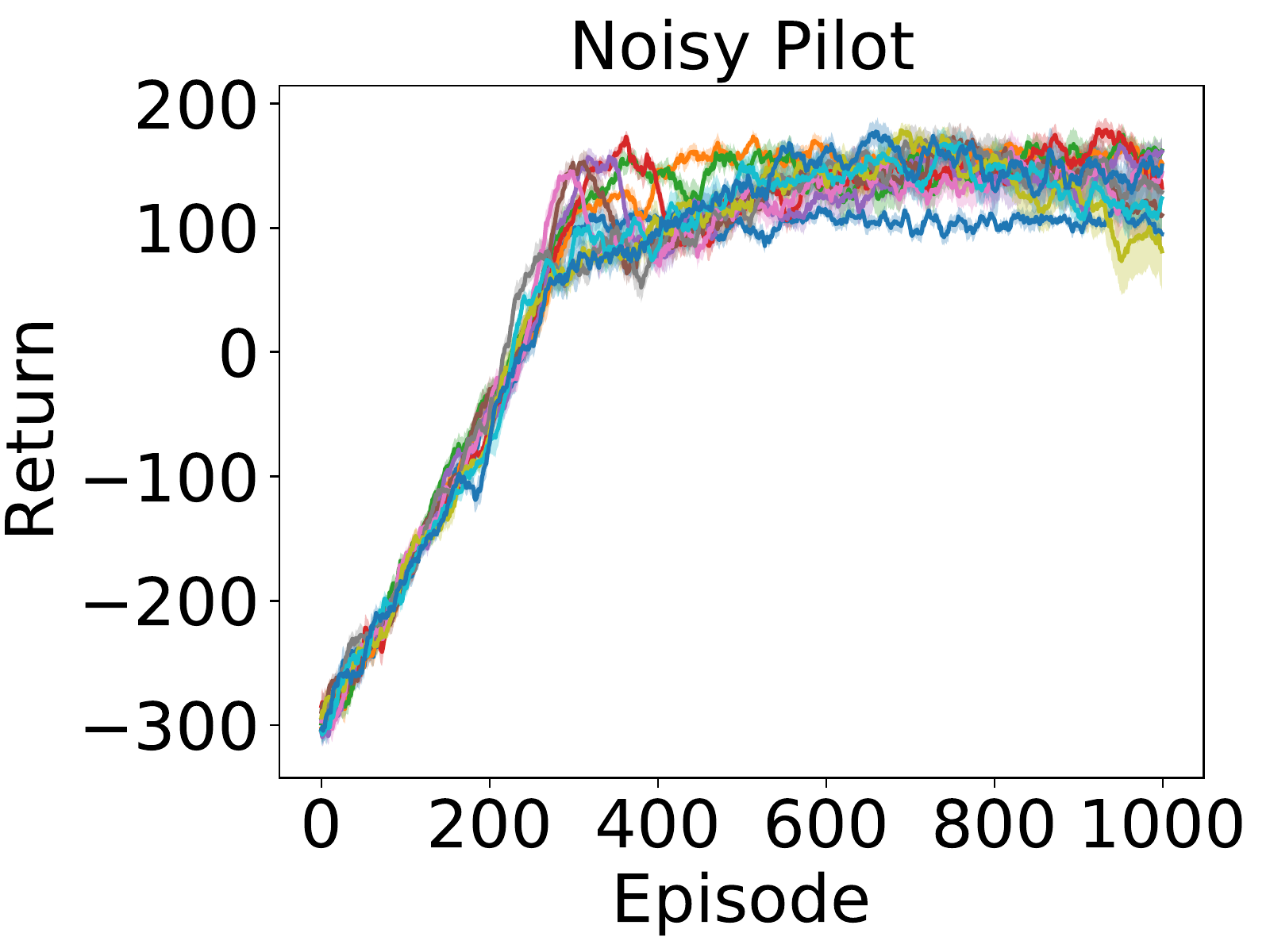}
         \end{subfigure}
         \begin{subfigure}[t]{0.24\textwidth}
             \centering
             \includegraphics[width=\textwidth]{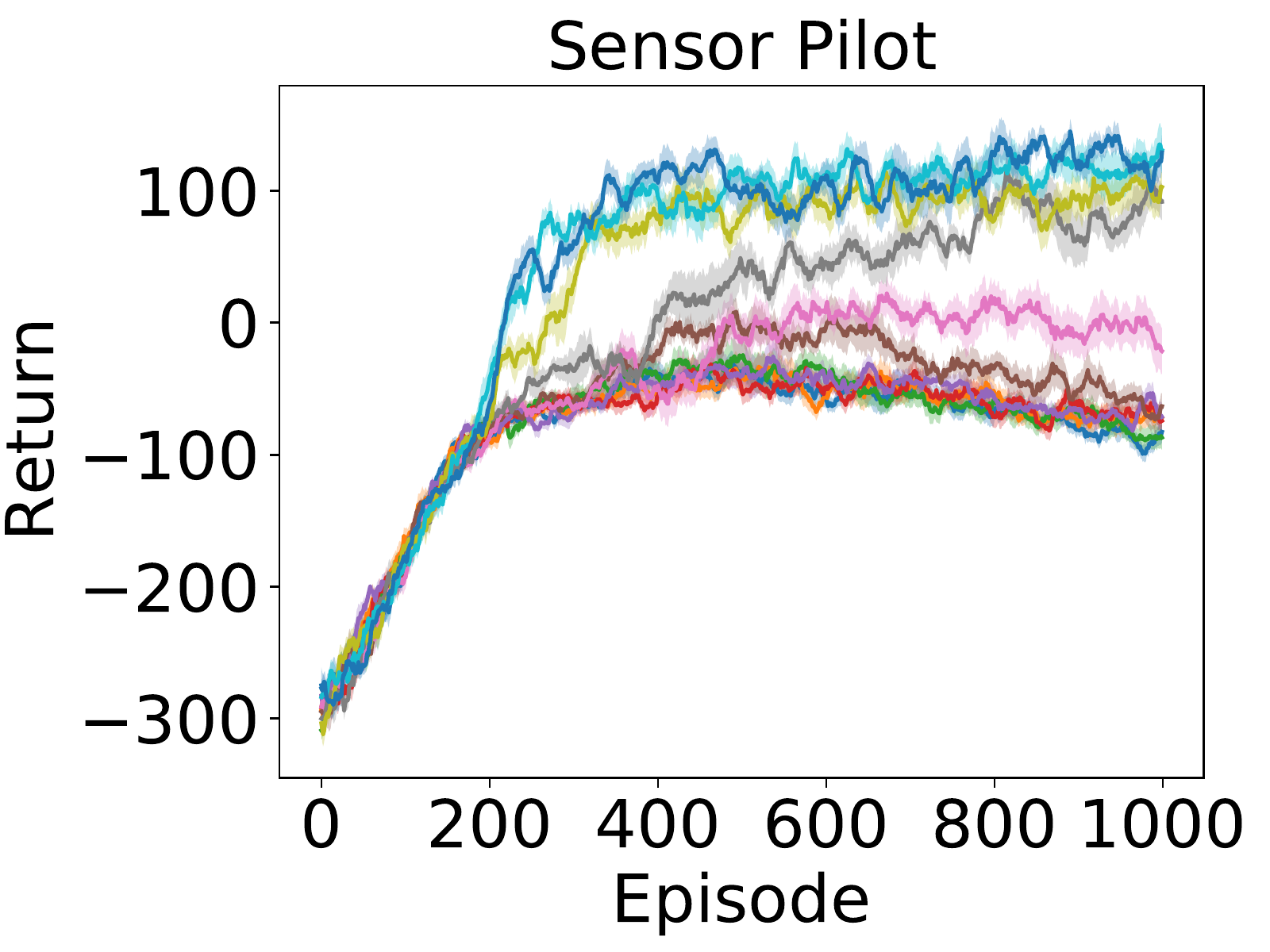}
         \end{subfigure}
         \begin{subfigure}[t]{0.24\textwidth}
             \centering
             \includegraphics[width=\textwidth]{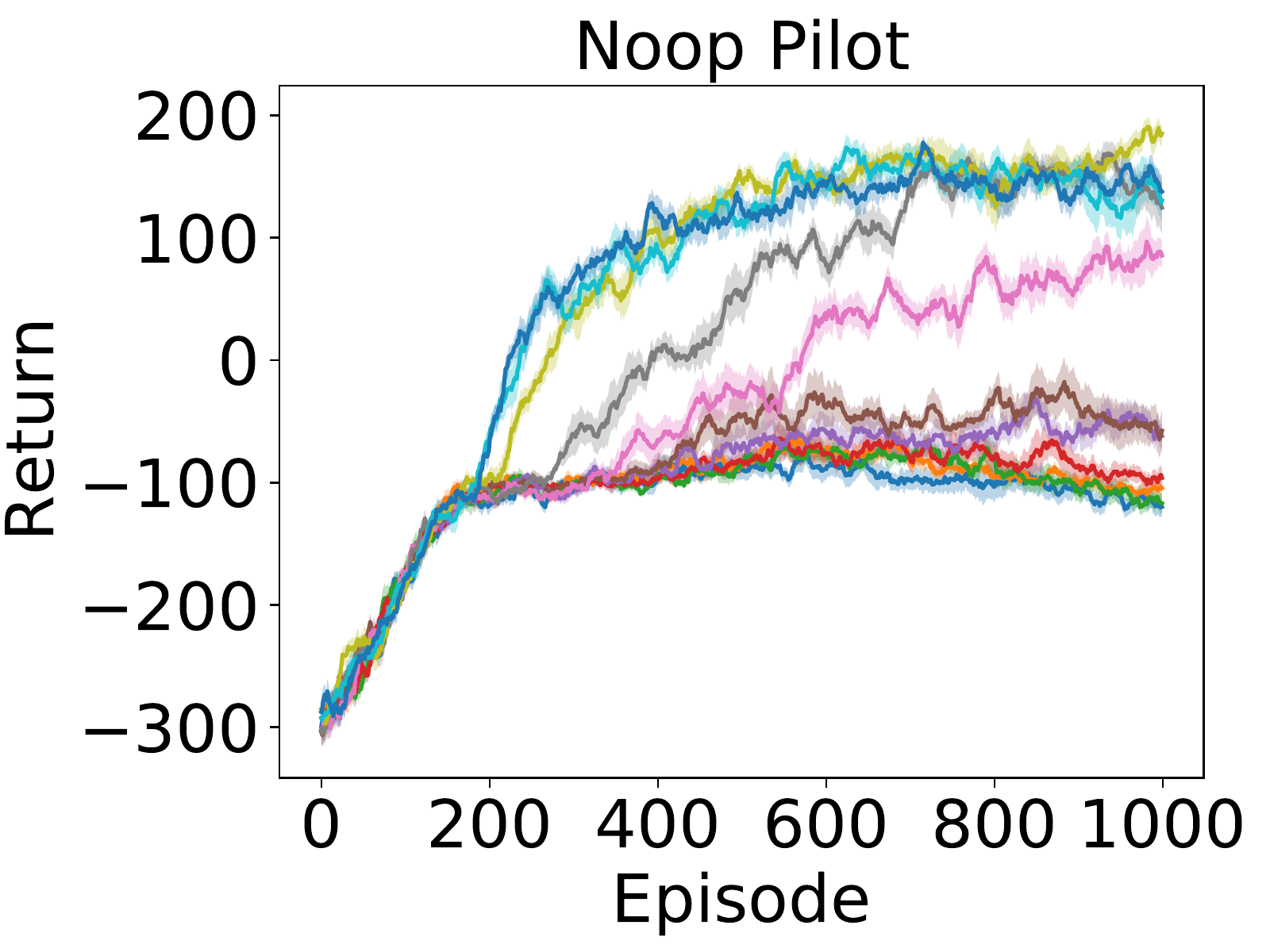}
         \end{subfigure}
         \begin{subfigure}[t]{0.24\textwidth}
             \centering
             \includegraphics[width=\textwidth]{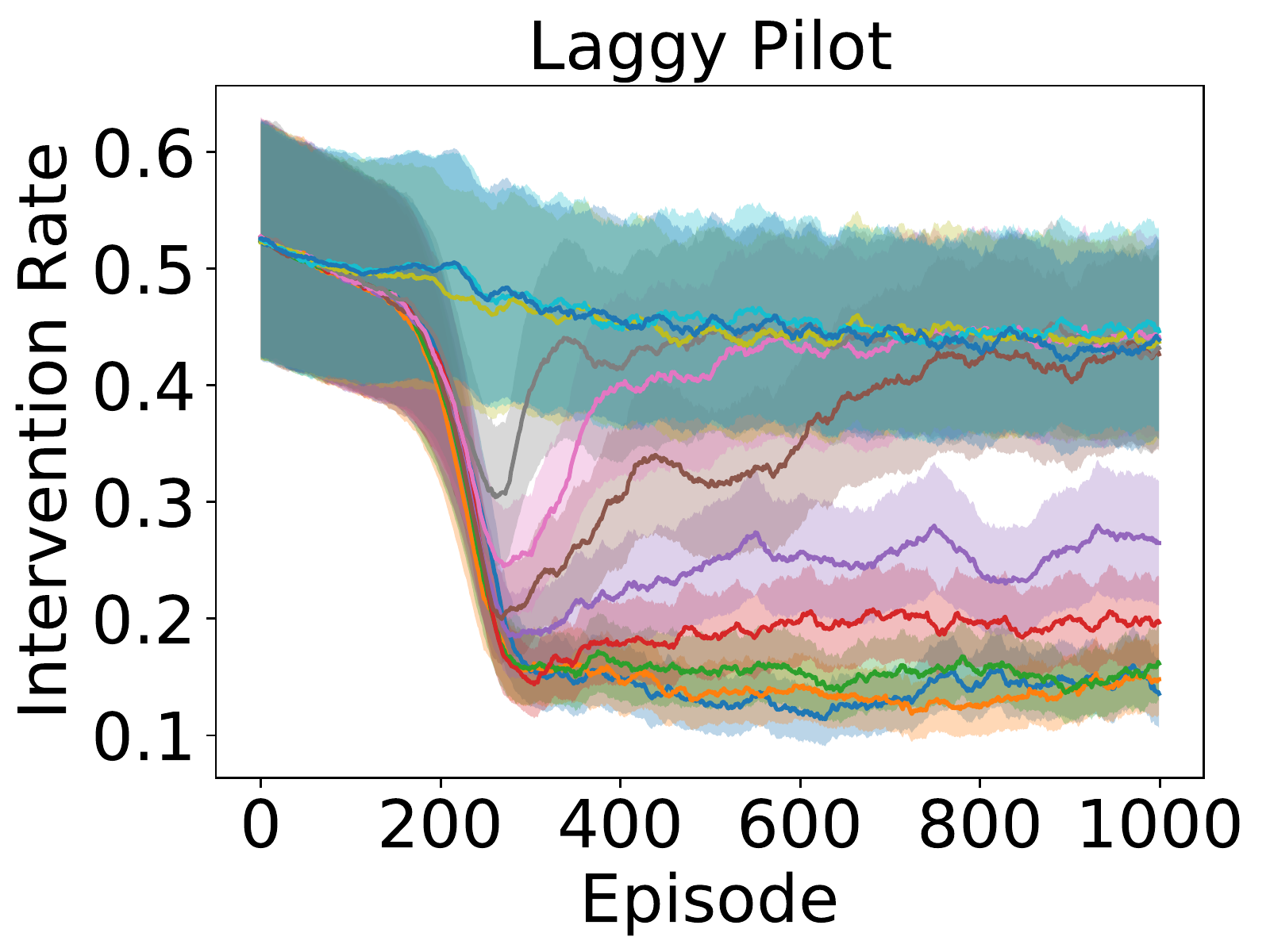}
         \end{subfigure}
         \begin{subfigure}[t]{0.24\textwidth}
             \centering
             \includegraphics[width=\textwidth]{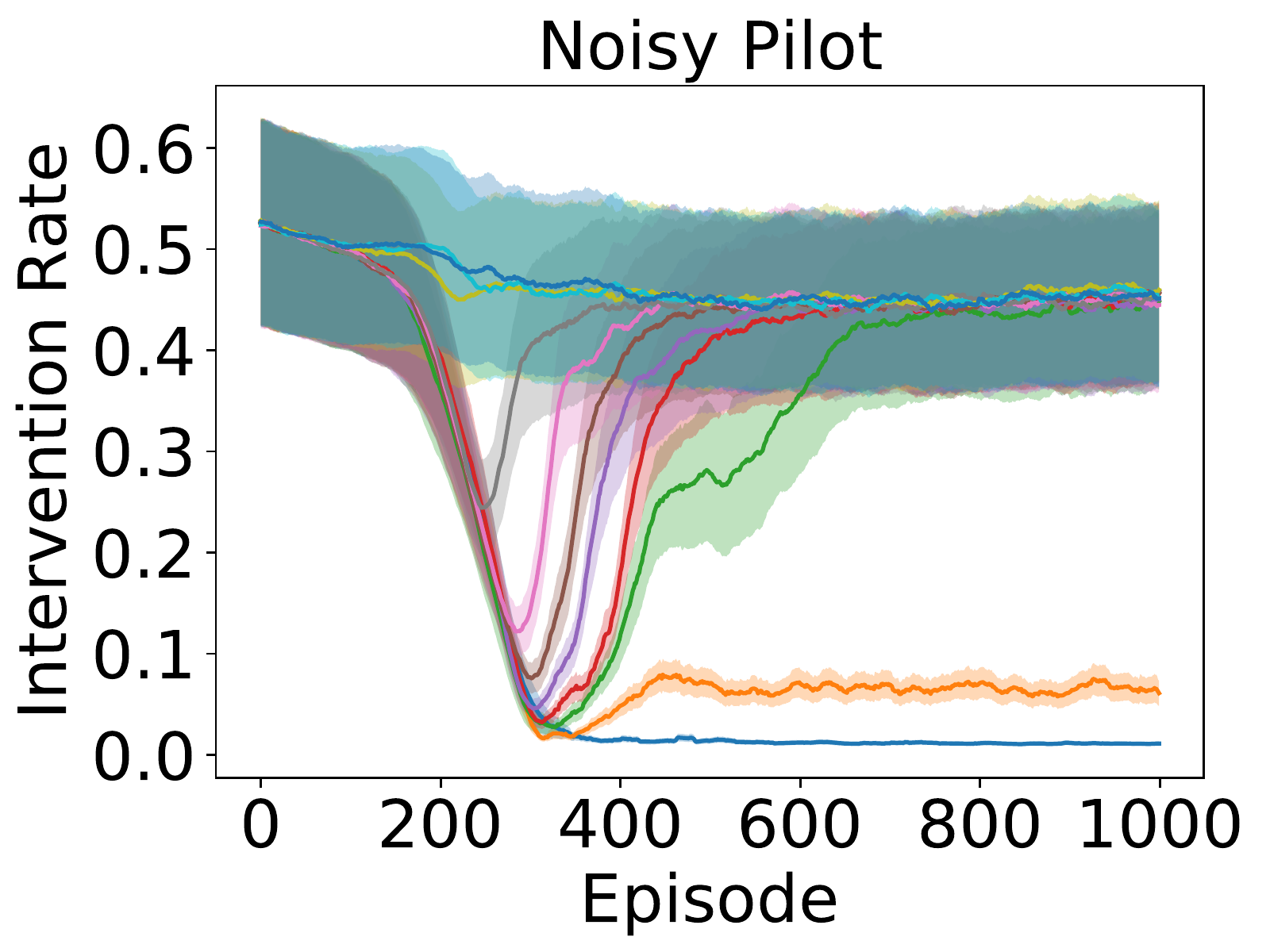}
         \end{subfigure}
         \begin{subfigure}[t]{0.24\textwidth}
             \centering
             \includegraphics[width=\textwidth]{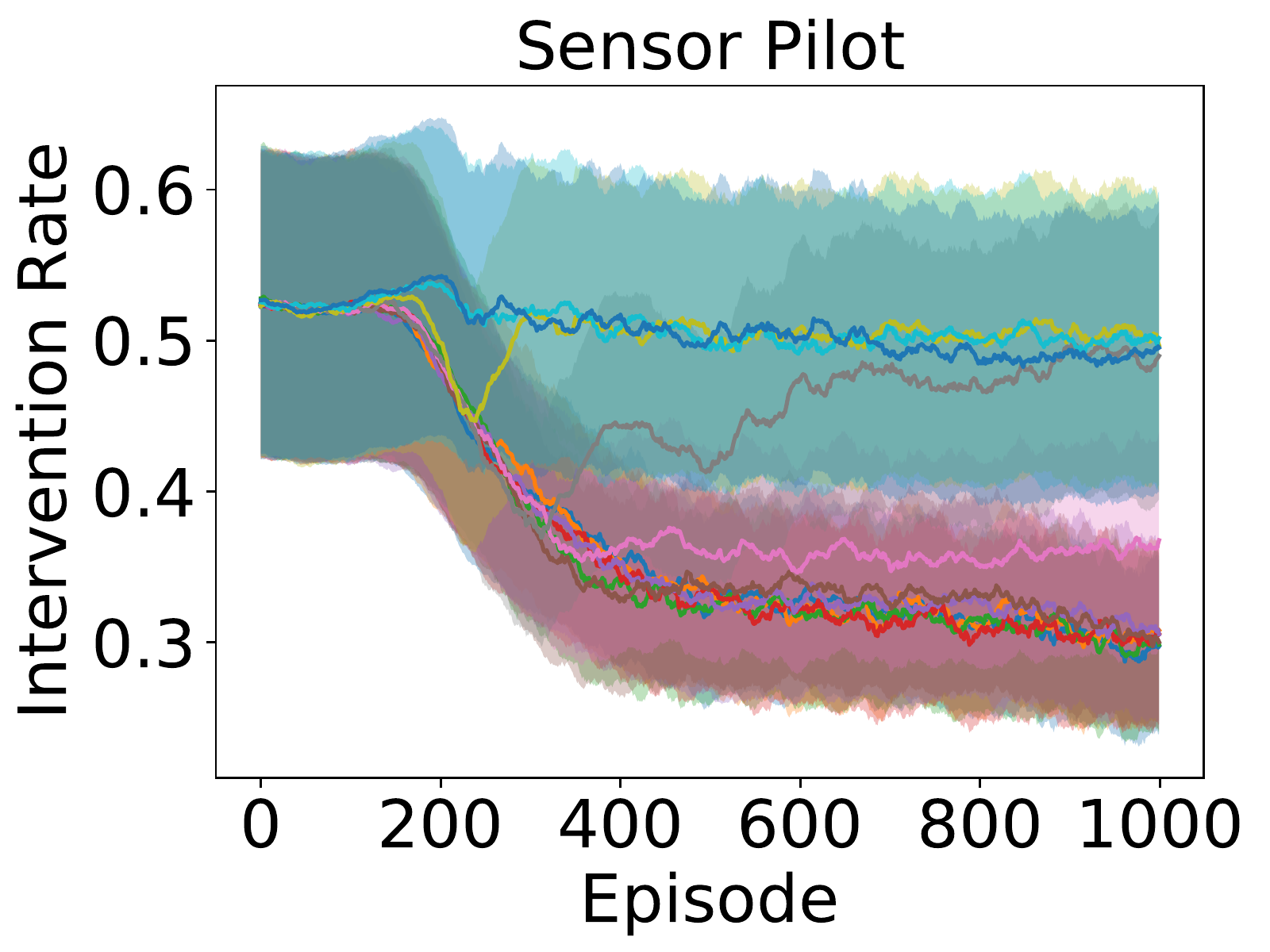}
         \end{subfigure}
         \begin{subfigure}[t]{0.24\textwidth}
             \centering
             \includegraphics[width=\textwidth]{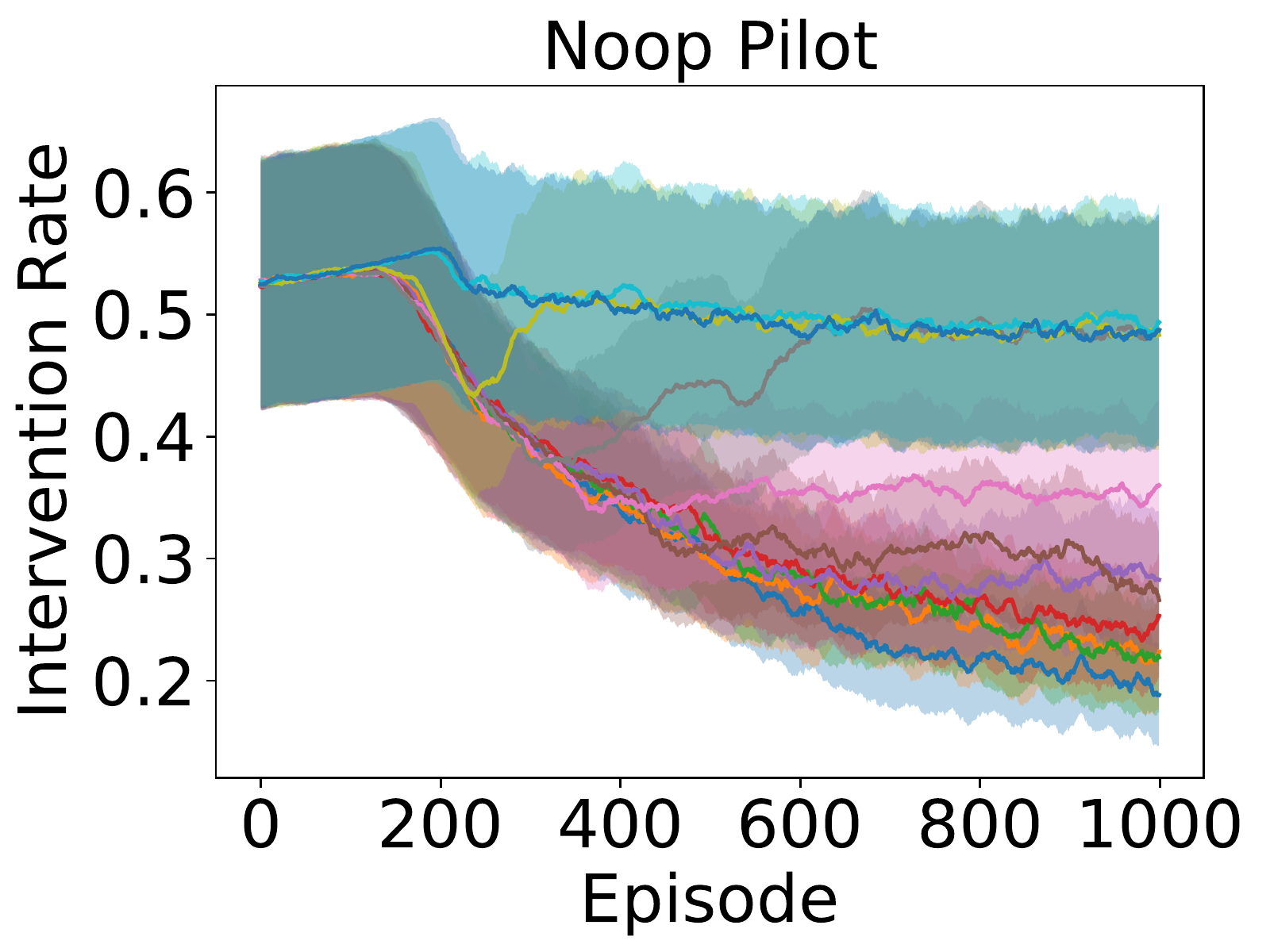}
         \end{subfigure}
         \begin{subfigure}[t]{0.24\textwidth}
             \centering
             \includegraphics[width=\textwidth]{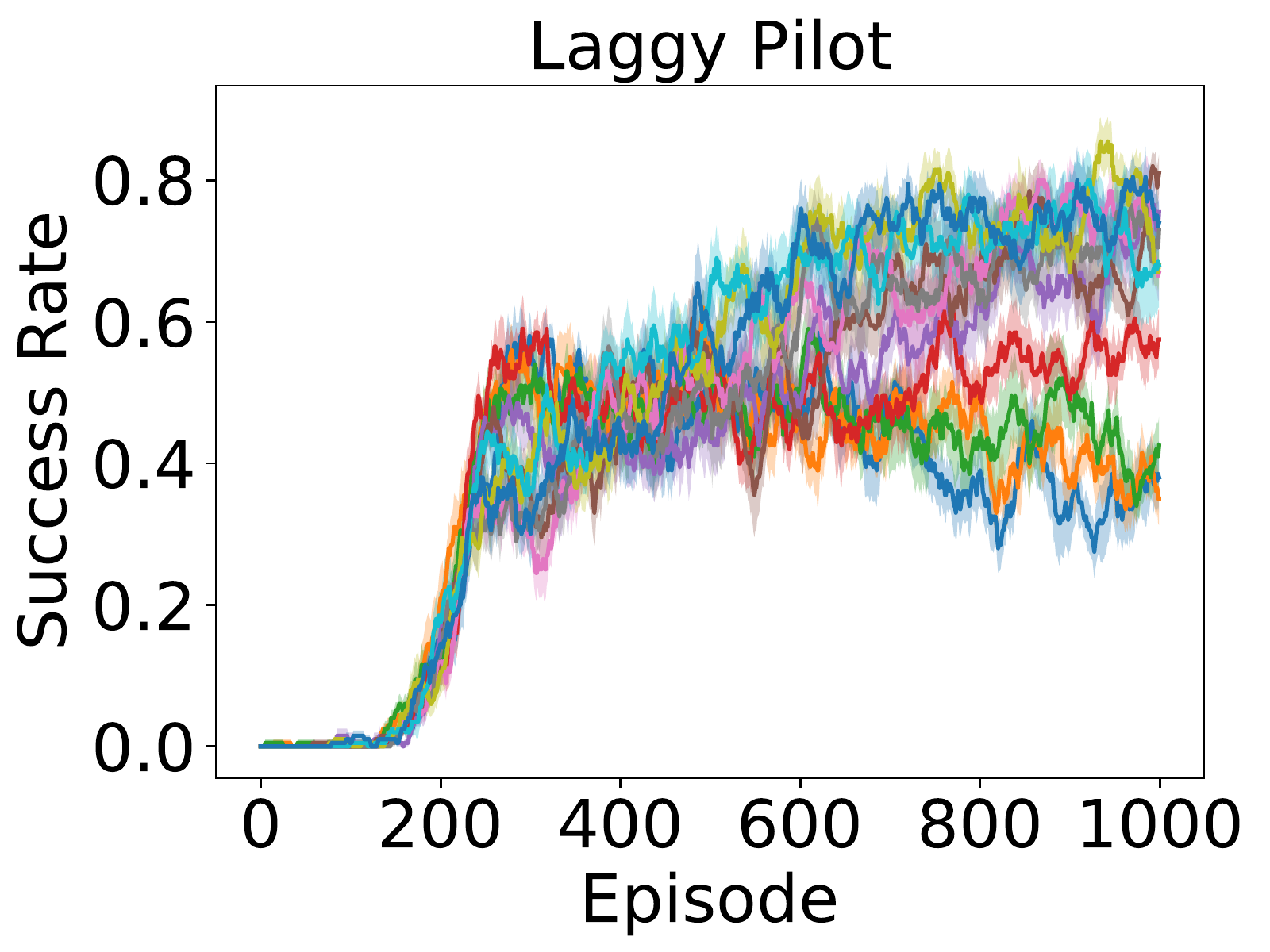}
         \end{subfigure}
         \begin{subfigure}[t]{0.24\textwidth}
             \centering
             \includegraphics[width=\textwidth]{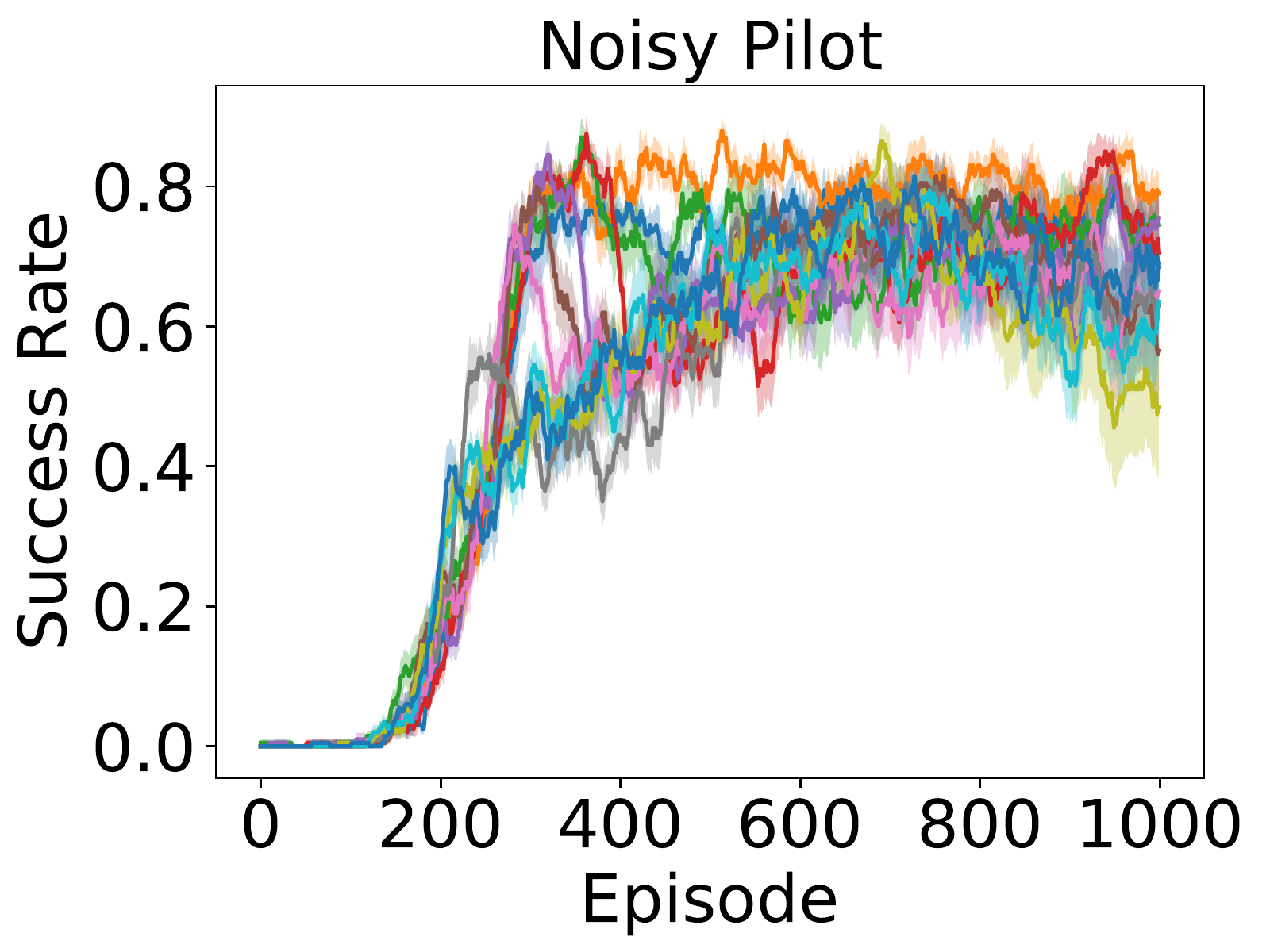}
         \end{subfigure}
         \begin{subfigure}[t]{0.24\textwidth}
             \centering
             \includegraphics[width=\textwidth]{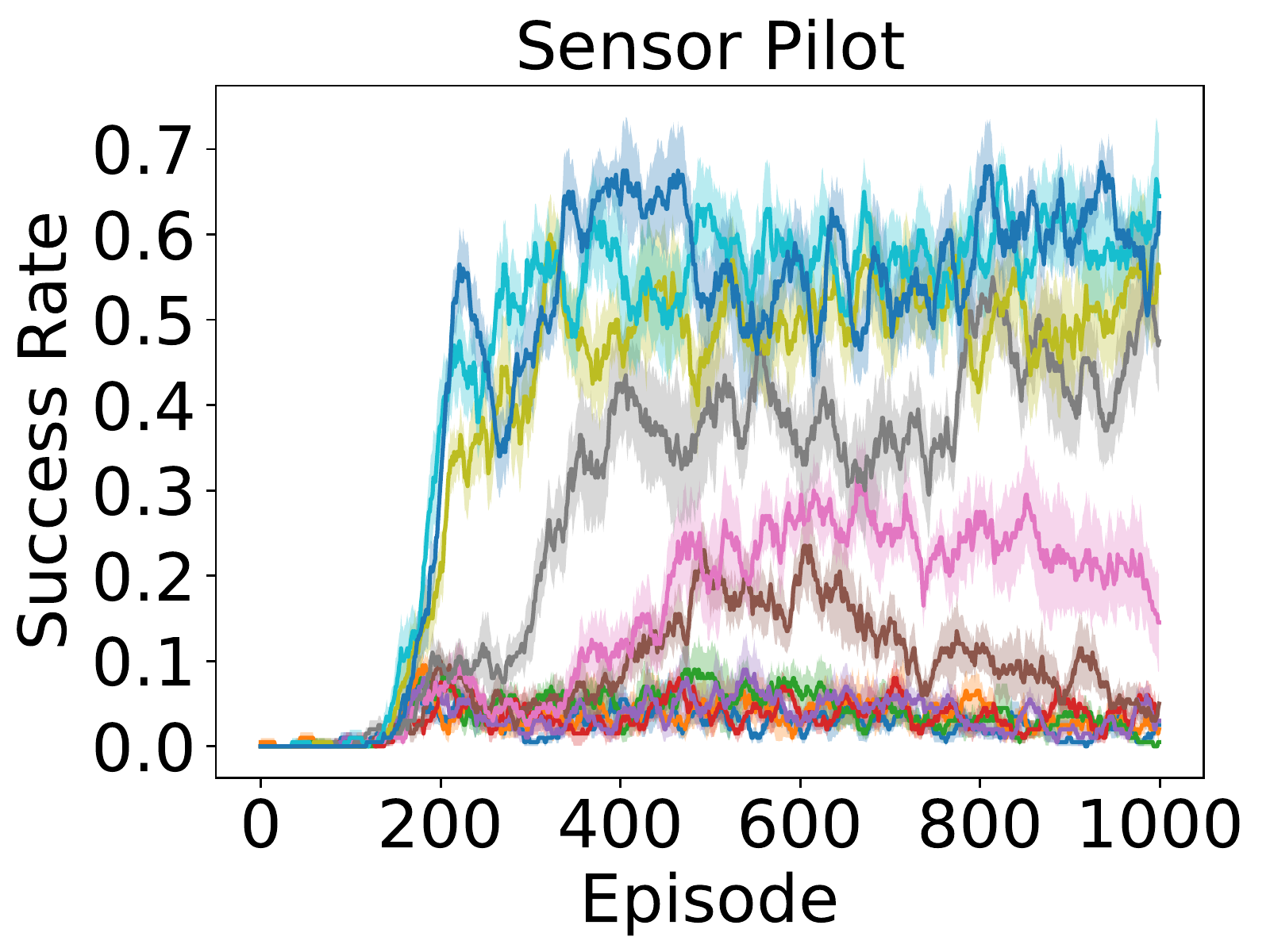}
         \end{subfigure}
         \begin{subfigure}[t]{0.24\textwidth}
             \centering
             \includegraphics[width=\textwidth]{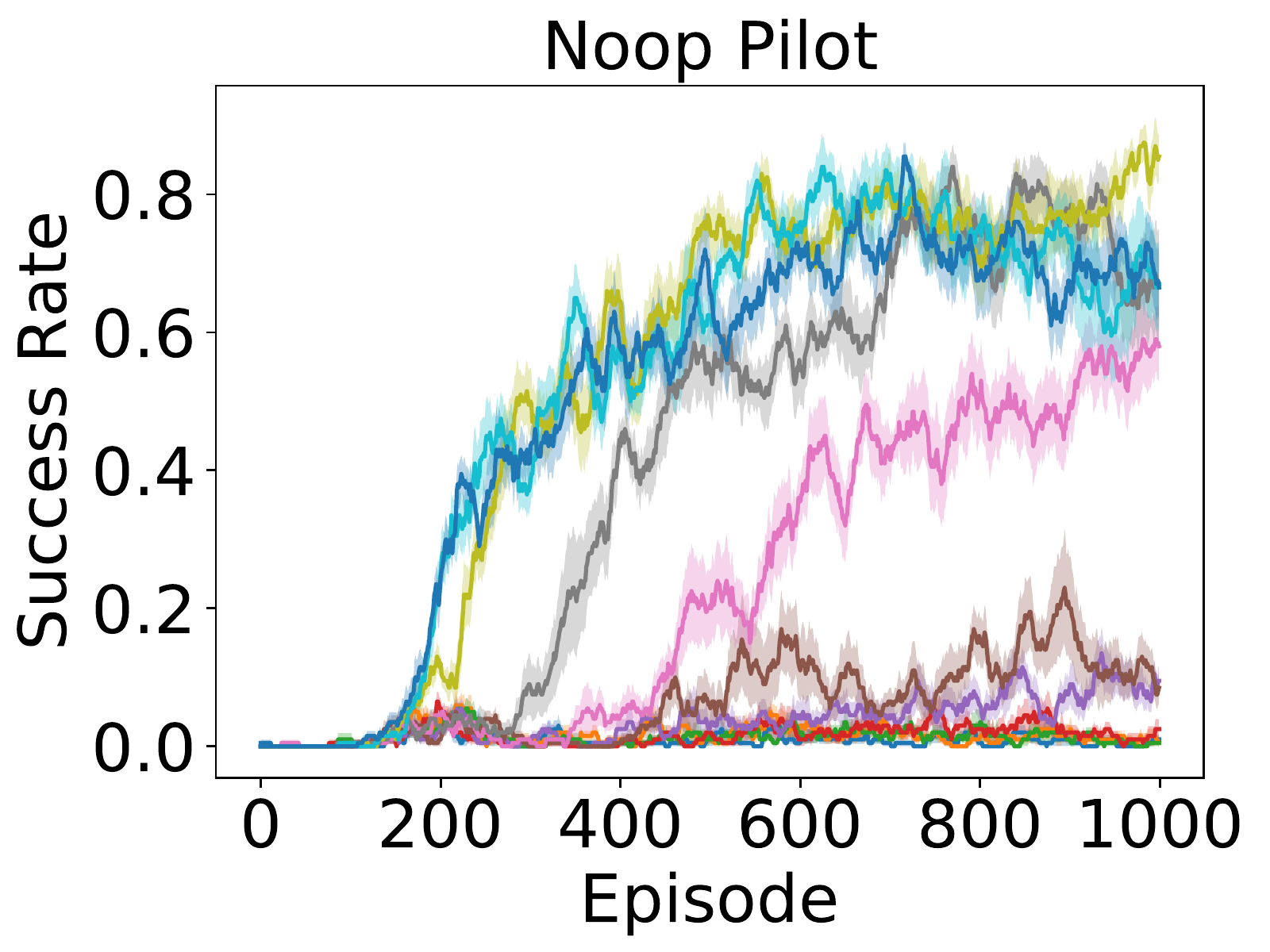}
         \end{subfigure}
         
         \begin{subfigure}[t]{\textwidth}
             \centering
             
             \includegraphics[width=\textwidth]{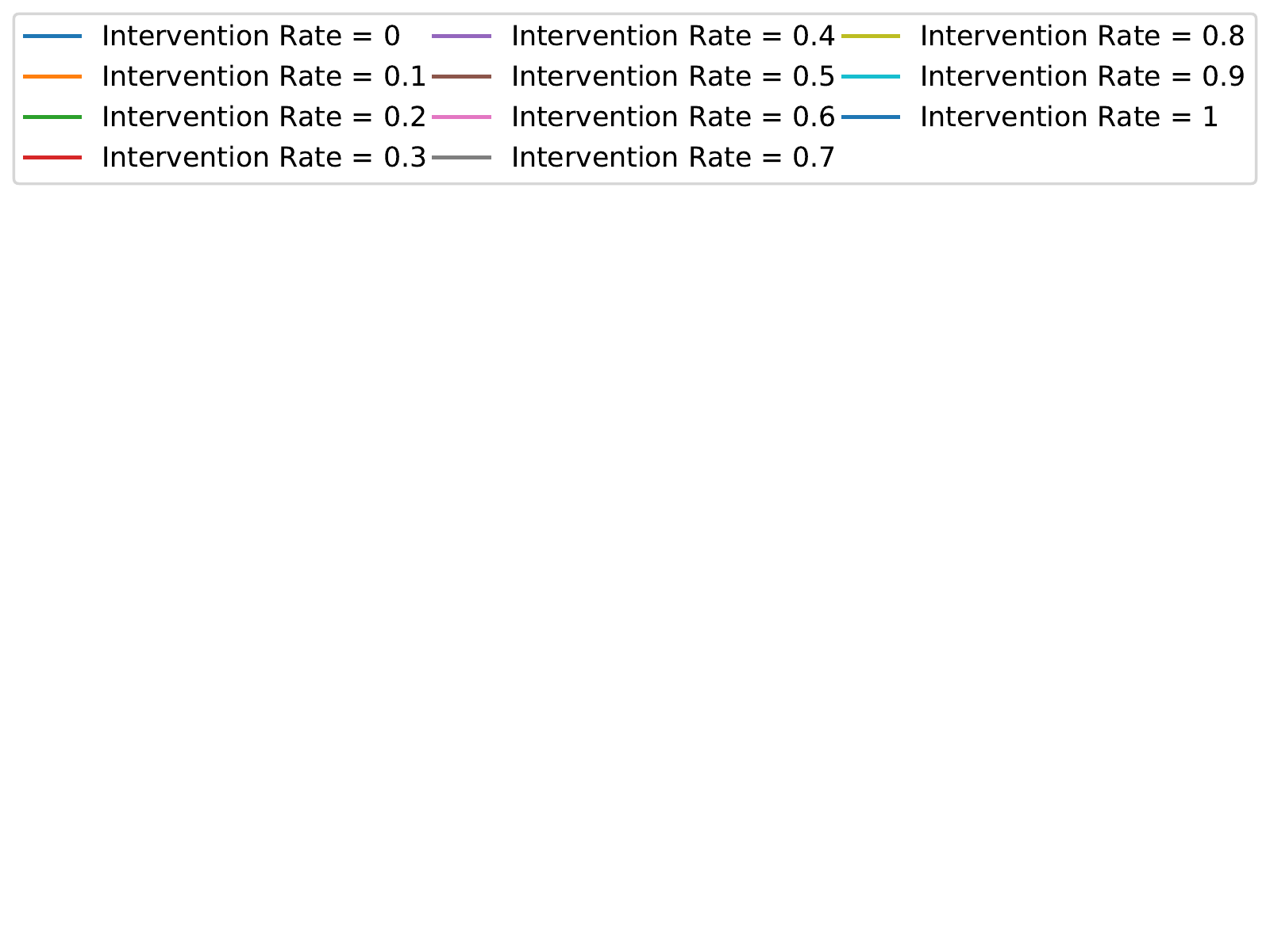}
          \end{subfigure}
     \end{subfigure}
     \vspace{-8.5cm}
     \caption{The learning curves for return, intervention rate, and success rate of the four types of pilots assisted by \textbf{penalty adapting} method copilot during training. Return, intervention rate, and success rate are smoothed using a moving average with a window size of 20 episodes.}
     \label{fig:learning_curve_adapt}
\end{figure*}

Figures \ref{fig:learning_curve_alpha}, \ref{fig:learning_curve_budget}, \ref{fig:learning_curve_penalty}, \ref{fig:learning_curve_adapt} show the learning curves of the baseline method and our budget method, penalty method and penalty adapting methods trained with DDQN in Lunar Lander with discrete action space (these results correspond with the experiments shown in Figure 2 in the main paper). Because we train each hyperparameter for 1000K frames, each experiment ends in different number of episodes, ranging from 1000 to 4000 episodes. We show the first 1000 episodes here. 

\clearpage

\subsection{Deployment} \label{deployalgo}

We briefly present the control loop for using an agent trained with our hard constrained method during evaluation time or a production setting. A very similar control loop can be derived for our soft constrained methods.

\begin{algorithm}[H]
\caption{Evaluation Loop for a Hard Constrained Agent}
\SetAlgoLined
\DontPrintSemicolon
    Initialize agent's policy: $\pi_a$, budget: $b \in \mathbb{Z}^{0+}$; \\
    \textup{Observe} $s_0$ \\
    \For{$t\gets1$ \KwTo $\infty$}{
        \textup{Sample} $a_t^h \sim \pi_h (a_t^h|s_t)$ \tcp*{get human action}
        $s_t^b = [s_t,a_t^h,b]$ \tcp*{modify state for agent policy} 
        \textup{Sample} $a_t^a \sim \pi_a (a_t^a|s_t^b)$ \tcp*{sample agent policy} 
        \uIf {$I(a_t^a, a_t^h) = 1$ and $b > 0$}{ 
                $b \leftarrow b - 1$ \tcp*{if an intervention occurred, update budget}
        }
        \uElse{
                $a_t^a \leftarrow a_t^h$ \tcp*{otherwise do not modify human action}
        }
        Execute action $a_t^a$ and observe $s_{t+1}$ \tcp*{take action and transition}
    }
\end{algorithm}

\subsection{Hyperparameters}
Table \ref{table:hyperparameter} shows the hyperparameters we use in the experiments.

\begin{table}[H]
\caption{The hyperparameters used to train on Lunar Lander and Super Mario. Alpha corresponds to the baseline method, budget for our budget method, penalty for our penalty method and intervention rate for our penalty adapting method.}
\label{table:hyperparameter}
\begin{tabular}{c|cc}
\hline
                                    & Lunar Lander                                                                                               & Super Mario Bros.                                                                       \\ \hline
Alpha                          & \begin{tabular}[c]{@{}c@{}}$[0,0.1,0.2,0.3,0.4,0.5,$\\ $0.6,0.7,0.8,0.9,1]$\end{tabular}                     & $[0,0.2, 0.4, 0.6, 0.8 ,1]$                                                           \\ \hline
Budget                              & \begin{tabular}[c]{@{}c@{}}$[0,25,50,75,100,125,150,175,200,225,$\\ $250,275,300,400,500,1000]$\end{tabular} & \begin{tabular}[c]{@{}c@{}}$[0, 50, 100, 150, 200,$\\ $250,300, 350, 400]$\end{tabular} \\ \hline
Penalty                             & \begin{tabular}[c]{@{}c@{}}$[0, 0.01, 0.02, 0.05, 0.1, 0.2, 0.5,$ \\ $1, 2, 5, 10, 20, 50]$\end{tabular}     & $[0, 0.05, 0.1, 0.2, 0.5, 1, 2, 5]$                                                   \\ \hline
Intervention rate                   & \begin{tabular}[c]{@{}c@{}}$[0,0.1,0.2, 0.3,0.4,0.5,$\\ $0.6,0.7,0.8,0.9,1]$\end{tabular}                    & $[0,0.2, 0.4, 0.6, 0.8 ,1]$                                                           \\ \hline
Discount factor                     & 0.99                                                                                                       & 0.9                                                                                   \\ \hline
Learning rate                       & 1e-3                                                                                                       & 1e-4                                                                                  \\ \hline
Target network update frequency     & 1500                                                                                                       & 1500                                                                                  \\ \hline
Final exploration rate              & 0.05                                                                                                       & 0.05                                                                                  \\ \hline
Final exploration frame             & 1e5                                                                                                        & 1e6                                                                                   \\ \hline
Training frame                      & 1e6                                                                                                        & 1e7                                                                                   \\ \hline
Replay start size                   & 1000                                                                                                       & 1000                                                                                  \\ \hline
Replay buffer size                  & 50000                                                                                                      & 10000                                                                                 \\ \hline
Prioritized experience replay alpha & $/$                                                                                                           & 0.4                                                                                   \\ \cline{2-3} 
Prioritized experience replay beta  & $/$                                                                                                           & 0.6                                                                                   \\ \hline
\end{tabular}

\end{table}

\end{document}